\documentclass{article}
\usepackage{float}
\usepackage{shapepar}
\usepackage{listings}
\usepackage{amsthm}
\usepackage{mathtools}
\usepackage{graphics}
\usepackage{amssymb}
\usepackage{mathrsfs}
\usepackage{framed}
\usepackage{diagbox}
\usepackage{booktabs}
\usepackage{fancybox}
\usepackage{geometry}
\usepackage{multirow}
\usepackage{enumerate}
\usepackage{caption}
\usepackage{subcaption}
\usepackage{hyperref}
\usepackage{boldline}
\usepackage[table]{xcolor}
\usepackage{slashbox}
\usepackage{tabularray}
\usepackage{vcell}
\usepackage{titlesec}
\usepackage[pagewise]{lineno}
\titleformat{\paragraph}
{\normalfont\normalsize\bfseries}{\theparagraph}{1em}{}
\titlespacing*{\paragraph}
{0pt}{3.25ex plus 1ex minus .2ex}{1.5ex plus .2ex}

\newtheorem{theorem}{Theorem}
\usepackage[ruled,linesnumbered]{algorithm2e}
\usepackage[square,numbers]{natbib}
\newtheorem{definition}{Definition}[section]
\newcommand \R {\mathbb{R}}
\newcommand \argmin {\mbox{argmin}}

\SetCommentSty{mycommfont}

\usepackage{authblk}

\usepackage{abstract}

\title{Deep Learning Methods for Partial Differential Equations and Related Parameter Identification Problems}

\author[$\ddagger \bowtie *$]{Derick Nganyu Tanyu}
\author[$\dagger \bowtie$]{Jianfeng Ning}
\author[$\ddagger$]{Tom Freudenberg}
\author[$\ddagger$]{Nick Heilenk{\"o}tter}
\author[$\ddagger$]{\\Andreas Rademacher}
\author[$\S$]{Uwe Iben}
\author[$\ddagger$]{Peter Maass}

\affil[$\ddagger$]{\small Centre for Industrial Mathematics (ZeTeM), University of Bremen, Germany}
\affil[$\dagger$]{\small School of Mathematics and Statistics, Wuhan University, China}
\affil[$\S$]{\small Robert Bosch GmbH, Germany}
\affil[$\bowtie$]{\small Both authors contributed equally}
\affil[*]{Corresponding author: derick@uni-bremen.de}
\begin{document}
	\maketitle
	%\tableofcontents
	%\footnotetext[*]{Corresponding author: derick@uni-bremen.de}
	\begin{abstract}
		Recent years have witnessed a growth in mathematics for deep learning\textemdash which seeks a deeper understanding of the concepts of deep learning with mathematics and explores how to make it more robust\textemdash and deep learning for mathematics, where deep learning algorithms are used to solve problems in mathematics. The latter has popularised the field of scientific machine learning where deep learning is applied to problems in scientific computing. Specifically, more and more neural network architectures have been developed to solve specific classes of partial differential equations (PDEs). Such methods exploit properties that are inherent to PDEs and thus solve the PDEs better than standard feed-forward neural networks, recurrent neural networks, or convolutional neural networks. This has had a great impact in the area of mathematical modeling where parametric PDEs are widely used to model most natural and physical processes arising in science and engineering. In this work, we review such methods as well as their extensions for parametric studies and for solving the related inverse problems. We equally proceed to show their relevance in some industrial applications.
	\end{abstract}
	
	\section{Motivation}
	Arguably, partial differential equations (PDEs) provide the most widely used models for a large variety of problems in natural sciences, engineering and industry. The ever-increasing complexity of these models such as digital twins for high-tech industrial applications requires the most efficient solvers. 
	
	These models follow the full life cycle of products from classical simulation and optimisation during the development phase to process monitoring and control during production.
	Such models typically rely on accurate calibration of critical parameters, which might be scalar parameters or even distributed spatial-time varying parameter functions. The calibration process itself requires multiple runs of the model and increases the need for efficient solvers even more.

	Not surprisingly, data-driven concepts and in particular neural network approaches have been studied extensively over the last few years as they have the potential to overcome three major obstacles in PDE simulations: First of all, no mathematical-physical model is ever complete, however, even the finest detail or tricky non-linearity is contained in a sufficient data set; secondly, the parameters to be determined most often are not arbitrary but follow an unknown, application specific distribution, which can be recovered and exploited from training data; thirdly, as already mentioned, the complexity of PDE-based modelling has reached a level such that conventional and most widely used methods such as finite elements or finite difference do require computational times beyond reasonable limits, while neural network concepts after training are most efficient.
	
	This is matched by our own experience over the last few years, where an increasing number of our industrial and engineering collaboration partners started to experiment with deep learning (DL) concepts based on neural networks for PDE problems. However, they as much as we were overwhelmed by the somewhat confusing variety of different DL concepts for these tasks. These span from general approaches for large classes of problems to very specifically tailored approaches for individual equations. The present paper is based on our endeavour to provide an unbiased overview of the most common general approaches and to provide limited, but scientifically based guidance on how to select appropriate methods. We are well aware of the many extension of the mentioned methods and the exploding body of literature dealing with such problems, and we readily apologise to those scientists, whose research we neglected to include.  
	%\end{align}
	To be more precise, we are considering second-order partial differential equations defined in a bounded domain $\Omega\subset \mathbb{R}^d$, which depend on a parameter function $\lambda$:  
	\begin{equation} \label{basicPDE} {\cal N}(u, \nabla u, \Delta u, u_t;\lambda) =  0 \end{equation}
	\begin{equation} \lambda : \Omega \to \R \end{equation}
	In this general notation, $\mathcal{N}$ encodes the differential equation as well as boundary conditions.  
	We always assume, that the parameter-to-state operator $F$, which maps
	a given parameter $\lambda$ to the solution of the PDE, is well-posed. I.e. we assume a function space setting such that the solution  $u$ of the PDE is unique and depends continuously on the parameter $\lambda$.   
	We will consider three related classes of problems:
	\begin{enumerate}[\bf Level 1:]
		\item  Forward problem: solving a single PDE: given $\lambda$, compute $u = F(\lambda)$
		\item  Parametric studies: given many parameters, $\lambda \in \{ \lambda_1, .. ,\lambda_N \}$, compute corresponding $u$'s
		\item  Parameter identification (inverse problem):    given a measured $u^\delta$ or its values $Pu^\delta$ under a measurement operator $P$, determine a corresponding $\lambda$, e.g. solve $F(\lambda) \sim  u^\delta$.
	\end{enumerate}
	
	Our main target is the third class of problems, i.e. inverse problems stated as parameter identification problems for PDEs. These problems are typically non-linear and ill-posed, even if the differential equation is linear and the forward operator is well-posed. Nevertheless, we start with the first problem and review the most common DL approaches for solving the forward problem. We then discuss their potential for parametric studies and parameter identification problems. As an underlying motivation for using DL concepts in this setting, we assume that an evaluation of the forward operator $F$ by classical methods, e.g. finite difference schemes or finite elements, is computationally expensive and not suitable for large-scale parametric studies.

	We have two remarks for clarifying the scope of the present survey.
	First of all, parameter identification problems are inverse problems and can be attacked by well-established general regularisation schemes for operator equations. For a recent overview of such data-driven concepts for inverse problems and their regularisation properties, see e.g. \cite{arridge_acta}. However, these concepts, e.g. unrolled iteration schemes, typically involve an evaluation of the forward operator or its adjoint. This will not lead to efficient schemes in our framework, where the evaluation of $F$ is assumed to be too costly for large-scale parametric studies. We will remark shortly on that in our section on the state of the art. Hence, in the present paper, we only consider DL concepts, where the forward operator itself, i.e. the parameter-to-state operator, or its inverse are replaced by a neural network.
	Secondly, there exists a growing number of highly optimised DL concepts for very specific PDE problems e.g. for coupled physics systems, molecular dynamics, or complex fluid dynamic problems. These approaches do have a limited potential for transfer to other problems, and we will not address these concepts. We will rather focus on general classes of DL concepts that apply to a larger variety of PDE-based problems. However, we will highlight some of those successful but specialised approaches in the section on the state of the art. All the codes and dataset used for the numerical experiment will equally be made available online on \href{https://github.com/dericknganyu/dl\_for\_pdes}{GitHub} (see also Section \ref{apdx:num_sett})
	
	\subsection{Classical methods for solving PDEs numerically}
	Let us first comment on the main competitors for data-driven concepts, namely the classical and well-established concepts of finite difference (FDM), finite element (FEM), boundary element Method (BEM), finite volume (FVM) or particle methods.
	
	The so-called finite difference method (FDM) is based on the replacement of the occurring derivatives by difference quotients. In this process, the function values of the solution are approximated at individual discrete grid points. The result is a so-called grid function. We refer, e.g., to \cite{Strikwerda04} for a more detailed description. The great advantage of the FDM is its straightforward realisation. However, its extension to more complex problems is limited and rather unrealistic smoothness assumptions are necessary for its analysis. 
	
	The finite element method (FEM) overcomes the mentioned disadvantages of FDM methods to a large extent. In contrast to the FDM, it is based on the weak formulation of the PDE under consideration and searches for the discrete solution over a finite-dimensional subspace of the underlying function space, see \cite{BrennerScott08} for more details. Especially in connection with adaptive and domain decomposition methods, it shows its full potential. An interesting variant is discontinuous Galerkin (dG) methods, in which the smoothness assumptions for the discrete solution, which stems from the weak formulation of the PDE, are only enforced by penalty terms.
	
	The boundary element method (BEM) follows a different idea. In this concept, the PDE is transformed into a boundary integral equation, which, however, is only possible in special cases. This reduces the dimension of the problem. The boundary integral equation is then discretised by means of FEM, for example. However, the resulting system of equations is dense and its numerical solution is often tricky and numerically very elaborate. Nevertheless, the BEM offers great advantages for exterior space problems in which an unbounded area is considered. We refer for instance to \cite{gwinner2018advanced} for more information.
	
	The Finite Volume Method (FVM) brings together ideas from the FEM and the FDM and combines them with arbitrary control volumes, see e.g. \cite{Leveque02}. The FVM is particularly well suited for the discretisation of hyperbolic equations and convection-dominated problems. However, its theoretical analysis is not yet that advanced. 
	
	Another approach are particle methods or discrete element methods (DEM), see e.g. \cite{meshfree}, which are based on a completely different approach. Here, individual particles and their interaction with each other are considered. By using different approaches to describe the interaction, different materials and processes can be simulated. They thus achieve very good accuracy in many applications, e.g. in the simulation of sand, which is difficult to simulate with the other methods. 
	
	The mentioned concepts are well-established for solving PDE-based forward problems and all of them have been adapted to inverse problems, in particular in the context tasks of optimal control or numerical parameter identification. They lead to accurate results in this context. However, two main difficulties have to be mentioned: Firstly and as said already in the previous section, solving these inverse problems requires solving the forward problem multiple times, which leads to unacceptable overall computing times at least for large problems. Furthermore, the optimisation algorithms for solving the above-mentioned parameter identification tasks require the first and, if possible, also the second derivative with respect to the unknown parameters. At first, it is often unclear whether the mentioned derivatives exist at all. In addition, their calculation is algorithmically very complex and time-consuming, especially for time-dependent problems. We refer, for example, to \cite{Troeltzsch2010:1} for a detailed discussion.
	
	In summary, the classical methods are well-researched and reliable and should be used whenever possible within the given restrictions of data and computer power. Hence, we once again stress, that the present paper refers to potential applications where solving the forward problem multiple times is simply too costly.

	\subsection{Deep learning for PDE solvers and scope of this survey}
	The main purpose of this section is to list the DL methods considered in this present survey. In the subsequent sections, we will then analyse their potential for parametric studies and inverse problems. 
	
	As already mentioned, mesh-based methods for solving PDEs such as finite element methods, finite difference methods, and finite volume methods, are the dominant techniques for obtaining numerical solutions of PDEs. When implementing these methods, the computational domain of interest should be first discretised using a set of mesh points and the solution is then approximated at the chosen grid points. The solutions are usually obtained from a finite linear space by solving a linear or non-linear system of equations. The classical methods are very stable and efficient for low-dimensional problems and regular geometries. They have been well understood in terms of existence, stability and error estimate, and we can usually achieve the desired accuracy by increasing the resolution of the discretisation.  In addition to the more general remarks above, the classical methods have several additional drawbacks and limitations. Firstly, the curse of dimensionality has limited the use of mesh-based methods, since the number of discretisation points will increase exponentially with the dimension, and there is no straightforward way to discretise irregular domains in high-dimensional spaces. Secondly, traditional methods are designed to model one specific instance of the PDE system, not the parameter-to-state operator. Thus, for any given new instance of the PDE system, the problem has to be solved again. Thirdly, the numerical solution is only computed at the mesh points and evaluation of the solution at any other point requires interpolation or some other reconstruction method. Fourthly, classical methods require the knowledge of the underlying analytic system of differential equations, but sometimes the exact physics is unknown. Lastly, solving inverse problems by classical problems is often prohibitively expensive and requires complex formulations, repeated computation of forward problems, new algorithms and elaborate computer coding. 
	
	Deep learning methods have shown great power in addressing the above difficulties. Since there are many different deep learning methods for PDEs, and each method has its particular properties, we will discuss them one by one. We should add, that the theoretical investigation of deep learning concepts for PDEs, beyond more or less direct applications of the universal approximation theorem, is still scarce.
	This is in contrast to the classical PDE theory, where there exists an exhaustive literature on the analytic properties of different types of PDEs. The theoretical characterisation of PDEs, as well as the type of theoretical background needed for proving solvability and uniqueness, differentiates between linear and non-linear elliptic, parabolic or hyperbolic equations in terms of the given boundary conditions. Analytic tools are different for these different classes, hence it is somewhat surprising that DL concepts for PDEs hardly ever use this classical classification but rather use a classification based on whether or not the PDE (no matter which type of PDE) is known, which type of data is given for training and which type of network architecture as well as which loss function for training is used.
	
	We follow this data-driven classification scheme. Our first level of characterisation is based on the input and output structure of the network. A first class of networks takes $x$ or $(t,x)$ as input and outputs a one- or low-dimensional scalar value or vector $u_\Theta (\cdot)$.
	During training the weights $\Theta$ of the network are optimised such that $u_\Theta (\cdot) = \hat u(\cdot)$ is an approximation
	to the solution $u$ of the PDE at the specified point $x$ or $(t,x)$, i.e. a  regression problem for the function $u_\Theta  \approx  u$ is solved.  These networks aim at a function approximation.
	Otherwise, one aims at an operator approximation for the parameter-to-state map and uses networks which take a parameter function, either in discretised form or as a vector of precomputed feature values,  as input and outputs the full solution function $u$. The solution $u$  again is delivered either in discrete form or as a related feature vector, that can be turned into a proper function representation in a post-processing step. The general idea of these two characterisations is summarised in the following Figures \ref{fig:pinn_general}-\ref{fig:pcann_general}: 
	\begin{itemize}
		
		\item Function evaluation, neural network $u_\Theta (t,x) \approx {u}(t,x)$ based on the PINN concept, \cite{raissi2019physics} \par
		\begin{minipage}{\linewidth}
			\centering
			\includegraphics[width=0.8  \linewidth]{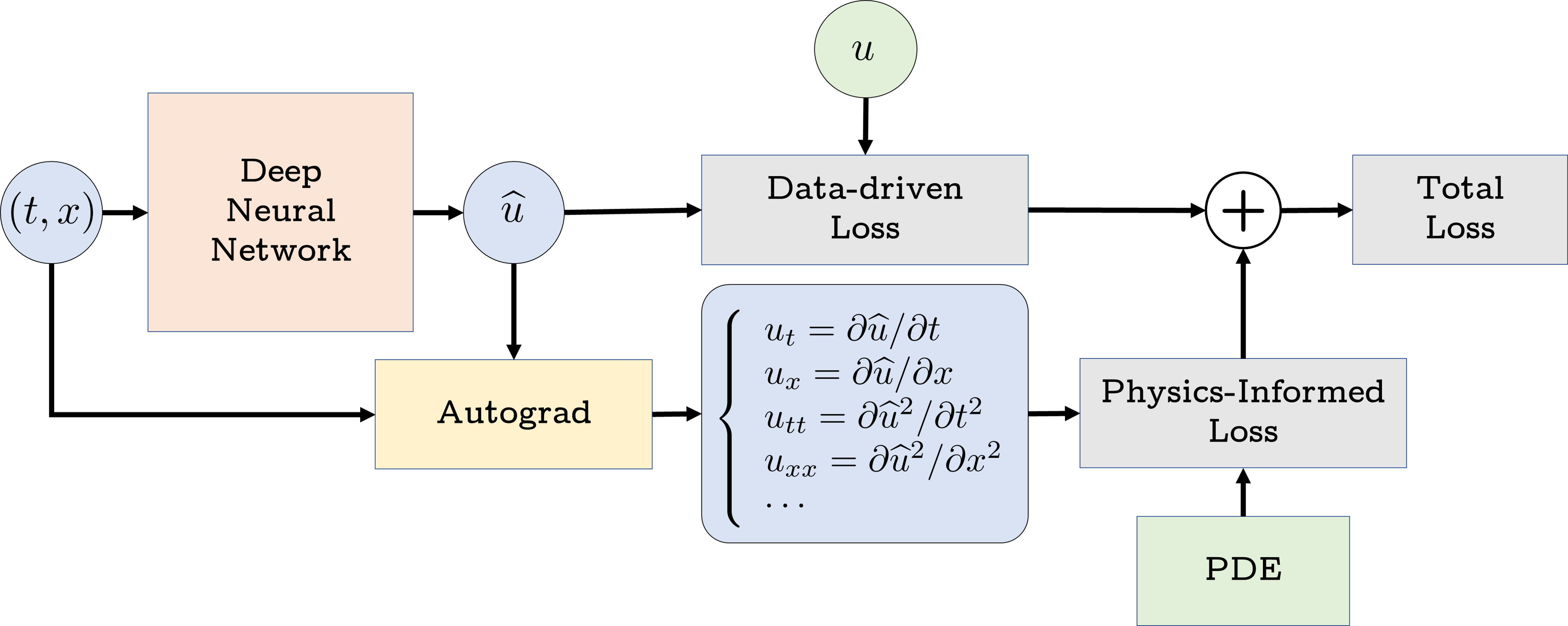}
			\captionsetup{type=figure}
			\captionof{figure}{A network with a low input dimension for $(t,x)$, which outputs a scalar or low-dimensional $\hat{u}(t,x) = u_\Theta (t,x)$. Hence, the neural network is trained to be a function approximation of the solution $\hat u(t,x) \approx u(t,x)$. During training, automatic differentiation by backpropagation is used to compute all necessary derivatives for checking, whether the PDE is satisfied. In addition, the loss function typically sums up over several values of $(t,x)$ and their corresponding function values. After
				training, the Deep Neural Network is then used and $\hat{u}$  can be evaluated at arbitrary points $(t,x)$.}
			\label{fig:pinn_general}
		\end{minipage}
		
		\item Operator evaluation, neural network  $\Phi_\Theta (\lambda)$ following the  PCANN concept \cite{bhattacharya2020model} \par
		\begin{minipage}{\linewidth}
			\centering
			\includegraphics[width=0.5\linewidth]{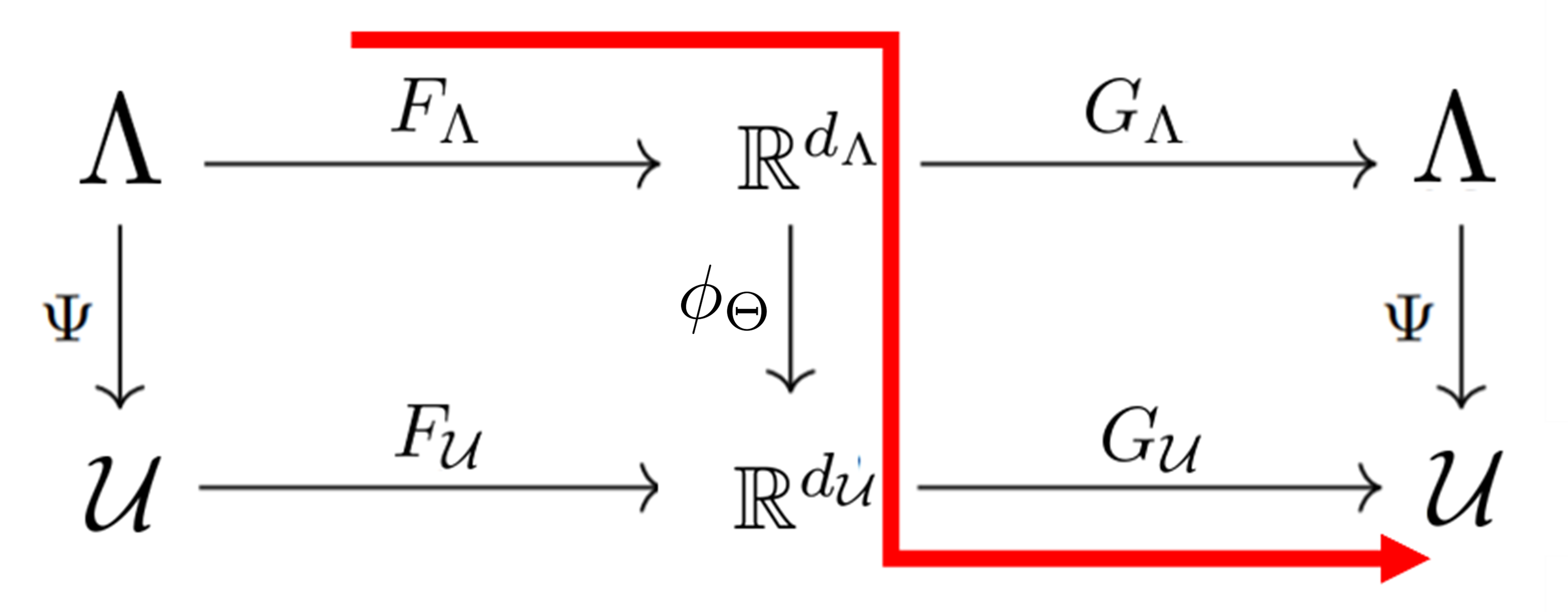}
			\captionsetup{type=figure}
			\captionof{figure}{A typical setup for an operator approximation scheme. Here an order reduction model (see \cite{schilders2008introduction,chinesta2016model,benner2020model} for an overview of reduction models) in combination with a neural network is employed. First, the parameter $\lambda \in \Lambda$ is reduced to a feature vector in $\mathbb{R}^{d_\Lambda}$ by determining its scalar products with a pre-computed reduced basis, i.e. $c_k=\langle  \lambda, b_k \rangle $ (which in case of an orthonormal system is equivalent to $\lambda \sim \sum c_k b_k    \in \Lambda$). Only the feature vectors are mapped by the neural network $\phi_\Theta: \mathbb{R}^{d_\Lambda} \ni c \mapsto \tilde c \in \mathbb{R}^{d_\mathcal{U}}$. The output features $\tilde c$ are then used to determine an expansion of the sought-for solution $ \hat u = \sum_\ell \tilde c_\ell  \tilde b_\ell$ in terms of a basis 
				$\{\tilde b_\ell \}$ in the solution space. In this way, we then approximate the parameter to solution function $\Psi : \Lambda \ni \lambda \mapsto u \in \mathcal{U}$ by $\Phi_{\Theta} = G_{\mathcal{U}} \circ \phi_\Theta \circ F_{\Lambda}$ as indicated by the red arrow.   $F_{\Lambda}$ and $F_{\mathcal{U}}$ are encoder functions (PCA maps) in the parameter and solution spaces respectively, while $G_{\Lambda}$ and $G_{\mathcal{U}}$ are their respective decoder functions. (see \ref{sec:pcann} and \cite{bhattacharya2020model}).}
			\label{fig:pcann_general}
		\end{minipage}
		
	\end{itemize}

	\begin{table}[ht]
		\centering
		\begin{tabular}{|c|c|c|c|cc|c|}
			\hline
			\multirow{2}{*}{\textbf{Networks}} & \multirow{2}{*}{PDE usage} & \multirow{2}{*}{Data usage}  &  Neural Network & \multicolumn{2}{c|}{discretisation} & \multirow{2}{*}{Reference} \\
			&  &  & approximation &  Parameter & Solution & \\ 
			\hlineB{3}\hline
			Deep Ritz  & yes & no & function & no & no & \cite{yu2017deep,liao2019deep} \\
			PINN & yes & no  & function & no & no  & \cite{raissi2019physics}\\
			WAN & yes & no  & function & no & no & \cite{zang2020weak, bao2020numerical}\\ 
			\hline
			PCANN & no & many & operator & free grids & free grids& \cite{bhattacharya2020model} \\
			FNO & no & many &operator & free grids & free grids  & \cite{li2020fourier}\\
			UFNO & no & many &operator & free grids & free grids& \cite{wen2022u}\\
			MWT & no & many &operator & free grids & free grids & \cite{gupta2021multiwavelet}\\
			DeepONet & no & many & operator & fixed & no & \cite{lu2019deeponet}\\ %\multicolumn{2}{c|}{input}
			\hline
			PINO & yes & if needed &operator & free grids & free grids  & \cite{li2021physics}\\
			PI-DeepONet & yes & if needed &operator & fixed & no &  \cite{wang2021learning}\\
			\hline
		\end{tabular}
		\caption{The list of DL concepts considered in this survey is based on a pre-selection mirroring our own interest in inverse problems. Hence, concepts for operator approximation are preferred. In their original papers, most of these concepts were designed for Level 1 problems only. Hence, we need to specify in later sections how to extend them for Level 2 or Level 3 problems. Also, we will later refine the features characterising the different methods in terms of limitations, e.g. which PDE equations are most suitable and which are not.}
		\label{tab:methods}
	\end{table}
	
	Typical examples of DL concepts aiming for a function approximation are Deep Ritz method \cite{yu2017deep,liao2019deep}, Physics-Informed Neural Network (PINN) \cite{raissi2019physics}, Deep Galerkin Method (DGM) \cite{sirignano2018dgm}, Weak Adversarial Networks \cite{zang2020weak}, and Deep Splitting Approximation methods \cite{beck2020overview}. In all these concepts, information of the underlying physics must be used when training the neural network, i.e. the PDE must be known analytically. These concepts are usually mesh-independent and accurate, while they require knowledge of the governing laws of the PDEs, and the optimisation problem needs to be solved for every new instance. This is similar to classical methods. 
	
	Examples for the second type, i.e. methods aiming at an operator approximation, include Model reduction neural network (PCANN) \cite{bhattacharya2020model}, Deep Operator Network (DeepONet) \cite{lu2019deeponet} and its extensions \cite{wang2021learning, goswami2021physics}, Fourier neural operator (FNO) \cite{li2020fourier} and its physics-informed extension--Physics Informed Neural Operator (PINO) \cite{li2021physics}, Graph Kernel Network \cite{li2020neural, li2020multipole}, Bayesian deep convolutional encoder-decoder networks \cite{zhu2018bayesian}, Wavelet Neural Operators \cite{tripura2022wavelet}, Multiwavelet based operator (MWT) \cite{gupta2021multiwavelet} and many more. These operator methods usually learn the neural network with some paired input-output (parameter-solution) observations with or without the knowledge of the physical system (using the physics information could sometimes alleviate the need for much data as in the case of PINO and PI-DeepONet) as highlighted in Table \ref{tab:methods}. Thus, the neural network only needs to be trained once, and a new parameter identification task can be directly solved by a forward pass of the network. Some of these deep learning methods still need to discretise the domain of interest, and the solutions are sought in a finite linear space \cite{zhu2018bayesian}.  A recently published paper \cite{de2022cost} aims at comparing some of these operator approximation concepts for solving various PDEs. The primary criterion for comparison in this well-written paper is the cost-to-accuracy curve.
	Accordingly, the authors compare results obtained with different network sizes. However, their comparison includes the basic methods PCANN, DeepONet and FNO which in our tests did not yield the best results. Also, the numerical test stays in a range - when compared with FEM methods - of rather large approximation errors. This matches our experience, it is hard to get to high precision for forward solvers with DL concepts. Anyway, our focus is on inverse problems, which are not covered in the mentioned paper.
	
	In terms of the involved computational load, the computational cost of classical methods for solving linear PDEs is mainly determined by the need to solve a large linear system. For deep learning methods, it is the training process (optimisation) of the neural network. As a general rule of thumb, for low dimensional problems with regular geometries, classical methods are in general more powerful and efficient than deep learning methods. Ignoring rounding errors, the main error of the classical methods arises from the step size of the discretisation. There exists a trade-off on the resolution: coarse grids are fast but less accurate; fine grids are more accurate but slower.
	On the other hand, for DL approaches one usually has to carefully choose the neural network, optimisation method and hyperparameters in consideration of the underlying differential equation.
	Additionally, the error of deep learning methods is usually difficult to estimate and consists of several factors. The accuracy can be characterised by dividing the total error into three main types: approximation, optimisation, and generalisation errors. More specifically, the approximation error measures the smallest difference between the neural networks and the underlying function or operator, which we aim to approximate. This error is influenced by the size and architecture of the neural networks. The optimisation error arises from the non-convexity property of the loss function and the limited number of iterations used for its minimisation. Stochastic Gradient Descent is an efficient algorithm for escaping local minima and reducing the optimisation error, nevertheless, in real-life applications, we never achieve a global minimum. The generalisation error refers to how the trained networks perform on unseen data and is mainly affected by the amount of training data, the modelling information of the PDEs, and the architecture of the neural networks.

	The selection of DL-based PDE solvers which are further investigated in this survey is based on our endeavour to cover a most complete range of different concepts.  Hence we included the basic concepts of learned DL solvers as well as some recent improvements of these. 
	
	A more detailed description of these approaches, their algorithmic implementation and their analytical properties will be given in the subsequent sections. There we will also highlight, how these methods, which were typically designed as PDE-forward solvers can be extended to parameter identification problems.
	
	\subsection{Some remarks on concepts not covered in this survey}
	In this section, we want to highlight those directions of research, which are not covered in this survey. 
	
	First of all, there exist several specialised and powerful DL concepts for complex but specific PDEs.
	Research on neural networks for such specialised cases of PDEs as well as related parametric studies and inverse problems is exploding and it is impossible to give an exhaustive list of references. Hence, we only want to highlight areas of research, which have reached a certain level of maturity in terms of experimental success as well as mathematical rigor. Again, this selection is based on personal preferences and we apologise to those, who quite rightly might want to see their own work included.
	
	A particularly successful area for DL applications are coupled-physics systems, which lead to some of the most complex inverse problems. Typically, those inverse problems cannot be solved by standard concepts, neither in the classical analytical regularisation setting nor in data-driven frameworks. Hence, such problems, e.g. opto-acoustic tomography, require specialised concepts, which typically integrate domain-specific expert knowledge or at least partially draw their motivation from analytic reconstruction formulae, see e.g.
	\cite{coupledphys_colombo,Thuerey2021,DL_photo_hauptmann,DL_diffusion_arrridge}. While we are at it, we also mention successes of DL methods in other sub-fields of tomography such as electrical impedance tomography \cite{hamilton2018deep, herzberg2021graph}, diffuse optical tomography \cite{mozumder2021model} and computed tomography \cite{leuschner2021quantitative, leuschner2021lodopab, leuschner2021lodopab}.
	Another class of papers do take numerical schemes for PDEs as a starting point for developing network architectures, see e.g. \cite{Ruthotto_stable,Ruthotto_special}. In these papers, the different layers of a neural network are regarded as time stepping in a discretised scheme. In particular, this allows to specifically mimic and improve numerical schemes for parabolic equations, which e.g. are the basis for many imaging problems based on diffusion processes.  
	Simulation of turbulent flows also is an area, where DL concepts have shown good success. We only list some papers, which in our opinion together with the cited literature therein allow an overview of this topic
	\cite{DL_turbulence_Rui,DL_turbulence_milani,DL_turbulence_Gottschalk}.  One should also mention the spectacular results on simulating molecular dynamics \cite{DL_moldyn_adler,DL_moldyn_chem} and many other fields of applications, where DL simulations lead to groundbreaking novel insights.  
	
	We also do not cover the topic of DL for stochastic PDEs and recent results, which we regard to be more on the side of PDE theory and simulation. Hence, we do not cover recent developments as e.g. described in \cite{DL_jentzen,DL_burger,Dittmer_2019}.

	Secondly,  we want to remark on general regularisation schemes for inverse problems using a data-driven approach, which can be applied to but are not particularly tailored for PDE-based parameter identification problems.
	Typically these general data-driven regularisation schemes for parameter identification problems do need an evaluation of $F$ or any form of its adjoints. For reference to this type of schemes we refer to \cite{arridge_acta} or to \cite{holler_2022} for a recent paper, which embeds the learning of the network together with the parameter identification into a regularisation scheme. We would like to remark that unrolled iteration schemes such as LISTA or unrolled primal-dual, are somewhat hybrid methods and could be included in our survey. However, LISTA is predominately successful for linear forward operators $A$ and it still needs the adjoint  $A^*$ for initialising the input of the network. One could extend such DL schemes, which are derived from classical regularisation theory, in such a way that the evaluation $F$ or its adjoint would be done by training an embedded network for every operator evaluation.  However, this leads to a  rather complex training task and this extension is not part of the scope of the original papers for e.g. primal-dual, NETT or DeepPrior or similar \cite{primal_dual,DL_NETT,Dittmer_2019}.

	\section{Characterisation of DL concepts for PDEs}
	
	Having specified the restricted scope of the present survey, we will now list the concepts under consideration in the following sections. Before starting the introduction of the individual methods we include some general remarks on the characterisation of deep learning methods for PDEs. As already mentioned, the characterisation of DL concepts for solving PDEs differs considerably from the analytic characterisation of PDEs. They are mainly characterised in terms of the information needed for applying them. The two main classes of information are data and physical models, where 'physical models' refers to the mathematical formulation of the differential equations and their boundary conditions.

	\subsection{Data} 
	Data-driven methods rely on the availability of sufficiently large data sets of good quality. 
	The data provided can be sampled values of single or multiple solutions, where sampling can be done either on a predefined grid or with arbitrary sampling points. Some of the most advanced theoretical investigations do start from assuming a sampling in function space, which opens the path for incorporating functional analytical machinery in their analysis, see e.g. \cite{bhattacharya2020model,raissi2018deep,holler_2022}.
	
	However, the volume of available data is usually rather scarce,e.g. useful experimental data is generally limited or even intractable for many practical scenarios and high-fidelity numerical simulations are often computationally expensive. Sometimes it is also a challenge to extract interpretable information and knowledge from the data deluge.  Moreover, purely data-driven methods may fit observations very well, but predictions may be physically inconsistent or implausible. Theoretically, finite data can not fully and correctly determine a mapping which maps an infinite set to an infinite set. However, we can still see the success of data-driven methods, which is mainly due to two reasons: first of all the search for a solution or parameter typically follows a certain prior distribution, which we might be able to  approximate well with few samples; and secondly
	most problems are continuous, which in connection with  suitable regularisation schemes leads to good generalisation properties even for limited sets of training data, \cite{raissi2019physics}
	
	In addition to sampled data, there exists a wealth of domain-specific expert knowledge for most applications.
	This information may come in the form of observational, physical or mathematical understanding of the system under consideration. Integrating data and this prior knowledge can yield more interpretable machine learning methods and can improve the accuracy and generalisation performances without large amounts of data. 
	
	In our context information from physics may come in different forms: most commonly, the system of partial differential equations as well as boundary conditions are specified. Otherwise, some energy functionals or a weak formulation may be given.  
	Also, certain approaches \cite{schildersmsode} incorporate conservation laws in the network architecture.
	Whenever such information is used, the concept is typically called 'physics informed'.
	
	Different from data-driven methods, the physics information may uniquely determine the systems, however, only within the limitations of the chosen model. For example, given the PDE formula and some initial and boundary equations, the solution is uniquely determined if the system is well-posed. In other words, mathematical formulae may inherently contain all information of the physical systems, while data are just some observations and reflections of the physical systems and thus can not fully reveal the whole information of the systems. 
	
	The most common approach is to use this physics information in the loss function during the training of the network. Hence, the type of physics information determines the loss function.

	\subsection{Network training and application}
	Training the resulting networks typically leads to additional problems. This may require extensive hyperparameter search,  data preprocessing or subtle parameter settings for controlling the convergence of the training scheme.  
	We have listed the respective training concepts for each method in the Appendix \ref{sec:appendix}.

	%In real applications, the information we have can be characterised by box1 from \cite{karniadakis2021physics}. The leading motivation is to develop an algorithm that can seamlessly integrate noisy data and abstract mathematical operators and thus can improve accuracy and generalisation performance.

	%\begin{figure}[htp]
	%	\centering
	%	\includegraphics[width=12cm]{figures/others/Box1}
	%\end{figure}

	As pointed out in \cite{karniadakis2021physics}, specific to physics-informed learning, there are currently three pathways that can be followed separately or in tandem to accelerate training and enhance the generalisation of ML models by embedding physics in them. The first one is \textbf{observational biases} from observational data, which is the foundation of data-driven methods in machine learning. The second is \textbf{inductive biases} by designing special NN architectures that can automatically and exactly satisfy some properties and conditions. The most common example are convolutional NNs which can be designed to respect properties, such as symmetry, rotations and reflections. Other examples are Graph neural networks \cite{li2020neural} which are adapted to graph-structured data. Another concept is based on Quadratic residual networks(QREs) \cite{bu2021quadratic}, which increases the potential of the neural networks to model non-linearities by adding a quadratic residual term to the weighted sum of inputs before applying activation functions. They were shown to have good performance in learning high-frequency patterns. The invertible neural network \cite{ardizzone2018analyzing} makes the inverse of the neural network well-defined. It has applications in inverse problems and generative models. Specific architectures can also be designed to satisfy the initial/boundary conditions. The last pathway is \textbf{learning biases}, where the physics constraints are imposed in a soft manner by appropriately penalising the loss function of conventional NN approximations. Representative examples include the Deep Ritz method and PINN. 
	
	We also want to comment on the arguably most important tools for physics-informed machine learning, namely automatic differentiation \cite{baydin2018automatic}.  In principle, this general concept us to compute exactly the derivatives of the network output with respect to the input variables. Hence, we can incorporate weak or strong formulations of the underlying PDE conveniently.  However, compared to conventional numerical gradients such as finite difference, the automatic differentiation method is usually slower and requires more memory.

	Moreover, after training, the methods will again differ substantially in their scope of applicability. Methods might allow the sampling of a solution of a PDE at arbitrary points. In this case, the method is called meshfree, which is one of the most desirable properties of most DL concepts for PDEs. Another important property, in particular for Level 2 and Level 3 problems, is the computational load needed for applying the network to a different parameter of the PDEs. This is most efficient for most operator learning schemes but might require re-training or the computation of specific feature vectors.

	In conclusion, different deep learning methods for PDEs are mainly different in terms of their neural network architecture and the particular choice of the loss function for training.

	\subsection{General DL concepts for PDE-based inverse problems}\label{subsec:Tikhonov_inverse}
	So far we have sketched two main approaches for solving PDEs (forward problems), namely function or operator approximation schemes. In the following sections, we will introduce these methods and their baseline implementation in more detail.  
	We will also include, whenever appropriate, the related concepts for the PDE-based inverse problems. 
	
	The extension of function approximations $u_\Theta (t,x)$ for solving inverse problems is not straightforward and differs from method to method.
	In contrast to that, the application of learned operator approximation schemes $\Phi_\Theta (\lambda)$ to inverse problems is always based on one of the two following concepts:
	\begin{itemize}
		\item The direct (or backward) method, where the network is trained with reversed input-output. In other words, the solution of the PDE is the input of a neural network which outputs the parameter function. This method is similar to the forward problem, but we learn the backward operator.
		
		\item Tikhonov Regularisation, for this case, the forward operator $\Phi_\Theta (\lambda)$ is trained as usual. For an inverse problem with given noisy data
		$u^{\delta}$ we then approximate a suitable parameter $\lambda$ by minimising a Tikhonov functional with respect to $\lambda$
		$$\| \Phi_\Theta (\lambda) - u^\delta \|^2 + \alpha R(\lambda) \ .$$
		
		The choice of the penalty term $R$ adds to the flexibility of the method and can encode further properties of the solution such as smoothness, sparsity or piece-wise constant structures. minimisation is done iteratively via gradient descent, where the differentiation with respect to $\lambda$ can be implemented using backpropagation with respect to the input but with fixed weights $\Theta$. 
		We exemplify the procedure for the model reduction concept PCANN in Algorithm \ref{algo:PCANN_inv}. A similar idea can be applied to the other solution operators mentioned in Sections \ref{sec:FNO} and \ref{sec:DeepONet}.
	\end{itemize}
	
	\section{Theoretical Results of Deep Learning for PDEs}
	
	One reason for the success of finite element methods is that the functions in Sobolev space can be approximated by piece-wise polynomials with a certain convergence rate depending on the mesh size. However, classical methods require degrees of freedom of the order of $\varepsilon^{-d}$ for a desired accuracy $\varepsilon$ in $d$ dimensional problems, which limits their applications to high-dimensional problems.  The task of deep learning for PDEs is to employ neural networks to approximate the solution or parameter-to-state operator with a limited amount of data or physical information. A fundamental question is therefore whether the neural network can effectively approximate the solution/parametric operator of a PDE system, and if so, how to estimate the complexity of a neural network in terms of the dimensionality growth and accuracy requirements. In this section, we will briefly review some approximation theoretical aspects of deep learning for PDEs. 
	
	\subsection{Approximation Results for Solution Learning in PDEs}
	Neural networks are universal approximators \cite{cybenko1989approximation,hornik1989multilayer} in the sense that  continuous functions on  compact domains can be uniformly approximated by neural networks with arbitrary accuracy, provided the number of neurons is sufficiently large. A lower bound on the size of the neural network for achieving a desired accuracy was not given in these papers. With some assumptions on the activation functions and the space of functions to be considered, more refined results on the relationship between the size and the approximation accuracy of neural networks have been reported in e.g., \cite{barron1994approximation,bolcskei2019optimal,mhaskar1996neural,shaham2018provable,lu2021deep,devore2021neural,gribonval2022approximation,guhring2020error,guhring2020expressivity}. The universal approximation theorem has also been generalised to convolutional neural networks in \cite{zhou2020universality} and complex-valued neural networks in \cite{voigtlaender2023universal}.
	These results concern general function approximations and do not refer to PDE solutions.

	\subsubsection{Neural Networks Approximation for High Dimensional PDEs}
	We now shortly discuss the theoretical foundations for approximating solutions of PDEs by neural networks. Recent years have witnessed great empirical success of various deep learning methods \cite{yu2017deep,han2018solving,hure2020deep,zang2020weak} in particular for solving high-dimensional PDEs. To explain these successes, several mathematical results have been proven which show that neural networks have the ability to approximate solutions of some PDEs without the curse of dimensionality (CoD). In particular, in \cite{grohs2018proof} it is shown that the solutions of linear Black-Scholes PDEs can be approximated by neural networks with the size increasing at most polynomially with respect both the reciprocal of the prescribed approximation accuracy and the PDE dimension $d$. A number of articles have appeared that significantly extend the results in \cite{grohs2018proof} to more classes of PDEs \cite{beneventano2020high,gonon2022uniform,hornung2020space,hutzenthaler2020proof,jentzen2018proof,reisinger2020rectified}. In particular, in \cite{hutzenthaler2020proof} it is proven that neural networks have the expressiveness to overcome the CoD for semilinear heat PDEs.
	
	On the other hand, there are also some articles, e.g. \cite{grohs2023lower,petersen2018optimal,yarotsky2017error,grohs2021proof}, which derive some lower bounds on the complexity of neural networks with ReLU activation function to achieve a certain accuracy and show that there are natural classes of functions for which deep neural networks with ReLU activation cannot escape the CoD.

	\subsubsection{Barron Spaces}
	
	Typical classical approximation results for PDE solutions use a setting in Sobolev or Besov spaces.
	These spaces can be defined by their approximation properties using piece-wise polynomials or wavelets. In the same sense, the Barron space \cite{chen2021representation,ma2022barron} is the analogue when we consider approximation by two-layer neural networks. Roughly speaking, the Barron space consists of the functions that can be approximated by two-layer neural networks, and the approximation error is governed by the norm of the Barron space. A nice property of Sobolev/Besov type spaces is that solutions to partial differential equations lie in these spaces. This is the core of regularity theory for PDEs. When introducing Barron spaces in PDEs, a natural corollary and fundamental question is whether the solutions of the dimensional partial differential equations (PDEs) we are interested in belong to a Barron space. We will briefly review the definition of the Barron space and its approximation theorem, as well as an example of its use in escaping CoD in high-dimensional PDEs. 
	
	Consider functions $g \subset \mathbb{R}^d: X \rightarrow R$ which allow the below representation:
	\begin{equation}
		g(x;\Theta)=\int_{\Omega}a\sigma(w^{T}x+b)\rho(da,dw,db), \quad x\in \mathbb{R}^{d},
	\end{equation}
	where $\rho$ is a probability distribution on $\Omega=\mathbb{R}\times \mathbb{R}^d\times \mathbb{R}$ and $\sigma:\mathbb{R}\rightarrow \mathbb{R}$ is some activation function. This representation can be thought of as a continuum realisation of two-layer neural networks, which are given as as
	\begin{equation}
		g_n(x,\Theta)=\frac{1}{n}\sum_{k=1}^{n}a_k\sigma(w_k^{T}x+b_k),\quad x\in \mathbb{R}^d \ .
	\end{equation}
	We present the following definition from \cite{chen2021representation} for  Barron functions defined in a domain $\Omega\subset \mathbb{R}^d$. This definition is an adaptation of the definition given in \cite{ma2022barron}, with some crucial modifications, for the purpose of analysing PDEs. 
	\begin{definition}
		For a domain $\Omega\subset \mathbb{R}^d$ and a fixed radius $R\in[0,+\infty]$, the corresponding Barron space with index $p$ is defined as
		\begin{equation}
			\mathcal{B}^p_R(\Omega)=\{f:\Vert f\Vert_{\mathcal{B}^p_R(\Omega)}<\infty\},
		\end{equation}
		where
		\begin{equation}
			\Vert f\Vert_{\mathcal{B}^p_R(\Omega)}:=\inf_{\rho\in\mathcal{P}_R}\bigg \{ \big(\int|a|^p\rho(da,dw,db) \big)^{1/p}:f =\int_{\Omega}a\sigma(w^{T}x+b)\rho(da,dw,db)\bigg\}
		\end{equation}
		and $\mathcal{P}_R:=\bigg\{\rho:\rho \quad \text{is supported on} \quad \mathbb{R}\times \{x\in \mathbb{R}^d:\Vert x\Vert\le R\}\times \mathbb{R} \bigg\}$.
	\end{definition}
	
	The most essential feature that distinguishes  Barron function spaces from  Sobolev or Besov spaces is that elements of the former can be approximated by a two-layer neural network with an approximation rate that is independent of the dimension, as shown in the next theorem \cite{chen2021representation}:
	\begin{theorem}
		Suppose that the activation function $\sigma$ is smooth with $C_0:=\sup _{z\in \mathbb{R}}|\sigma(z)|\le\infty, C_1:=\sup _{z\in \mathbb{R}}|\sigma '(z)|\le\infty$ and $\sup _{z\in \mathbb{R}}|\sigma ''(z)|\le\infty$, and that $g\in \mathcal{B}_R^1(\Omega)$. Then for any open bounded subset $\Omega_0\subset \Omega$ and any $n\in \mathbb{N}_{+}$, there exists $\{(a_k,w_k,b_k)\}_{k=1}^{n}$ satisfying 
		\begin{equation}
			\bigg\Vert \frac{1}{n}\sum_{k=1}^{n}a_k\sigma(w_k^T x+b_k)-f\bigg\Vert_{H^{1}(\Omega_0)}\le \dfrac{2(C_0^2+R^2C_1^2)m(\Omega_0)\Vert g\Vert_{\mathcal{B}^p_R(\Omega)}}{n} \ ,
		\end{equation}
		where $m(\Omega_0)$ is the Lebesgue measure of of $\Omega_0$.
	\end{theorem}
	
	The above theorem gives an $H^1$-approximation rate for the Barron space defined by its integral representation. Furthermore, for a family of second-order elliptic PDEs with some assumptions on the coefficients and the source term, it is proved in \cite{chen2021representation} that for any exact solution $u$ and $\varepsilon\in(0,1/2)$, there exists $u^{*} \in \mathcal{B}^{1}_R(\mathbb{R}^d)$ with $R\le c_1(\frac{1}{\varepsilon})^{c_2}$ and $\Vert u^{*}\Vert_{\mathcal{B}^{1}_R(\mathbb{R}^d)}\le c_3(\frac{d}{\varepsilon})^{c_4|\ln\varepsilon|}$ so that $\Vert u- u^{*}\Vert_{H^{1}(\mathbb{R}^d)}\le \varepsilon$. Thus, it is easy to conclude that there is a two-layer neural network $u_m(x; \Theta)$ with $m\le c_5(\frac{d}{\varepsilon})^{c_6|\ln \varepsilon|}$ such that $
	\Vert u_m-u\Vert\le \varepsilon$, where $c_1,c_2,c_3,c_4$ are constants independent of $u^{*},\varepsilon$ and $d$. Therefore, these results prove that even a simple two-layer neural network with a single activation function has sufficient representational power to approximate the solution of an elliptic PDE without CoD. 
	
	In \cite{chen2023regularity}, the regularity theory of solutions to the static Schr\"odinger equation in spectral Barron spaces was studied. In \cite{ma2022barron} the flow-induced function space was introduced and analysed by considering the residual neural networks. And the analysis in \cite{jentzen2018proof} shows that the solutions of some linear parabolic PDEs belong to a close analogue of the flow-induced spaces.
	
	\subsubsection{The Monte Carlo Approach}
	
	The approximation power of neural networks is the basic requirement for escaping CoD in high dimensional PDEs, while the algorithms for finding the optimal parameters of the neural networks are also equivalently important when considering computational complexity. The approximation of high dimensional integrals plays an important role in such high dimensional problems. Let $f:D=[0,1]^d \rightarrow \mathbb{R}$ be a function in $L^2(D) $and let 
	\begin{equation}
		I(f)=\int_{D} f(x)dx.
		\label{intMCA}
	\end{equation}
	To approximate the integral (\ref{intMCA}), the Monte Carlo algorithm \cite{DL_jentzen} randomly samples independent, continuous, uniformly distributed points $\{x_i\}_{i=1}^{N}$ from $D$ and let
	\begin{equation}
		\mathcal{I}_N(f)=\frac{1}{N}\sum_{i}^{N}f(x_i).
	\end{equation} 
	Then by a simple calculation, we have
	\begin{equation}
		\mathbb{E}[|I(f)-\mathcal{I}(g)|^2]=\frac{\text{Var}(f)}{N} \quad \text{and}\quad \text{Var}(f)=\int_{X}|f(x)|^2dx-\bigg[\int_{x}f(x)dx\bigg]^2. 
	\end{equation}
	Thus, the Monte Carlo algorithm, which approximates the integral in high-dimensional spaces, can escape the CoD in a probabilistic sense. The convergence rate of the Monte Carlo algorithm plays an important role in the theory of machine learning for high-dimensional PDEs. As we can see, the deep Ritz and WAN loss functions discussed later for PDEs all involve the evaluation of an integral over the domain of interest, and the PINN loss function can also be viewed as the integral of the squared residual of the PDE over its domain of definition.  In addition, the paper \cite{grohs2022deep} proved that if a function can be approximated by a proper discrete approximation algorithm without CoD, and if there are neural networks that satisfy certain regularity properties and approximate this discrete approximation algorithm without CoD, then the function itself can be approximated by neural networks without suffering the CoD. Full history recursive multilevel Picard approximation methods(MLP) \cite{hutzenthaler2019multilevel,DL_jentzen} are some recursive nonlinear extensions of the classical Monte-Carlo approximation methods, and it has been shown that such approximation schemes also escape the CoD in the numerical computation of semilinear PDEs with general time horizons. 
	
	In summary there exist a growing number of strict mathematical results which prove that deep neural networks have the expressive capacity to approximate the solutions of some specific PDEs without the CoD and that the Monte Carlo algorithm can provide an efficient method for computing the associated loss functions. Nevertheless, the conjecture that there is a deep learning-based approximation method that overcomes the CoD in terms of computational complexity in the numerical approximation of PDEs has not yet been fully analysed, for example, the convergence rate of the optimisation procedure to learn the solution is also required to overcome the CoD. 
	
	\subsection{Neural Network Approximations for Parametric PDEs}

	Operator learning methods aim to approximate the parametric map connecting a parameter space with a solution state space (parameter-to-state map). In the literature, the proposed operator learning methods usually consider low-dimensional PDEs. However, the inputs and outputs of neural networks for operator learning are usually high dimensional vectors, which requires the neural networks to have sufficient expressiveness to approximate the parametric map. The first successful use of neural networks in the context of operator learning appeared in \cite{chen1995universal}, where the authors designed a novel learning architecture based on neural networks and proved that these neural networks yield a surprising universal approximation property for infinite-dimensional nonlinear operators. This was later extended to deeper networks, see \cite{pinkus1999approximation}.
	
	\subsubsection{A Theoretical Analysis of Neural Networks for  Parametric PDEs with Reduced Basis Assumption}
	
	In this subsection, we will briefly review a theoretical result \cite{kutyniok2022theoretical} for parametric PDEs, which is based on the assumption of an inherent low dimensionality of the solution manifold. We will present a brief overview of the arguments that lead to the approximation theorem obtained in \cite{kutyniok2022theoretical}. The lemmas in this paper and the arguments used for proving them are constructive and have important value in the analysis of neural networks.
	
	In \cite{kutyniok2022theoretical}, the authors consider parameter-dependent equations in their variational form:
	\begin{equation}
		\mathcal{A}_\lambda(u_\lambda, v)=f_\lambda(v),\quad \text{for all} \quad \lambda \in \mathcal{Y}, v\in \mathcal{H}, 
	\end{equation}
	where the parameter  set $\mathcal{Y}$ is a compact subset of $\mathbb{R}^p$ with a fixed and potentially large $p$. $\mathcal{H}$ is a Hilbert space and $\mathcal{A}_\lambda$ is a symmetric, uniformly continuous and coercive bilinear form, $f_\lambda\in \mathcal{H}^{*}$ is bounded by a constant $C$ for all $\lambda\in \mathcal{Y}$, $u_\lambda\in \mathcal{H}$ is the solution and the solution manifold is assumed to be compact in $\mathcal{A}$.
	
	The following is a simplified outline of the arguments of \cite{kutyniok2022theoretical}, which derived upper bounds on the size of neural networks with activation ReLU approximating the solution operator of parametric partial differential equations under the assumption that the solution space is inherently lying in a low-dimensional space.  
	\begin{itemize}
		\item [1.] Firstly, it recalled the result shown in \cite{yarotsky2017error} that the scalar multiplication $f(x,y)=xy$ for $x,y\in [0,1]$ can be constructed by a ReLU NN of size $\mathcal{O}(\log_2(1/\varepsilon))$ up to an error of $\varepsilon>0$.
		\item [2.] As a next step, the approximate scalar multiplication is used to show that a matrix multiplication of two matrices of size $d\times d$ with entries bounded by 1 can be performed by NN of size $\mathcal{O}(d^3\log_2(1/\varepsilon))$ up to an error of $\varepsilon>0$.
		\item [3.] For $A\in \mathbb{R}^{d\times d}$ such that $\Vert A\Vert_2\le 1-\delta$ for some $\delta\in (0,1)$, the map $A\rightarrow \sum_{s=0}^{n}A^s$ can be approximated by a ReLU NN with an accuracy of $\varepsilon>0$ and having a size of $\mathcal{O}(n\log_2^2(m)d^3\cdot(\log_2(1/\varepsilon)+\log_2(n)))$. Furthermore, with the fact that the Neumann series $\sum_{s=0}^{n}A^s$ converges exponentially fast to $(\text{Id}-A)^{-1}$, a ReLU NN can be constructed that approximates the inversion operator $B\rightarrow B^{-1}$ to accuracy $\varepsilon >0$ under suitable conditions on the matrix $B$. This NN has the size $\mathcal{O}(d^3\log_2^q(1/\varepsilon))$ for a constant $q>0$.
		\item [4.] Next, two assumptions are required, which are satisfied in many applications. The first is that the map from the parameter space to the associated stiffness matrices of the Galerkin discretisation of the parametric PDE with respect to a reduced basis can be well approximated by neural networkss. The second is that the map from the parameters to the right-hand side of the variational form of the parametric PDEs discretised with based on the reduced basis can be effectively approximated by neural networks. Then there exists a neural network that approximates the operator from parameters to the corresponding discretised solution with respect to the reduced basis. If the reduced basis is of size $d$ and the implementations of the map obtaining the stiffness matrix and the right-hand side are adequately efficient, then the corresponding NN is of size $\mathcal{O}(d^3\log_2^q(1/\varepsilon))$. Finally, if $D$ is the size of the high-fidelity basis, then one can approximate a base change by applying a linear map $\mathbf{V}\in\mathbb{R}^{D\times d}$ to a vector with respect to the reduced basis. This procedure increases the size of the NN to $\mathcal{O}(d^3\log_2^q(1/\varepsilon) + dD)$.
	\end{itemize}
	\label{ATANN}

	\subsubsection{Approximation Results for Operator Learning Methods with State-to-the-Art Architectures}
	In contrast to neural networks aiming at function approximations, several operator learning methods for PDEs aiming at learning the full parameter-to-state map have been proposed in recent years. These operator learning methods have their own specific and somewhat complex architectures. For example, the networks of Fourier neural operators are mainly defined in Fourier space or the DeepONets consists of a branch network that approximates the map and a trunk network that approximates the solution basis. Thus, the network approximation result in (\ref{ATANN}) cannot be directly applied to these state-of-the-art operator learning methods. However, there are also some approximation results available for some operator learning methods.

	In \cite{kovachki2021universal} it is shown that in the worst case, the network size of the Fourier neural operator can grow exponentially in terms of accuracy when approximating general operators. However, the author also proved that, under suitable hypotheses, the size of the network in FNOs for approximating the parametric map for a Darcy-type elliptic equation or for the incompressible Navier-Stokes equations of fluid dynamics scales only polynomially in the error bound.
	
	For PCANN, under some assumptions on the probability measure of the parameter and solution spaces, in \cite{bhattacharya2020model} it is proven that for any error level $\varepsilon>0$ there exist dimensions for the reduced basis of the parameter and solution spaces, and a network with maximum layers and widths, such that the neural network can approximate the operator up to a certain error $\varepsilon$ associated with the probability distribution.
	
	In DeepONet \cite{lu2019deeponet}, the basis functions of the solution space are represented by the trunk net. Thus, not only is the branch net required to have approximation capability to learn the map connecting the parameter functions and the coefficients of the solutions with respect to the basis of the trunk net, but also the trunk net should efficiently approximate the basis of the solutions. A first answer to this question lies in a remarkable universal approximation theorem for the operator network first proven in \cite{chen1995universal}. More refined results are given in \cite{lu2019deeponet, lanthaler2022error}. In particular, in \cite{lanthaler2022error} the authors extended the universal approximation theorem from continuous to measurable functions, while removing the compactness requirements. Upper and lower bounds on the DeepONet error are given in \cite{lanthaler2022error} with respect to the number of sensors, the number of branch and trunk networks $p$, and the sizes and depths of the networks.
	
	Some approximation results for operator learning with convolutional neural networks have also been derived. In the paper \cite{franco2023approximation}, the authors have established and verified theoretical error bounds for the approximation of nonlinear operators using convolutional neural networks. The results shed light on the role of convolutional layers and their hyperparameters, such as input and output channels, depth and others.

	\section{DL concepts based on function approximation}
	In this section, we list the most common DL concepts based on function approximation for PDE forward solvers and their related inverse problems.
	\subsection{Deep Ritz Method}
	\subsubsection{Motivation}
	The Deep Ritz method \cite{yu2017deep} is a function evaluation concept based on a combination of the classical Ritz method (variational method) and deep learning. The Deep Ritz method assumes that a variational formulation of a PDE as stated in Equation  (\ref{basicPDE}) exists, i.e. there exists a $\Psi: \R \rightarrow \R$ such that the unique solution $u$ of the PDE is given by
	\begin{equation}
		u =	\argmin_{v\in H}I(v)
	\end{equation}
	where
	\begin{equation}
		I(u)=\int_{\Omega}\Psi(u(x))dx
		\label{loss}
	\end{equation}
	and $H$ is the set of admissible functions.
	Hence, the Deep Ritz method is a physics-informed method.
	
	The idea of the Deep Ritz method is to replace $u$ with a neural network  $u_\Theta$ which has a scalar or vectorial $x$ as input. The network is trained by choosing suitable collocation points $\{x_i\}_{i=1}^{N}\subset \Omega$ and by replacing the integral in the variational formulation by a finite sum%, i.e. $\Theta$ is obtained by
	\begin{equation}
		\min_\Theta \ \sum\limits_{i =1}^N \ \Psi(u_\Theta(x_i))  \ .
	\end{equation}
	After training, $u_\Theta$ can be evaluated efficiently at arbitrary points in the domain of definition.
	
	\subsubsection{Network architecture}
\begin{figure}[ht]
		\centering
		\includegraphics[width=0.9\linewidth]{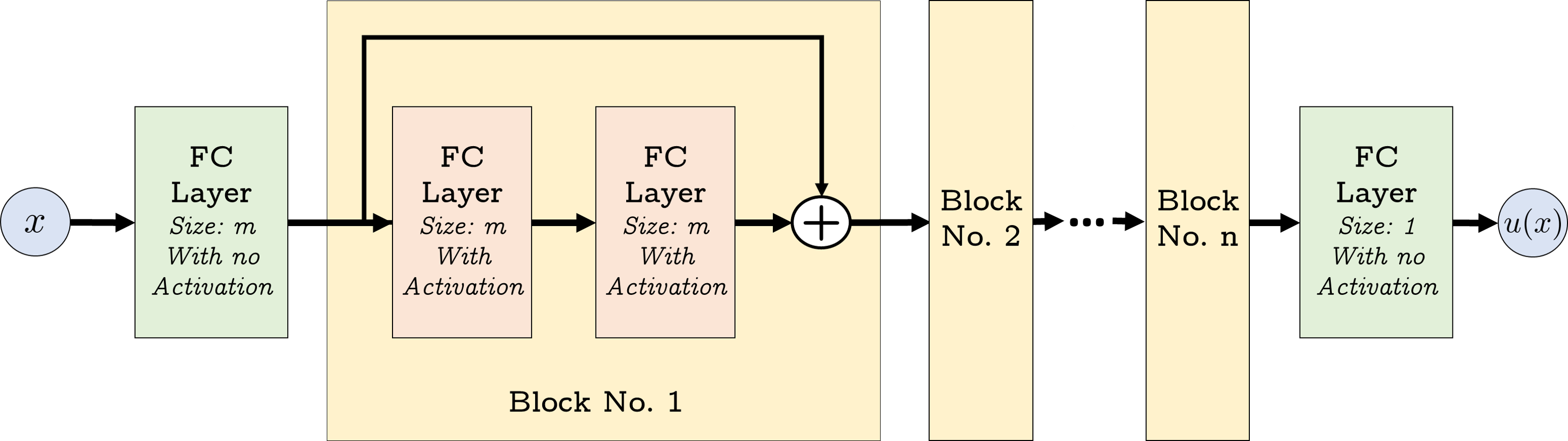}
		\caption{Deep Ritz network with $n$ blocks and with appropriate input and output FC linear layers. The first FC linear layer can be replaced by zero padding} \label{fig:deepritz_archi}
	\end{figure}

	For the numerical examples for our standard test problems, see Section \ref{sec:Numerical experiments}, we used the network shown in Figure \ref{fig:deepritz_archi}.
	The main building block of the Deep Ritz neural network consists of two stacked fully connected (FC) layers which are each followed by a non-linear somewhat "smooth"  activation function. A residual connection, which can help to avoid the vanishing gradient, links the input of the first FC layer to the output of the second FC layer. Several blocks are then stacked to complete the architecture of the network. 
	
	Consider the $i$-th block of the neural net. Let $W_{i, 1}, W_{i, 2} \in \mathbb{R}^{m \times m}$ and $b_{i, 1}, b_{i, 2} \in \mathbb{R}^m$ be the respective weight and bias of the first and second FC layers. A block which receives an input $s$, performs the operation
	\begin{eqnarray}
		B_{i}(s)=\sigma \left(W_{i, 2} \cdot \sigma \left(W_{i, 1} s+b_{i, 1}\right)+b_{i, 2}\right)+s, 
	\end{eqnarray}
	where $\sigma$ is the activation function. From experience, 
	\begin{eqnarray}
		\sigma(x)=\max \left\{x^{3}, 0\right\},
	\end{eqnarray} provides a good balance between accuracy and simplicity. Its smoothness contributes to the accuracy of the Deep Ritz method. 
	To match the input, resp. output, dimensions of the considered problem, an appropriate FC layer is used at the entrance (resp. exit) of the first (resp. last) layer of the first (resp. last) block of the neural network. Figure \ref{fig:deepritz_archi} shows the architecture of the Deep Ritz network.
	
	\subsubsection{The algorithm}\label{ch:DeepRitz_Algo}
	The training of the network does not need any information or data of the solution $u$, hence, with this respect, it is equivalent to classical numerical PDE solvers. It, however, needs the PDE in its variational formulation, hence, it is a physics-informed concept.
	The network architecture will be of the type `function approximation', i.e. a rather small network with two or three nodes in the input level and a single value for output. This can be adjusted for higher-dimensional PDEs accordingly.
	The training of the network uses a loss function, which is simply a discretisation of the variational formulation. A second term is added for ensuring the boundary conditions. Hence, first of all, it requires sampling a suitable set of evaluation points $x_i, i=1,.., N$ in the domain of definition and on the boundary. This can be done via stochastic sampling of a uniform distribution over the domain of definition (Monte Carlo integration) on a fixed grid. The training itself is then done by classical stochastic gradient descent. The sampling points may be changed during training after a fixed number of epochs.
	We specify the approach for the classical Poisson problem
	\begin{eqnarray}
		\begin{aligned}
			\Delta u &=\lambda \text { on } \Omega \subset \mathbb{R}^{2} \\
			u &=g \text { on } \partial \Omega
		\end{aligned}
		\label{eqn:poisson0}
	\end{eqnarray}

	This has the variational form: $ I(u)=\int_{\Omega}\left(\frac{1}{2}|\nabla u(x)|^{2}-\lambda (x) u(x)\right) d x,$  see  Algorithm \ref{algo:DRM}.	
	
	\begin{algorithm}[htp]
		\caption{Deep Ritz Method}
		\label{algo:DRM}
		\KwIn{\\
			\qquad$\bullet$ $N_\mathcal{L}/N_\mathcal{B}$: number of inner/boundary collocation points\\
			\qquad$\bullet$ $\tau$: learning rate
		}
		Initialise the network architecture $u_{\Theta}$\\
		\While{\textnormal{not converged}}{
			Sample collocation points $\{x_\mathcal{L}^{i}\in \Omega:i = 1,\dots,N_\mathcal{L}]\}$ and  $\{x_\mathcal{B}^{j}\in \partial\Omega:j=1,\dots, N_\mathcal{B}\}$\\
			
			Compute $E_\mathcal{L}(u_\Theta) = \frac{1}{N_{\mathcal{L}}}\sum_{i=1}^{N_{\mathcal{L}}}\left(\frac{1}{2}|\nabla u_\Theta (x_\mathcal{L}^{i})|^{2}-\lambda (x_\mathcal{L}^{i}) u_\Theta(x_\mathcal{L}^{i})\right)  $, via backpropagation\\
			
			Compute $E_\mathcal{B}(u_\Theta) = \frac{1}{N_{\mathcal{B}}}\sum_{j=1}^{N_{\mathcal{B}}}\left(u_\Theta(x_\mathcal{B}^{j})-g(x_\mathcal{B}^{j})\right)^{2}$ \\
			
			Add loss values $L = E_\mathcal{L}(u_\Theta) + E_\mathcal{B}(u_\Theta)$ \\ 
			
			Optimise Loss, L using the appropriate optimisation algorithm.
			
			Update $\Theta \leftarrow \Theta - \tau \nabla_\Theta L$ 
		} 
	\end{algorithm}

	It has been numerically shown that the Deep Ritz has the power to solve high-dimensional problems, this is mainly due to the use of the Monte Carlo algorithm for approximating the integral. In addition, Deep Ritz is potentially a naturally adaptive algorithm that can solve problems with corner singularities. One drawback of the Deep Ritz method is that the neural network parameterises the solution function instead of the parameter-solution operator. Hence, the Deep Ritz method is less suited for parametric studies, since it requires a new training of the full network for every new parameter. Note that the loss function of the Deep Ritz method can be negative and the minimal value is usually unknown, this also causes some challenges in the training process. In addition, the minimisation problem that results from Deep Ritz is usually not convex even when the original problem is. The treatment of the essential boundary condition is not as simple as for traditional
	methods.

	\subsubsection{Theoretical Background}
	The convergence properties of the Deep Ritz methods have been analysed intensively over the last years  \cite{DeepRitz_analysis_2022, muller2019deep, duan2021convergence, jiao2021error}. The most far-reaching results - to the best of our knowledge - are described in \cite{DeepRitz_analysis_2022}. This paper also contains a nicely written survey on the state of the art of other results concerning convergence properties. This paper use techniques form $\Gamma$-convergence for proving, that under rather mild assumptions on the network architectures, the loss function of the network training $\Gamma$-converges to the true variational formulation and also the minimisers converge weakly to the true solution of the PDE, see Theorem 7 \cite{DeepRitz_analysis_2022}.
	%\subsection{Profile of the method}
	%physics informed, function approximation, no data, meshfree.
	
	\subsection{Physics-informed Neural Network (PINN)}\label{ch:PINNs}
	\subsubsection{Motivation}
\begin{figure}[ht]
		\centering
		\includegraphics[width=0.90\linewidth]{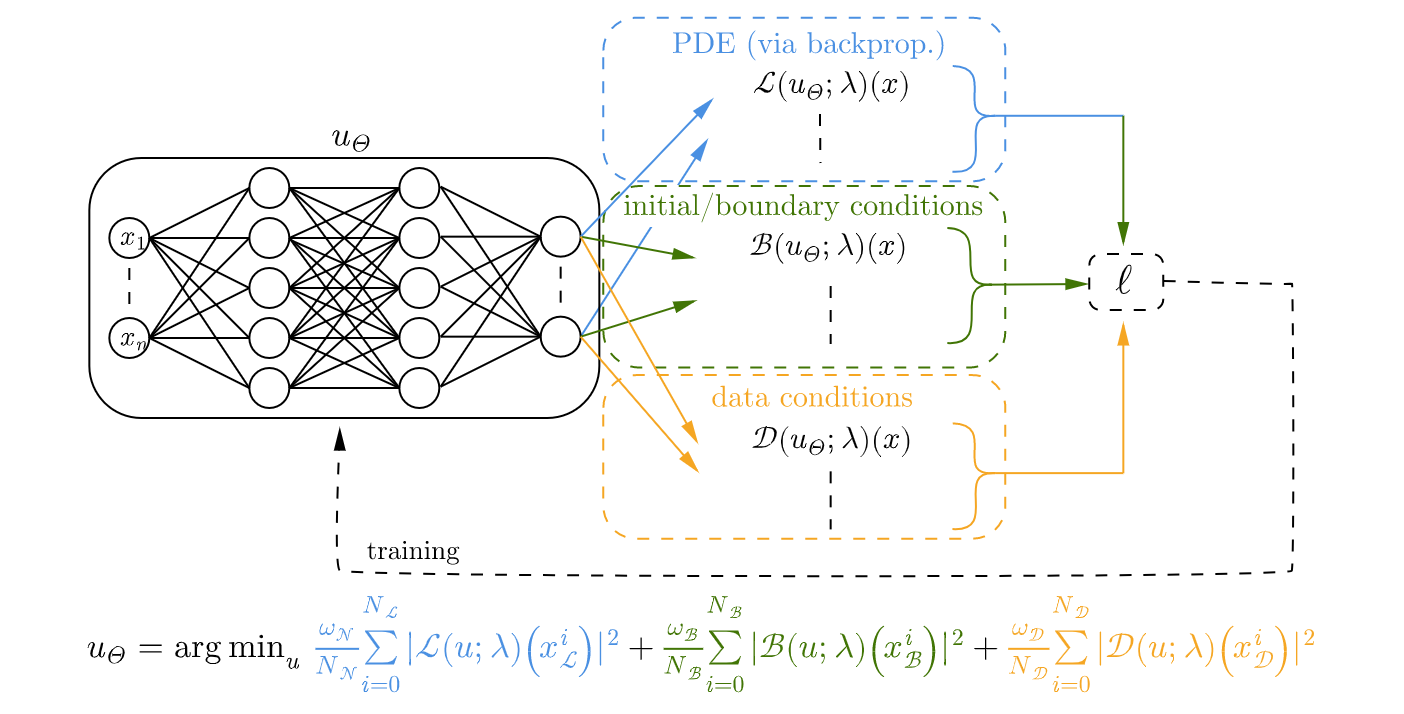}
		\caption{General training procedure of the PINN approach. As compared with the conditions in (\ref{eq:pinn_MSE}), in this figure we included a data condition that refers to observations of the solution on some points.}
		\label{fig:pinn_procedure}
	\end{figure}

	The notion of 'physics-informed neural networks' is now used in more general terms. In this section, however, we discuss the baseline version of a PINN as introduced in the original paper \cite{raissi2019physics}.
	Similar to the previously described Deep Ritz method, this original PINN concept parameterises the solution function as a neural network and requires knowledge of the underlying  In contrast to the Deep Ritz method, PINNs use the strong form of the PDE, thus their application to general PDEs is straightforward. To describe the method, we split up the operator $\mathcal{N}$ in equation (\ref{basicPDE}) into a differential operator $\mathcal{L}$ encoding the PDE and the initial/boundary operator $\mathcal{B}$. The problem then reads
	\begin{equation}
		\begin{split}
			&\mathcal{L}(u; \lambda)=0 \quad \text{in}\quad\Omega\\
			&\mathcal{B}(u; \lambda)=0   \quad \text{on} \quad \partial\Omega.
		\end{split}
	\end{equation}
	The neural network $u_\Theta$ is again substituted into the PDE, where the differential operator $\mathcal{L}$ can be applied to $u_\Theta$ via automatic differentiation (backpropagation).
	We note that $\mathcal{L}(u_\Theta; \cdot)$ and $\mathcal{B}(u_\Theta; \cdot)$ rely on the same set of parameters $\Theta$ as $u_\Theta$, which is crucial for the optimisation of PINNs. To fit $u_\Theta$ to the PDE, we penalise the residuals of both operators by their mean squared error on previously sampled points. This leads to the training objective
	\begin{equation}
		\min_\Theta \left(MSE_{\mathcal{L}}(u_\Theta) + MSE_{\mathcal{B}}(u_\Theta) \right), 
	\end{equation}
	where
	\begin{equation}\label{eq:pinn_MSE}
		MSE_{\mathcal{L}}(u_\Theta) =\frac{1}{N_{\mathcal{L}}}\sum_{i=1}^{N_{\mathcal{L}}}|\mathcal{L}(u_\Theta; \lambda)(x_{\mathcal{L}}^{i})|^{2} \quad \text{and} \quad
		MSE_{\mathcal{B}}(u_\Theta) =\frac{1}{N_{\mathcal{B}}}\sum_{j=1}^{N_{\mathcal{B}}}|\mathcal{B}(u_\Theta ; \lambda)(x_{\mathcal{B}}^{j})|^{2}.
	\end{equation}
	During optimisation, $MSE_{\mathcal{B}}$ enforces the initial or boundary conditions of a given problem and $MSE_{\mathcal{L}}$ checks the differential equation at the collocation points. A visual representation of the general idea is shown in Figure \ref{fig:pinn_procedure}.
	\subsubsection{Network architecture}
	The architecture of the neural network $u_\Theta$ allows a lot of freedom, the original paper \cite{raissi2019physics} uses a simple fully-connected architecture. Depending on the underlying differential equation, different network structures can lead to better approximation results. This promotes the existence of a great variety of extensions of the PINN approach, either focused on individual problems or designed for broader classes of equations. In Section \ref{ch:PINN_extensions} we will mention a few of these methods.
	
	\subsubsection{The algorithm}\label{Ch:PINN_Algo}
	The general training procedure is comparable to the Deep Ritz algorithm \ref{ch:DeepRitz_Algo}. Again, the network itself is optimised to approximate the solution as a function, therefore the number of input nodes corresponds to the dimensionality of $\Omega$ and the amount of output nodes denotes the dimensionality of the possibly vector-valued solution. The training requires the underlying PDE and some evaluation points $x_{\mathcal{L}}^{i}, x_{\mathcal{B}}^{j}$, $i=1,\dots, N_\mathcal{L}, $ ${j=1,\dots, N_\mathcal{B}}$ to compute the residuals. The sampling strategy and therefore the distribution of these points can be chosen freely and may be changed throughout the training.
	Summarising the approach for the Poisson problem of Equation \ref{eqn:poisson0}, the general algorithm is shown in \ref{alg:AlgorithmPINN}.
	
	%\begin{itemize}
	%\item Sample $x_{\mathcal{L}}^{i} \in \Omega $, resp. $x_{\mathcal{B}}^{i} \in \partial \Omega $ 
	%\item Network $u_\Theta (x) \sim u(x)$, compute $\Delta u_\Theta(x)$ via backprop.
	%\item  Minimise, via gradient descent, the loss function
	%$$ \frac{1}{N_{\mathcal{B}}}\sum_{j=1}^{N_{\mathcal{B}}}|u_\Theta(x_{\mathcal{B}}^{j}) - g(x_{\mathcal{B}}^{i}))|^{2} + \frac{1}{N_{\mathcal{L}}}\sum_{i=1}^{N_{\mathcal{L}}}|\Delta u_\Theta(x_{\mathcal{L}}^{i}) - \lambda(x_{\mathcal{L}}^{i}))|^{2}$$  
	%\end{itemize}
	\begin{algorithm}[ht]
		\caption{Physics-Informed Neural Networks}
		\label{alg:AlgorithmPINN}
		\KwIn{$N_\mathcal{L}/N_\mathcal{B}$: number of inner/boundary collocation points, \\
			$\tau$: learning rate }
		Initialise the network architecture $u_{\Theta}$\\
		\While{\textnormal{not converged}}
		{Sample collocation points $\{x_\mathcal{L}^{i}\in \Omega:i = 1,\dots,N_\mathcal{L}]\}$ and  $\{x_\mathcal{B}^{j}\in \partial\Omega:j=1,\dots, N_\mathcal{B}\}$\\
			Compute $MSE_\mathcal{L}(u_\Theta) = \frac{1}{N_{\mathcal{L}}}\sum_{i=1}^{N_{\mathcal{L}}}|\Delta u_\Theta(x_{\mathcal{L}}^{i}) - \lambda(x_{\mathcal{L}}^{i}))|^{2}$, via backpropagation\\
			Compute $MSE_\mathcal{B}(u_\Theta) = \frac{1}{N_{\mathcal{B}}}\sum_{j=1}^{N_{\mathcal{B}}}|u_\Theta(x_{\mathcal{B}}^{j}) - g(x_{\mathcal{B}}^{i}))|^{2}$ \\
			Add loss values $L = MSE_\mathcal{L}(u_\Theta) + MSE_\mathcal{B}(u_\Theta)$ \\ 
			Update $\Theta \leftarrow \Theta - \tau \nabla_\Theta L$}
	\end{algorithm}
	The generality of PINN allows the extension of the loss function to include several additional conditions. For example, if some observed data is available, it can be seamlessly incorporated into the loss function.  In addition, PINN can be extended to solve integro-differential equations \cite{lu2021deepxde}, fractional PDEs \cite{pang2019fpinns} and stochastic PDEs \cite{zhang2019quantifying}. Furthermore, by using a discrete-time model (Runge-Kutta methods) to link two distinct temporal snapshots, the number of collocation points for time-dependent PDEs can be reduced significantly. 
	
	\subsubsection{Application to parameter identification} \label{pinns-param-iden}
	Notably, PINNs are also suitable for the parameter identification of PDEs. Given a set of observed data points, the unknown parameters of the PDE can be optimised alongside the parameters of the solution network $u_\Theta$. From an implementation point of view, PINNs can therefore be applied to parameter identification problems requiring minimal additional work, see \cite{raissi2019physics, chen2020physics, raissi2018deep, lu2021physics, jagtap2020conservative, yu2022gradient}.
	
	While the identification/learning of scalar parameters is quite straightforward, that of function-valued (e.g. space-dependent and/or time-dependent) parameters requires some additional effort. For function-valued parameters such as the right side, $\lambda$  in equation \ref{eqn:poisson0} or the coefficient $\lambda$ in \ref{eqn:darcy} of a certain PDE, one usually introduces an additional neural network $\lambda_{\Theta}$ for the searched functions. The idea behind these networks is comparable to $u_\Theta$ which parameterises the PDE solution as described in \ref{alg:AlgorithmPINN}, i.e. we input the space and/or time variables and the network should output an approximation of the searched function at these values. The training process then consists of a combination of physics-informed loss and data loss. Via the data loss, the network $u_\Theta$ learns an interpolation of the given (measured) data. A physics-informed loss, like the $MSE_{\mathcal{L}}$ term in \ref{eq:pinn_MSE}, then enables us to learn the searched functions, by using the learned interpolation $u_\Theta$. 
	During the minimisation of the sum of both loss terms, the corresponding weights of all appearing networks then are optimised simultaneously via automatic differentiation. The resulting loss function which is optimised, e.g. in the case of the Poisson problem \ref{eqn:poisson0} is thus
	\begin{eqnarray}
		L = \underbrace{\frac{1}{N_{\mathcal{P}}}\sum_{i=1}^{N_{\mathcal{P}}}|\Delta u_\Theta(x_{\mathcal{P}}^{i}) - \lambda_{\Theta}(x_{\mathcal{P}}^{i}))|^{2}}_\text{\normalsize Physics loss} + \underbrace{\frac{1}{N_{\mathcal{D}}}\sum_{j=1}^{N_{\mathcal{D}}}|u_\Theta(x_{\mathcal{D}}^{j}) - u(x_{\mathcal{D}}^{j}))|^{2}}_\text{\normalsize Data loss},
	\end{eqnarray}
	where $N_\mathcal{P}$ is the number of collocation points, and  $N_\mathcal{D}$ is the number of known data points, such that $\{x_\mathcal{P}^{i}\in \Omega:i = 1,\dots,N_\mathcal{P}]\}$ and  $\{x_\mathcal{D}^{j}\in \partial\Omega:j=1,\dots, N_\mathcal{D}\}$.
	
	In the case of noisy data, the physics can also act as a regularisation for the learned interpolation, since the loss also influences $u_\Theta$. Even when only discrete data points of the solution are available, the physics loss may be evaluated on arbitrary points, because of the interpolation properties of $u_\Theta$.
	Moreover, the flexible framework of PINNs allows for further inclusion of a-priori knowledge on the solution or parameter functions, such as their regime or boundary values, by similar incorporation into the loss.

	In the numerical experiments in Section \ref{sec:Numerical experiments}, we will use simple fully connected neural networks (FCNNs), consisting of multiple FC layers. As an activation function, we will use $\phi(x) = \tanh(x)$.  
	
	\subsubsection{Theoretical Background}
	In contrast to the large number of applications of PINNs, they still lack a rigorous theoretical background. First consistency results under additional assumptions for linear elliptic and parabolic PDEs, can be found in \cite{Shin_2020}. Some estimates on the generalisation error of PINNs approximating solutions for forward and inverse problems are proven in \cite{pinn_theo_forward,  pinn_theo_inverse}. The authors derive an error estimation in terms of training error and training points by exploiting the stability properties of the underlying equation. More general convergence results of the PINN approach still have to be found.
	
	Some studies of the convergence rate were carried out in \cite{pinn_ntk_fourier, pinn_ntk}. There the Neural Tangent Kernel (NTK) theory \cite{jacot2020ntk_theory} was applied to the PINN framework, which gave the first insight into the convergence behaviour of the different loss terms and spectral bias of PINNs. With the NTK theory special weights for the different terms can be found, to achieve more robust and accurate results. In \cite{ntk_deeponet}, the same approach was also applied to the DeepONet architecture, which will be introduced in Section \ref{sec:DeepONet}. 
	
	%\subsection{Profile of the Method}
	%The general properties of PINN and Deep Ritz are the same. They are both physics-informed methods, needing the underlying PDE, can only learn parameter dependencies, not the forward operator w.r.t functions and do not need previously know data or a domain discretisation. 
	
	%Some advantages of PINNs are the easy extension to additional conditions and inverse problems and the avoidance of the requirement to derive the variational formulation by hand. Additionally, the loss function is bounded from below by zero which improves the observability of the optimisation performance, especially for inverse problems.
	
	%On the other hand, PINNs require higher regularity of the PDE
	%For a PDE which has discontinuous parameters, the problem may be difficult to train and even fail. The Deep Ritz method only has to find the weak solution. In this regard, the PINN approach
	%and the computation of higher order derivatives, can %get expensive for large networks and complex problems. 
	
	\subsubsection{Generalisations and extensions of PINN}\label{ch:PINN_extensions}
\begin{figure}[ht]
		\centering
		\includegraphics[width=0.95\linewidth]{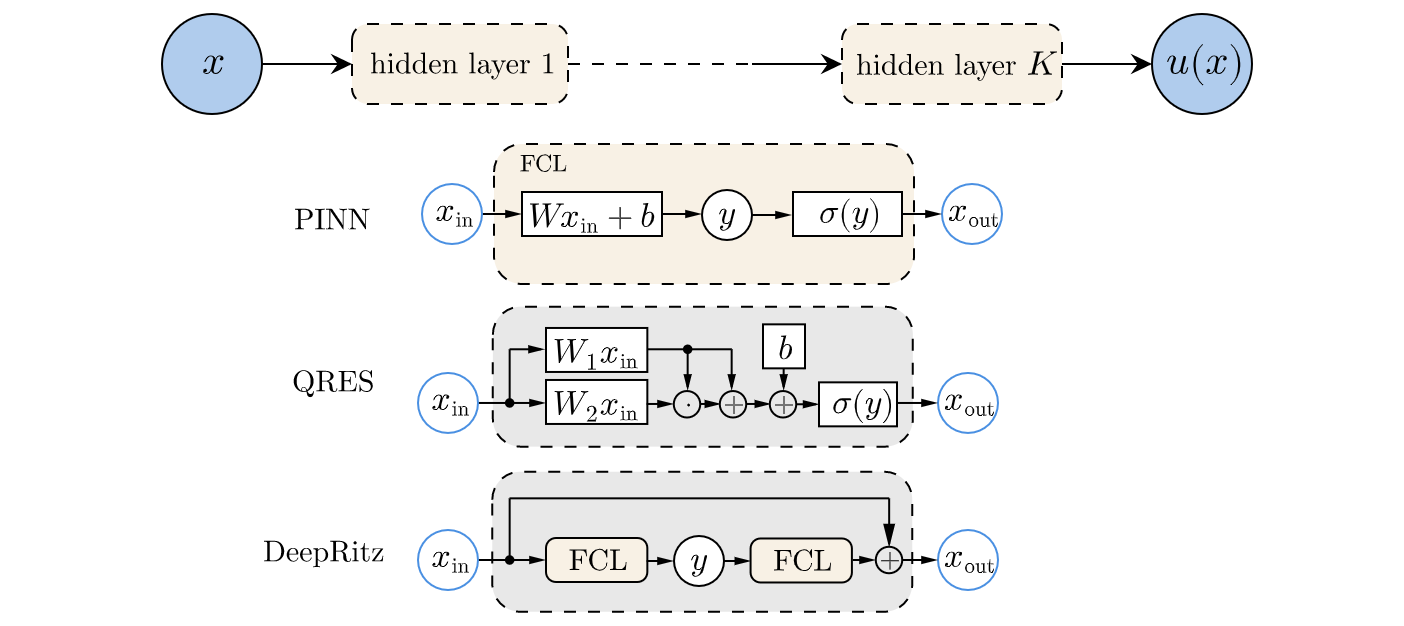}
		\caption{Overview and comparison of network architectures used in PINN, QRES and Deep Ritz. With weight matrix $W$, bias vector $b$ and activation function $\sigma$.}
		\label{fig:network_overview}
	\end{figure}

	Lastly, we want to give a short overview of possible extensions and generalisations for PINNs. Since the literature in this regard is vast, we can not represent every approach.
	\begin{itemize}
		
		\item \textbf{Quadratic Residual Networks (QRES)} \cite{bu2021quadratic}: Extends PINNs by including quadratic terms in each layer. The used architecture consists of FC layers with additional weight matrices, see Figure \ref{fig:network_overview}. In each Layer, The output of the FC layer matrix and additional matrix are point-wise multiplicated and added to the FC layer output, to create a quadratic term. This may lead to faster convergence and better parameter efficiency with respect to network width and depth. 
		
		\item \textbf{Residual-based adaptive refinement} \cite{lu2021deepxde}: The idea of RAR is to add more residual points in the locations where the PDE residual is large during the training process, conceptually similar to FEM refinement methods.
		\item \textbf{Gradient-enhanced PINN (gPINN)} \cite{yu2022gradient}: If the PDE residual $\mathcal{L}u$ is zero, then it is clear that the gradient of the PDE residual, i.e $\triangledown\mathcal{L}u$, should also be
		zero. Thus, the gradient-enhanced PINN (gPINN) uses a new type of loss function by leveraging the gradient information of the PDE residual
		to improve the accuracy and training efficiency of PINNs.
		
		\item \textbf{XPINN and cPINN} \cite{jagtap2020extended, jagtap2020conservative}: The XPINN and cPINN apply domain decomposition, and then different subdomains use independent neural networks to approximate the solutions. In addition to the loss function used in PINN, interface conditions are introduced to stitch the decomposed subdomains together in order to obtain a solution for the governing PDEs over the complete domain. These extensions efficiently lend themselves to paralleled computation and are able to solve more general PDEs, e.g. PDEs with jump parameter functions.
	\end{itemize}
	For further details on the PINN approach, we suggest the paper \cite{pinn_survey}, which contains a comprehensive and nicely written survey for the current state of the art of the PINNs. Mentioning different applications, extensions, theories, general challenges and currently open questions.
	
	\subsection{Weak Adversarial Networks}
	\subsubsection{Motivation}
	Similar to PINN and Deep Ritz, the method of weak adversarial networks \cite{zang2020weak} also parameterises the solution as a neural network and is a physics-informed method. However, the weak adversarial network is based on the weak formulation of the PDE and the loss function is defined as an operator norm to be minimised. For simplicity, consider the following weak formulation of the Poisson problem with either Dirichlet or Neumann boundary conditions on an arbitrary domain $\Omega\subset \mathbb{R}^{d}$:
	\begin{equation}
		\langle \mathcal{L}[u, \lambda],\varphi\rangle := a(u,\varphi;\lambda)= \int_{\Omega}(\nabla u \cdot\nabla\varphi - \lambda\varphi)dx = 0,\quad \text{for all } \varphi \text{ in } H_{0}^{1}(\Omega),		
	\end{equation}
	\begin{equation}
		\mathcal{B}(u,\lambda) = 0 \quad \text{on}\quad\partial\Omega.
	\end{equation}
	From the above formula, we see that for each $u\in H^{1}(\Omega)$, it defines a linear functional (operator) $\mathcal{L}[u,\lambda]$ such that  $\mathcal{L}[u,\lambda](\varphi)=a(u,\varphi;\lambda)$. A function $u\in H^{1}(\Omega)$ satisfying the boundary condition is the weak solution of the equation if and only if the operator norm:
	\begin{equation}
		\Vert \mathcal{L}[u,\lambda]\Vert_{op}:=\max\{\langle \mathcal{L}[u,\lambda],\varphi\rangle/\Vert \varphi\Vert_{2}\bigg| \varphi\in H_{0}^{1}(\Omega), \varphi\ne 0\}
	\end{equation} 
	is equal to $0$. Hence, the solution satisfies the boundary condition and solves the following minmax problem:
	\begin{equation}
		\min_{u}\max_{\varphi}\vert\langle \mathcal{L}[u,\lambda],\varphi\rangle\vert^{2}/\Vert \varphi\Vert_{2}^{2}\label{Minmax}.
	\end{equation}
	This weak formulation inspires an adversarial approach, where a network 
	$u_{\Theta}$ with parameter $\Theta$ is used to parameterise the solution of the PDE and another network $\varphi_{\eta}$ with parameter $\eta$ approximates the test function. 
	\subsubsection{Network architecture}
	As just stated, in weak adversarial networks, the solution is parameterised as a neural network $u_{\Theta}$ with parameter $\Theta$, which is learned by minimising the operator norm $\Vert \mathcal{L}[u_\Theta]\Vert_{op}$.  The test function, however, is parameterised by an adversarial network $\varphi_{\eta}$ with parameter $\eta$ and is trained by maximising $\langle \mathcal{L}[u_{\Theta}],\varphi_{\eta}\rangle/\Vert \varphi_{\eta}\Vert$, hence it approximates the operator norm $\Vert \mathcal{L}[u_\Theta,\lambda]\Vert_{op}$. Thus, it is an
	adversarial training in a way that the test function critics on the solution network where the PDE is violated,
	and the solution network corrects itself at those spots until the PDE system is satisfied (almost) everywhere in the
	domain. Thus the network version of (\ref{Minmax}) can be formulated as:
	\begin{equation}
		\min_{\Theta}\max_{\eta}\vert\langle \mathcal{L}[u_{\Theta},\lambda],\varphi_{\eta}\rangle\vert^{2}/\Vert \varphi_{\eta}\Vert_{2}^{2}.
		\label{minmaxNet}
	\end{equation}
	Both networks $u_{\Theta}$ and $\varphi_{\eta}$ can be general architectures, e.g., fully connected networks(FCN), residual neural networks (ResNet), convolutional neural networks(CNN), recurrent neural networks(RNN). However, the choice of neural networks can affect efficiency and accuracy.  For example, using a ResNet for parameterising the trial and test function usually has better performance than using a fully connected network.
	
	\begin{figure}[htp]
		\centering
		\includegraphics[width=0.8\linewidth]{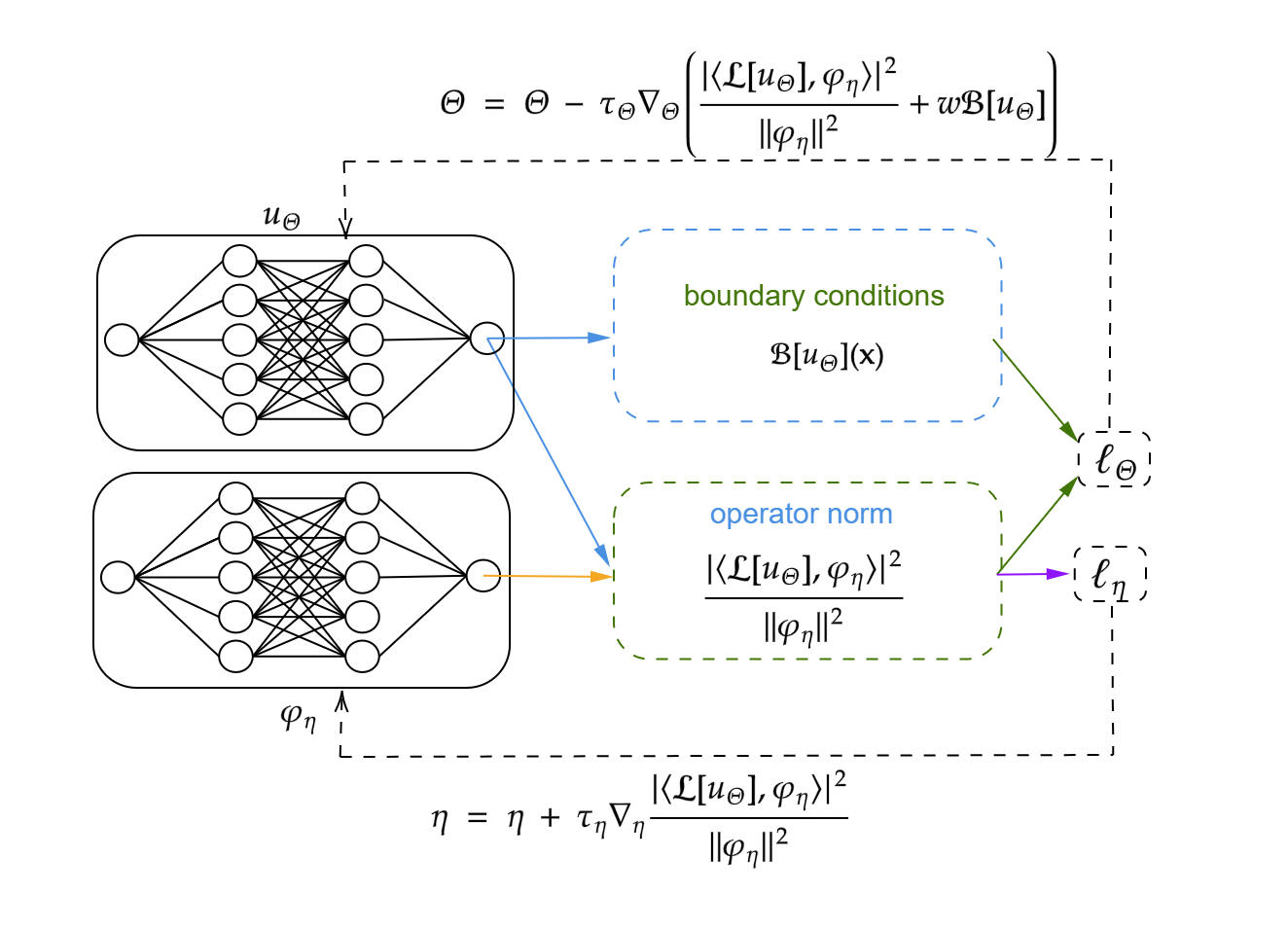}
		\caption{General training procedure of the WAN approach, where $u_{\Theta}$ and $\varphi_{\eta}$ denote the neural network's approximations for the solution and the test function, respectively. $\ell_\Theta$ and $\ell_\eta$ denote the loss functions for updating $u_{\Theta}$ and $\varphi_{\eta}$, respectively.}
		\label{fig:wan_archi}
	\end{figure}
	
	\subsubsection{The algorithm}
	The main challenge of the WAN method  is how to design an efficient algorithm for solving the minmax problem (\ref{minmaxNet}), 
	or equivalently for 
	\begin{equation}
		\min_{\Theta}\max_{\eta} L_{int}(\Theta,\eta)\quad \text{where} \quad L_{int}(\Theta,\eta) = \log\vert\langle \mathcal{L}[u_{\Theta},\lambda],\varphi_{\eta}\rangle\vert^{2}-\log\Vert \varphi_{\eta}\Vert_{2}^{2} .
	\end{equation}
	In addition, the weak solution $u_{\Theta}$ also needs to satisfy the boundary condition, which can be realised by minimising
	\begin{equation}
		L_{bdry}(\Theta)=\int_{\partial\Omega}\vert u_{\Theta}(x)-g(x)\vert^{2}dx.
	\end{equation}
	The total objective function is the weighted sum of the two objectives and  we seek to compute a saddle point that solves the minimax problem:
	\begin{equation}
		\min_{\Theta}\max_{\eta}L(\Theta,\eta)=	\min_{\Theta}\max_{\eta}L_{int}(\Theta,\eta) + \alpha L_{bdry}(\Theta).
	\end{equation}
	where $\alpha>0$ is a user-chosen balancing parameter. Typical algorithms use alternating updates, i.e. the objective function is alternatively minimised and maximised and the parameters $\Theta$ and $\eta$ are alternately updated such that the primal network can approximate the solution gradually.

	Given the objective function (8), the key components of the network training are the computation of $L(\Theta, \eta)$ and its gradients with respect to $\Theta$ and $\eta$, respectively.
	By denoting 
	\begin{equation}
		I(x,\Theta,\eta)=\nabla u_{\Theta}(x) \cdot\nabla\varphi_{\eta}(x) - \lambda(x)\varphi_{\eta}(x),
	\end{equation}
	we have 
	\begin{equation}
		\nabla_{\Theta}I(x,\Theta,\eta)=\nabla (\nabla_{\Theta}u_{\Theta}(x)) \cdot\nabla\varphi_{\eta}(x),
	\end{equation}
	\begin{equation}
		\nabla_{\eta}I(x,\Theta,\eta)=\nabla u_{\Theta}(x) \cdot\nabla(\nabla_{\eta}\varphi_{\eta}(x)) - \lambda(x)(\nabla_{\eta}\varphi_{\eta}(x)),
	\end{equation}
	and 
	\begin{equation}
		\nabla_{\Theta} L_{int}(\Theta) = 2\left(\int_{\Omega}I(x;\Theta,\eta)dx\right)^{-1}\left(\int_{\Omega}I(x;\Theta,\eta)dx\right) \ .
	\end{equation}
	Monte-Carlo (MC) approximations can be used to approximate the integrals, i.e., we randomly sample $N_{r}$ collocation points $\{x_{r}^{(j)}\}_{j=1}^{N_{r}}$ uniformly in the interior of the region and approximate the gradient $\nabla_{\Theta} L_{int}(\Theta) \approx 2\cdot(\sum_{j=1}^{N_{r}}I(x_{r}^{(j)};\Theta,\eta))^{-1}(\sum_{j=1}^{N_{r}}\nabla_{\Theta}I(x_{r}^{(j)};\Theta,\eta))$. The gradients $\nabla_{\eta}L_{int}$ and $\nabla_{\Theta}L_{bdry}$ can be approximated by the same procedure. Using these gradients of $\nabla_{\Theta}L$ and $\nabla_{\eta}L$, we can apply alternating updates to optimise the parameters $\Theta$ and $\eta$.  The resulting algorithm is then summarised in Algorithm \ref{AlgorithmWAN}.  Details on how to adapt weak adversarial networks for parabolic PDEs involving time can be found in the original paper. 
	\begin{algorithm}[t]
		\caption{Weak Adversarial Network (WAN) for Solving static PDEs.}
		\label{AlgorithmWAN}
		\KwIn{$N_{r}/N_{b}$:number of region/boundary collocation points;\\
			$K_{u}/K_{\varphi}$:number of solution/adversarial network parameter updates per iteration;\\
			$\tau_{\Theta}/\tau_{\eta}$:learning rates for $\Theta$ and $\eta$;\\
			$\alpha$: balancing weight}
		Initialise the network architectures $u_{\Theta}, \varphi_{\eta}$;\\
		\While{\textnormal{not converged}}
		{Sample collocation points $\{x_{r}^{j}\in \Omega:j\in[N_{r}]\}$ and  $\{x_{b}^{j}\in \partial\Omega:j\in[N_{b}]\}$\\
			Update weak solution parameter\\
			\For{$k=1,...,K_{u}$}{Update $\Theta \leftarrow \Theta-\tau_{\Theta}\nabla_{\Theta} L$ where $\nabla_{\Theta}L $ is approximated using $\{x_{r}^{j}\}$ and $\{x_{b}^{j}\}$.} 
			\For{$k=1,...,K_{\varphi}$}{Update $\eta \leftarrow \eta+\tau_{\eta}\nabla_{\eta} L$ where $\nabla_{\eta}L $ is approximated using $\{x_{r}^{j}\}.$}}
	\end{algorithm}
	\subsubsection{Weak Adversarial Network for inverse problems}%autorefnameection
	The weak adversarial network can be extended to solve inverse problems directly \cite{bao2020numerical}. An inverse problem defined on an open and bounded domain in $\mathbb{R}^{d}$ may be presented in a general form as 
	\begin{equation}
		\mathcal{A}[u,\lambda]=0, \quad \text{in}\quad \Omega,
	\end{equation}
	\begin{equation}
		\mathcal{B}[u,\lambda]=0, \quad \text{on}\quad \partial \Omega,
	\end{equation}
	where $u$ is the solution and $\lambda$ can be the coefficient in the inverse medium problem or the source function in the inverse source problem. $\mathcal{B}[u,\lambda]=0$ are some observations on the boundary. Suppose that the variational form is given by 
	\begin{equation}
		\langle \mathcal{L}[u,\lambda],\varphi\rangle:=\int_{\Omega}\mathcal{L}[u,\lambda](x)\varphi(x) dx \quad \text{for all } \varphi \text{ in } H_{0}^{1}(\Omega).
	\end{equation}
	Take the inverse conductivity problem as an example, we have
	\begin{equation}
		\langle \mathcal{L}[u,\lambda],\varphi\rangle=\int_{\Omega}(\lambda\nabla u \cdot\nabla\varphi - f\varphi)dx=0.
	\end{equation}
	By defining the norm of $\mathcal{A}[u,\lambda]$ as
	\begin{equation}
		\Vert\mathcal{L}[u,\lambda]\Vert_{op}^{2} := \text{sup}_{\varphi\in H_{0}^{1},\varphi\ne 0}\dfrac{\vert\langle  \mathcal{L}[u,\lambda],\varphi\rangle\vert^{2}}{\Vert\varphi\Vert^{2}}.
	\end{equation}
	Then, the weak formulation and the boundary condition induce a minmax problem:
	\begin{equation}
		\min_{u,\lambda}\max_{\varphi}\dfrac{\vert\langle \mathcal{L}[u,\lambda],\varphi\rangle\vert^{2}}{\Vert\varphi\Vert^{2}} + \int_{\partial\Omega}\vert\mathcal{B}[u,\lambda]\vert^{2}dx.
	\end{equation}
	By parameterising $u$,  $\lambda$, and $\varphi$ as neural networks $u_{\theta_{1}}$,$\lambda_{\theta_{2}}$ and  $\varphi_{\theta_{3}}$, respectively, we have the network version of the minmax problem:
	\begin{equation}
		\min_{\theta_{1},\theta_{2}}\max_{\theta_{3}}\dfrac{\vert \langle\mathcal{L}[u_{\theta_{1}},\lambda_{\theta_{2}}],\varphi_{\theta_{3}}\rangle\vert^{2}}{\Vert\varphi_{\theta_{3}}\Vert^{2}} + \alpha\int_{\partial\Omega}\vert\mathcal{B}[u_{\theta_{1}},\lambda_{\theta_{2}}]\vert^{2}dx
	\end{equation}
	The algorithm is then similar to that of the direct problems by updating $\theta = (\theta_{1}, \theta_{2})$ and $\eta=\theta_{3}$ alternately.
	
	\section{DL concepts based on operator approximation}
	In this section, we introduce the basic concepts for DL approaches which aim at approximating the parameter-to-state operator. By construction, these methods are perfectly suited for parametric studies and can be adapted directly to inverse problems. The two basic concepts for adaptation to inverse problems have already been described in Section \ref{subsec:Tikhonov_inverse}.

	\subsection{Model Reduction And Neural Networks For Parametric PDEs (PCANN)} \label{sec:pcann}
	
	\subsubsection{Motivation}
	PCANN, as we call it in this paper, seeks to provide a meshless operator for the evaluation of the solution of a PDE by combining ideas of model order reduction with deep learning.
	For given training data $(\lambda_i,u_i)$ one first obtains a model reduction by use of the principal component analysis (PCA) for both the input (parameter $\lambda$) and output (solution $u$) functions. Only the coefficients of a finite number of PCA components are retained. The PCA thus reduces the dimensions of both the input and output spaces to finite latent dimensional latent spaces.
	A neural network then maps the coefficients of the respective representations in these latent spaces as shown in Figure \ref{fig:pcann_general}.
	The evaluation of this operator approximation for a novel parameter $\lambda$ is then most efficient: one only needs to compute the scalar products with the specified finite number of PCA components, and the neural network then maps these coefficients to the latent coefficients of the output space and an expansion using these coefficients and the PCA on the output side gives a function approximation to the solution of the PDE.
	
	The formulation of this approach is in a function space setting and hence mesh-free.
	For implementation purposes, however, one has to specify how to compute the scalar products with the PCA components.
	These functions are only given numerically, usually by their values on a specified grid.  The red arrow in Figure \ref{fig:pcann_general} shows the flow during testing. 
	The overall PCANN can be used to evaluate the solution of the PDE for any given grid size as illustrated in Algorithm \ref{algo:PCANN}. In line \ref{PCA1}-\ref{PCA2} of Algorithm \ref{algo:PCANN}, it is possible to use one of the numerous PCA algorithms available \cite{wold1987principal}. For the implementations in this work, we make use of randomised PCA proposed in \cite{halko2011algorithm, halko2011finding}.
	
	In a situation where the input-to-output operator is linear, like in the case of the Poisson problem considered, it might be beneficial to use a simple linear map for the mapping of the reduced/latent dimensions. Section  \ref{sec:Numerical experiments}, equally shows results for this case, PCALin, where a single linear layer with no activation functions is used to map the latent dimensions. PCALin is similar to the linear method proposed in \cite{bhattacharya2020model}, only, we use a neural network instead of the normal equations. The results in both cases are similar as we later show in \ref{sec:Numerical experiments}. The combination of model order reduction with deep learning has recorded success in a number of application problems ranging from cardiac electrophysiology \cite{fresca2020deep, fresca2021comprehensive}, where the use of proper orthogonal decomposition (POD) further improves the results \cite{fresca2022pod}, to fluid flow \cite{fresca2021real} and non-linear models \cite{cicci2022deep, fresca2022deep}. Specifically, the PCANN operator has been used in \cite{liu2022learning, kovachki2022multiscale} in a multiscale plasticity problem to map strain to stress. Worth highlighting is an earlier work \cite{hesthaven2018non}, which equally combines model order reduction with neural networks to solve PDEs.

	\subsubsection{Network Architecture}
\begin{figure}[htp]
		\centering
		\includegraphics[width=0.6\linewidth]{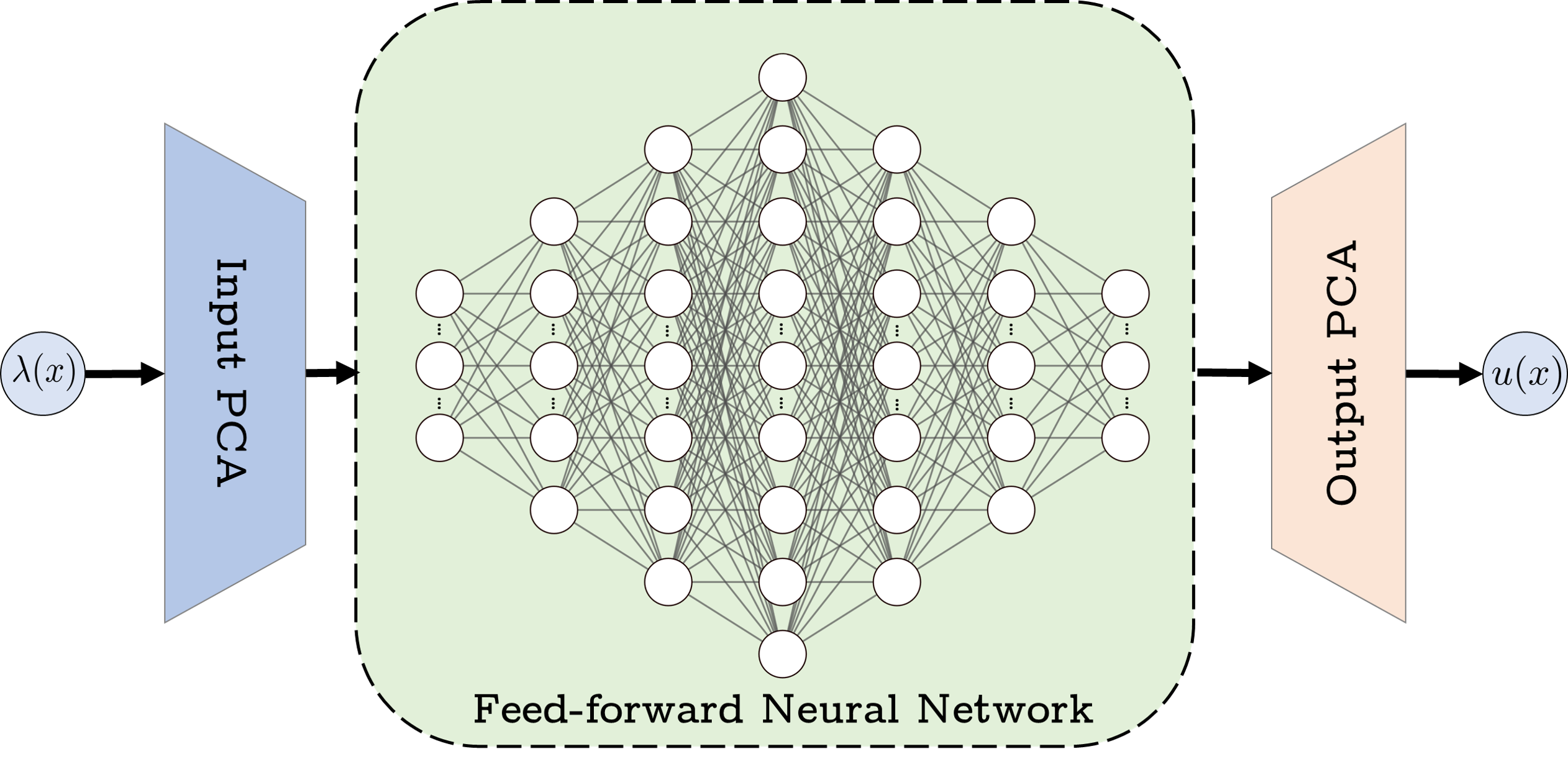}
		\caption{Architecture of the PCANN.}
		\label{fig:pcann_archi}
	\end{figure}

	In our numerical examples, we followed the outline of the original paper \cite{bhattacharya2020model} and used a fully connected feed-forward neural network for the mapping of the latent spaces. The number of nodes per layer is as follows $d_{\mathcal{X}}, 500, 1000, 2000, 1000, 500, d_{\mathcal{Y}}$. $d_{\mathcal{X}}$ and $d_{\mathcal{Y}}$ are considered to be hyperparameters, and for convenience, are chosen to be equal. As activation function we use the Scaled Exponential Linear Unit (SELU)  activation function which that induce self-normalizing properties in feed-forward neural networks \cite{klambauer2017self}.  Figure \ref{fig:pcann_archi} shows a simplified schematic of the overall architecture.
	
	\subsubsection{The Algorithm}
	As a purely data-driven method, no physics or PDE is needed in the training of the model. However,  the data used for the training of the network is obtained from a classical FDM for solving the underlying  By training the network with these input-output (numerically-given) function pairs, we obtain a neural operator which solves the PDE for various instances irrespective of the mesh size.
	
	Once again, we specify the algorithm for the Poisson problem in Equation \ref{eqn:poisson0}. The training data is the pair $\left(\lambda_{i}, u_{i}\right)$, with each $\lambda_i, u_{i} \in \mathbb{R}^{s^2}$. Originally, $\lambda_i, u_{i}$ are functions given in a (square) grid of dimensions $s \times s$. By flattening these functions, we now have them in $\mathbb{R}^{s^2}$. Training then proceeds as in Algorithm \ref{algo:PCANN}, while testing/usage of the trained network proceeds as in Algorithm \ref{algo:PCANN_test}.

	\subsubsection{Theoretical Background}
	As the PCANN method combines both the PCA and Neural network (NN) to achieve the task of operator learning, the strength of its approximation, therefore, comes from that of both the PCA and NNs. For a good approximation, one thus needs not only a good neural network but also, appropriate PCA truncation parameters (resulting from choosing sufficient amounts of data). The existence of all these factors is shown in Theorem 3.1 of \cite{bhattacharya2020model} which is a consequence of Theorem 3.5 of the same work. A recent work \cite{lanthaler2023operator} further extends the approximation theory of PCANN. Notably, it derives novel approximation results, with minimal assumptions \cite{lanthaler2023operator}.

	%\subsection{Profile of the Method}
	\begin{algorithm}[htp]
		\SetNoFillComment
		\caption{PCANN}
		\label{algo:PCANN}
		\KwIn{\\
			\qquad$\bullet$ $\left(\lambda_{i}, u_{i}\right)$: Training Data pair with $i = 0 \ldots N_{\textnormal{train}} $\\ 
			\qquad$\bullet$ $\tau$: learning rate
		}
		Initialise the network $\Phi_{\Theta}$\\
		Compute PCA of $\left(\lambda_{i}\right)$, store PCA $\left(a_{k}\right), k=1,\ldots, d_{\mathcal{X}}$ \label{PCA1}\\
		Compute PCA of $(u_{i})$, store PCA $(b_{\ell}), \ell=1, \ldots, d_{\mathcal{Y}}$\label{PCA2}\\
		\For{$i \leftarrow 0, \ldots, N_{\textnormal{train}} $}{
			compute $c_{i}^{k}=\langle\lambda_{i}, a_{k}\rangle \in R^{d_{\mathcal{X}}}$\\
			compute $d_{i}^{\ell}=\langle u_{i}, b_{\ell}\rangle \in R^{d_{\mathcal{Y}}}$
		}
		\While{\textnormal{not converged}}{ 
			\tcc{Train network}
			\For{$i \leftarrow 0, \ldots, N_{\textnormal{train}} $}{
				Compute Loss, $L_i = \sum_{\ell=1}^{d_{\mathcal{Y}}}\left\|\dfrac{\Phi_{\Theta}\left(c_{i}^{k}\right)-d_{i}^{\ell}}{d_{i}^{\ell}}\right\|_{2}$ 
			}
			Compute the Loss, $L = \sum_{i=1}^{N_{\textnormal{train}}} L_i$\\
			Optimise L using the appropriate/chosen optimisation algorithm.\\
			Update $\Theta \leftarrow \Theta - \tau \nabla_\Theta L$ 
		} 
	\end{algorithm}
	\begin{algorithm}[htp]
		\SetNoFillComment
		\caption{Using/Testing the PCANN}
		\label{algo:PCANN_test}
		\KwIn{\\
			\qquad$\bullet$ Input parameter functions $\lambda_i$ with $i = 0 \ldots N_{\textnormal{test}}$\\ 
			\qquad$\bullet$ Input PCA basis $\left(a_{k}\right)_{k=1,\ldots, d_{\mathcal{X}}}$\\
			%\qquad$\bullet$ Inverse PCA basis $(\tilde{b}_{p})_{p=1,\ldots,  s^2}$ of the 
			\qquad$\bullet$ Output PCA basis $(b_{\ell})_{\ell=1,\ldots,  d_{\mathcal{Y}}}$\\
			\qquad$\bullet$ Trained network $\Phi_{\Theta}$.
		}
		\KwResult{Output solution functions $\tilde{u_i}$, with $i = 0 \ldots N_{\textnormal{test}}$}
		\For{$i \leftarrow 0, \ldots, N_{\textnormal{test}}$}{
			$\tilde{u_i} = \sum_{\ell} \Phi_{\Theta} \left( \langle\lambda_i, a_{k}\rangle \right)_{\ell} b_{\ell} $\\
		}
	\end{algorithm}
	
	\subsubsection{PCANN adaptation for inverse problems}
	Following the general concepts as discussed in Section \ref{subsec:Tikhonov_inverse} we have two principle options. We either train a PCANN network $\Psi_\Theta(u)$ with training data $(u^\delta_i, \lambda_i)$, where we reverse the input-output order. I.e. the $(u^\delta_i)$ are used as input and
	$( \lambda_i)$ as output.
	After training we can directly use a new measurement $(u^\delta)$ and compute a corresponding parameter $\hat \lambda = \Psi_{\Theta}(u^ \delta)$.
	
	The other option is to use a trained network for the forward problem $\Phi_\Theta$ and include this into a Tikhonov functional.
	The resulting algorithm is summarized as follows:
	\begin{algorithm}[htp]
		\SetNoFillComment
		\caption{Using trained forward PCANN for Inverse Problem}
		\label{algo:PCANN_inv}
		\KwIn{\\
			\qquad$\bullet$ Solution function $u^{\delta}$\\ 
			\qquad$\bullet$ Input PCA basis $\left(a_{k}\right)_{k=1,\ldots, d_{\mathcal{X}}}$\\
			%\qquad$\bullet$ Inverse PCA basis $(\tilde{a}_{j})_{j=1,\ldots,  s^2}$ of the Input PCA basis.\\ 
			\qquad$\bullet$ Output PCA basis $\left(b_{\ell}\right)_{\ell=1, \ldots, d_{\mathcal{Y}}}$\\
			\qquad$\bullet$ Trained network $\Phi_{\Theta}: c=\left(\left\langle\lambda, a_{k}\right\rangle\right) \mapsto d=\left(\left\langle u, b_{\ell}\right\rangle\right)$.
		}
		\KwResult{Output parameter function $\hat{\lambda}$}
		Initialise $c^0$\\
		\For{$m \leftarrow 0, \ldots, N$ (\text{large integer})}{
			Compute $J_{\alpha}(c) =\left\|\Phi_{\Theta}(c) - \langle u^{\delta}, b_{\ell} \rangle \right\|_{2}$\\
			Compute the loss $J_{\alpha}(c):= J_{\alpha}(c) +\alpha R(c).$\\
			Optimise Loss $J_{\alpha}(c)$ using gradient descent\\
			Update $c^{m+1} \leftarrow \left(c^{m}-\sigma_{m} \partial J / \partial c\left(c^{m}\right)\right)$
		}
		Compute $\hat{\lambda}= \sum_k c_{k}^N a_{k}$\\
	\end{algorithm}

	\subsection{Fourier Neural Operator} \label{sec:FNO}
	
	\subsubsection{Motivation}
	Fourier neural operators (FNO) \cite{li2020fourier} are designed as deep learning architectures for learning mappings between infinite-dimensional function spaces. The Fourier neural network is formulated as an iterative architecture, where each iteration (hidden layer), inspired by the convolution theorem, is a Fourier integral operator defined in Fourier space. The main network parameters are therefore defined and learned in Fourier space rather than in physical space, i.e., the coefficients of the Fourier series of the output function are learned from the data. 
	
	While standard feed-forward neural networks (FNN) and CNNs combine linear multiplications with non-linear activations in order to learn highly nonlinear functions, FNOs combine linear integral operators with non-linear activations in each layer to learn non-linear operators. The fast Fourier transform (FFT) makes the implementation even more efficient.
	
	Different from DeepONet, and similar to the PCANN methods, FNO discretises both the input function $\lambda(x)$ and the output function $u(x)$ by using point-wise evaluations in an equispaced mesh. In addition, FNO requires that $\lambda$ and $u$ be defined on the same domain and are discretised by the same discretisation. 
	The function space formulation of the approach allows training an FNO  on a dataset with low resolution and applying it directly on a dataset with higher resolution, thus achieving the so-called zero-shot super-resolution.

	\subsubsection{Theoretical Background}
	
	The idea of FNO, stems from an earlier paper by the same team, see \cite{li2020neural}, which also seeks to approximate a neural operator by concatenating multiple hidden layers. While standard artificial neural networks (FCN, CNN) have affine functions (weights and biases) with scalar nonlinear activation as hidden layers, neural operators rather have affine operators (usually, in addition to affine functions with weights and biases \cite{kovachki2021universal} \cite{FAIR_karniadakis}) with scalar nonlinear activation functions. A hidden layer numbered $j+1$ thus performs the update of input $v_j$ as follows
	\begin{eqnarray}
		v_{j+1}(x):=\sigma\left(W v_{j}(x) + \left(\mathcal{K}(\lambda ; \theta) v_{j}\right)(x)\right), \quad \forall x \in \Omega,
		\label{eqn:fourier_update}
	\end{eqnarray}
	where $W, b$ and $\mathcal{K}$ are the respective weights, biases and neural operators. These neural operators are chosen to be integral operators of the form
	\begin{eqnarray}
		(\mathcal{K}(\lambda ; \theta) v)(x)=\int_{\Omega} \kappa_{\theta}(x, y ; \lambda(x), \lambda(y)) v(y) d y, \quad \forall x \in \Omega.
		\label{eqn:int_kernel}
	\end{eqnarray}
	These operators can be realised by different concepts such as graph kernels \cite{li2020neural}, multipole expansions \cite{li2020multipole}, non-local kernels \cite{you2022nonlocal}, low-rank kernels \cite{kovachki2021neural}, wavelet transforms \cite{tripura2022wavelet}, multiwavelet transforms \cite{gupta2021multiwavelet}, or Laplace transforms \cite{chen2023laplace}. A recent work even uses some famous convolutional neural networks architectures \cite{raonic2023convolutional}.  
	In the special case where $\kappa_{\theta} = \kappa_{\theta} (x- y)$, Equation \ref{eqn:int_kernel} becomes 
	\begin{eqnarray*}
		(\mathcal{K}(\theta) v)(x) &=& \int_{\Omega} \kappa_{\theta} (x -y) v(y) d y, \quad \forall x \in \Omega\\
		& = & (\kappa_{\theta} * v )(x) , \quad \forall x \in \Omega\\
		& = & \left(\mathcal{F}^{-1}( \mathcal{F}(\kappa_{\theta}) \cdot \mathcal{F}(v)) \right)(x) , \quad \forall x \in \Omega\\
		& = & \left(\mathcal{F}^{-1}( K_{\theta} \cdot \mathcal{F}(v)) \right)(x) , \quad \forall x \in \Omega,
		%\label{eqn:int_kernel_special}
	\end{eqnarray*}
	%\begin{eqnarray*}
	%    (\mathcal{K}(\theta) v)(x) = \int_{\Omega} \kappa_{\theta} (x -y) v(y) d y =  (\kappa_{\theta} * v )(x) = \left(\mathcal{F}^{-1}( \mathcal{F}(\kappa_{\theta}) * \mathcal{F}(v)) \right)(x) = \left(\mathcal{F}^{-1}( K_{\theta} \cdot \mathcal{F}(v)) \right)(x) ,\forall x \in \Omega,
	%    \label{eqn:int_kernel_special}
	%\end{eqnarray*}
	by the use of the convolution theorem. This then leads to the parameterisation of the neural network given by $\kappa_\theta$ directly in the Fourier space $K_\theta$. \cite{kovachki2021universal} provides a good theoretical background of FNOs and \cite{kovachki2021neural} offers a good overview of neural operators using different integral operators along with some theoretical justification.
	%\subsection{Profile of the Method}.
	
	It is worth mentioning the recent work in \cite{lanthaler2023nonlocal}, which introduces the Nonlocal Neural Operator (NNO), that generalises over arbitrary geometries. FNO is a special case of the NNO, and this work highlights how increasing the number of channels in FNO as opposed to increasing the number of Fourier modes benefits the FNO. The same work  also introduces the averaging neural operator which is a subclass of NNO but also happens to be at the core of many neural operator frameworks \cite{lanthaler2023nonlocal}.

	\subsubsection{The algorithm} \label{subsec:FNO-algo}
	
	The classical FNO is only possible with a uniform grid (cartesian domain), but can be extended for any mesh. We note that, unlike the PCANN method, the input and output functions are inputted in their unflattened states. Once again, as an example, we consider the problem in Equation \ref{eqn:poisson0} and we seek to learn the operator  
	\begin{eqnarray}
		F:  \Lambda (\Omega; \mathbb{R}^{d_\lambda}) \ni \lambda(x) \mapsto u(x) \in \mathcal{U}(\Omega;  \mathbb{R}^{d_u}),
	\end{eqnarray}
	where $d_\lambda$ and $d_u $ depend on the discretisation used ($d_\lambda = d_u  = 513^2$ if we use functions with a resolution of $513 \times 513$),  $\Omega$ is a subset of $\mathbb{R}$ while $\Lambda$ and $\mathcal{U}$ are Banach spaces of functions taking values in $\mathbb{R}$.

	\subsubsection{Network Architecture} \label{subsec:FNO-archi}
	
	The architecture of the FNO follows from the algorithm described in Section \ref{subsec:FNO-algo}. %as seen in \cite{li2020fourier}. in Figure \ref{fig:fno_archi}. 
	For the special case in Equation \ref{eqn:poisson0}, we consider a mesh of resolution $N \times M$. We denote each point in the domain $\Omega$ as  $x_k^l$, with input $\lambda(x_k^l)$, and output $u(x_k^l)$, with $k \in \{1, \ldots, N\}$ and $l \in \{1, \ldots, M\}$. We equally denote $\lambda(x)$ and $u(x)$ as the input and output for the whole domain.  
	\begin{itemize}
		\item The FCN $P_\theta$ takes each $\lambda(x_k^l) \in \mathbb{R}$ and maps it to a higher dimension $d_{v_0}$, i.e. 
		\begin{eqnarray*}
			v_0(x_k^l) &=& P_\theta (\lambda(x_k^l)) \in \mathbb{R}^{d_{v_0}},\\
			v_0(x) &=& P_\theta (\lambda(x)) \in \mathbb{R}^{d_{v_0}  \times N \times M}.
		\end{eqnarray*} 
		$v_0(x)$ is thus an $N \times M$ `image' with $d_{v_0}$ channels.
		
		\item To obtain $v_1(x)$ from $v_0(x)$, the following operations are applied.
		
		\begin{itemize}
			%\item The Fourier transform $\mathcal{F}$ is applied to each channel of $v_0 (x)$. We then obtain, $\mathcal{F} (v_0 (x)) \in \mathbb{R}^{d_{v_0} \times k}$
			\item The (2D) Fourier transform $\mathcal{F}$ is applied to each channel of $v_0 (x)$, and the first $k = k_1 \cdot k_2$ modes are kept ($k_1, k_2$, being the number of modes to be kept in the first and second dimensions respectively). As a result we obtain $\mathcal{F} (v_0 (x)) \in \mathbb{C}^{d_{v_0} \times k}$.
			
			%\item The linear transform and the first $k = k_1 \cdot k_2$ modes are kept ($k_1, k_2$, being the number of modes to be kept in the first and second dimensions respectively)
			\item The linear transform $\mathcal{R}_\phi$, which is realised as a weight matrix in $\mathbb{C}^{k \times d_{v_0} \times d_{v_0}}$ is applied to $\mathcal{F} (v_0 (x))$. The result is  $\mathcal{R}_\phi \cdot \mathcal{F} (v_0 (x)) \in \mathbb{R}^{d_{v_0} \times k}$ .
			
			\item Apply the inverse FFT to $\mathcal{R}_\phi \cdot \mathcal{F} (v_0 (x))$, (appended with zeros to make up for the truncated/unselected modes). The outcome is $\mathcal{F}^{-1}\left(\mathcal{R}_\phi \cdot \mathcal{F} (v_0 (x))\right) \in \mathbb{R}^{N \times M \times d_{v_0}}$
			
			\item Finally, $v_1(x)$ is obtained from the above together with a residual connection with a weight matrix $W_{\Phi} \in \mathbb{R}^{d_{v_0} \times d_{v_0}}$, as follows $$v_1(x) = \sigma (W_{\Phi} \cdot v_0(x) + \mathcal{F}^{-1}\left(\mathcal{R}_\phi \cdot \mathcal{F} (v_0 (x))\right)) \in \mathbb{R}^{N \times M \times d_{v_0}}$$
			
		\end{itemize}
		\item $v_2, v_3, \ldots, v_T$ are obtained from $v_1, v_2, \ldots, v_{T-1}$ respectively in a similar manner as above. In this way, we have $T$ Fourier layers.
		
		\item At the end of the last Fourier layer another FCN $Q_\psi$  is applied, which projects $v_T(x)$ to the output dimension. We therefore obtain the output function $$u(x) = Q_\psi(v_T(x)) \in \mathbb{R}^{N \times M}$$
	\end{itemize}
	
	In the FNO architecture, we use the Gaussian Error Linear Units (GELUs) \cite{hendrycks2016gaussian, hendrycks2016bridging} as activation functions.
	
	\begin{figure}[!ht]
		\centering
		\includegraphics[width=\linewidth]{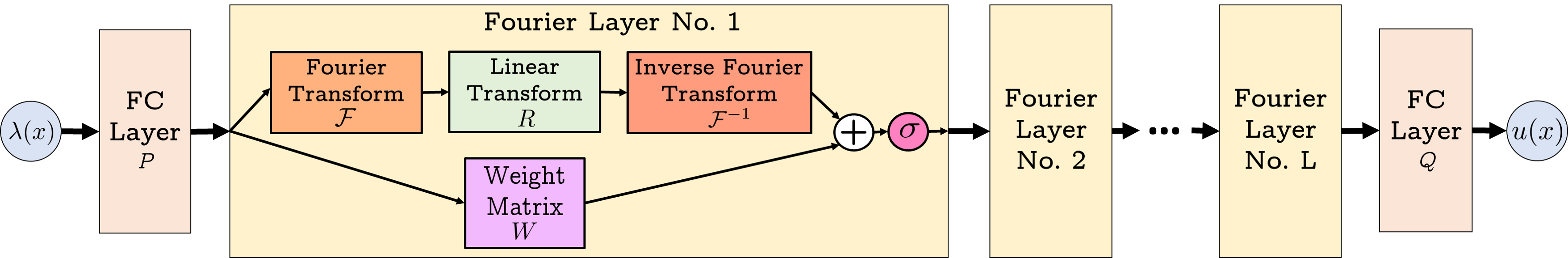}
		\caption{Architecture of the FNO.}
		\label{fig:fno_archi}
	\end{figure}
	
	\begin{algorithm}[htp]
		\SetNoFillComment
		\caption{FNO}
		\label{algo:FNO}
		\KwIn{\\
			\qquad$\bullet$ $\left(\lambda_{i}, u_{i}\right)$: Training data pair with $i = 0 \ldots N_{\textnormal{train}} $\\ 
			\qquad$\bullet$ $\tau$: learning rate     
		}
		Initialise the fully connected networks (FCN) $P_\theta$ and $Q_{\psi}$\\
		Initialise the linear transforms $W_{\Phi}$ and $R_{\phi}$\\
		\tcc{Section \ref{subsec:FNO-archi} defines these FCN and linear transform networks}
		Lift $\lambda_i$ to a higher dimension by computing $v_{0}=P_\theta(\lambda_i)$\\
		\For{$j \leftarrow 0, \ldots, T-1$}{
			$v_{j+1}:=\sigma\left(W_\Phi v_{j}+[\mathcal{F}^{-1}\left(R_{\phi} \cdot\left(\mathcal{F} v_{j}\right)\right)]\right)$, output of Fourier layers\\
		}
		Obtain the target dimension by using $Q_\psi$ to get $\tilde{u_i} = Q_\psi(v_T)$\\
		\While{\textnormal{not converged}}{ 
			\tcc{Train network}
			\For{$i \leftarrow 0, \ldots, N_{\textnormal{train}} $}{
				Compute loss $L_i = \left\|\dfrac{\tilde{u_i}-u_i}{u_i}\right\|_{2}$     	}
			Compute the Loss, $L = \sum_{i=1}^{N_{\textnormal{train}}} L_i$\\
			Optimise L using the appropriate/chosen optimisation algorithm.\\
			Update $\Theta \leftarrow \Theta - \tau \nabla_\Theta L$  where $\Theta=(\theta, \psi)$.
		} 
	\end{algorithm}
	\subsubsection{Generalisation and Extensions of FNOs}
	A detailed guide for implementation and an illustration of how to extend FNO to operators with inputs and outputs defined on different domains can be found in \cite{FAIR_karniadakis}. There, also two cases of generalisation are considered:
	
	\begin{enumerate}[\bf C 1:]
		\item As motivation, take a parabolic PDE, where the initial condition at time $t=0$ is the input for the parameter-to-state operator. The output is the solution for all times $t \in [0, T]$.
		This leads to the general setting where the output space is that of functions defined on a product space, where the first component is the same as for the input space and the second component is arbitrary. i.e.
		
		$$ F:  \Lambda (\Omega; \mathbb{R}^{d_\lambda}) \ni \lambda(x) = u(x, 0) \mapsto u(x, t) \in \mathcal{U}(\Omega \times [0, t]; \mathbb{R}^{d_u}),$$
		
		where $\Omega$ is a subset of $\mathbb{R}$ while $\Lambda$ and $\mathcal{U}$ are Banach spaces taking values in $\mathbb{R}^{d_\lambda}$ and $\mathbb{R}^{d_u}$ (with $d_\lambda = d_u$, depending on the discretisation of the functions) respectively.
		In order to match dimensions before applying an FNO, the output domain could be shrunk by use of a Recurrent Neural Network (RNN)  or the input domain could be extended to incorporate the time, $t$ as an extra coordinate. %( CK, please define RNN.) || Resp: Done
		
		\item The input space is a subset of the output space, like in the case of the operator $F$, mapping the boundary conditions of the domain to the solution of the PDE in the whole domain. 
		
		$$ F:  \mathcal{G} (\partial \Omega; \mathbb{R}^{d_g}) \ni g(x) = \left.u\right|_{\delta \Omega} \mapsto u(x) \in \mathcal{U}(\Omega; \mathbb{R}^{d^u}),$$
		
		where $\partial \Omega \subset \Omega \subset \mathbb{R}$, while $\mathcal{G}$ and $\mathcal{U}$ are Banach spaces taking values in $\mathbb{R}^{d_g}$ and $\mathbb{R}^{d_u}$ (defined in a similar way as above in the first case) respectively.
		For this situation, zero padding could be applied or better still, an appropriate transformation could be used to map the boundary condition to a lower dimension. This then brings us back to the previous case.
	\end{enumerate}
	
	It is worth noting that the use of the FFT in the algorithm of the FNO restricts its applicability to situations where the input and output are defined on a cartesian domain (square, rectangle domain in 2D and cube, cuboid in 3D). This introduces a challenge when dealing with complex geometries. A practical workaround is as followed.
	
	\begin{enumerate}[\bf W 1:]
		\item Define a new domain which is the minimum bounding box of the complex domain, and perform a zero padding for the area within this bounding box, but out of the complex domain. However, in practice, this makes the resulting function discontinuous in the `rectangle' and usually yields larger errors. 
		\item Use the same minimum bounding of the complex domain, as previously described, but padding is done with the nearest function values within the complex domain, instead of zeros.
	\end{enumerate}
	
	Though somewhat successful, the extensions provided in \cite{FAIR_karniadakis} do not seem to favour FNO. A better approach to complex domains is by transforming/deforming this complex domain to a cartesian domain by use of neural networks (NN), positioned before and after the FNO as proposed in \cite{li2022fourier}. These neural networks could either be fixed or learned together with the FNO parameters. The resulting NN-FNO-NN network is said to be `geometry aware' and is called Geometry-Aware FNO (Geo-FNO). 
	
\begin{figure}[htp]
		\centering
		\includegraphics[width=\linewidth]{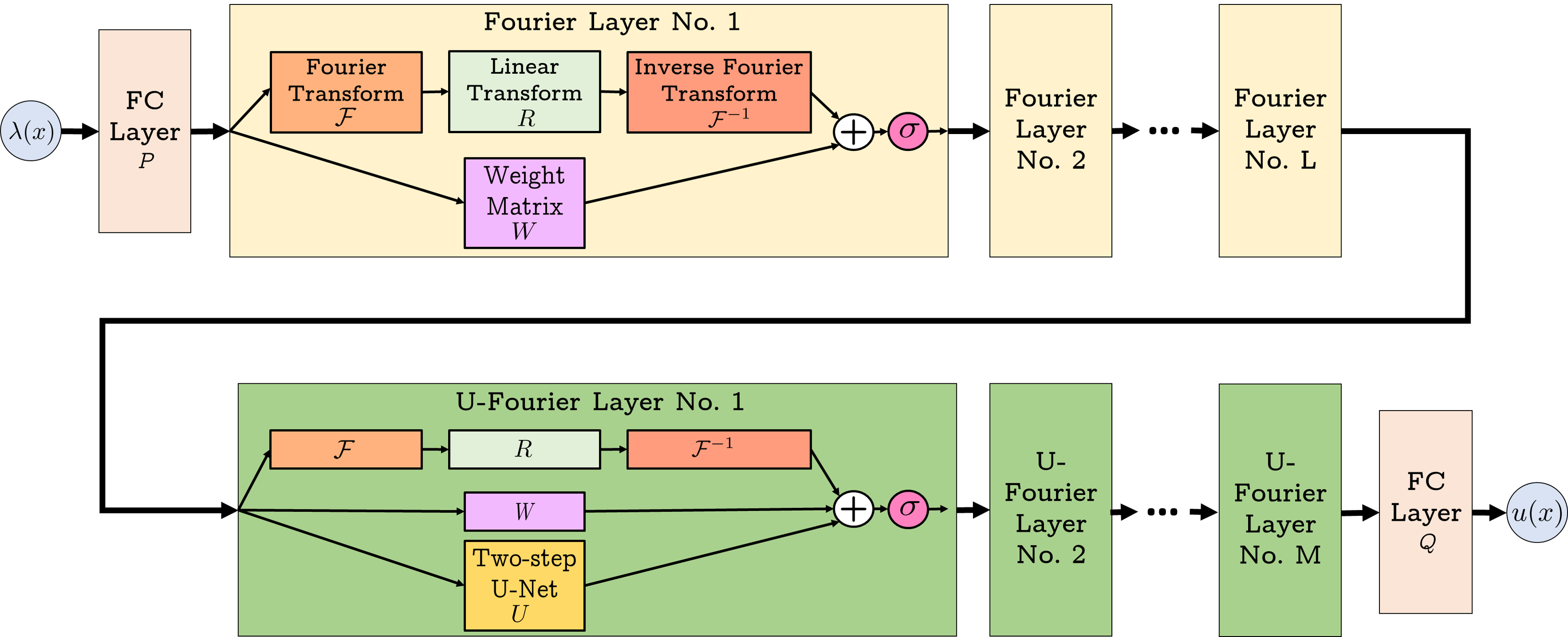}
		\caption{Architecture of the U-FNO.}
		\label{fig:ufno_archi}
	\end{figure}

	Besides these structural considerations, the following generalisations have recently gained increasing attention:
	
	%\begin{itemize}
	
	\paragraph{From conv-FNO to U-FNO:} 
	Possibly the most recent variant of the FNO, the U-FNO \cite{wen2022u} introduces the famous U-Net to the architecture in the so-called U-Fourier layer. The U-FNO thus has both Fourier and U-net layers: it starts off with the former and ends with the latter. Essentially, a U-Fourier layer is similar to the Fourier layer, but has both a weight matrix and a two-step U-Net layer in the residual as illustrated in Figure \ref{fig:ufno_archi}. In contrast to the update performed by the Fourier layer in Equation \ref{eqn:fourier_update}, the U-Fourier layer performs the update in Equation \ref{eqn:u-fourier_update}, with $U$ being the two-step U-Net.
	\begin{eqnarray}
		v_{t+1}(x):=\sigma\left(U v_{t}(x) + W v_{t}(x)+ \left(\mathcal{K}(a ; \theta) v_{t}\right)(x)\right), \quad \forall x \in \Omega .
		\label{eqn:u-fourier_update}
	\end{eqnarray}
	As compared with methods based on CNNs, the baseline FNO achieves a good accuracy for single-phase flow problems, but it doesn't seem to do so well with multiphase flows  \cite{wen2021ccsnet}. On the other hand, conv-FNO does better than CNN-based methods, and the U-FNO even improves on that. conv-FNO is implemented by using a standard convolution in place of the U-Net. 
	\paragraph{Multiwavelet based operator (MWT):}
	In a more recent work \cite{gupta2021multiwavelet}, a neural operator which evaluates the kernel operator in (\ref{eqn:int_kernel}) using a different approach is introduced. The idea here is to leverage the successes in signal processes of both orthogonal polynomials and wavelet basis (notably, vanishing moments and orthogonality). Multiwavelets, therefore, do not just project the function onto a single wavelet function as wavelets do, but rather, they project the function onto a subspace of degree-restricted polynomials.
	
	For an understanding of the concept, the space of piece-wise polynomials of degree up to $k \in \mathbb{N}$ with $n \in \mathbb{Z} \cup \{0\}$ is defined on $[0,1]$: $$\mathbf{V}_n^k=\bigcup_{l=0}^{2^n-1}\left\{f \mid \operatorname{deg}(f)<k  \text{ for } x \in\left(2^{-n} l, 2^{-n}(l+1)\right) \wedge 0, \text{ elsewhere} \right\}.$$
	For each $k \in \mathbb{N}$ with $n \in \mathbb{Z} \cup \{0\}$, we have that $\text{dim}(\mathbf{V}_n^k) = 2^n k$ and
	\begin{eqnarray}
		\mathbf{V}_{n-1}^k \subset \mathbf{V}_n^k.
		\label{eqn:ops}  
	\end{eqnarray}
	We note that, as $n$ increases, so does $l$, and $\mathbf{V}_n^k$ is defined from a lower(coarser) to a higher(finer) resolution; these resolutions being each time a power of 2. As a result, the method is restricted to discretisations which are a power of 2. Appropriate padding or interpolation could be applied to the function if its resolution is not a power of 2.
	
	Given the basis $\phi_j, j= 0, 1 \ldots , k-1$ w.r.t a measure $\mu_0$ of $\mathbf{V}_0^k$, it is possible to obtain the basis of subsequent $\mathbf{V}_n^k, n>0$ by appropriate shifts and scales of $\phi_j$ : 
	\begin{equation}
		\phi_{j l}^n(x)=2^{n / 2} \phi_j\left(2^n x-l\right), \quad j=0,1, \ldots, k-1, \quad l=0,1, \ldots, 2^n-1 \text{, w.r.t. }\mu_n
		\label{eqn:basis-rel}  
	\end{equation}
	One then defines the multiwavelet subspace $\mathbf{W}_n^k$, which is related to the spaces of orthogonal polynomials as below:
	\begin{eqnarray}
		\mathbf{V}_{n+1}^k=\mathbf{V}_n^k \bigoplus \mathbf{W}_n^k, \quad \mathbf{V}_n^k \perp \mathbf{W}_n^k .
		\label{eqn:op-mw}
	\end{eqnarray}
	Similarly, the basis of $\mathbf{W}_n^k, n>0$ can be obtained by appropriate shifts and scales if the basis $\psi_j, j= 0, 1 \ldots, k-1$ w.r.t a measure $\mu_0$ of $\mathbf{W}_0^k$ is known. A similar expression as in Equation \ref{eqn:basis-rel} can thus be obtained.
	
	Equations \ref{eqn:ops} and \ref{eqn:op-mw} inform us that the basis of $\mathbf{V}_n^k$ and ${W}_n^k$ can be written as linear combinations of that of $\mathbf{V}_{n+1}^k$.
	As a result, for a given function it is possible to obtain a relationship between its coefficients in these bases. Specifically, the function $f \in \mathbb{R}^d$ has coefficients in the space of orthogonal polynomials (multiscale coefficients) $\mathbf{s}_l^n = [\left\langle f, \phi_{i l}^n\right\rangle_{\mu_n}]_{i=0}^{k-1} \in \mathbb{R}^{k^d \times 2^n}$ and coefficients in the multiwavelet subspace (multiwavelet coefficients) $\mathbf{d}_l^n = [\left\langle f, \psi_{i l}^n\right\rangle_{\mu_n}]_{i=0}^{k-1} \in \mathbb{R}^{k^d \times 2^n}$, which are related by the \textit{decomposition} equations: 
	\begin{eqnarray}
		\mathbf{s}_l^n &=& H^{(0)} \mathbf{s}_{2 l}^{n+1}+H^{(1)} \mathbf{s}_{2 l+1}^{n+1}, 
		\label{eqn:dec1}\\
		\mathbf{d}_l^n &=& G^{(0)} \mathbf{s}_{2 l}^{n+1}+G^{(1)} \mathbf{s}_{2 l+1}^{n+1},
		\label{eqn:dec2}
	\end{eqnarray}
	and \textit{reconstruction} equations
	\begin{eqnarray}
		\mathbf{s}_{2 l}^{n+1} &=& \Sigma^{(0)}(H^{(0) T} \mathbf{s}_l^n+G^{(0) T} \mathbf{d}_l^n),
		\label{eqn:rec1} \\
		\mathbf{s}_{2 l+1}^{n+1} &=& \Sigma^{(1)}(H^{(1) T} \mathbf{s}_l^n+G^{(1) T} \mathbf{d}_l^n).
		\label{eqn:rec2}
	\end{eqnarray}
	$H^{(0)}, H^{(1)}, G^{(0)}, G^{(1)} \in \mathbb{R}^{k^d \times k^d}$ are the reconstruction filters while $\Sigma^{(0)}, \Sigma^{(1)}$ are the correction terms. The theory and derivation of these terms are well covered in  \cite{gupta2021multiwavelet}, we only highlight it here for better explanation of the steps involved in the method.
	In Equations \ref{eqn:dec1}-\ref{eqn:dec2}, $H^{(0)}$ and $G^{(0)}$ act on the even terms of the multiscale coefficient while $H^{(1)}$ and $G^{(1)}$  act on the odd terms. The result of these operations is visibly a down-sampling from the higher to a lower resolution (half the original resolution), thus the term decomposition. On the other hand, Equations \ref{eqn:rec1}-\ref{eqn:rec2} perform the reverse of this operation, leading to the recovery of the higher resolution.
	
\begin{figure}[htp]
		\centering
		\includegraphics[width=\linewidth]{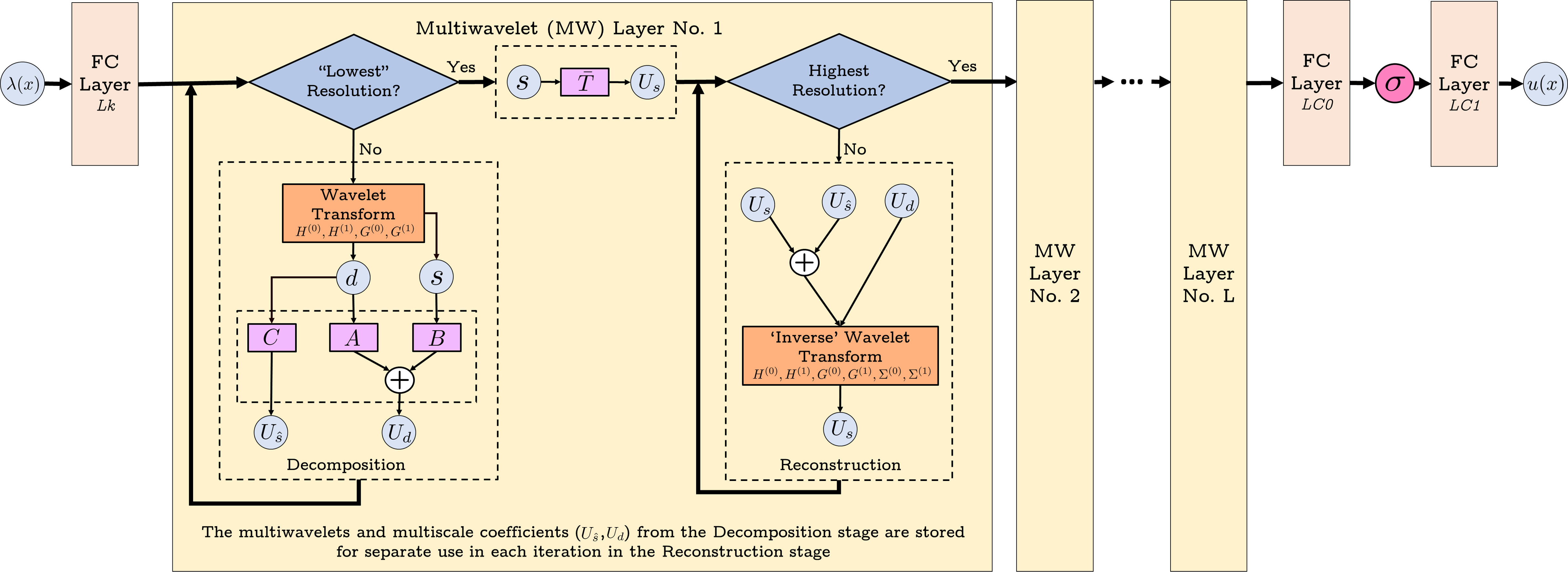}
		\caption{Architecture of the MWT Operator.}
		\label{fig:mwt_archi}
	\end{figure}

	So far, we walked through multiwavelet representation of a function $f \in \mathbb{R}^d$. This notion will be applied to the input and output functions of the neural network/operator. A different notion, known as the Non-Standard Form is used to obtain the multiwavelet representation of the kernel function. This method essentially reduces the operator kernel in Equation \ref{eqn:int_kernel} $\mathcal{K}v = w$ to the set of equations. 
	\begin{eqnarray*}
		U_{d l}^n &=& A_n d_l^n + B_n s_l^n, \\
		U_{\hat{s} l}^n &=& C_n d_l^n,\\
		U_{s l}^L &=& \bar{T} s_l^L,
	\end{eqnarray*}
	with $U_{d l}$ and $U_{s l}$ (respectively $d_l^n$ and $s_l^n$) being the respective multiscale and multiwavelet coefficients of $w$ (respectively $v$). $L$ here is the index corresponding to the coefficients of the lowest resolution (output of the last decomposition). $A_n, B_n, C_n$ and $\bar{T}$ are then parameterised with a CNN (which could be done in Fourier space as in FNO or not) network followed by a ReLU activation then linear layer which we learn during training. In practice we use same neural networks for all $n$: thus $A_n = A_{\theta_A}, B_n := B_{\theta_B}, C_n := C_{\theta_C}$ and $\bar{T} := \bar{T}_{\theta_T}$. For a given input, the multiwavelet operator performs a series of operations as shown in Figure \ref{fig:mwt_archi}.
	
	\paragraph{Physics Informed (Fourier) Neural Operator - PINO} 
	
	This method combines the operator learning of FNOs and function learning PINNs, in a quest to alleviate the challenges faced in the individual cases \cite{li2021physics}. While PINNs face a challenge in optimisation when dealing with complex problems as in multiscale dynamics, FNO needs a wealth of data which might not be readily available. PINO combines both concepts by introducing two stages in the learning process:
	\begin{itemize}
		\item Firstly, the operator is learned using either/both a \textbf{data loss} (same loss as in FNO) or/and a \textbf{PDE loss} (physics informed loss), in a phase termed Pre-training. This case reduces to the normal FNO method if no PDE loss is used. As an example, the PDE loss for Poisson's equation in \ref{eqn:poisson0} is given by Equation \ref{eqn:pdeloss-pino} where $\alpha$ is a hyperparameter.
		\begin{eqnarray}
			L_{\text{PDE}} = \underbrace{\sum_{i = 1}^{N_{\textnormal{train}}}\|\Delta u_{i\Theta} - \lambda_{i\Theta}\|^{2}_{L^2(\Omega)}}_\text{\normalsize Domain ($\Omega$) loss} + \underbrace{\alpha \sum_{i = 1}^{N_{\textnormal{train}}}\|u_{i\Theta} - g_{i}\|^{2}_{L^2(\partial \Omega)}}_\text{\normalsize Boundary ($\partial \Omega$) loss},
			\label{eqn:pdeloss-pino}
		\end{eqnarray}
		\item Secondly, for a specific instance of the input (parameter, for forward problem), the learned operator from the first stage is further fine-tuned, using the \textbf{PDE loss} and an \textbf{operator loss}. The latter minimises the difference between the initial network, $F_{\Theta_0}$ resulting from the pre-training phase and the further optimised networks $F_{\Theta_i}, i >0$. This phase is the test-time optimisation stage. For a specific parameter instance $\lambda$, the operator Loss is thus given by
		\begin{eqnarray}
			L_{\text{OP}} = \|F_{\Theta_i}-F_{\Theta_0}\|^2.
			\label{eqn:pdeloss}
		\end{eqnarray}
		
	\end{itemize}
	
	A strength of the PINO is in its ability to achieve competitive errors with fewer data as demonstrated in \cite{li2021physics}. This is mainly due to the introduction of the PDE loss. Notably, this PDE loss is evaluated in a not-so-usual way as the input of the network is the function and not a set of collocation points. Automatic differentiation as we know it, is thus not possible. Three methods for gradient descent are outlined in \cite{li2021physics}.  One option is numerical differentiation, i.e. a finite difference method (FDM), which we use for our experiments in later sections.
	%\end{itemize}
	
	\subsubsection{FNO usage for inverse problems}
	FNO is an operator learning concept, hence both general concepts as outlined in Section \ref{subsec:Tikhonov_inverse} can be used for solving inverse problems. We will report on the achieved results in our section on numerical examples. FNO does remarkably well, despite its rather linear and Fourier-centric approach. However, it allows the incorporation of fine details related to high frequencies in a consistent way, nevertheless understanding the success of FNO methods for solving inverse problems should be an important direction for future research.
	
	\subsection{DeepONet} \label{sec:DeepONet}
	\subsubsection{Motivation}
	It is widely known that deep neural networks are universal approximators, i.e., they can approximate any finite-dimensional function to arbitrary accuracy \cite{cybenko1989approximation,hornik1989multilayer}. This has been extended to a universal approximation theorem for operators \cite{chen1995universal}, which states that a neural network with a single hidden layer can approximate accurately any nonlinear continuous operator. This theorem and its extension to multi-layer networks, see \cite{lu2019deeponet}, provides the motivation for the concept of DeepONet.
	
	%\begin{theorem}[\textbf{Universal Approximation Theorem for Operator}]
	%Suppose that $\theta$ is a continuous nonpolynomial
	%function, $X$ is a Banach Space, $K_{1}\in X, K_{2}\in \mathbb{R}^{d}$ are two compact sets in $X$ and $\mathbb{R}^{d}$,respectively, $V$ is a compact set in $C(K_{1})$, $G$ is a nonlinear continuous operator, which maps $V$ into $C(K_{2})$.Then for any  $\epsilon>0$, there are positive integers $n, p, m,$ constants $c_{i}^{k},\xi_{i,j}^{k},\theta_{i}^{k},\zeta_{k}\in \mathbb{R},w_{k}\in R^{d}, x_{j}\in K_{1}, i=1,...,n, k=1,...,p, j=1,...,m,$ such that
	%\end{theorem}
	
	\subsubsection{Network architecture}
	The DeepONet \cite{lu2019deeponet} mirrors the structure of the universal approximation theorem of operators with a novel network architecture. Let us consider an operator $\cal F$ that maps an input function $\lambda$ to an output function $u$, i.e. $u={\cal F}(\lambda)$. A DeepONet $G_{\Theta}$ takes an input function $\lambda$, which is sampled at fixed predefined collocation points, and provides an approximation for $u(x)$ at arbitrary points $x$ by the combination of a trunk and branch net in the following way:
	\begin{equation}
		u(x)\approx G_{\Theta}(\lambda)(x)=\sum_{k=1}^{p}\underbrace{b_{k}(\lambda(s_{1}),\lambda(s_{2}),...,\lambda(s_{m}))}_{\text{branch}}\underbrace{t_{k}(x)}_{\text{trunk}}=\sum_{k=1}^{p}b_{k}(\mathbf{\lambda})t_{k}(x).
	\end{equation}
	
	The trunk net $\mathbf{t}=\mathbf{T}(x;\theta_{t})$ takes the continuous coordinates $x$ as the input, and outputs a feature vector $\mathbf{t}=[t_{1},...,t_{p}]\in \mathbb{R}^{p}$, which can be considered as $p$ functions of $x$. The branch net $\mathbf{b}=\mathbf{B}(\mathbf{\lambda};\Theta_{b})$ takes $\mathbf{\lambda}=[\lambda(s_{1}),\lambda(s_{2}),...,\lambda(s_{m})]$, the discretisation of the input function, as input and returns a feature vector $\mathbf{b}=[b_{1},...,b_{p}]\in \mathbb{R}^{p}$ as output. The branch and trunk nets are then combined by an inner product to approximate the underlying operator. 
	
	This DeepONet is named as "unstacked DeepONet", while it is called a "stacked DeepONet" if each $b_{k}(\mathbf{\lambda}), k=1,...,p$ is an individual neural network, i.e.,$
	[b_{1},...,b_{m}]=[\mathbf{B}(\mathbf{\lambda};\Theta_{b}^{1}),...,\mathbf{B}(\mathbf{\lambda};\Theta_{b}^{m})].$
	Several numerical results have shown that unstacked DeepONets typically have larger training errors as compared with stacked DeepONets, but the test error is smaller and unstacked DeepONets lead to smaller generalisation errors.
	Both branch net and trunk net can have general architectures, e.g., fully connected neural network(FCN), recurrent neural network(RNN), and convolutional neural network(CNN). However, as $x$ is usually in low dimensional space, a standard FNN is commonly used as the trunk net. A bias can also be added to the last stage of the DeepONet to improve performance.
	
	\begin{figure}[htp]
		\centering
		\includegraphics[width=0.8\linewidth]{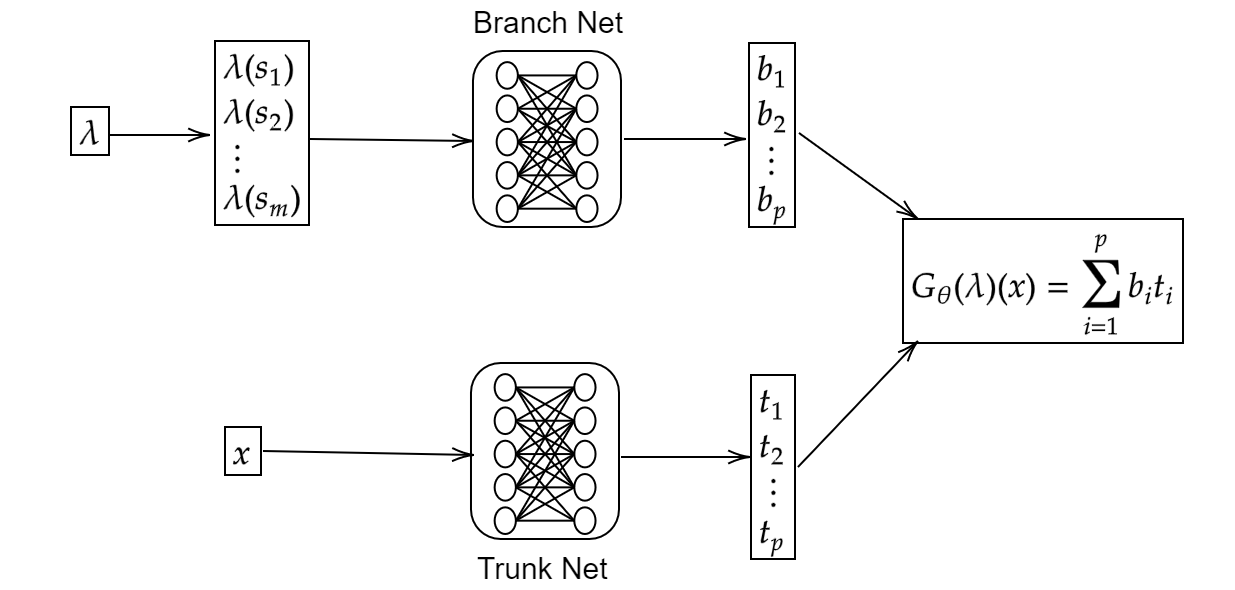}
		\caption{Architecture of the DeepONet}
		\label{fig:deeponet_archi}
	\end{figure}
	
	The novelty of the DeepONet is that its network architecture is composed of these two sub-networks, which treat the input $\mathbf{\lambda}$ and $x$ differently, thus it is consistent with prior knowledge. In addition, there is no need to discrete the domain to approximate the solution space, and the evaluated solution functions are defined in the whole domain. The only condition required is that the sensor locations  $[s_{1},s_{2},...,s_{m}]$ should be consistent for the training dataset. There also are some generalisations which encode the input functions to the branch net by a feature vector, for example, we can use the coefficients of $\lambda$ with respect to some chosen basis as input.

	Once the DeepONet is trained, it is easy to see that the numerical solutions will lie in the linear space $\text{Span}\{t_{1}(x), ..., t_{p}(x)\}$, i.e., $\{t_{k}(x)\}_{k=1}^{p}$ are actually the trained basis to approximate the solution space, and $\{{b_{k}(\mathbf{\lambda})}\}_{k=1}^{p}$ are the corresponding coefficients. Since usually, $p$ is not large, e.g. around $100$ for our $2$ dimensional problems, the DeepONet can be regarded as a model reduction method in which the reduced basis is obtained by training. 
	
	\subsubsection{The algorithm}
	Based on the  architecture of the DeepONet, the 
	parameters can be optimised by minimising the following mean square error loss:
	\begin{equation}
		L(\Theta) = \frac{1}{NM}\sum_{i=1}^{N}\sum_{j=1}^{M}|\mathcal{F}(\lambda_{i})(x_{i,j})-G_{\Theta}(\lambda_{i})(x_{i,j})|^{2}= \frac{1}{NM}\sum_{i=1}^{N}\sum_{j=1}^{M}|u_{i}(x_{i,j})-\sum_{k=1}^{p}b_{k}(\lambda_{i})t_{k}(x_{i,j})|^{2} \ .
	\end{equation}
	There are two important hyper-parameters which should be determined before training: the positive integers $m$ and $p$, i.e., the number of sensors for encoding the input functions and the number of the basis used to approximate solutions. Larger $m$ and $p$ means smaller encoding error and reconstruction error, respectively. However, increasing them usually does not necessarily reduce the total error due to the increased complexity of the optimisation problem. The locations of the sensors are not necessarily equispaced. However, they should be consistent with the training dataset.
	
	Note that a pair of the training data in DeepONet is $\{\lambda_{i},x_{i,j},u_{i}(x_{i,j})\}$, thus it is slightly different from most neural operator methods, i.e. the full field observation of solutions is not necessary for the DeepONet and the DeepONet can work with only partially observed solutions. This is of particular importance for real applications where more often than not the available data is not complete. In addition, for different types of data sets, we can use different implementations for DeepONet to dramatically reduce the computational cost and memory usage by orders of magnitudes. For example, different branch input $\lambda$ may share the same trunk net input $x$, and different input $x$ may share the same input $\lambda$,  see \cite{FAIR_karniadakis} for details. This computation technique can also be applied to the extension PI-DeepONet \cite{wang2021learning} when computing the derivatives of the output functions. 
	
	\begin{algorithm}[t]
		\caption{Fast implementation of DeepONet}
		\label{AlgorithmDeepONet}
		\KwIn{ data: $\{(\lambda_{i},x_{j},u_{i}(x_{j}))|i=1,2,...,N;j=1,2,...,M\}$\\
			Number of sensors: $m$, and the locations of the sensors: $S=(s_{1},...,s_{m})$\\
			Let $\mathbf{\lambda}=[\lambda_{1}(S),...,\lambda_{N}(S)]\in \mathbb{R}^{N\times m}; X = [x_{1},...,x_{M}]\in R^{M}, U=[u_{1},u_{2},...,u_{M}]\in \mathbb{R}^{N\times M}$\\
			$p$: the number of basis to be learned.\\
			$\tau_{b}/\tau_{t}:$ learning rates of the branch net and trunk net}
		Initialise the branch net $B_{\Theta_{b}}$ and trunk net $T_{\Theta_{t}}$\\
		\While{\textnormal{not converged}}
		{$B = B_{\Theta_{b}}(\lambda)\in \mathbb{R}^{N\times p}$ and $T = T_{\Theta_{t}}(X)\in \mathbb{R}^{M\times p}$\\
			Output $= BT^{T}\in \mathbb{R}^{N\times M}$\\
			$L(\Theta)=\dfrac{1}{MN}\|U-\text{Output}\|$\\
			Update $\Theta \leftarrow \Theta - \tau \nabla_\Theta L$\\
			\tcc{The fast algorithm can be easily extended to  the mini-batch case and the computation of the derivatives in PI-DeepONet.}}
	\end{algorithm}
	
	\subsubsection{Generalisations of DeepONet}
	Several extensions of DeepONet have also been developed. 
	\begin{itemize}
		\item In \cite{FAIR_karniadakis}, a feature expansion for the trunk net input is proposed in order to satisfy some desirable properties of the output function, e.g., oscillating structures
		or decay properties. Feature expansion for the branch net can also be used to incorporate a feature which is a function of $x$. E.g. the POD-DeepONet as proposed in \cite{FAIR_karniadakis} precomputes a basis by performing proper orthogonal decomposition (POD) on the training data. Thus, POD-DeepONet shares the same idea with PCANN for the output space. Employing such feature vectors might be particularly advantageous for non-smooth parameters or solutions, i.e.,
		discontinuous or highly oscillatory functions.
		\item The Bayesian B-DeepONet \cite{lin2021accelerated} uses the Bayesian framework of replica-exchange Langevin diffusion to enable DeepONets training with noisy data. The B-DeepONet and Prob-DeepONet as proposed in \cite{moya2022deeponet} were shown to have good predictive power along with uncertainty quantification. In \cite{garg2022variational}, a further generalization 'Variational Bayes DeepONet' was introduced, in which the weights and biases of the neural
		network are treated as probability distributions instead of point estimates, and their prior distributions are updated by Bayesian inference. 
		\item A universal approximation theorem of continuous multiple-input operators was proved and the corresponding MIONet was proposed to learn multiple-input operators in \cite{jin2022mionet}. Similarly, The authors in \cite{tan2022enhanced} proposed a new enhanced DeepONet, in which multiple input functions are represented by multiple branch DNN sub-networks, which are then combined with an output trunk network via inner products to generate the output of the whole neural network. 
		\item Several extensions for dealing with particular function properties were recently proposed. E.g., a multi-scale DeepONet \cite{liu2021multiscale} was proposed to approximate a nonlinear operator between Banach spaces of highly oscillatory continuous functions. The shift-DeepONets proposed in \cite{hadorn2022shift} extends Deep Operator Networks for
		discontinuous output functions by elevating the linear basis expansion of the classical DeepONet architecture to a
		non-linear combination. This can make the basis functions themselves dependent on the input function by exposing explicit shift and scale parameters. In the paper \cite{sun2022deepgraphonet}, the authors introduce the Deep Graph Operator Network, a combination of DeepONet and Graph neural networks, by using GNN-Branch Net to exploit spatially correlated graph information. The authors in \cite{moya2022fed} developed a framework named Fed-DeepONet to allow multiple clients to train
		DeepONets collaboratively under the coordination of a centralised server.

		\item The Variable-input Deep Operator Network (VIDON) \cite{prasthofer2022variable}, is peculiar in that it allows for sensor points to be queried from any point in the domain during training. In this way, no prior discretisation of the domain is needed. 
	\end{itemize}
	
	\subsubsection{Theoretical Background}
	
	In addition to the universal theorem, several analytic results have been published for  DeepONet. In the original paper, a theoretical analysis was presented which allows estimation of the approximation properties for ODE operators with respect to an underlying probability distribution. The analysis depends on the number of sensors and the related approximation of the input functions.

	In \cite{lanthaler2022error}, the universal approximation property of DeepONets was extended to include measurable mappings in non-compact spaces. By decomposition of the error into encoding, approximation and reconstruction errors, both lower and upper bounds on the total error were derived, relating it to the spectral decay properties of the covariance operators associated with the underlying measures. For four prototypical examples of nonlinear operators, it was proved that DeepONets can break the curse of dimensionality. 
	
	In \cite{deng2021convergence} the convergence rates of the DeepONet were considered for both linear and non-linear advection-diffusion equations with or without reaction. The conclusion is that the convergence rates depend on the architecture of the branch network as well as on the smoothness of the inputs and outputs of the operators. The paper \cite{gopalani2022capacity} gives a bound on the Rademacher complexity for a large class of DeepONets.
	%\section{Physics-Informed DeepONet}
	
	%\section{Physics-Informed Neural Operators?}
	
	\section{Numerical experiments}\label{sec:Numerical experiments}
	The core of this paper is a numerical comparison of the described methods. The aim is to develop a guideline for scientists who want to start working with DL methods for PDE-based problems and face the problem of determining suitable methods.

	We will perform numerical experiments on different levels.  First of all, we will investigate the performance for solving the forward Poisson problem. This linear problem is most commonly used and -despite its limited value for generalisations to non-linear problems - already yields some insight into the performance of the chosen methods.
	We then perform tests for the inverse Poisson problem, where the source term is the sought-after parameter.
	
	After that, we turn to analyse the behaviour of the chosen methods for forward and inverse Darcy flows. A particular emphasis is on evaluating and, comparing the respective numerical schemes in terms of accuracy but also computational time and storage needed for parametric studies.

	After these basic numerical tests, we do report on how learned PDE solvers behave for two industrial projects.

	We want to acknowledge that there already exists a growing list of publications devoted to testing DL concepts for the solution of PDEs, see e.g. \cite{FAIR_karniadakis, goswami2022pidnon, grossmann2023can}. However, these studies focus on PDE-based forward problems, in contrast, our experiments for comparing forward solvers is a preliminary step towards our main goal, namely a comparison of  DL methods for PDE-based parameter identification problems.
	
	We start this section with an outline of our data generation procedure, which follows \cite{bhattacharya2020model, li2020neural, li2020multipole}.
	In summary, this section is organised as follows
	\begin{itemize}
		\item Generating training and test data
		\item Poisson problem
		\begin{itemize}
			\item Forward problem
			\item Inverse problem trained with exact data
			\item Inverse problem trained with noisy data
		\end{itemize}
		\item Darcy flow
		\begin{itemize}
			\item Forward problem
			\item Inverse problem trained with exact data
			\item Inverse problem trained with noisy data
		\end{itemize}
		\item Industrial applications
		\begin{itemize}
			\item A moving domain problem (Bosch)
			\item Identification of material model parameters from stress tensor (Volkswagen)
		\end{itemize}
	\end{itemize}

	\subsection{ Generating training and test data} \label{subsec:data_gen}
	The outline of our numerical experiments for the linear differential equations $\Delta u =  \lambda$ on $\Omega$, $u=g$ on $\partial \Omega$ (Poisson problem) as well as simulations for the Darcy flow follows the approach of \cite{bhattacharya2020model}. Hence, we do a performance analysis in terms of accuracy as well as computational load and the parameter $\lambda$ is randomly generated as a Gaussian random field. As a gold standard for comparison, we use a standard FDM code, which is tuned to high precision and also provides the ground truth data for training the networks.
	
\begin{figure}[htp]
		\centering
		\includegraphics[width=0.8\linewidth]{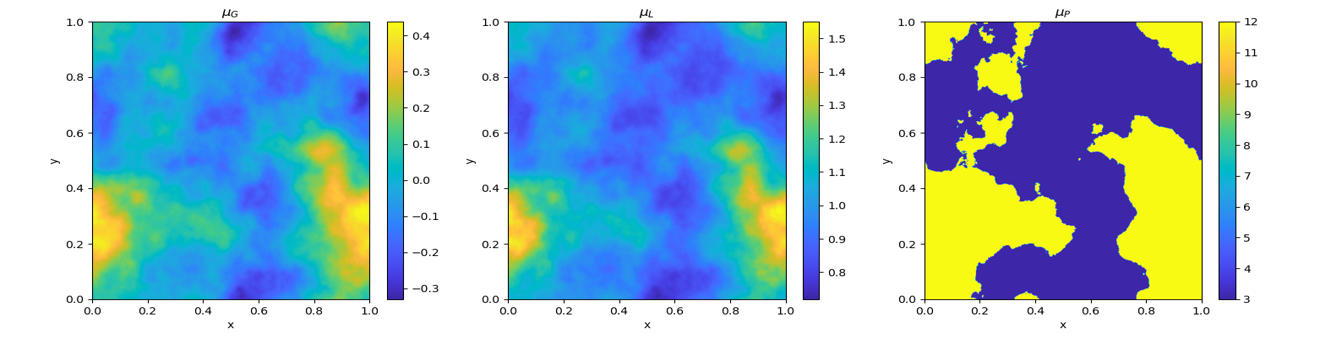}
		\caption{Example of samples from the GRF probability measures $\mu_G$, $\mu_{\mathrm{L}}$ and $\mu_{\mathrm{P}}$. }\label{fig:grf-samples}
	\end{figure}

	Learning operators implies learning mappings between function spaces. For a supervised learning task, we need input-output data pairs for training, which in our case are the parameter and solution functions. We start with Gaussian Random Fields, commonly used in the stochastic modelling of physical phenomena as e.g. described in \cite{bhattacharya2020model, li2020neural, li2020multipole, FAIR_karniadakis}; more precisely we use as base measure the Gaussian distribution $$\mu_{\mathrm{G}}=\mathcal{N}\left(0,(-\Delta+9 I)^{-2}\right),$$ with a zero Neumann boundary condition on the operator $\Delta$, which yields random but smooth test data for the Poisson problem. Other interesting measures are $\mu_{\mathrm{L}}$ and $\mu_{\mathrm{P}}$ which are the push-forwards of  $\mu_{\mathrm{G}}$ under the exponential and piece-wise constant maps respectively, so that $\mu_{\mathrm{L}}=\exp _{\sharp } \mu_{\mathrm{G}}$ and $\mu_{\mathrm{P}}=T _{\sharp } \mu_{\mathrm{G}}$, where $\exp_{\sharp }$ and $T_{\sharp}$ represent the respective push-forward functions, with \begin{eqnarray*}
		%\exp(s) &=& e^{s}           & \forall s \in \mathrm{R}\\
		T(s)    &=& \begin{cases}12 & s \geq 0 \\ 3 & s<0\end{cases}
	\end{eqnarray*}
	For our experiments with the Darcy flow, we used piece-wise constant parameters. 
	We show some examples in Figure \ref{fig:grf-samples}.
	
\begin{figure}[htp]
		\centering
		\includegraphics[width=0.6\linewidth]{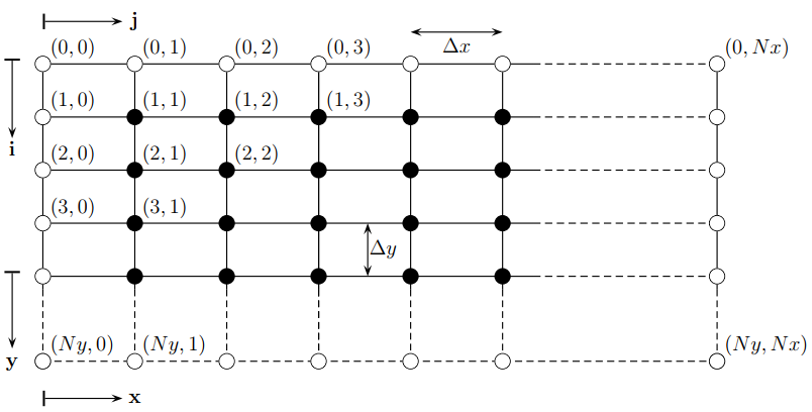}
		\caption{Computational grid showing interior grid points (black) and boundary grid points (white).}
		\label{fig:discretisation_FDM}
	\end{figure}

	To complete the data pair, we use the generated samples as right-hand sides of the Poisson problem and use a finite difference method (FDM) to get the solution for the considered problem: As an example, we consider the Poisson problem given by 	
	\begin{equation}%\label{Poisson_basic}
		\begin{aligned}
			-\Delta u(s) &=\lambda (s) & & s \in (0,1)^{2} \\
			u(s) &=0 & & s \in \partial(0,1)^{2}, 
		\end{aligned}
		\label{eqn:poisson}
	\end{equation}
	where the forcing term $\lambda(s)$ is sampled from the GRF $\lambda \sim \mu_{\mathrm{G}}$. Then, the domain $(0,1)^{2}$ is discretised as shown in Figure \ref{fig:discretisation_FDM}, and a second-order FDM scheme is used to evaluate the Laplacian, reducing the Poisson equation to a system of linear equations given by Equation \ref{eqn:poisson_FDM}, where $i = 0, \ldots N_y$ and $j = 0, \ldots N_x$ for a resolution of $(N_x + 1) \times (N_y + 1) $
	%Algorithm \ref{alg:poisson_FDM_algo} shows how we solve this system of equations to get $u$ in $(0,1)^{2}$
	
	\begin{eqnarray}
		\frac{\Delta y}{\Delta x}\left(-u_{i, j-1}+2 u_{i, j}-u_{i, j+1}\right)+\frac{\Delta x}{\Delta y}\left(-u_{i-1, j}+2 u_{i, j}-u_{i+1, j}\right)=\Delta x \Delta y \lambda\left(x_{j}, y_{i}\right)
		\label{eqn:poisson_FDM}
	\end{eqnarray}
	
	A similar discretisation scheme is used for the Darcy Flow equation \ref{eqn:darcy} which is equally of interest to us in this work.
	\begin{equation}
		\begin{aligned}
			-\nabla \cdot(\lambda(s) \nabla u(s)) &=f(s) & & s \in(0,1)^{2} \\
			u(s) &=0 & & s \in \partial(0,1)^{2}
		\end{aligned}
		\label{eqn:darcy}
	\end{equation}
	
	Our numerical tests are done using baseline implementations of the mentioned DL algorithms and we did some extensive hyperparameter search for every method. For details see the appendices. Nevertheless, hyperparameter search and fine-tuning the optimisation scheme in search of global minima is a never-ending story. Hence, the presented results should be understood as the best we could achieve within the given time and with the available computing capacity (specified in \ref{apdx:num_sett}). All methods were treated equally and we believe, that this leads to a fair basis for comparison.
	
	%DERICK Hardware computing hours (Resp: Available in the appendix and now referred to in the previous sentence)
	
	In addition, we are well aware of the extensive body of literature dealing with sometimes rather refined extensions and improvements of these methods, e.g. there are several extensions of the PINN approach and 
	a survey on different Physics-informed neural operator networks variants has been published recently, \cite{goswami2022pidnon}. However, these extensions most often come with the necessity to fine-tune additional hyperparameters or they do apply only to special cases.
	Hence, as said before we focus on a comparison of the basic implementations in this section.
	To give proper credit to the different approaches, we assigned different co-authors to different concepts and everybody was doing his best to ‘defend’ the respective concepts.
	
	So far we have addressed the general outline of our testing scenario and our data generation for the academic test examples. This is in line with procedures used by our collaborating industrial partners. Research and development departments in the industry also seem to rely mostly on simulated data for during the first development cycles for new products.   In contrast to simulated data, real-life data for industrial applications typically is very scarce. Almost often this is only available in very limited numbers. This leads to the problem of data enrichment and data augmentation, which is outside the scope of the present paper. Hence, the two industrial examples included in this survey only use simulated data, which were obtained at least partially by the software of our collaboration partners.

	Furthermore, let us clarify the criteria for our evaluation. There are several obvious categories such as achievable mean accuracy, training time, and time for solving the forward Poisson problem after training, but also degrees of freedom of the networks used, stability concerning hyperparameter tuning, etc. Nevertheless, in the following, we will mainly discuss two criteria, namely accuracy and the potential for parametric studies/inverse problems, i.e. the computational time needed to run the algorithm after training for novel sets of parameters.
	
	\subsection{Poisson problem}
	\subsubsection{Forward problem}
	Testing different DL concepts for training a {\bf forward solver} for the Poisson problem (\ref{eqn:poisson}) is the most basic academic example which is used by almost all relevant publications in this context. This linear problem can be solved by all other methods considered and the errors are as reported in Table \ref{tab:errors-513-poisson}. These errors do not differ too much, they are below $1 \%$ relative error except for DRM, which has an error of approximately $2,5 \%$. Still, there are some notable differences. The overall winner is PCALin both in terms of achievable accuracy as well as in terms of complexity of the network. However, none of the methods was able to achieve an accuracy comparable with fine-tuned FDM or FEM methods. The advantage of DL concepts, in this case, lies in the potential for large-scale parametric studies, where the execution time for testing novel parameters has to be minimal.
	
	However, not all DL concepts are suitable for parametric studies. In general DL concepts aiming for a function approximation do require expensive retraining of the network for every additional parameter and do have limited value for large-scale parameter variations. 
	Hence, we focus on DL concepts for operator approximations, see Table \ref{tab:methods}. In our experiments, DeepONet and their physics-informed versions performed best in terms of efficiency (run time for testing). The differences in testing time for these methods are below the variance introduced by the random sampling of test data.

	The comparison with respect to different resolutions of $u$ does show the expected, but interesting result, that operator approximations, which are based on a functional analytic reasoning in function spaces rather than discrete settings, indeed show an independent accuracy across scales, see Table \ref{tab:errors-poisson}. This cannot be matched by function approximations where the error increases by several orders of magnitude for coarse discretisations.
	
	These numerical findings for the linear Poisson problem in a standard domain do have little value for generalisations to other PDE problems. Nevertheless, as a punch line for simple linear PDE problems, we would stress the old saying ‘keep it simple’ and suggest applying ‘easy to use’ concepts such as PCALin with a suitable but comparatively small network. 
	
	PINNs do also offer an easily accessible concept, which is easy to adapt to other PDE, but it requires retraining, i.e. PINN in their original version are not suitable for parametric studies, and they did exhibit a slightly larger $L_2$-error in our scenario.
	
	As an alternative, MWT requires a more advanced implementation but seems to be more efficient for parametric studies, however, training times are rather high.
	
	% || Resp: it should be PINN
	
	\subsubsection{Inverse problem}

	% These numerical findings for the linear Poisson problem in a standard domain do have little value for generalisations to other PDE problems. Nevertheless, as a punch line, we would stress the old saying ‘keep it simple’ and would suggest applying ‘easy to use’ concepts such as PCALin with a suitable but comparatively small network for simple linear PDE problems. FNO seem to be more efficient for parametric studies, but training times are rather high.
	% PINNs do also offer an easily accessible concept, which nicely adapts to other PDE, but it requires retraining, i.e. PINN in their original version are not suitable for parametric studies, and they did exhibit a slightly larger $L_2$-error in our scenario.

	The investigation of the performance of DL concepts for PDE-based inverse problems for parameter identification is the core of this survey. In this subsection, we summarise the results for the inverse Poisson problem, i.e. determining $\lambda$ in Equation \ref{eqn:poisson0} from a measured version of $u$.
	Iterative methods for solving such inverse problems as well as parametric studies require multiple evaluations of the parameter-to-state map $F$.
	As already mentioned, DL concepts based on operator approximations are better suited in this context and we will focus on these methods. 
	The only exception is the extension to inverse problems of PINNs and QRES as described in Section \ref{pinns-param-iden}
	
	Our tests for PDE-based inverse problems are organised in terms of how we attack the inverse problem and which data is used for training.
	First of all, there are two primary approaches for dealing with inverse problems, see Section \ref{subsec:Tikhonov_inverse}, i.e. we can either train the inverse problem with a reversed input-output structure or we can integrate a learned forward solver in a Tikhonov approach.
	Secondly, we can train the network with either noiseless data or noisy data with different noise levels. Both tests are meant for applications depending on whether some clean data obtained in research labs or measured data from field experiments are available. As it is common for inverse problems, after training these networks will be evaluated for reconstruction problems with noisy data. The case of testing with noise-free data is included for reason of completeness.
	
	Training is always done with 1000 training samples, where $\lambda$ is computed as a smoothed random Gaussian field, see Section \ref{subsec:data_gen}. The parameter-to-state map, $u=F(\lambda)$, is then computed with a high-precision finite difference scheme. The resulting solution $u$ is then perturbed with normally distributed random noise of different levels, $\delta$.  
	$$u^\delta (x) = u(x) + \delta \cdot ||u|| \cdot N(0,1)$$
	
	For the evaluation of the methods we use additional  $N= 5000$ samples of $u^\delta$, which are computed by the same procedure, i.e. drawing
	additional random samples for $\lambda_{true}$,  computing the corresponding solution $u$ and adding noise.

	These perturbed solutions are the input for the inverse problem during evaluation.
	The resulting estimation of the parameter $\hat \lambda$ is then compared with the original, true parameter $\lambda_{true}$. 
	As a standard measure of success we average over the different evaluation samples and take the mean $L_2$-error, $E \left( \|\hat \lambda - \lambda_{true}\| \right)$. 
	
	For certain applications, i.e. control problems the output error is also of importance and for some experiments we also report the difference between the solution obtained with $\hat \lambda$ and the true, unperturbed solution $u$ as shown in Tables \ref{tab:unoisy-noise-poisson-65} and \ref{tab:unoiseless-noise-poisson-65}.
	
	Let us comment on the different test scenarios (inverse learning or Tikhonov, noise-free or noisy data for training) in more detail.

	The tests with unperturbed data, i.e. the case where the networks were trained and evaluated with perfect noise-free data are reported in Table \ref{tab:errors-513-poisson} and Table \ref{tab:errors-poisson} for different resolutions. For operator approximation methods the inverse problem in these tables is always solved by inverse learning (backward operator training). Also, PINN and QRES  can be extended to parameter learning directly by a doubling of the network, see Section \ref{pinns-param-iden}.
	This results in a doubling of the network parameters as shown in the second column of both tables. As a general observation, we remark that the achievable accuracy for the inverse problem is considerably below the accuracy of the forward problem, which reflects the ill-posedness of the inverse problem. Also, the baseline implementations of PCANN, PINN and DeepONet do perform less reliably as compared with their extensions such as PCALin, U-FNO, PINO or MWT. After training run times of the different methods are the same as for the forward problem.
	
	Due to the linear nature of the problem and the missing noise in the data, one should not overestimate the value of these tests. In particular testing with noisy data is essential for inverse problems.
	
	To this end, we have done experiments where the networks were trained with noiseless data but the evaluation was done with noisy data, see Table \ref{tab:noise-backward-operator-65}. We clearly see the networks trained with noise-free data do not generalise to the case of noisy data. The error exceeds $100 \%$ in most cases.
	The only notable exception is surprisingly DeepONet and PI-DeepOnet.  Even for high noise levels, this method still produces errors in the reconstructed parameter which are approx. 4 – times larger than the noise in the data, which is the range of errors one would expect for optimal regularisation schemes for ill-posed problems.
	
	Tables \ref{tab:noise-backward-operator-trained-noise-65} and \ref{tab:xxnoise-poisson-65} are the most meaningful ones for inverse problems. Here we report results on networks, which are trained and evaluated with noisy data. Table \ref{tab:noise-backward-operator-trained-noise-65} reports results obtained by reversed/backward operator training, i.e. the network is trained to take $u^\delta$ as input and to return directly an estimate for the parameter $\lambda$.
	Table \ref{tab:xxnoise-poisson-65} reports results obtained by first training a forward network $\Phi_\Theta (\lambda)$ and then computing an approximation to the parameter by solving a Tikhonov minimisation problem
	$$ \min_\lambda \ \|\Phi_\Theta (\lambda) - u^\delta\|^2 + \alpha R(\lambda) \ .$$
	In these experiments, $R$ was always taken as a discretisation of the $L_2$-norm and $\alpha$ was optimised by numerical experiments. Depending on the problem, one could add an additional regularisation term as shown in Equations \ref{eqn:TikLoss1} and \ref{eqn:TikLoss2}.

	Both approaches, i.e. inverse learning and embedding a learned forward operator into a Tikhonov scheme do give comparable results, see Tables
	\ref{tab:noise-backward-operator-trained-noise-65} and \ref{tab:xxnoise-poisson-65}. All errors in the reconstructions are naturally larger than the error in the input data, as is to be expected for ill-posed problems.  Given the small differences achieved by the different methods, it is challenging to suggest a particular method. However, in our opinion, the range of errors $1 \%$ or $5 \%$ might be most important for applications and a closer look at the numbers reported in Table
	\ref{tab:noise-backward-operator-trained-noise-65} for inverse training of the inverse Poisson problem  shows that FNO and PINO seem to perform best, with PCALin and MWT as runner-ups.
	For Tikhonov-based training and giving higher importance to larger error rates of $1 \%$ or $5 \%$, the best results are achieved by MWT with PINO as runner-up.  
	Hence comparing the winners of either approach (FNO/PINO and MWT) for this range of errors yields a slight advantage for using the Tikhonov approach in connection with the MWT concept.
	However, we should remark that the Tikhonov approach requires solving a minimisation problem for each new data set and is computationally  more expensive.

	Finally, Tables \ref{tab:unoisy-noise-poisson-65} and \ref{tab:unoiseless-noise-poisson-65} are based on experiments, where the reconstructed parameter $\hat \lambda$ was used to compute the corresponding solution $\hat u = F (\hat \lambda)$.
	This is then compared with either the true solution $u$, see Table \ref{tab:unoiseless-noise-poisson-65}), or the noisy data, which was used for training, see Table \ref{tab:unoisy-noise-poisson-65}). This is considered as a test for how well these parameter estimation problems can be used to solve control problems, which however is not the focus of this paper and this case is not investigated further.

	\begin{table}[!ht]\small
		\centering
		%\resizebox{\textwidth}{!}{
			\begin{tabular}{r|c|crc|crc} 
				\multicolumn{2}{c|}{} & \multicolumn{3}{c|}{\textbf{Forward Problem}} & \multicolumn{3}{c}{\textbf{Inverse Problem}}\\
				\cline{2-8}
				\multirow{2}{*}{\textbf{Networks}} & $\#$ \textit{ of} & \textit{Rel. L2} & \textit{Training} & \textit{Testing} & \textit{Rel. L2} & \textit{Training} & \textit{Testing}\\  
				& \textit{Parameters%\footnotemark
				} & \textit{Error} & \textit{(s/epoch)} & \textit{(s)} & \textit{Error} & \textit{(s/epoch)} & \textit{(s)} \\
				\hlineB{3}\midrule%\hline 
				DRM   & $6,721$ & 0.0251 & 0.0514 & $\sim 480 $& - & - & - \\
				PINN & $5,301~|~10,602$ & $0.0075$ & $0.1034$ & $\sim 1,315 $ & $0.1650$ & $0.2368$ & $\sim 3,370 $ \\
				QRES  & $5,509~|~11,018$ & $0.0076$ & $0.1581$ & $\sim 2,150 $ & $0.1549$ & $0.4186$ & $\sim 5,730 $ \\
				%WAN & & & & & & &  \\
				\midrule%\hline \hline
				PCANN & $5,155,150~|~5,205,200$ & $0.0073$ & $0.0142$ & $0.6063$ & $0.0981$ & $0.0161$ & $0.6179$   \\
				PCALin & $62,750$ & \cellcolor{gray!20}$ \mathbf{0.0013}$ & $0.0077$ & $0.7453$ & $\mathbf{0.0244}$ & $0.0090$ & $0.7640$   \\
				FNO & $2,368,001$ & $0.0066$ & $42.6249$ & $0.0168$ & $0.0415$ & $42.6245$ & $0.0174$  \\
				U-FNO & $3,990,401$& $0.0053$ & $97.6730$ & $0.0346$ & $0.0254$ & $97.6434$ & $0.0343$ \\
				MWT &  $9,807,873$ & $0.0036$ & $114.2043$ & $0.0488$ & \cellcolor{gray!20}$\mathbf{0.0159}$ & $113.2655$ & $0.0483$ \\
				DeepONet & $640,256~|~768,128$ & $0.0042$ & $0.1098$ &$0.0002$ & $0.1035$ &$0.1201$ &   $0.0002$\\
				\midrule
				PINO & $2,368,001$ & $\mathbf{0.0031}$ & $42.6289$ & $0.0166$ & $0.0301$ & $42.8172$ & $!0.0143$ \\
				PI-DeepONet & $640,256~|~739,712$ & $0.0061$ & $0.4637$ & $0.0002$ & $0.1068$ & $0.4462$ &  $0.0002$ \\
				\bottomrule%\hline
			\end{tabular}
			%}
		\caption{Performance of different methods for the Poisson Problem, using a $513 \times 513$ resolution. In the second column, for the forward and inverse problems, if the same amount of parameters is used, only one number is specified. If different amounts are used, two numbers are given. Where the left one corresponds to the forward problem and the right one to the inverse case. The networks were trained with noiseless data and backward operator training was used for solving the inverse problem.} %For each method, but for PINN and QRES (where the parameter is learned together with the solution), the inverse problem specified here is that where the same network architecture is trained with PDE solutions as input and parameter as input. We refer to this as the reverse or backward method in most parts of this work.}
	\label{tab:errors-513-poisson}
\end{table}
%\footnotetext{If for the forward and inverse problem, the same amount of parameters is used, only one number is specified. If different amounts are used, two numbers are given. Where the left one corresponds to the forward problem and the right one to the inverse case.}

\begin{figure}[ht!]
	\begin{subfigure}{0.24\textwidth}
		\centering
		\includegraphics[width=0.51\textwidth]{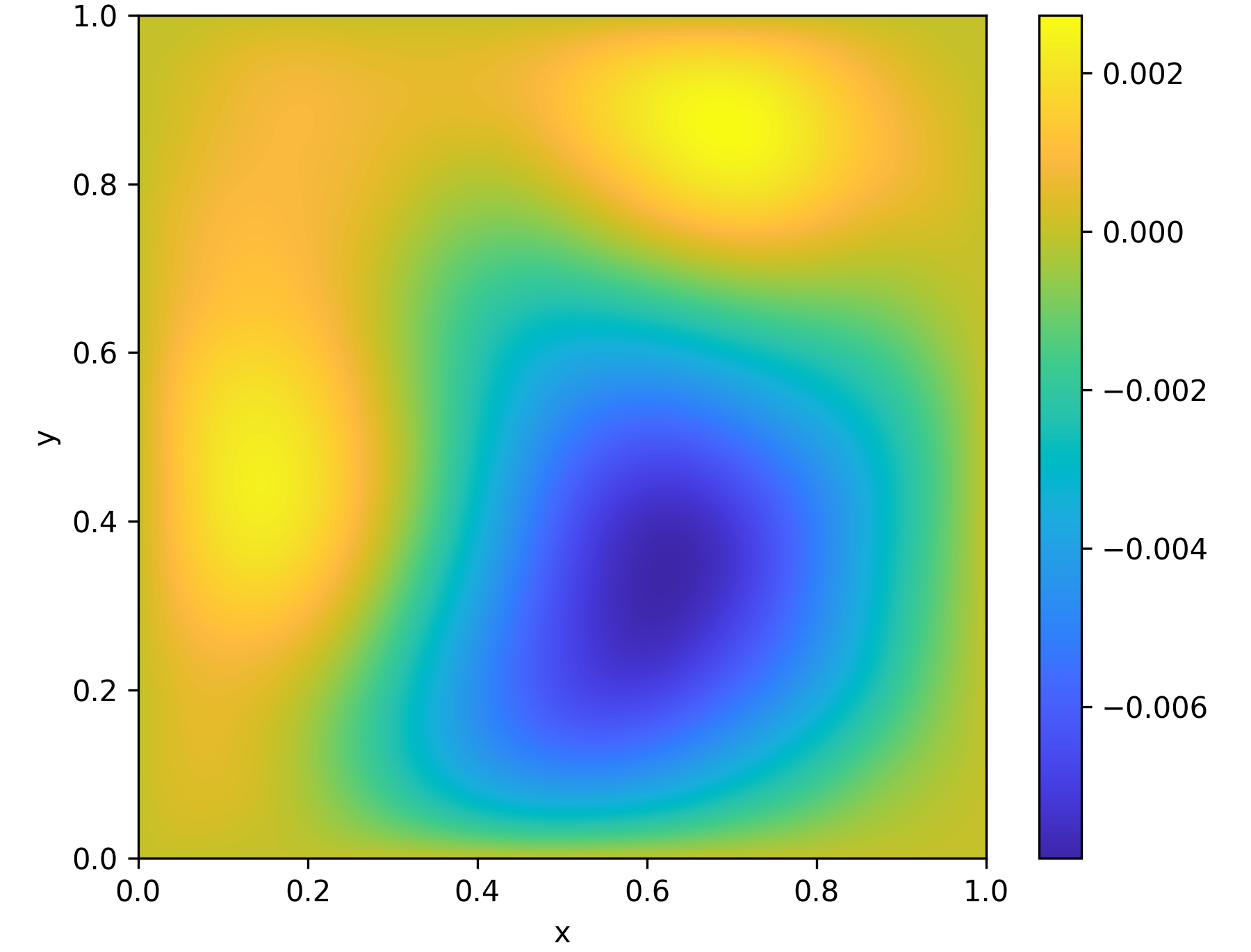}
		\caption{Ground Truth}
		\label{fig:groundtruth-poisson-forward-ug}
	\end{subfigure}\hfill 
	\begin{subfigure}{0.24\textwidth}
		\centering
		\includegraphics[width=\textwidth]{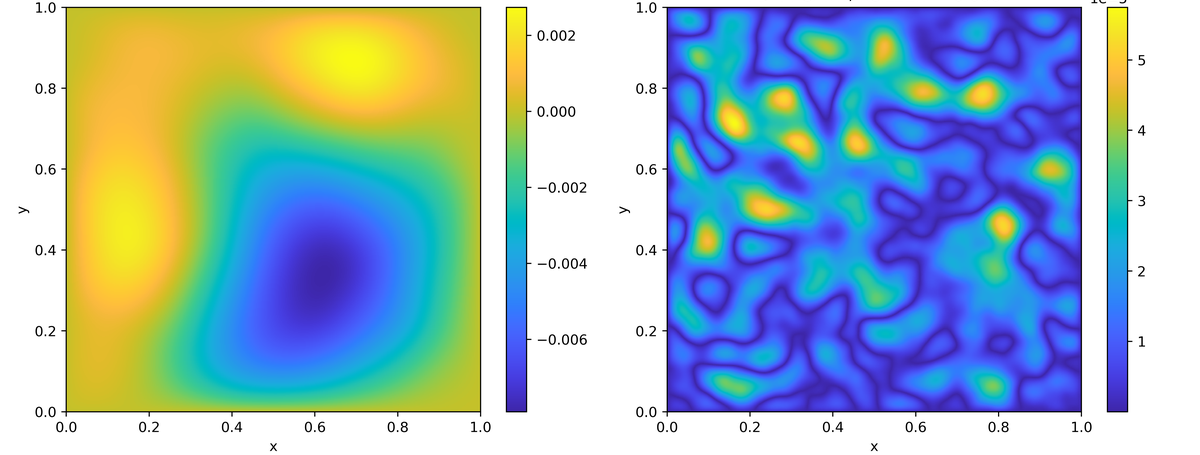}
		\caption{PCANN}
		\label{fig:pcann-poisson-forward-ug}
	\end{subfigure}\hfill 
	\begin{subfigure}{0.24\textwidth}
		\centering
		\includegraphics[width=\textwidth]{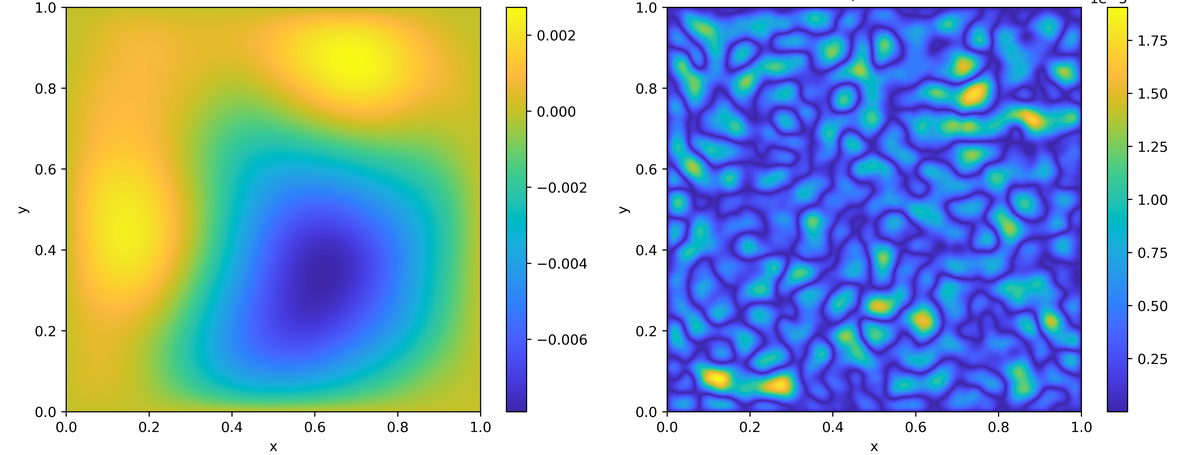}
		\caption{PCALin}
		\label{fig:pcalin-poisson-forward-ug}
	\end{subfigure}\hfill
	\begin{subfigure}{0.24\textwidth}
		\centering
		\includegraphics[width=\textwidth]{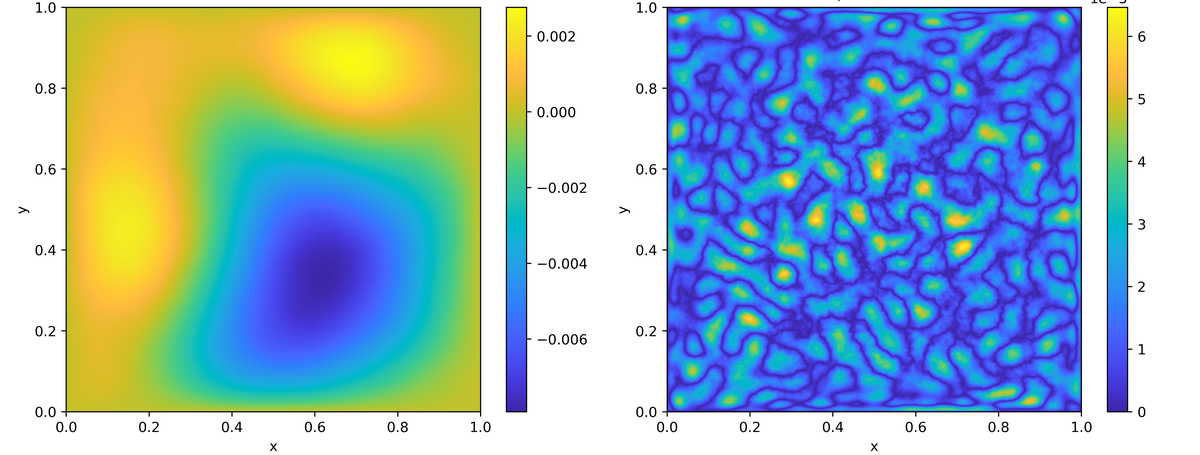}
		\caption{FNO}
		\label{fig:fno-poisson-forward-ug}
	\end{subfigure}\\
	\begin{subfigure}{0.24\textwidth}
		\centering
		\includegraphics[width=\textwidth]{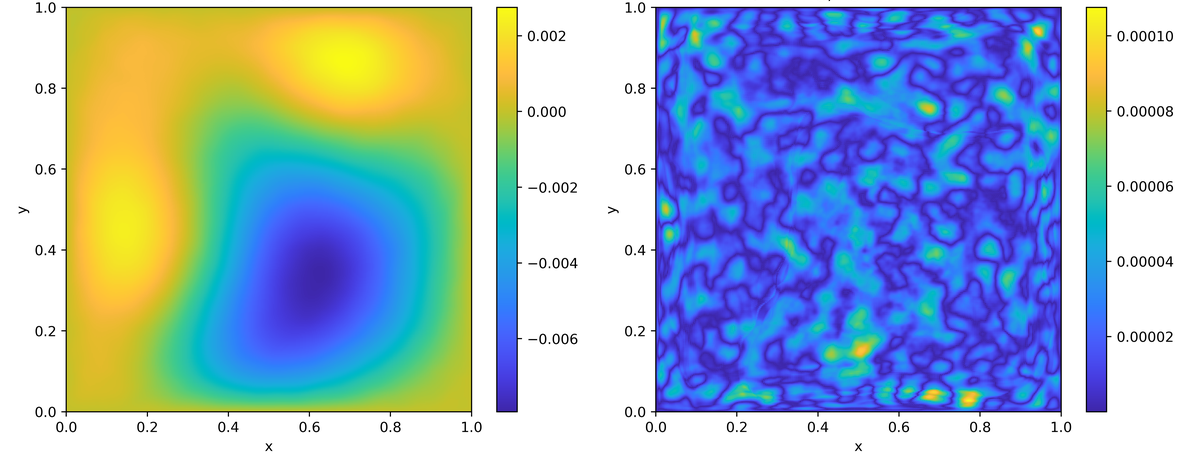}
		\caption{U-FNO}
		\label{fig:ufno-poisson-forward-ug}
	\end{subfigure}\hfill
	\begin{subfigure}{0.24\textwidth}
		\centering
		\includegraphics[width=\textwidth]{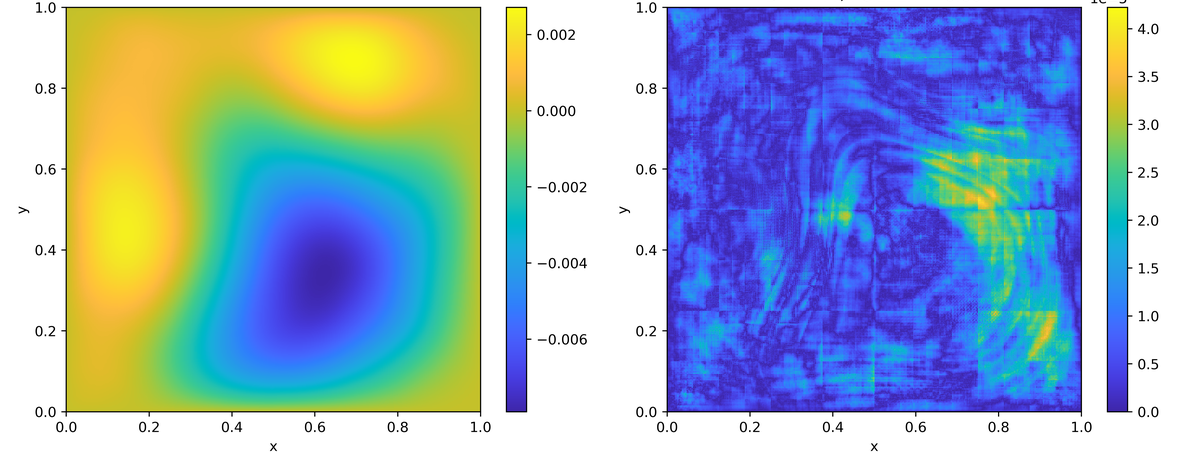}
		\caption{MWT}
		\label{fig:mwt-poisson-forward-ug}
	\end{subfigure}\hfill
	\begin{subfigure}{0.24\textwidth}
		\centering
		\includegraphics[width=\textwidth]{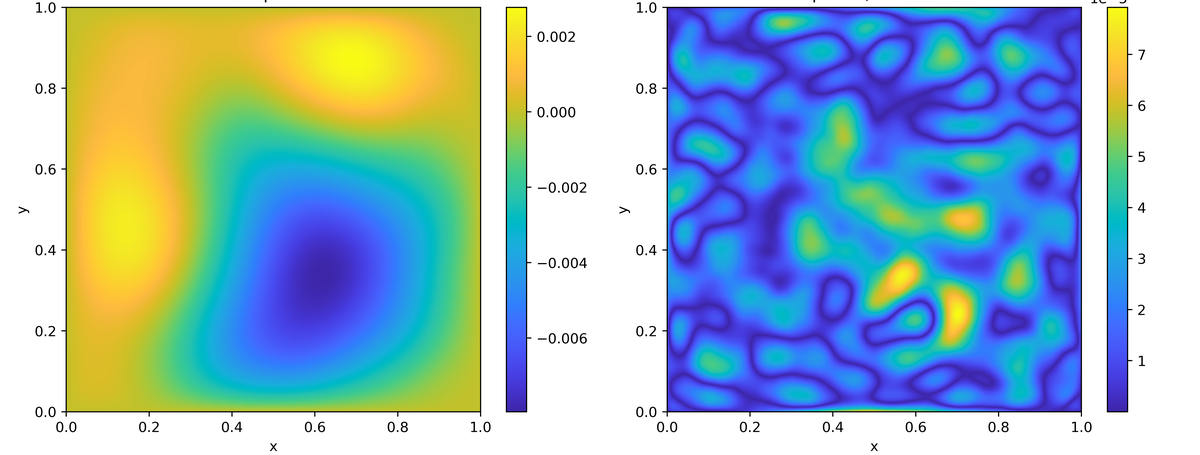}
		\caption{DeepONet}
		\label{fig:deeponet-poisson-forward-ug}
	\end{subfigure}\hfill
	\begin{subfigure}{0.24\textwidth}
		\centering
		\includegraphics[width=\textwidth]{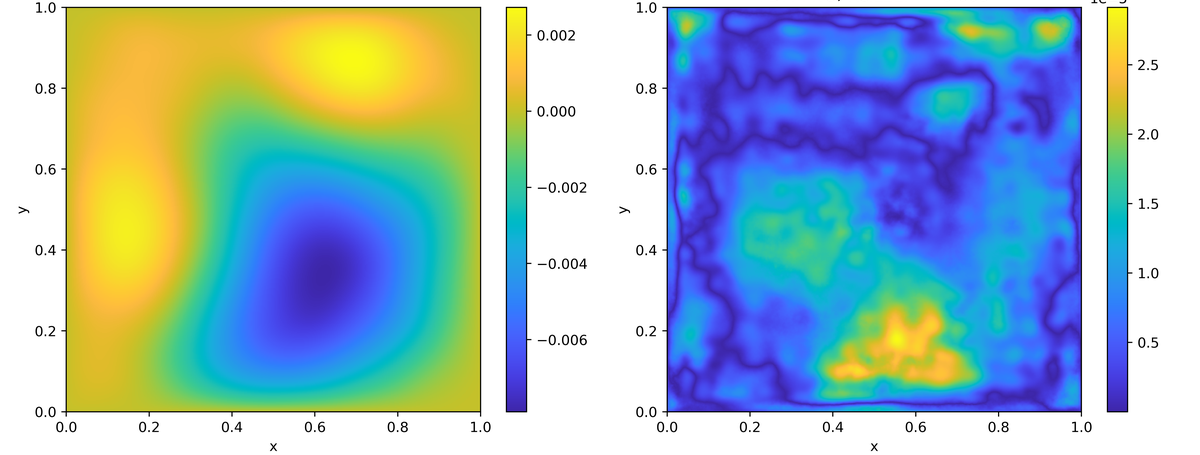}
		\caption{PINO}
		\label{fig:pino-poisson-forward-ug}
		\label{fig:pi-deeponet-poisson-forward-ug}
	\end{subfigure}	\\
	\begin{subfigure}{0.24\textwidth}
		\centering
		\includegraphics[width=\textwidth]{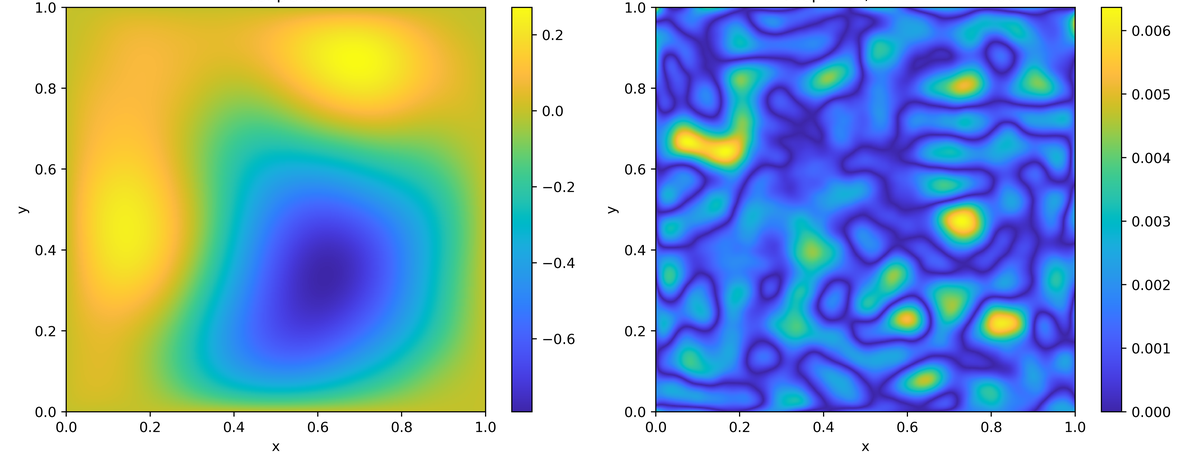}
		\caption{PI-Deeponet}
	\end{subfigure}\hfill
	\begin{subfigure}{0.24\textwidth}
		\centering
		\includegraphics[width=\textwidth]{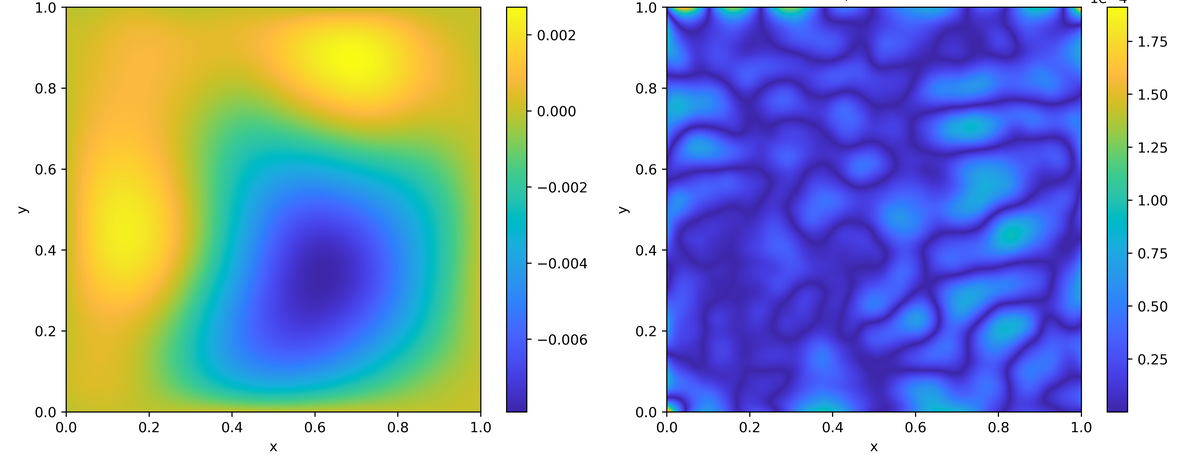}
		\caption{PINN}
		\label{fig:pinns-poisson-forward-ug}
	\end{subfigure} \hfill
	\begin{subfigure}{0.24\textwidth}
		\centering
		\includegraphics[width=\textwidth]{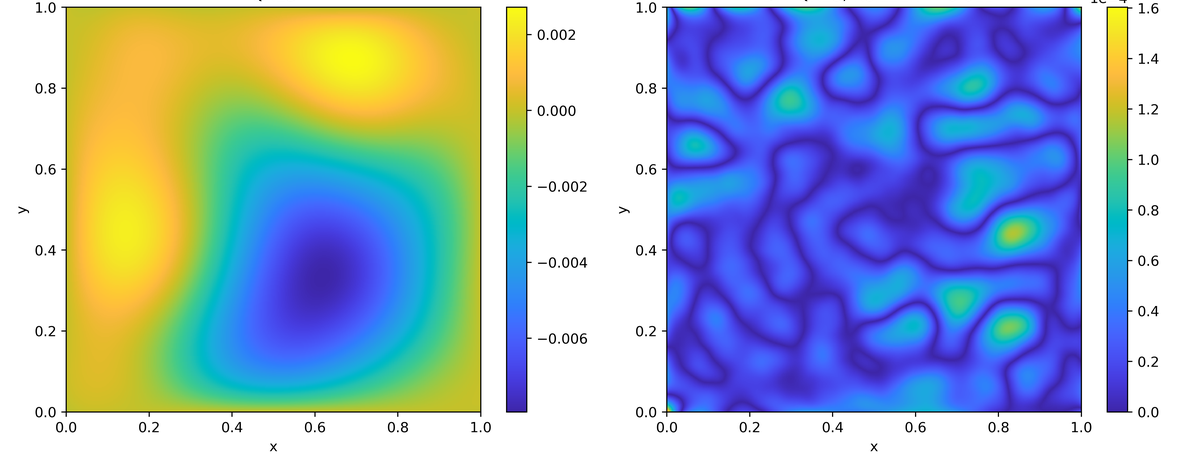}
		\caption{QRES}
		\label{fig:qres-poisson-forward-ug}
	\end{subfigure}\hfill
	\begin{subfigure}{0.24\textwidth}
		\centering
		\includegraphics[width=\textwidth]{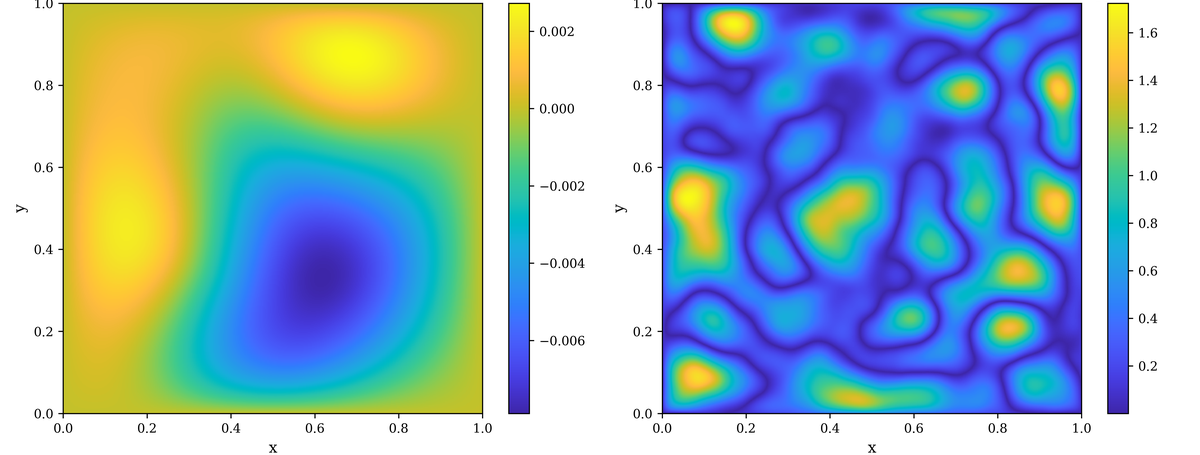}
		\caption{DRM}
		\label{fig:deepritz-poisson-forward-ug}
	\end{subfigure}
	\caption{Test examples for Poisson forward problem using resolution of $513 \times 513$. (b)-(l), shows the specified neural network's approximation of the solution (left-hand side) and the absolute difference between the Ground truth in (a) with the approximation (right-hand side).}
	\label{fig:poisson-forward-ug}
\end{figure}

\begin{figure}[!ht]
	\begin{subfigure}{0.24\textwidth}
		\centering
		\includegraphics[width=0.51\textwidth]{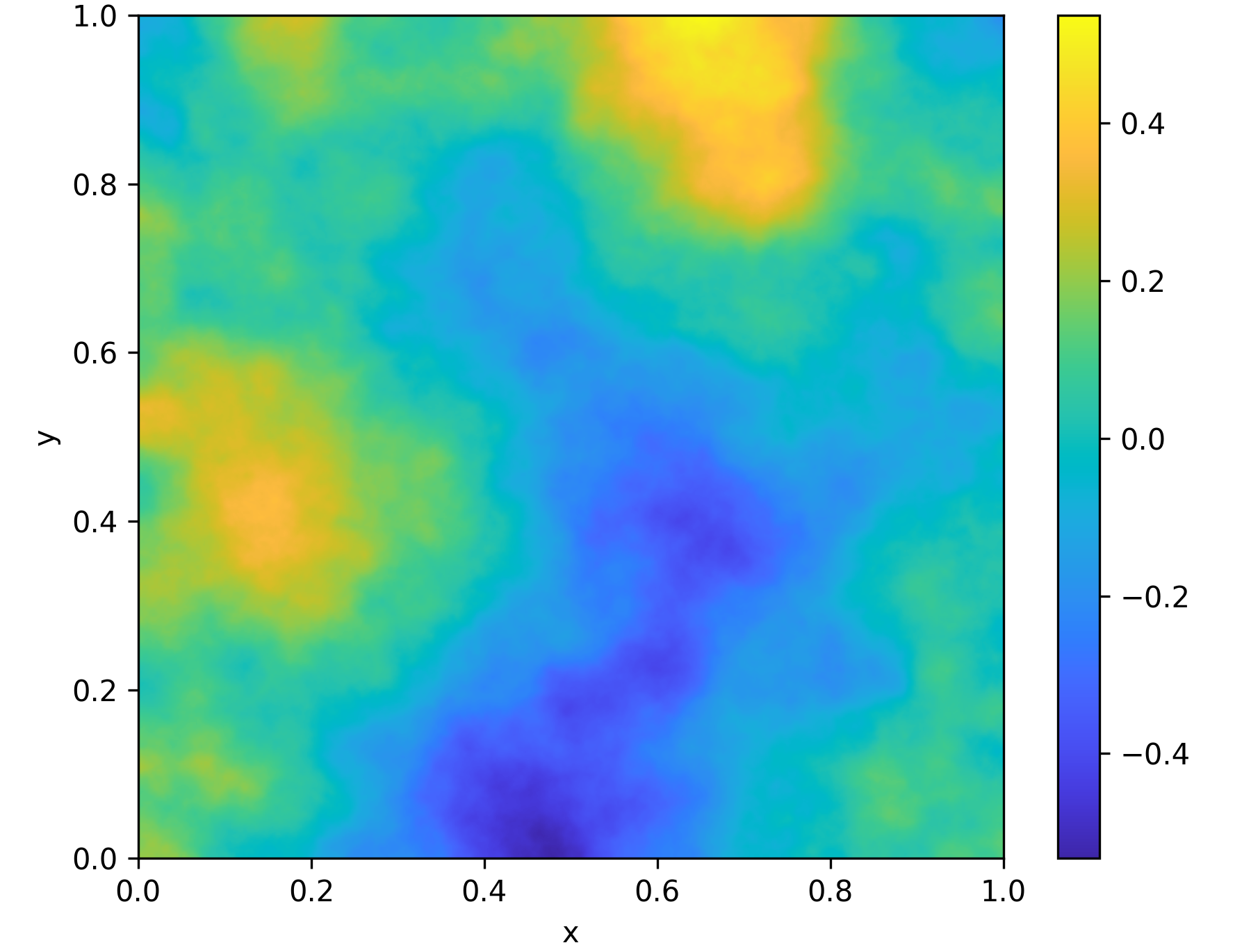}
		\caption{Ground Truth}
		\label{fig:groundtruth-poisson-inverse-ug}
	\end{subfigure}\hfill 
	\begin{subfigure}{0.24\textwidth}
		\centering
		\includegraphics[width=\textwidth]{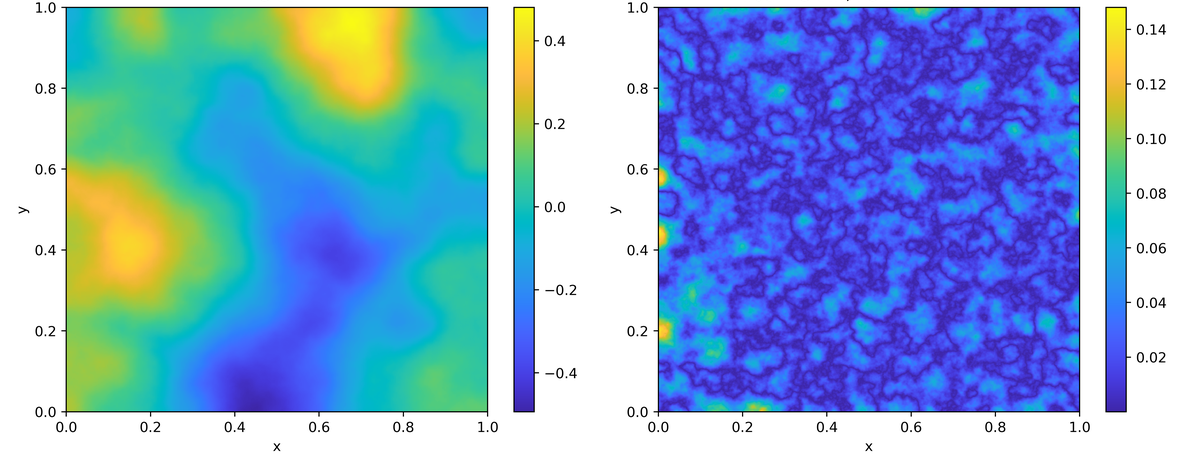}
		\caption{PCANN}
		\label{fig:pcann-poisson-inverse-ug}
	\end{subfigure} \hfill
	\begin{subfigure}{0.24\textwidth}
		\centering
		\includegraphics[width=\textwidth]{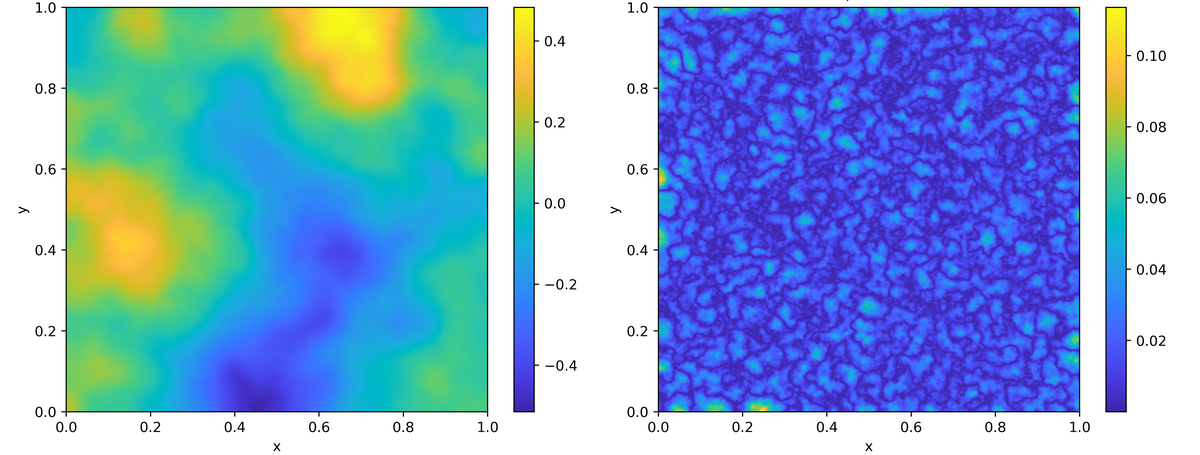}
		\caption{PCALin}
		\label{fig:pcalin-poisson-inverse-ug}
	\end{subfigure}\hfill 
	\begin{subfigure}{0.24\textwidth}
		\centering
		\includegraphics[width=\textwidth]{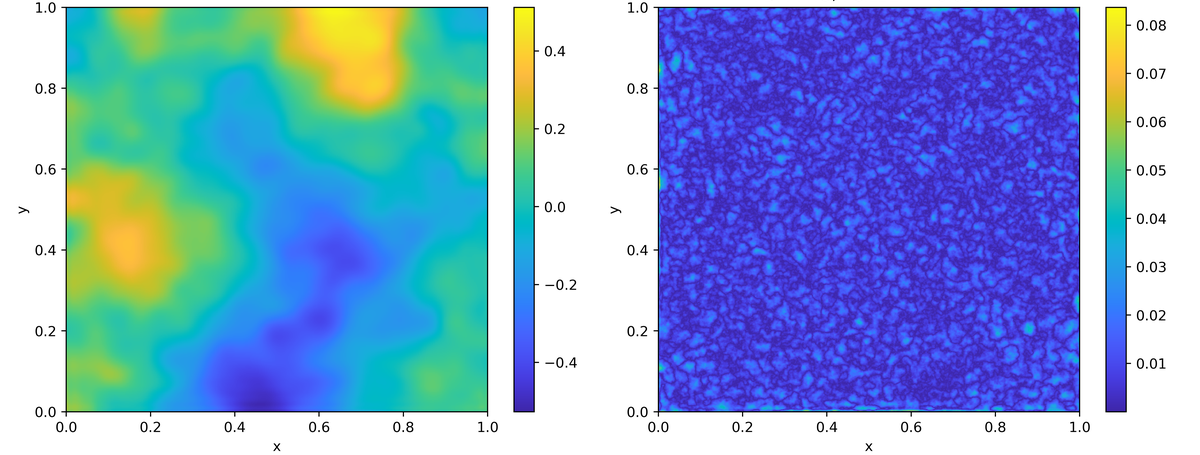}
		\caption{FNO}
		\label{fig:fno-poisson-inverse-ug}
	\end{subfigure}\\
	\begin{subfigure}{0.24\textwidth}
		\centering
		\includegraphics[width=\textwidth]{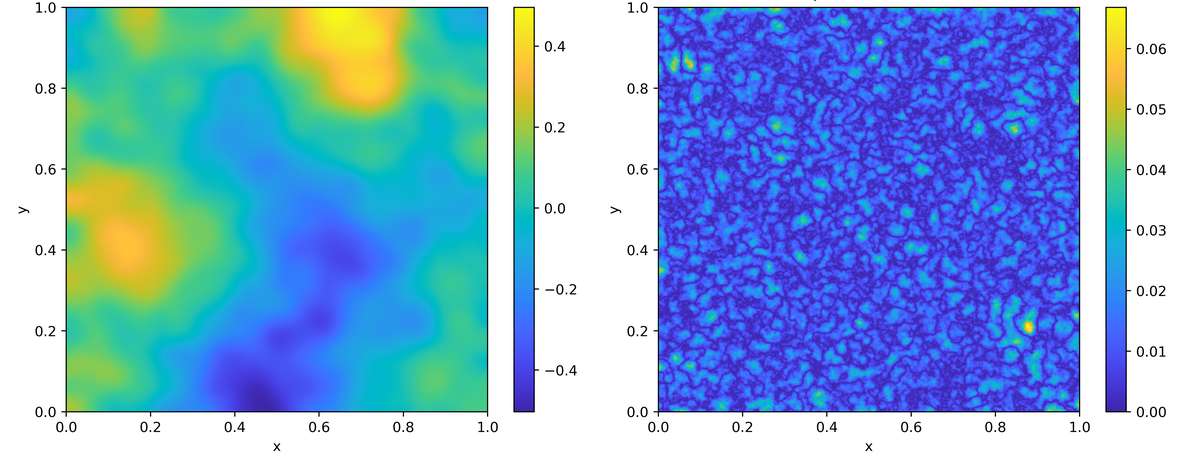}
		\caption{U-FNO}
		\label{fig:ufno-poisson-inverse-ug}
	\end{subfigure}\hfill
	\begin{subfigure}{0.24\textwidth}
		\centering
		\includegraphics[width=\textwidth]{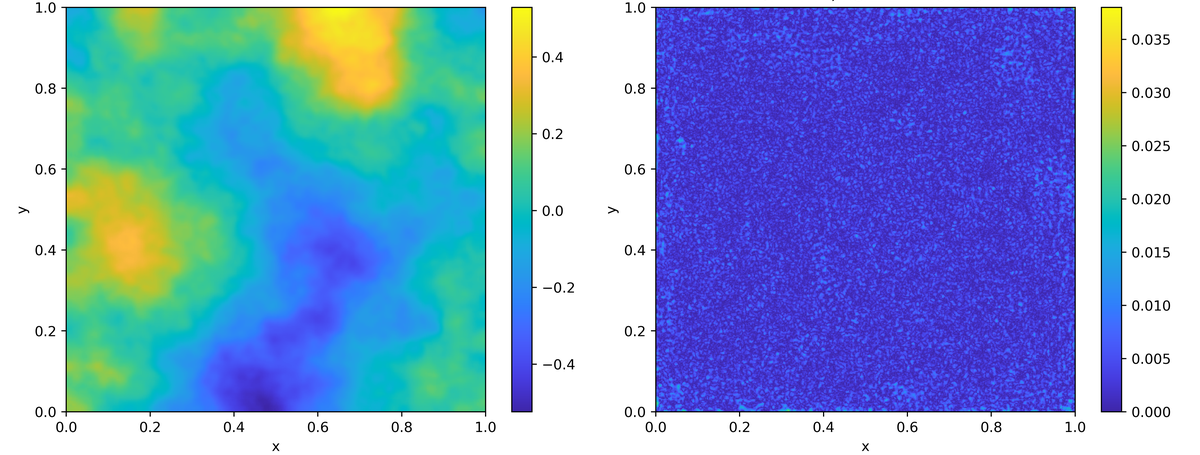}
		\caption{MWT}
		\label{fig:mwt-poisson-inverse-ug}
	\end{subfigure}\hfill 
	\begin{subfigure}{0.24\textwidth}
		\centering
		\includegraphics[width=\textwidth]{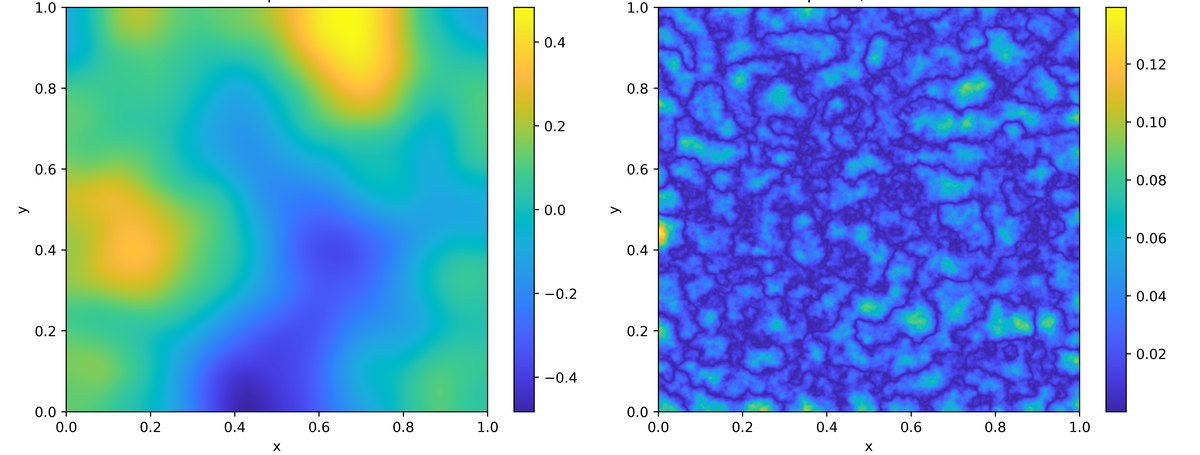}
		\caption{DeepONet}
		\label{fig:deeponet-poisson-inverse-ug}
	\end{subfigure}\hfill 
	\begin{subfigure}{0.24\textwidth}
		\centering
		\includegraphics[width=\textwidth]{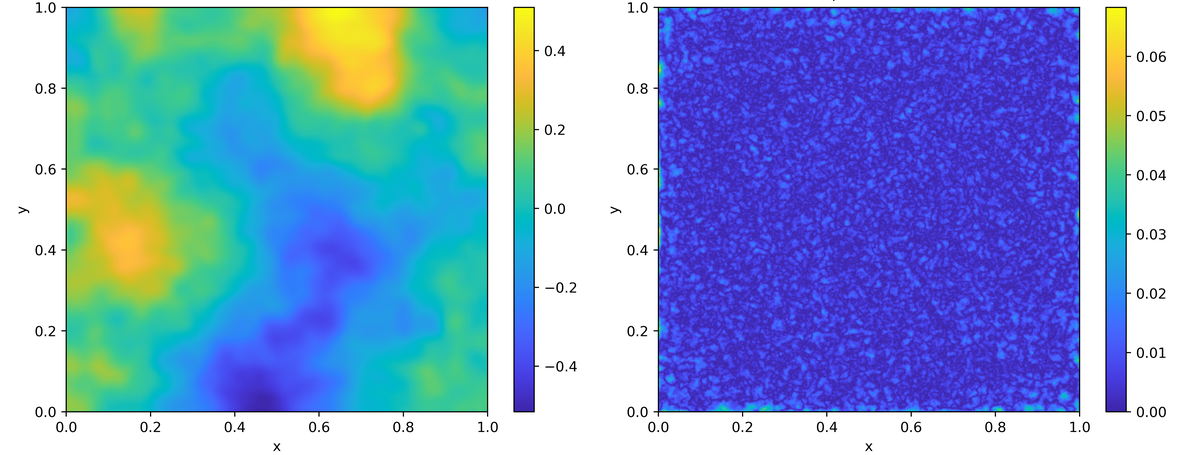}
		\caption{PINO}
		\label{fig:pino-poisson-inverse-ug}
	\end{subfigure} \\
	\begin{subfigure}{0.24\textwidth}
		\centering
		\includegraphics[width=\textwidth]{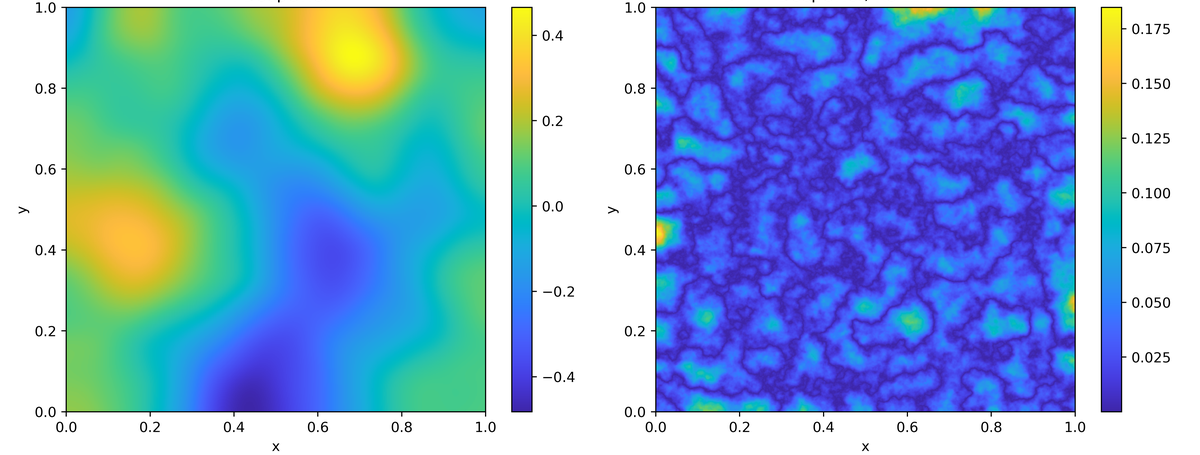}
		\caption{PI-Deeponet}
		\label{fig:pi-deeponet-poisson-inverse-ug}
	\end{subfigure}\hfill 
	\begin{subfigure}{0.24\textwidth}
		\centering
		\includegraphics[width=\textwidth]{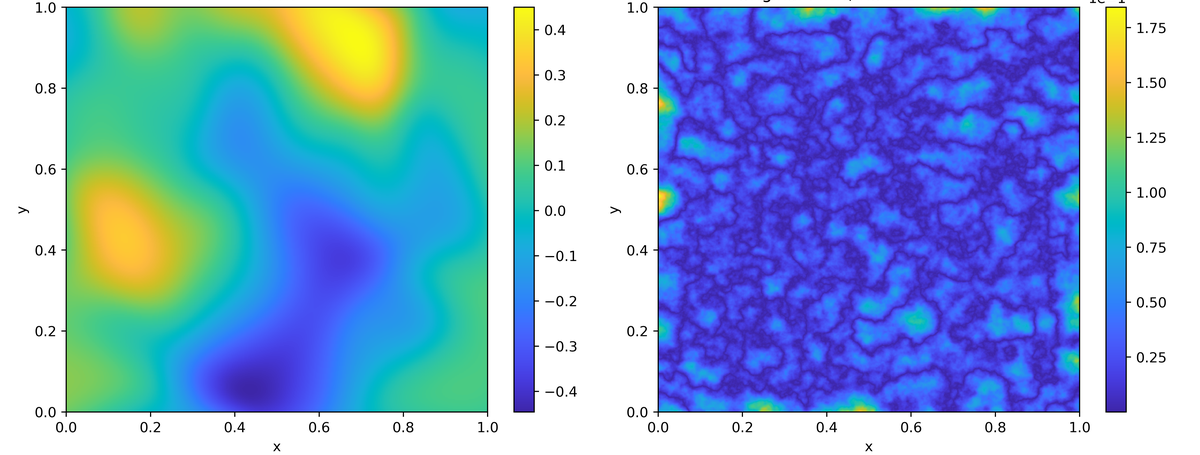}
		\caption{PINN}
		\label{fig:pinns-poisson-inverse-ug}
	\end{subfigure} \hfill
	\begin{subfigure}{0.24\textwidth}
		\centering
		\includegraphics[width=\textwidth]{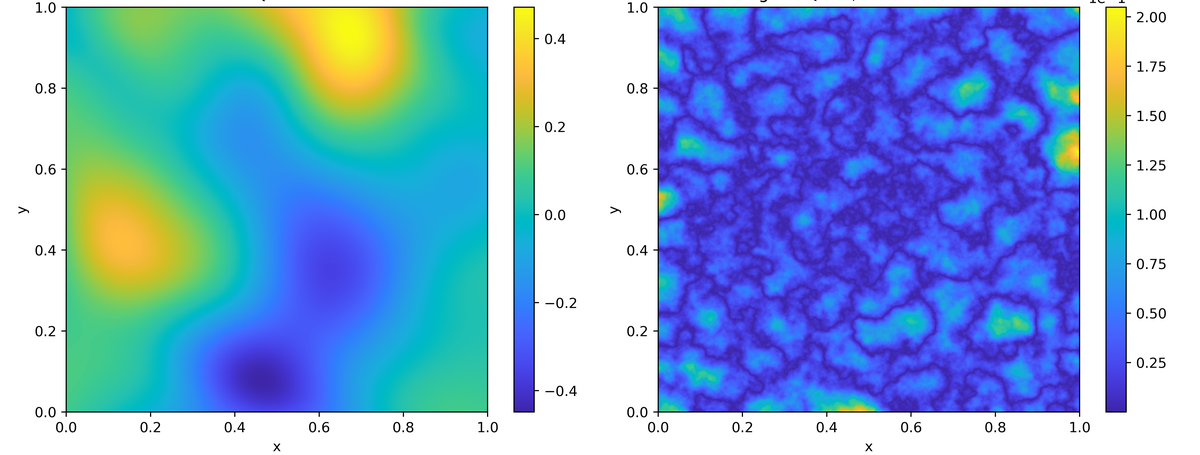}
		\caption{QRES}
		\label{fig:qres-poisson-inverse-ug}
	\end{subfigure}\hfill
	% \begin{subfigure}{0.24\textwidth}
		% 	\centering
		% 	\includegraphics[width=\textwidth]{figures/Poisson_inverse_1_5-th_res/DeepRitz-Poisson-Inverse-UG.png}
		% 	\caption{DRM}
		% 	\label{fig:deepritz-poisson-inverse-ug}
		% \end{subfigure}
	\caption{Test examples with backward operator training for inverse Poisson problem using a resolution of $513 \times 513$. (b)-(k), shows the specified neural network's approximation of the solution (left-hand side) and the absolute difference between the Ground truth in (a) with the approximation (right-hand side). All examples are computed with noise free data. For reconstructions with noisy data see Figure \ref{tab:fig-xxnoise-poisson-65}.}
	\label{fig:poisson-inverse-ug}
\end{figure}

\begin{table}[!ht]\small
	\centering
	%\resizebox{\textwidth}{!}{
		\begin{tabular}{r|cccc|cccc} 
			\multicolumn{1}{c}{} & \multicolumn{4}{c|}{\textit{Forward Problem Rel. Errors}} & \multicolumn{4}{c}{\textit{Inverse Problem Rel Errors}}\\
			\cline{2-9}
			Grid size, $s$ & $65$ & $129$ & $257$ & $513$ & $65$ & $129$ & $257$ & $513$  \\
			\hlineB{3}\midrule
			DRM      & 0.0397 & 0.0289 & 0.0244 & 0.0251 & - & - & - & - \\
			PINN    & 0.0245 & 0.0088 & 0.0084 & 0.0075 &  0.1715 & 0.1694 & 0.1653 &
			0.1650
			\\
			QRES     & 0.0288 & 0.0082 & 0.0088 & 0.0076 & 0.1598 & 0.1565 & 0.1542 & 0.1549\\
			%WAN      &  &  &  &  &  &  &  &  \\
			\midrule%\hline \hline
			PCANN     & $0.0078$ & $0.0075$ & $0.0073$ & $0.0073$ & $0.0989$ & $0.0979$ & $0.0981$ & $0.0981$ \\
			PCALin   & \cellcolor{gray!20}$0.0016$ & \cellcolor{gray!20}$0.0013$ & \cellcolor{gray!20}$0.0013$ & \cellcolor{gray!20}$0.0013$ & $0.0303$ & $0.0271$ & $0.0253$ & $0.0244$\\
			FNO      & $0.0066$ & $0.0061$ & $0.0067$ & $0.0066$ & $0.0299$ & $0.0341$ & $0.0366$ & $0.0416$ \\
			U-FNO    & $0.0063$ & $0.0056$ & $0.0062$ & $0.0054$ & $0.0292$ & $0.0277$ & $0.0288$ & $0.0254$\\
			MWT & $0.0047$ & $0.0047$ & $0.0048$ & $0.0036$ & $0.0357$ & \cellcolor{gray!20}$0.0202$ & \cellcolor{gray!20}$0.0198$ & \cellcolor{gray!20}$0.0159$\\
			DeepONet & 0.0047 & 0.0041 & 0.0041 & 0.0042 &0.0983  & 0.0992 &0.1041  &  0.1044\\
			\midrule%\hline 
			PINO     & $0.0031$ & $0.0030$ & $0.0034$ & $0.0031$ & \cellcolor{gray!20}$0.0263$ & $0.0273$ & $0.0319$ & $0.0301$ \\
			PI-DeepONet & $0.0081$  & $0.0057$ & $0.0060$ & $0.0061$ & $0.1074$ & $0.1068$ & $0.1067$ & $0.1068$ \\
			\bottomrule%\hline
		\end{tabular}
		%}
	\caption{Error variation with the resolution for the Poisson problem. The networks were trained with noiseless data and backward operator training was used for solving the inverse problem.}
	\label{tab:errors-poisson}
\end{table} 

\subsubsection{Summary Poisson problem}

In summary, for the forward problem, the investigated DL concepts are less accurate as compared with finite difference methods on a fine grid. However, these DL concepts do produce competitive results for solving inverse problems with different noise levels. DL concepts for operator approximation perform best and offer a significant advantage in run time. Hence, either large-scale parametric studies or iterative solvers for parameter identification decisively benefit from well-trained DL concepts.

For the somewhat simple Poisson problem with vanishing boundary data, where both the forward as well as the inverse operator are linear, there is not much difference between the DL concepts. Overall, we propose to use PCALin, PCANN or PINN for first testing due to their ‘easy-to-use’ structure. They produce acceptable results for a wide range of hyper-parameters and are comparatively easy to train.
For more accurate results, we propose to use MWT, which however is somewhat more involved, for implementation details see the respective appendices.

\subsection{Darcy flow}
\subsubsection{Forward problem}
We now consider the steady-state of the 2-d Darcy Flow equation on the unit square which is the second
order, linear, elliptic PDE

\begin{equation}
	\begin{aligned}
		-\nabla \cdot(\lambda(s) \nabla u(s)) &=f(s) & & s \in(0,1)^{2} \\
		u(s) &=0 & & s \in \partial(0,1)^{2}
	\end{aligned}
	%\label{eqn:darcy}
\end{equation}
with a Dirichlet boundary.  In this equation $a \in L^{\infty}\left((0,1)^{2} ; \mathbb{R}_{+}\right)$ is the diffusion coefficient,  $f = 1 \in L^{2}\left((0,1)^{2} ; \mathbb{R}_{+}\right)$ is the forcing function and $u \in H_{0}^{1}((0,1)^{2} ; \mathbb{R})$ is the unique solution of the   

We start with a discussion of the forward problem, i.e. computing the solution $u$ in Equation \ref{eqn:darcy}, for a given piece-wise constant diffusion coefficient $\lambda$. In all experiments, $\lambda(x) \in \{3,12\}$, i.e. the domain of definition is segmented randomly into two regions with known values.

Surprisingly, the errors for PINN and QRES are relatively high, even if the network is trained and evaluated with noise-free data, see Table  \ref{tab:errors-513-darcypwc}.
All other methods solve this non-linear forward problem reliably.   MWT is the overall winner for solving the forward problem with noise-free data. This also holds for most levels of resolution.

Again, the comparison with respect to different resolutions of $u$ does show the expected result that operator approximations based on functional analytic reasoning in function spaces indeed show an independent accuracy across scales, see Table \ref{tab:errors-darcypwc}.

\subsubsection{Inverse problem}
The investigation of the performance of DL concepts for inverse Darcy flows again either uses inverse training or embedding of a forward solver into a Tikhonov functional. The testing scenarios are similar to those of the Poisson problem, i.e. $1000$ random samples were used for training and $5000$ samples for evaluation.

\begin{figure}[!ht]
	\begin{subfigure}{0.24\textwidth}
		\centering
		\includegraphics[width=0.51\textwidth]{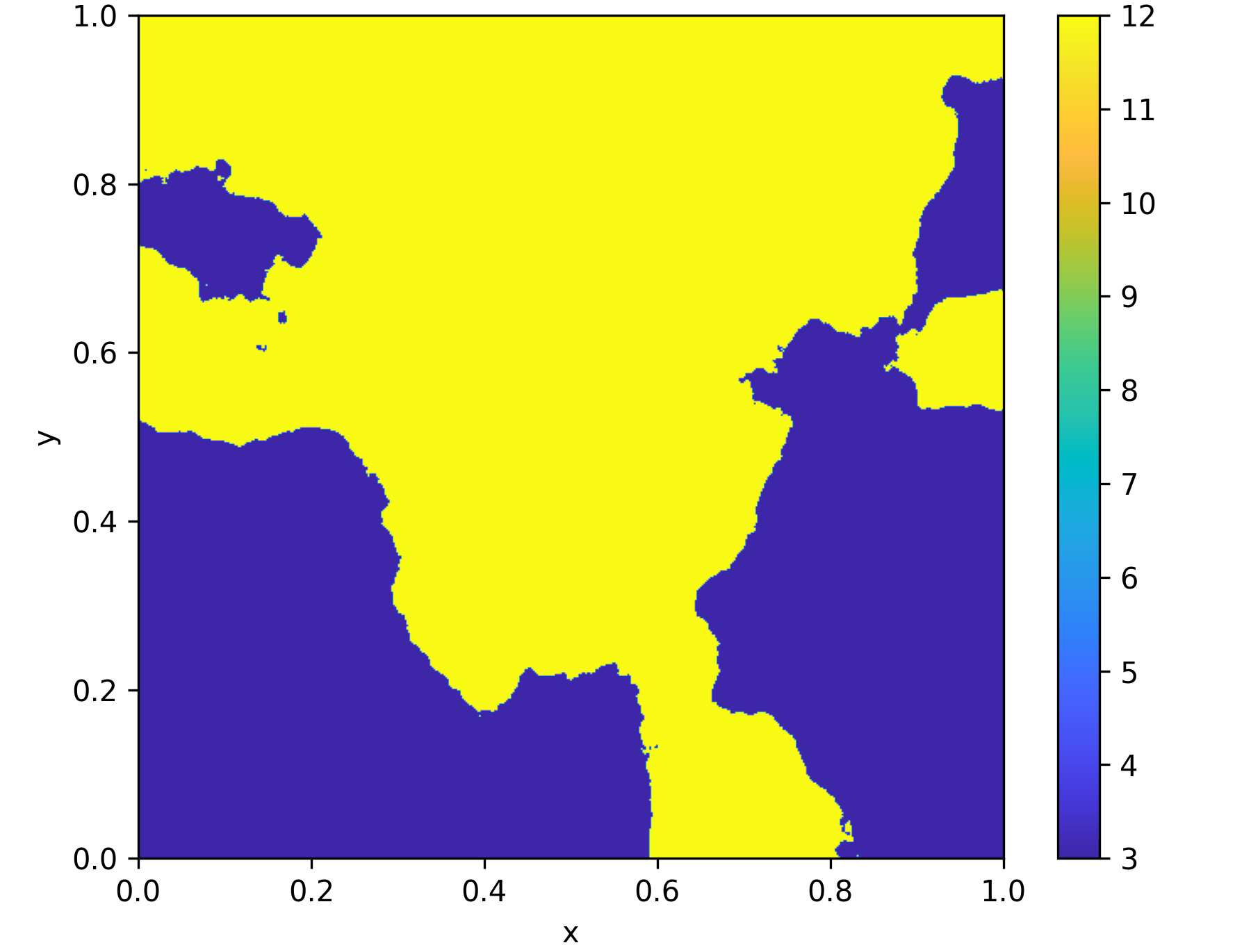}
		\caption{Ground Truth}
		\label{fig:groundtruth-darcypwc-inverse-ug}
	\end{subfigure}\hfill 
	\begin{subfigure}{0.24\textwidth}
		\centering
		\includegraphics[width=\textwidth]{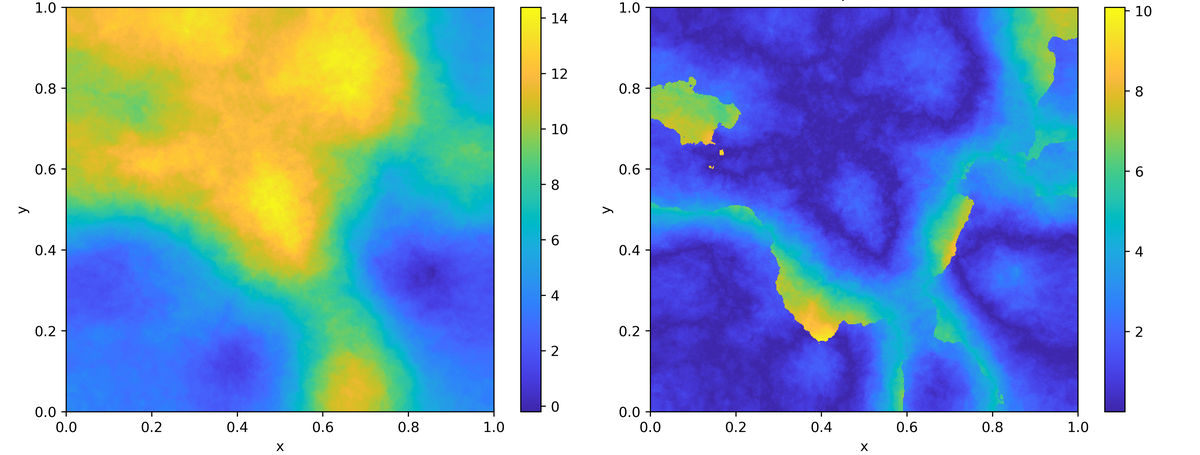}
		\caption{PCANN}
		\label{fig:pcann-darcypwc-inverse-ug}
	\end{subfigure} \hfill
	\begin{subfigure}{0.24\textwidth}
		\centering
		\includegraphics[width=\textwidth]{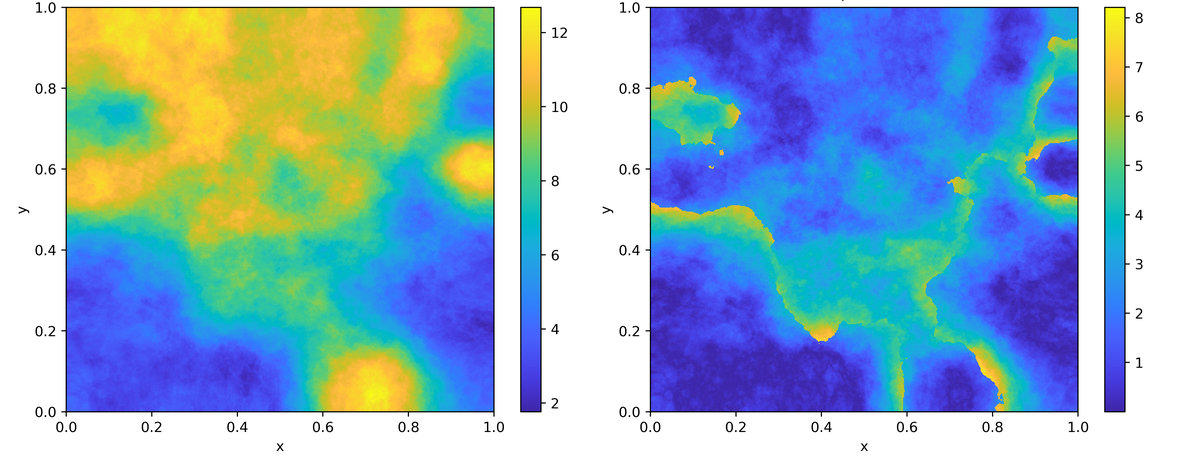} %CHANGE
		\caption{PCALin}
		\label{fig:pcalin-darcypwc-inverse-ug}
	\end{subfigure}\hfill 
	\begin{subfigure}{0.24\textwidth}
		\centering
		\includegraphics[width=\textwidth]{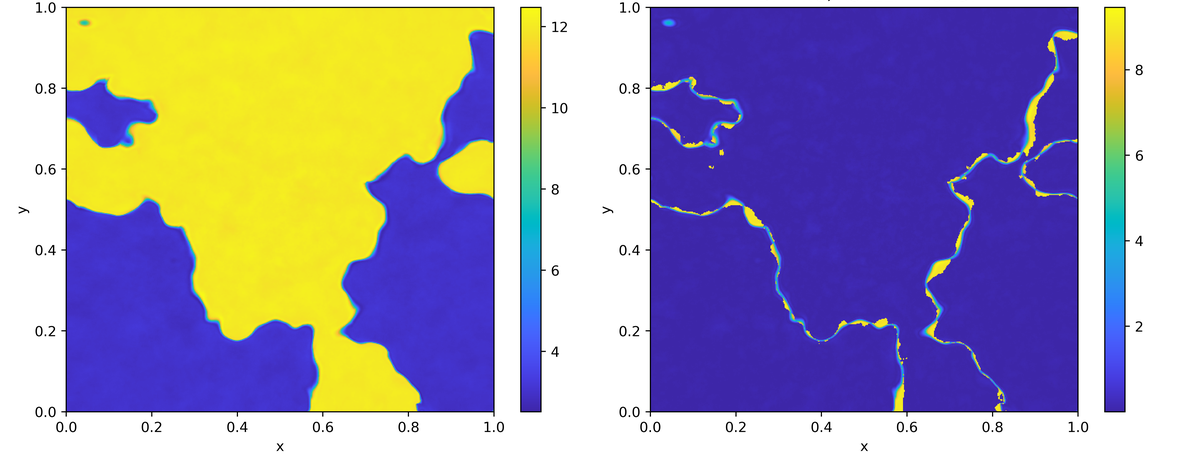}
		\caption{FNO}
		\label{fig:fno-darcypwc-inverse-ug}
	\end{subfigure}\\
	\begin{subfigure}{0.24\textwidth}
		\centering
		\includegraphics[width=\textwidth]{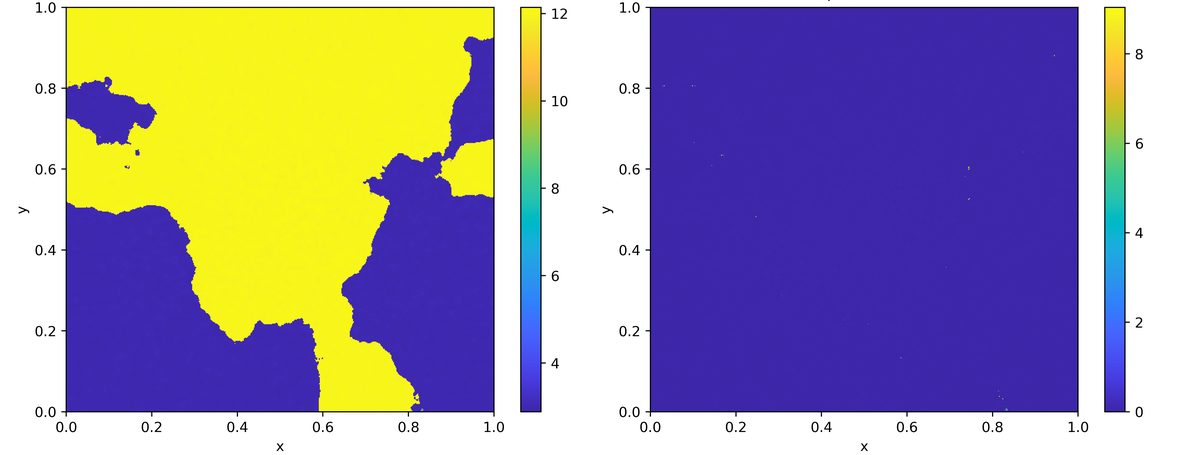}
		\caption{U-FNO}
		\label{fig:ufno-darcypwc-inverse-ug}
	\end{subfigure}\hfill
	\begin{subfigure}{0.24\textwidth}
		\centering
		\includegraphics[width=\textwidth]{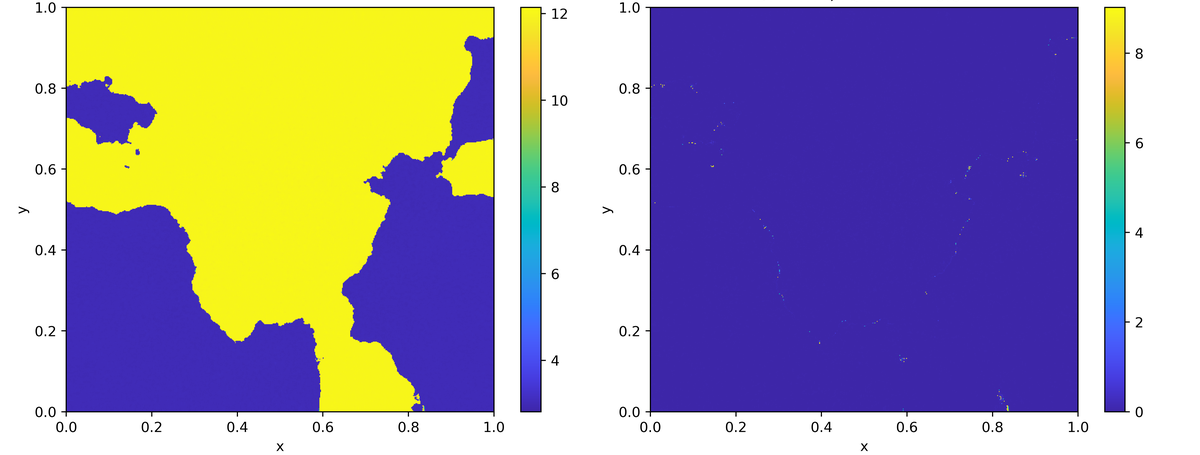}
		\caption{MWT}
		\label{fig:mwt-darcypwc-inverse-ug}
	\end{subfigure}\hfill 
	\begin{subfigure}{0.24\textwidth}
		\centering
		\includegraphics[width=\textwidth]{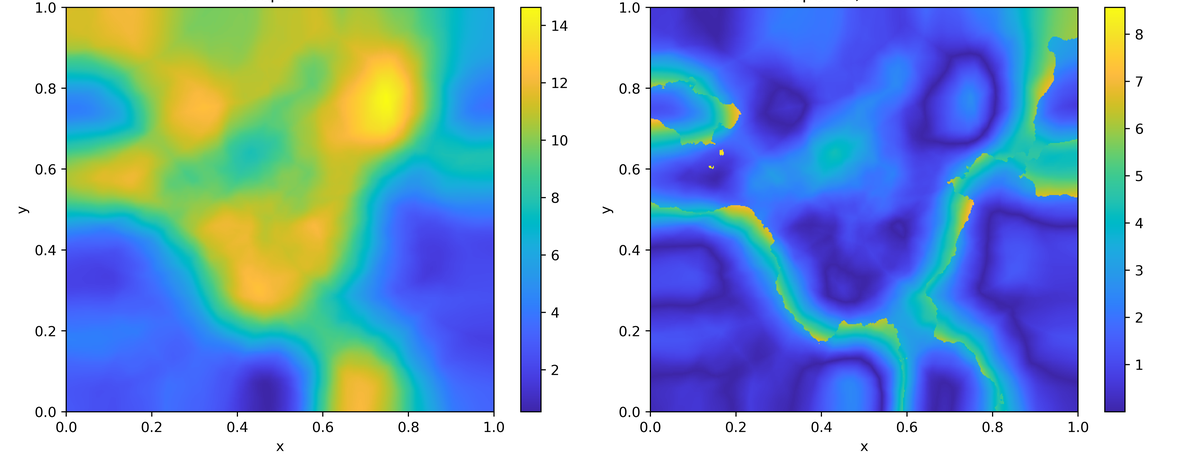}
		\caption{DeepONet}
		\label{fig:deeponet-darcypwc-inverse-ug}
	\end{subfigure}\hfill 
	\begin{subfigure}{0.24\textwidth}
		\centering
		\includegraphics[width=\textwidth]{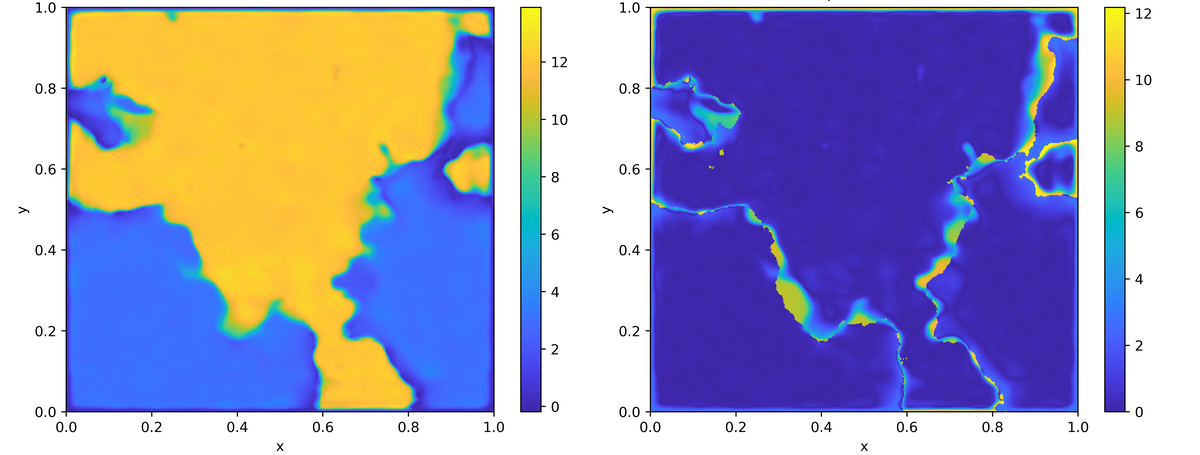} %CHANGE
		\caption{PINO}
		\label{fig:pino-darcypwc-inverse-ug}
	\end{subfigure} \\
	\begin{subfigure}{0.24\textwidth}
		\centering
		\includegraphics[width=\textwidth]{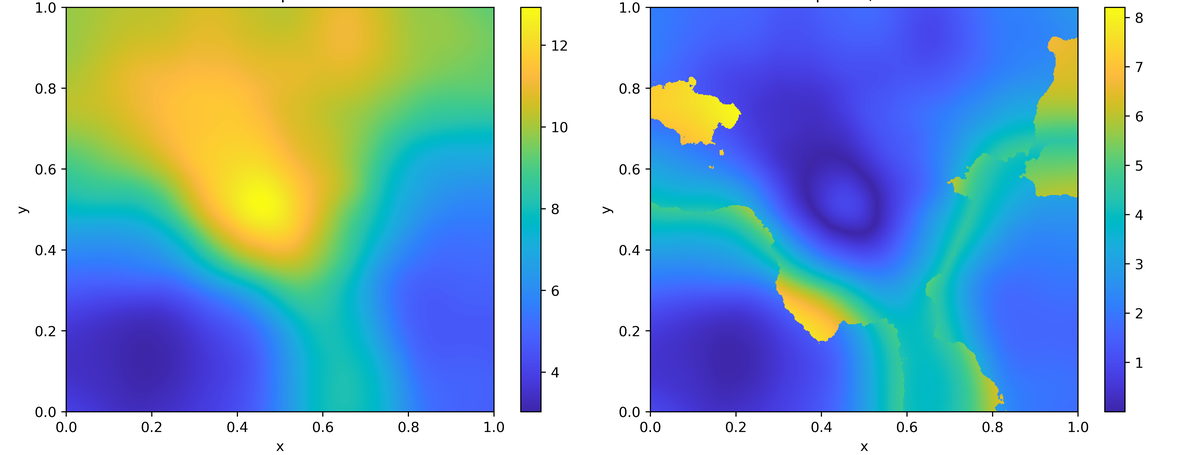}
		\caption{PI-Deeponet}
		\label{fig:pi-deeponet-darcypwc-inverse-ug}
	\end{subfigure}\hfill 
	\begin{subfigure}{0.24\textwidth}
		\centering
		\includegraphics[width=\textwidth]{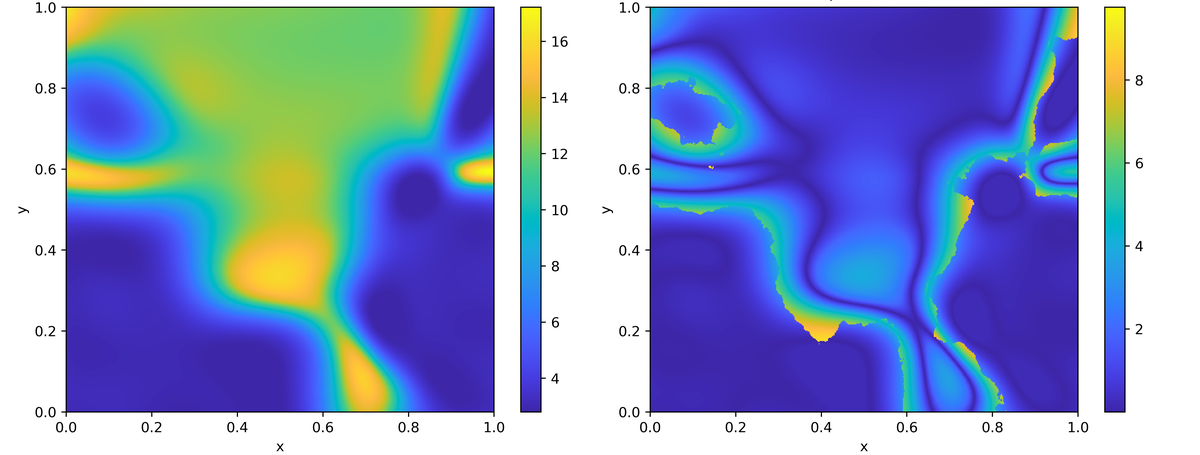}
		\caption{PINN}
		\label{fig:pinns-darcypwc-inverse-ug}
	\end{subfigure} \hfill
	\begin{subfigure}{0.24\textwidth}
		\centering
		\includegraphics[width=\textwidth]{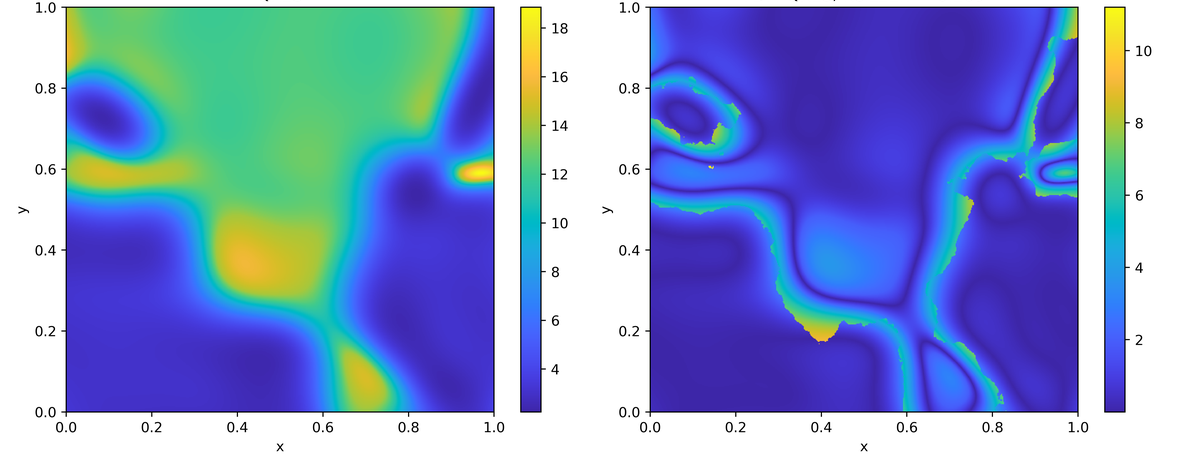}
		\caption{QRES}
		\label{fig:qres-darcypwc-inverse-ug}
	\end{subfigure}\hfill
	% \begin{subfigure}{0.24\textwidth}
		% 	\centering
		% 	\includegraphics[width=\textwidth]{figures/DarcyPWC_inverse_1_5-th_res/DeepRitz-DarcyPWC-Inverse-UP.png}
		% 	\caption{DRM}
		% 	\label{fig:deepritz-darcypwc-inverse-ug}
		% \end{subfigure}
	\caption{Test examples with backward operator training for inverse Darcy Flow problem using a resolution of $513 \times 513$. (b)-(k), shows the specified neural network's approximation of the solution (left-hand side) and the absolute difference between the Ground truth in (a) with the approximation (right-hand side). All examples are computed with noise free data. For reconstructions with noisy data see Figure \ref{tab:fig-xxnoise-darcypwc-65}. }
	\label{fig:darcypwc-inverse-ug}
\end{figure}

In view of different applications we have either used no further assumption on  $\lambda$, see Figure \ref{fig:darcypwc-inverse-ug}, or we have assumed that we know, that $\lambda$ is a segmentation of the domain of definition into two components with a known value, see Table \ref{tab:fig-xxnoise-darcypwc-65}. This e.g. models the problem of mixing two liquids with known densities.

For a closer investigation, we start with a basic experiment for solving the Darcy inverse problems by inverse training with noise-free data. Results using a $513 \times 513$ resolution and without knowing the values o$\lambda$ a priori are shown in Figure \ref{fig:darcypwc-inverse-ug}.  We notice the strength of the FNO, U-FNO, MWT, and PINO to identify the position of the discontinuities with backward operator training. PINO, however, computes a mismatch in the values of the densities. The network architecture seems to play a crucial role in these results. For e.g. DeepONet always provides a reconstruction, which is obtained by a linear combination of smooth functions(trunk nets), and in solution learning methods $u_\Theta(x)$ is naturally continuous, hence discontinuities are not in the range of these methods. Other methods based on discretisation, i.e. reconstructing the densities in the last layer of the networks at prescribed positions allow for sharper transitions of the densities.

As before, we have done experiments where the networks were trained with noiseless or noisy data.
The results of inverse training with perfect data for training and evaluation are stated in Tables \ref{tab:errors-513-darcypwc} and \ref{tab:errors-darcypwc}, they mimic the results of the forward training and do not yield additional insights. 
As for the Poisson problem, training with noise-free data does not generalise to noisy data, see Table \ref{tab:noise-backward-operator-65}, even only $1 \%$ error in the evaluation data yields reconstruction errors of $30 \%$ and larger. 
We should remark on the sub-optimal results obtained by DeepONet, which shows errors of about $25 \%$ for all noise levels tested. Here the error margins are considerably larger but almost constant over different noise levels. This might indicate, that the dominant error is not stemming from inversion and noise, but rather from a structural disadvantage of the network architecture or sub-optimal training.
We have tested several larger DeepONet network architectures, which however did not cure the problem. A major reason for the large error of DeepONet for this inverse problem is that DeepONet has to learn the basis of the output space via the trunk net. Thus, the trunk net must have sufficient expressiveness to approximate the basis of the output space. In addition, even if the size of the trunk net is large enough, finding the optimal parameters of the trunk net is another challenge. Thus, if the number of reduced basis of the output space is large or the output functions are very complicated (e.g. highly oscillatory or discontinuous), it would be very difficult for the trunk net to learn the basis accurately. For both Poisson and Darcy problems, the parameter spaces are more complicated than the solution spaces, which explains the large error of DeepONet for inverse training.

Tables \ref{tab:noise-backward-operator-trained-noise-65} and \ref{tab:xxnoise-darcypwc-65} state the results most important for inverse problems. They show some remarkable differences between the considered concepts. Most methods (PINN, QRES, PCANN, PCALin, DeepONet, PI-DeepONet) do show approximately the same errors for all noise levels. This indicates that a structural approximation error dominates the influence of the data noise. We have tested different hyperparameter settings for all of these methods, which, however, did not change the result. As a consequence, these methods do not yield competitive results for small levels of data noise. 
Nevertheless, PCANN exhibits about $10\%$ error in the reconstructions even for 
$5\%$ data error, which is remarkably good. In our test scenarios, there are clear winners: U-FNO and MWT for small noise levels and PCANN for larger noise levels. While the
DeepONet, its physics-informed variant PI-DeepONet as well as PINO  do not seem to be suitable for Darcy inverse problems. As mentioned earlier, PINO is capable of detecting the discontinuities but it fails to provide accurate density estimates. 

When comparing the best methods for inverse training (MWT for small noise levels, PCANN for larger noise levels, see Table \ref{tab:noise-backward-operator-trained-noise-65}  ) with the best method for Tikhonov minimisation, see Table \ref{tab:xxnoise-darcypwc-65}, we see an advantage for inverse training, which generally yields better results.  That errors are consistently larger for Tikhonov learning might be partially explainable by the difficulty of choosing a suitable regularisation parameter $\alpha$. We have also included the accuracy metric which measures the segmentation error rather than the functional $L_2$ error. Here results again are similar for all methods, see Table \ref{tab:lambdaAcc-noise-darcypwc-65}.

We finally estimated the output error by feeding the reconstructed $\hat \lambda$ into the respective neural networks and comparing the resulting $\hat u$ with the exact data $u$ and its noisy version $u^\delta$, see Figures \ref{tab:unoisy-noise-darcypwc-65} and \ref{tab:xxnoise-darcypwc-65}.

\subsubsection{Summary Darcy flow}

In summary, the investigated DL concepts do produce competitive results for solving inverse problems with different noise levels. DL concepts for operator approximation perform best and offer a significant advantage in run time. Hence, either large-scale parametric studies or iterative solvers for parameter identification decisively benefit from well-trained DL concepts.
This training has to be done with the same noise level, training with noise-free data does not generalise to noisy data.
As a general procedure, we would recommend favouring backward operator training over Tikhonov minimisation.

For choosing appropriate methods, we propose to use inverse training in combination with  
U-FNO and MWT  for small noise levels and, the concept PCANN was best for solving the non-linear inverse Darcy flow problem with higher noise levels.

As extensive as our numerical tests have been, they only provide a snapshot of the diverse landscape of PDE problems and testing scenarios. In particular, it might be interesting e.g. to test Darcy inverse problems with continuous parameters and inverse Poisson problems with piece-wise constant functions. However, for the sake of an overview and the already somewhat lengthy list of tables presented, we decided to restrict ourselves to the tests presented in the paper.

\begin{table}[!ht]\small
	\centering
	%\resizebox{\textwidth}{!}{
		\begin{tabular}{r|c|crc|crc} 
			\multicolumn{2}{c|}{} & \multicolumn{3}{c|}{\textbf{Forward Problem}} & \multicolumn{3}{c}{\textbf{Inverse Problem}}\\
			\cline{2-8}
			\multirow{2}{*}{\textbf{Networks}} & $\#$ \textit{ of} & \textit{Rel. L2} & \textit{Training} & \textit{Testing} & \textit{Rel. L2}  &\textit{Training} & \textit{Testing}\\  
			& \textit{Parameters} & \textit{Error} & \textit{(s/epoch)} & \textit{(s)} & \textit{Error} &\textit{(s/epoch)} & \textit{(s)} \\
			\hlineB{3}\midrule%\hline 
			DRM   & $22,201$ & $0.0369$ &  $0.0859$ & $\sim 1,050$  &  - & - & - \\
			PINN & $6,291 ~|~7,972 $  & $0.1995$  &  $0.1164$ & $\sim2,010$ &  $0.1988$ & $0.2528$ & $\sim4,790$ \\
			QRES  & $6,562 ~|~ 8,895$  &  $0.2017$ & $0.2108$ & $\sim3,320$ & $0.1975$ & $0.4221$ & $\sim6,420$ \\
			%WAN   &   &   &   &   &   &   & \\
			\midrule%\hline \hline
			PCANN  &  $5,155,150~|~5,035,030$ & $0.0253$  & $0.0136$  & $0.6108$  & $0.0988$   & $0.0487$ &  $0.1548$ \\
			PCALin & $10,100$ & $0.0656$  & $0.0075$  & $0.3290$  & $0.2203$  &  $0.0139$ &  $0.3108$ \\
			FNO    & $2,368,001$  & $0.0109$  & $41.6857$  & $0.0165$ &  $0.1490$  & $42.0596$ & $0.0163$  \\
			U-FNO  & $3,990,401$ & $0.0095$ & $97.9580$ & $0.0344$ & \cellcolor{gray!20}$\mathbf{0.0085}$  & $98.8170$ &  $0.0345$ \\
			MWT    & $9,807,873$ & \cellcolor{gray!20}$ \mathbf{0.0058}$ & $112.4575$ & $0.0637$ & $\mathbf{0.0156}$  &  $112.8750$ & $0.0403$ \\
			DeepONet & $568,320 ~|~ 1,047,644$  & $0.0295$ &$0.0606$ & $0.0011$ & $0.2220$ & $0.0659$ & $0.0011$ \\
			\midrule
			PINO & $2,368,001$  & $\mathbf{0.0084}$  & $42.8685$  & $0.0168$ & $0.2345$ & $42.7417$ & $0.0142$ \\
			PI-DeepONet & $568,320 ~|~ 684,288$  & $0.0384$  & $0.2513$   &  $0.0011$ & $0.2720$ &  $0.4905$ &  $0.0011$ \\
			\bottomrule%\hline
		\end{tabular}
		%}
	\caption{Performance of different methods for the  Darcy flow problem with piece-wise constant coefficients and using a $513 \times 513$ resolution. A similar convention is used for the second column as in Table \ref{tab:errors-513-darcypwc}. }
	\label{tab:errors-513-darcypwc}
\end{table}

\begin{figure}[!ht]
	\begin{subfigure}{0.24\textwidth}
		\centering
		\includegraphics[width=0.51\textwidth]{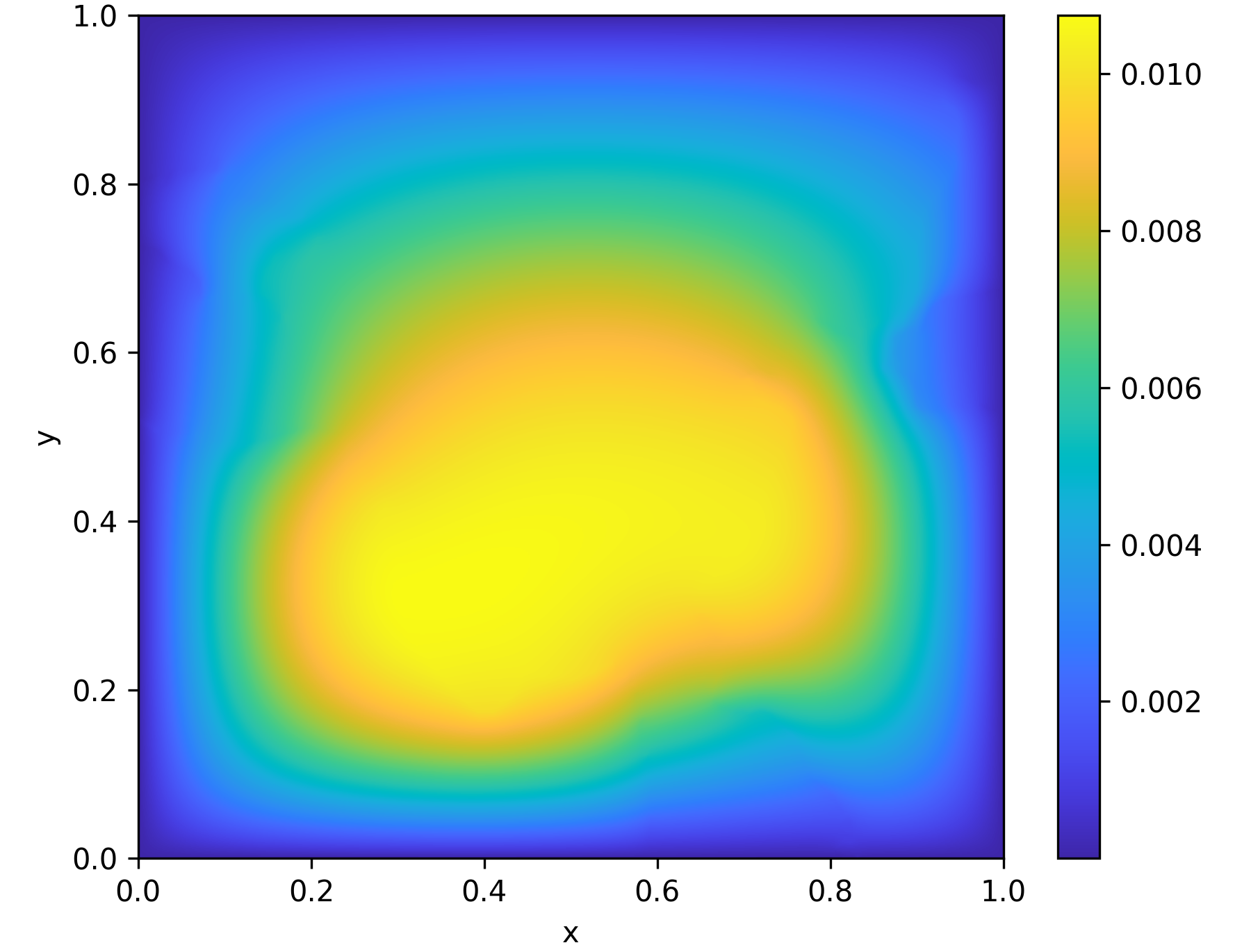}
		\caption{Ground Truth}
		\label{fig:groundtruth-darcypwc-forward-ug}
	\end{subfigure}\hfill
	\begin{subfigure}{0.24\textwidth}
		\centering
		\includegraphics[width=\textwidth]{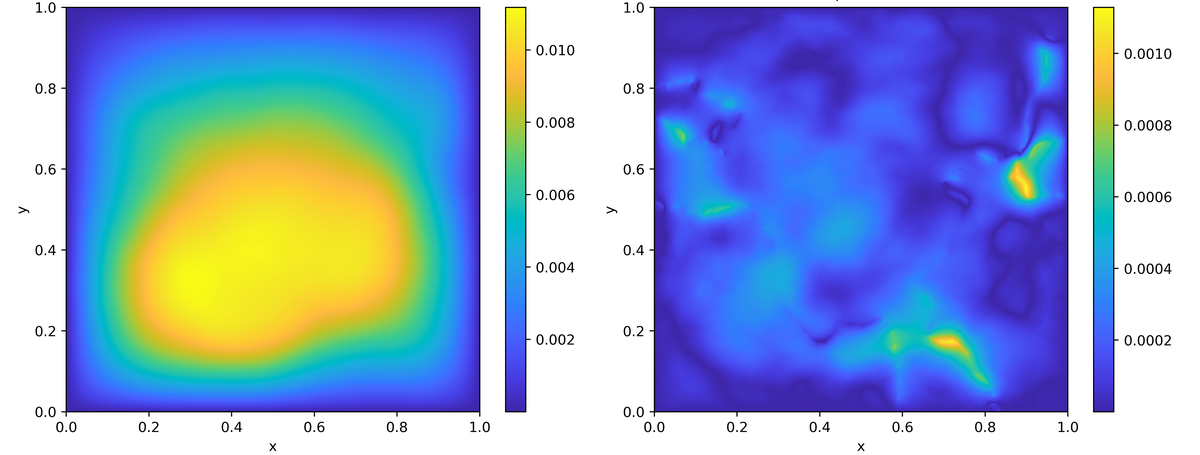}
		\caption{PCANN}
		\label{fig:pcann-darcypwc-forward-ug}
	\end{subfigure}\hfill
	\begin{subfigure}{0.24\textwidth}
		\centering
		\includegraphics[width=\textwidth]{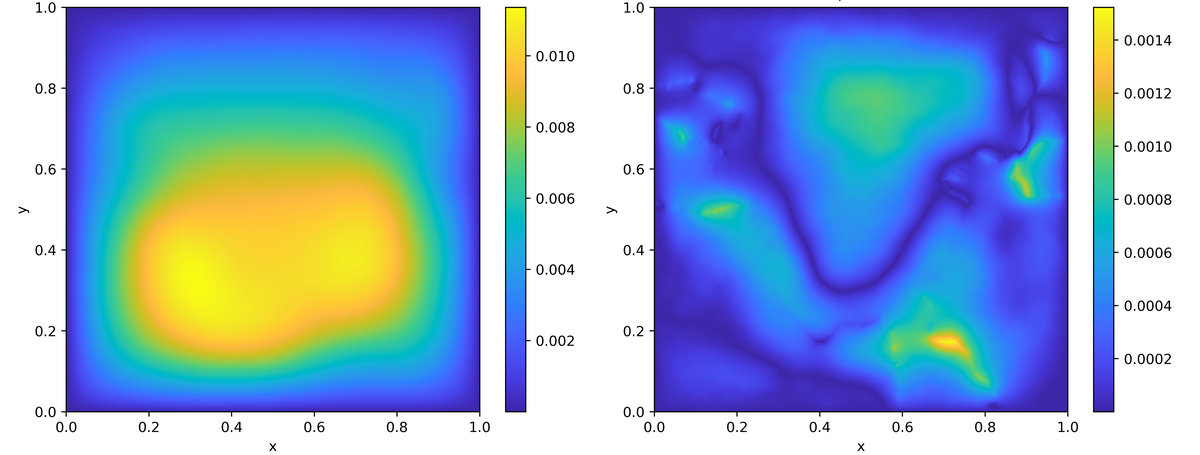}
		\caption{PCALin}
		\label{fig:pcalin-darcypwc-forward-ug}
	\end{subfigure}\hfill 
	\begin{subfigure}{0.24\textwidth}
		\centering
		\includegraphics[width=\textwidth]{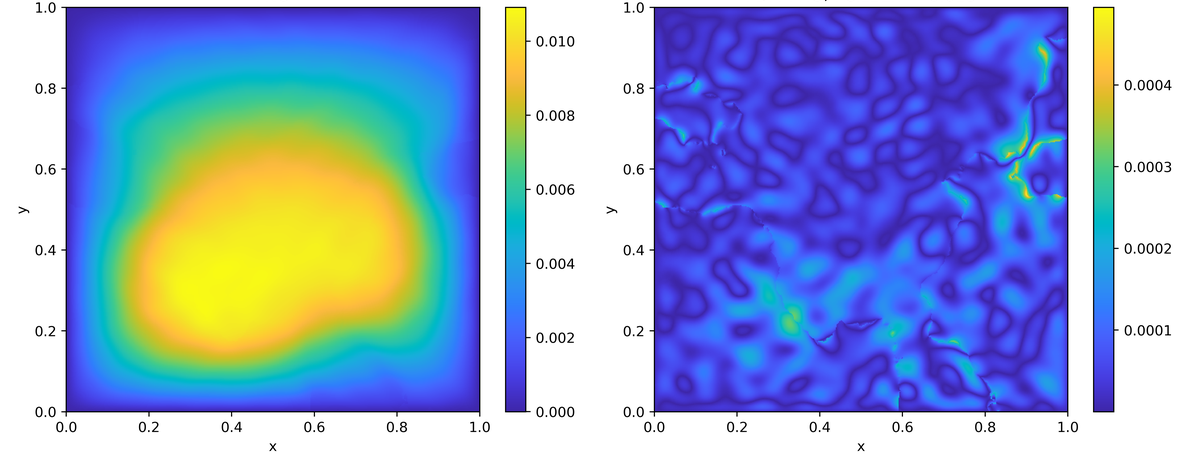}
		\caption{FNO}
		\label{fig:fno-darcypwc-forward-ug}
	\end{subfigure}\\
	\begin{subfigure}{0.24\textwidth}
		\centering
		\includegraphics[width=\textwidth]{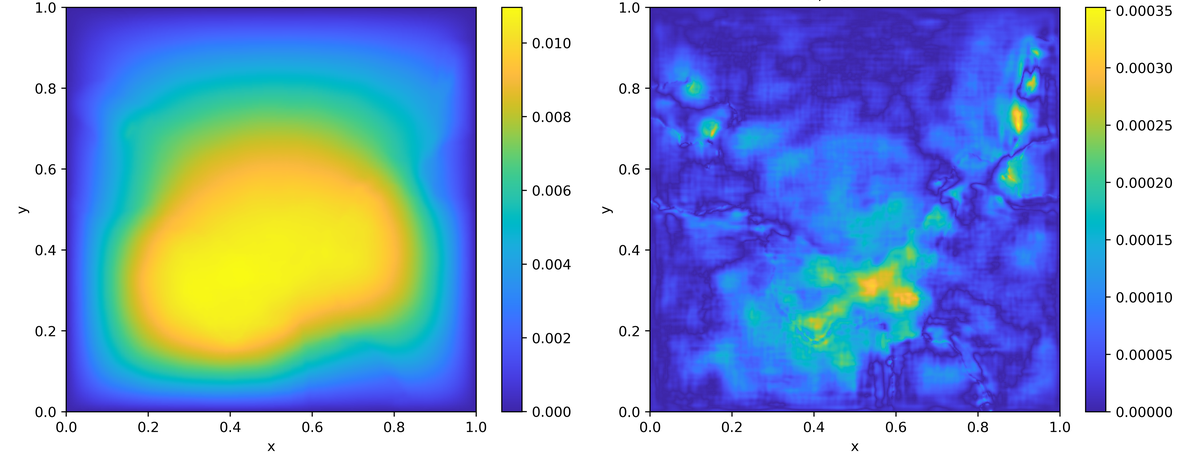}
		\caption{U-FNO}
		\label{fig:ufno-darcypwc-forward-ug}
	\end{subfigure}\hfill
	\begin{subfigure}{0.24\textwidth}
		\centering
		\includegraphics[width=\textwidth]{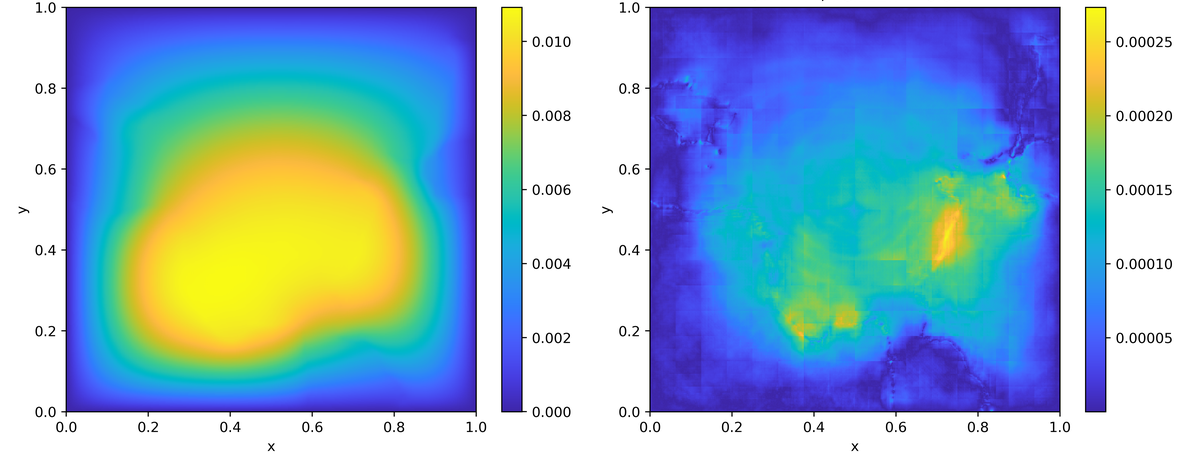}
		\caption{MWT}
		\label{fig:mwt-darcypwc-forward-ug}
	\end{subfigure}\hfill
	\begin{subfigure}{0.24\textwidth}
		\centering
		\includegraphics[width=\textwidth]{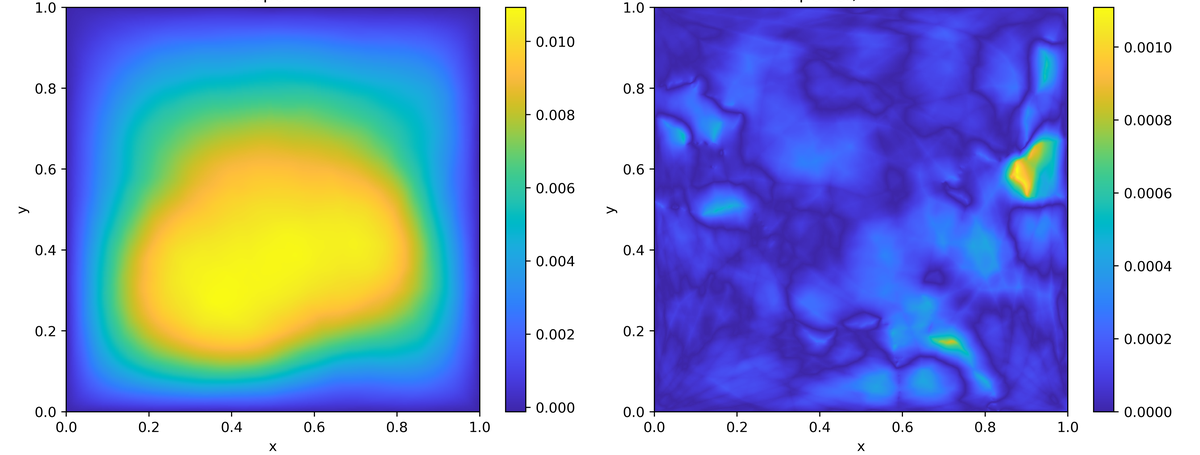}
		\caption{DeepONet}
		\label{fig:deeponet-darcypwc-forward-ug}
	\end{subfigure}\hfill
	\begin{subfigure}{0.24\textwidth}
		\centering
		\includegraphics[width=\textwidth]{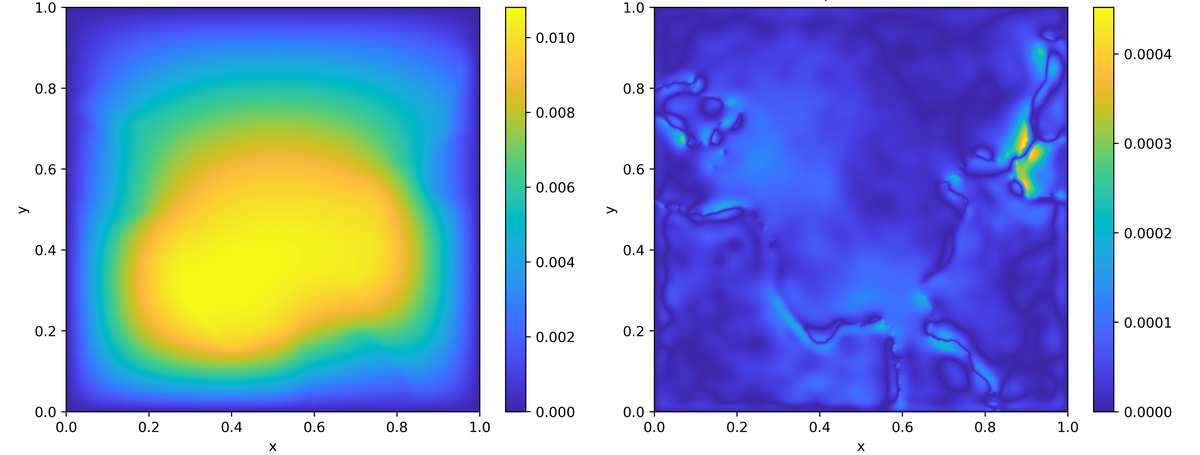}
		\caption{PINO}
		\label{fig:pino-darcypwc-forward-ug}
	\end{subfigure}	\\
	\begin{subfigure}{0.24\textwidth}
		\centering
		\includegraphics[width=\textwidth]{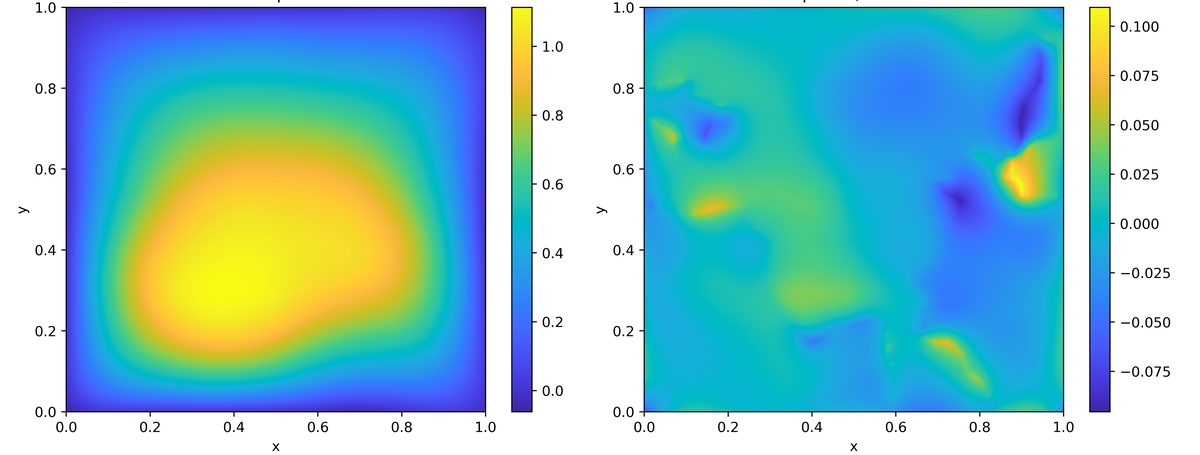}
		\caption{PI-Deeponet}
		\label{fig:pi-deeponet-darcypwc-forward-ug}
	\end{subfigure}\hfill
	\begin{subfigure}{0.24\textwidth}
		\centering
		\includegraphics[width=\textwidth]{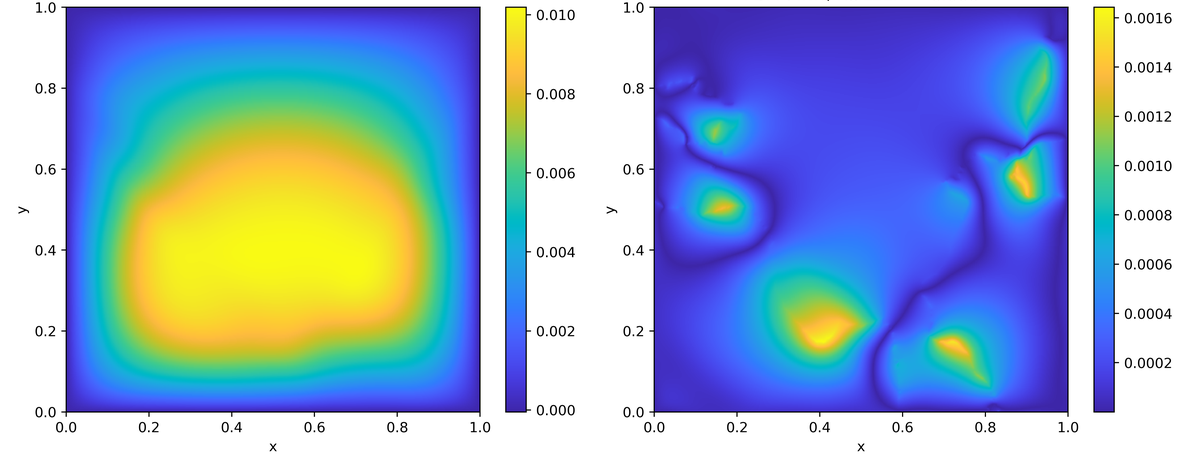}
		\caption{PINN}
		\label{fig:pinns-darcypwc-forward-ug}
	\end{subfigure} \hfill
	\begin{subfigure}{0.24\textwidth}
		\centering
		\includegraphics[width=\textwidth]{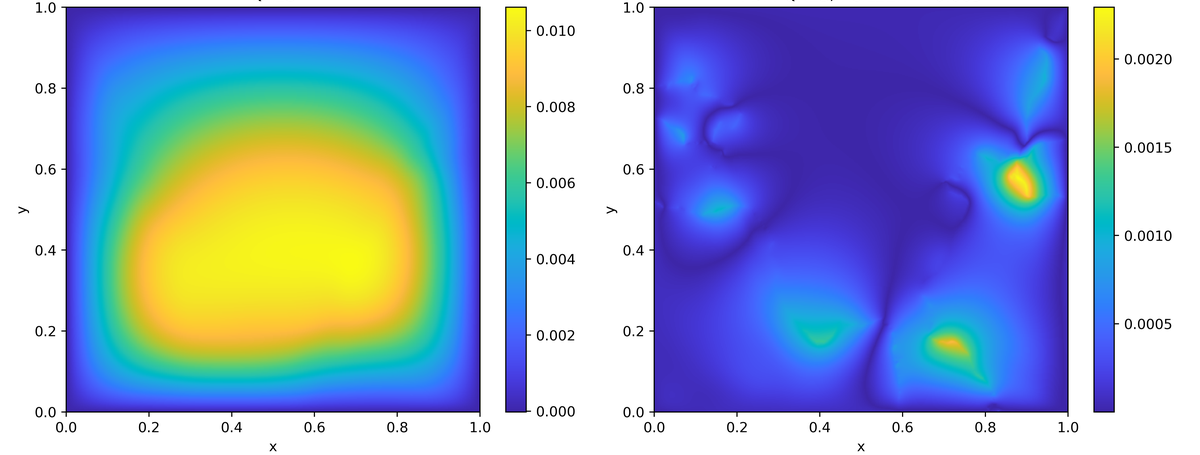}
		\caption{QRES}
		\label{fig:qres-darcypwc-forward-ug}
	\end{subfigure}\hfill
	\begin{subfigure}{0.24\textwidth}
		\centering
		\includegraphics[width=\textwidth]{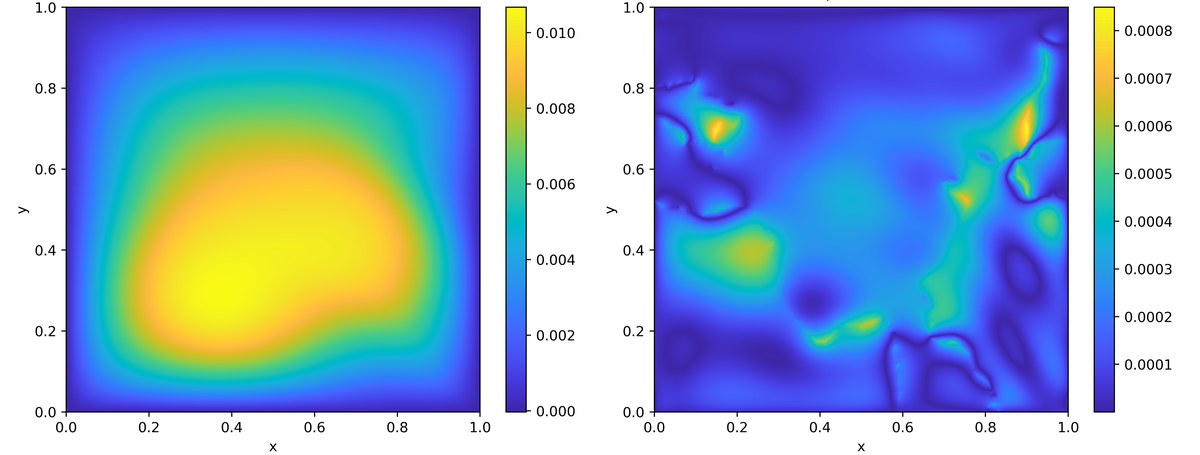}
		\caption{DRM}
		\label{fig:deepritz-darcypwc-forward-ug}
	\end{subfigure}
	\caption{Test examples for forward Darcy problem using a resolution of $513 \times 513$. (b)-(l), shows the specified neural network's approximation of the solution (left-hand side) and the absolute difference between the Ground truth in (a) with the approximation (right-hand side).}
\end{figure}

\begin{table}[!ht]\small
	\centering
	%\resizebox{\textwidth}{!}{
		\begin{tabular}{r|cccc|cccc} 
			\multicolumn{1}{c}{} & \multicolumn{4}{c|}{\textit{Forward Problem Rel. Errors}} & \multicolumn{4}{c}{\textit{Inverse Problem Rel Errors}}\\
			\cline{2-9}
			Grid size, $s$ & $65$ & $129$ & $257$ & $513$ & $65$ & $129$ & $257$ & $513$  \\
			\hlineB{3}\midrule
			DRM      & $0.0501$ & $0.0375$ & $0.0358$ & $0.0369$ & - & - & - & - \\
			PINN    &  $0.2254$ & $0.2043$ & $0.2071$ & $0.1995$ & $0.2413$ & $0.2014$ & $0.2032$ & $0.1988$\\
			QRES     & $0.2596$ & $0.2077$ & $0.1993$ & $0.2017$ & $0.2371$ & $0.2117$ & $0.2036$ & $0.1975$ \\
			%WAN      &  &  &  &  &  &  &  &  \\
			\midrule%\hline \hline
			PCANN     & $0.0256$ & $0.0254$ & $0.0253$ & $0.0253$ & $0.0983$ & $0.0985$ & $0.0987$ & $0.0988$ \\
			PCALin   & $0.0656$ & $0.0656$ &  $0.0656$ &  $0.0656$ & $0.2191$  & $0.2198$ & $0.2201$ & $0.2203$ \\
			FNO      & $0.0113$ & $0.0106$ & $0.0107$ & $0.0109$ & $0.1661$ & $0.1536$ & $0.1501$ & $0.1490$  \\
			U-FNO    & \cellcolor{gray!20}$0.0078$ & $0.0071$ & $0.0074$ & $0.0095$ & \cellcolor{gray!20}$0.0928$ & \cellcolor{gray!20}$0.0640$ & \cellcolor{gray!20}$0.0322$ & \cellcolor{gray!20}$0.0085$ \\
			MWT & $0.0080$ & \cellcolor{gray!20} $0.0060$ & \cellcolor{gray!20}$0.0059$ & \cellcolor{gray!20}$0.0058$ & $0.1047$ & $0.0800$ & $0.0461$ & $0.0156$ \\
			DeepONet & 0.0290 & 0.0292 & 0.0289  & 0.0295 & 0.2219 &0.2261  &0.2221  & 0.2220 \\
			\midrule%\hline 
			PINO     & \cellcolor{gray!20}$0.0078$ & $0.0066$ & $0.0071$ & $0.0084$ & $0.1703$ & $0.1997$ & $0.2216$ & $0.2345$ \\
			PI-DeepONet &$0.0390$  & $0.0382$ & $0.0381$ & $0.0384$ & $0.2706$ & $0.2711$ & $0.2682$ & $0.2720$ \\
			\bottomrule%\hline
		\end{tabular}
		%}
	\caption{Error variation with resolution for the Darcy Flow problem with piece-wise constant coefficients. The networks were trained with noiseless data and backward operator training was used for solving the inverse problem.}
	\label{tab:errors-darcypwc}
\end{table}

\begin{table}[!ht]\small
	\centering
	%\resizebox{\textwidth}{!}{
		\begin{tabular}{r|ccccc|ccccc}
			\multicolumn{1}{c}{} & \multicolumn{5}{c|}{\textit{Poisson}}& \multicolumn{5}{c}{\textit{Darcy Flow PWC}}\\ \cline{2-11}
			Noise level & $0 \%$ & $0.1 \%$ & $1 \%$ & $5 \%$ & $10 \%$ & $0 \%$ & $0.1 \%$ & $1 \%$ & $5 \%$  & $10 \%$\\
			\hlineB{3}\midrule%\hline
			% PINN & $0.1715$ & $0.1727$& $0.1734$& $0.2110$ & $0.2378$ & & & & \\
			% QRES & $0.1598$ & $0.1658$ & $0.1714$& $0.2007$ & $0.2256$ & & & & \\\midrule 
			PCANN & $0.0988$ & $0.0993$ & $0.1171$ & $0.3525$ & $0.5645$ & $0.0983$ & \cellcolor{gray!20}$0.0983$ & \cellcolor{gray!20}$0.0984$ & \cellcolor{gray!20}$0.0998$ & \cellcolor{gray!20}$0.1039$ \\
			PCALin & $0.0303$ & \cellcolor{gray!20}$0.0379$ & \cellcolor{gray!20}$0.0737$ & $0.3134$ & $0.6228$ & $0.2191$ & $0.2191$ & $0.2192$ & $0.2231$ & $0.2348$ \\
			FNO & $0.0299$ & $0.2966$ & $2.6444$ & $10.5499$ & $19.9952$ & $0.1341$ & $0.1743$ & $0.5100$ & $3.3888$ & $7.5197$ \\
			U-FNO & $0.0292$ & $0.2102$ & $1.1645$ & $4.3941$ & $8.3548$ & $0.0896$ & $0.1202$ & $0.5896$ & $1.4194$ & $2.2052$ \\
			MWT & $0.0316$ & $0.2201$ & $1.2716$ & $4.1969$ & $7.6609$ & \cellcolor{gray!20}$0.0865$ & $0.1123$ & $0.3668$ & $0.6949$ & $0.8620$ \\
			DeepONet & $0.0983$ &$0.0985$ &$0.1048$ &$0.2059$ &$0.3748$ & $0.2219$ &$0.2223$ & $0.2224$&$0.2262$ & $0.2375$ \\\midrule%\hline
			PINO & \cellcolor{gray!20}$0.0263$ & $0.3457$ & $3.3960$ & $16.8352$ & $33.8227$ & $0.1703$ & $0.1789$ & $1.0438$ & $6.1264$ & $12.3570$ \\
			PI-DeepONet & $0.1074$ & $0.1075$ & $0.1120$ & \cellcolor{gray!20}$0.1917$ & \cellcolor{gray!20}$0.3352$ & $0.2706$ & $0.2706$ & $0.2706$ & $0.2714$ & $0.2737$\\
			\bottomrule%\hline
		\end{tabular}
		%}
	\caption{Effects of noise on the solution for the inverse problems on a $65 \times 65$ resolution. The network is trained with noise-free data, but evaluated with noisy data. Backward operator training is used for solving the inverse problem.}
	\label{tab:noise-backward-operator-65}
\end{table}

\begin{table}[!ht]\small
	\centering
	%\resizebox{\textwidth}{!}{
		\begin{tabular}{r|ccccc|ccccc}
			\multicolumn{1}{c}{} & \multicolumn{5}{c|}{\textit{Poisson}}& \multicolumn{5}{c}{\textit{Darcy Flow PWC}}\\ \cline{2-11}
			Noise level & $0 \%$ & $0.1 \%$ & $1 \%$ & $5 \%$ & $10 \%$ & $0 \%$ & $0.1 \%$ & $1 \%$ & $5 \%$  & $10 \%$\\
			\hlineB{3}\midrule%\hline
			PINN & $0.1715$ & $0.1727$& $0.1734$& $0.2110$ & $0.2378$ & $0.2413$ & $0.2407$& $0.2496$& $0.2564$ & $0.2893$ \\
			QRES & $0.1598$ & $0.1658$ & $0.1714$& $0.2007$ & $0.2256$ & $0.2371$ & $0.2421$ & $0.2617$ & $0.2789$ & $0.2951$\\\midrule 
			PCANN & $0.0988$ & $0.0991$ & $0.1293$ & $0.1876$ & $0.2273$ & $0.0983$ & $0.0987$ & \cellcolor{gray!20}$0.0990$ & \cellcolor{gray!20}$0.1012$ & \cellcolor{gray!20}$0.1093$ \\
			PCALin & $0.0303$ & \cellcolor{gray!20}$0.0318$ & \cellcolor{gray!20}$0.0840$ & $0.1478$ & $0.1846$ & $0.2191$ & $0.2191$ & $0.2191$ & $0.2217$ & $0.2315$ \\
			FNO & $0.0299$ & $0.0554$ & $0.0957$ & \cellcolor{gray!20}$0.1389$ & \cellcolor{gray!20}$0.1678$ & $0.1341$ & $0.1344$ & $0.1449$ & $0.1770$ & $0.2019$ \\
			U-FNO & $0.0292$ & $0.0594$ & $0.1036$ & $0.1633$ & $0.1839$ & $0.0896$ & $0.0902$ & $0.1219$ & $0.1649$ & $0.1932$ \\
			MWT & $0.0316$ & $0.0533$ & $0.0966$ & $0.1525$ & $0.1867$ & \cellcolor{gray!20}$0.0865$ & \cellcolor{gray!20}$0.0893$ & $0.1115$ & $0.1634$ & $0.1956$ \\
			DeepONet & $0.0983$ & $0.1042$& $0.1084$ &$0.1537$ &$0.1870$ & $0.2219$ & $0.2283$ &$0.2273$ &$0.2379$ &$0.2522$ \\\midrule%\hline
			PINO & \cellcolor{gray!20}$0.0263$ & $0.0570$ & $0.0957$ & \cellcolor{gray!20}$0.1389$ & \cellcolor{gray!20}$0.1678$  & $0.1703$ & $0.1737$ & $0.2083$ & $0.5147$ & $0.9699$ \\
			PI-DeepONet & $0.1074$ & $0.1086$ & $0.1137$ & $0.1516$ & $0.1834$ & $0.2706$ & $0.2680$ & $0.2703$ & $0.2735$ & $0.2742$ \\
			\bottomrule%\hline
		\end{tabular}
		%}
	\caption{Effects of noise on backward operator inverse problems on a $65 \times 65$ resolution. The network is trained with noisy data and datasets with the same noise level are used for testing.}
	\label{tab:noise-backward-operator-trained-noise-65}
\end{table}

\begin{table}[htp]\small %\footnotesize%
	%\hfill
	\begin{subtable}[htp]{\textwidth}
		\centering    
		\begin{tabular}{@{\hskip 0.0in} r @{\hskip 0.05in} | @{\hskip 0.05in} c @{\hskip 0.08in} c @{\hskip 0.08in} c @{\hskip 0.08in} c @{\hskip 0.08in} c @{\hskip 0.0in}}
			& $0 \%$ & $0.1 \%$ & $1 \%$ & $5 \%$& $10 \%$ \\
			\hlineB{3}\midrule%\hline
			% PINN &  &  &  & &\\
			% QRES &  &  &  & &\\ \midrule 
			PCANN & $0.2128$ & $0.2128$ & $0.2137$ & $0.2975$ & $0.4468$ \\
			PCALin & $0.0803$ & $0.0805$ & $0.0918$ & $0.1534$ & $0.2132$ \\
			FNO & $0.0931$ & $0.0936$ & $0.1082$ & $0.1450$ & \cellcolor{gray!20}$0.1693$ \\
			U-FNO & $0.0687$ & $0.0697$ & $0.0909$ & $0.1397$ & $0.1802$ \\
			MWT & $0.0708$ & $0.0714$ & \cellcolor{gray!20}$0.0878$ & \cellcolor{gray!20}$0.1368$ & $0.1796$ \\
			DeepONet & $0.1341$ & $0.1330$ & $0.1350$ & $0.1660$ & $0.2088$  \\\midrule%\hline
			PINO & \cellcolor{gray!20} $0.0598$ & \cellcolor{gray!20} $0.0623$ & $0.0890$ & $0.1441$ & $0.1915$ \\
			PI-DeepONet &$0.1573$ &$0.1580$ &$0.1727$ & $0.2040$& $0.2298$\\
			%PI-DeepONet-Pi &$0.1090$ &$0.1092$ &$0.1159$ & $0.1424$& $0.1707$\\
			\bottomrule%\hline
		\end{tabular}
		\caption{$\lambda_{\text{err}}$, the relative error of the learned parameter}
		\label{tab:lambdaErr-noise-poisson-65}
	\end{subtable}
	\newline
	\vspace*{0.5cm}
	\newline
	\begin{subtable}[htp]{0.48\textwidth}
		\centering    
		\begin{tabular}{@{\hskip 0.0in} r @{\hskip 0.05in} | @{\hskip 0.05in} c @{\hskip 0.08in} c @{\hskip 0.08in} c @{\hskip 0.08in} c @{\hskip 0.08in} c @{\hskip 0.0in}}
			& $0 \%$ & $0.1 \%$ & $1 \%$ & $5 \%$& $10 \%$ \\
			\hlineB{3}\midrule%\hline
			% PINN &  &  &  & &\\
			% QRES &  &  &  & &\\ \midrule 
			PCANN & $0.0133$ & $0.0128$ & $0.0130$ & $0.0123$ & \cellcolor{gray!20}$0.0122$ \\
			PCALin & $0.0010$ & \cellcolor{gray!20}$0.0010$ & \cellcolor{gray!20}$0.0016$ & \cellcolor{gray!20}$0.0074$ & $0.0158$ \\
			FNO & $0.0022$ & $0.0025$ & $0.0097$ & $0.0464$ & $0.0924$ \\
			U-FNO & $0.0009$ & $0.0013$ & $0.0091$ & $0.0457$ & $0.0914$ \\
			MWT & $0.0014$ & $0.0017$ & $0.0093$ & $0.0459$ & $0.0917$ \\
			DeepONet &$0.0053$ &$0.0054$ &$0.0113$ & $0.0497$&$0.0988$ \\\midrule%\hline
			PINO & \cellcolor{gray!20}$0.0006$ & \cellcolor{gray!20}$0.0010$ & $0.0088$ & $0.0454$ & $0.0911$ \\
			PI-DeepONet &$0.0045$ &$0.0047$ &$0.0135$ &$0.0520$&$0.1013$\\
			%PI-DeepONet-Pi &$0.0042$ &$0.0043$ &$0.0110$ &$0.0500$&$0.0993$\\
			\bottomrule%\hline
		\end{tabular}
		\caption{$\tilde u_{\text{err}}$, the relative error between the solution of the learned parameter and the noisy solution.}
		\label{tab:unoisy-noise-poisson-65}
	\end{subtable}
	\hfill
	\begin{tabular}{|llllll|}
	\end{tabular}
	\begin{subtable}[htp]{0.48\textwidth}
		\centering    
		\begin{tabular}{@{\hskip 0.0in} r @{\hskip 0.05in} | @{\hskip 0.05in} c @{\hskip 0.08in} c @{\hskip 0.08in} c @{\hskip 0.08in} c @{\hskip 0.08in} c @{\hskip 0.0in}}
			& $0 \%$ & $0.1 \%$ & $1 \%$ & $5 \%$& $10 \%$ \\
			\hlineB{3}\midrule%\hline
			% PINN &  &  &  & &\\
			% QRES &  &  &  & &\\ \midrule 
			PCANN & $0.0133$ & $0.0128$ & $0.0131$ & $0.0150$ & $0.0209$ \\
			PCALin & $0.0010$ & $0.0010$ & \cellcolor{gray!20}$0.0020$ & $0.0074$ & $0.0133$ \\
			FNO & $0.0022$ & $0.0023$ & $0.0040$ & $0.0089$ & $0.0133$ \\
			U-FNO & $0.0009$ & $0.0010$ & $0.0024$ & \cellcolor{gray!20}$0.0072$ & \cellcolor{gray!20}$0.0124$ \\
			MWT & $0.0014$ & $0.0015$ & $0.0025$ & \cellcolor{gray!20}$0.0072$ & \cellcolor{gray!20}$0.0124$ \\
			DeepONet &$0.0053$ &$0.0053$ &$0.0055$ &$0.0085$& $0.0137$ \\\midrule%\hline
			PINO & \cellcolor{gray!20}$0.0006$ & \cellcolor{gray!20}$0.0008$ & $0.0023$ & \cellcolor{gray!20}$0.0072$ & $0.0126$ \\
			PI-DeepONet &$0.0045$&$0.0046$ &$0.0058$ &$0.0147$ &$0.0152$\\
			%PI-DeepONet-Pi &$0.0042$&$0.0042$ &$0.0048$ &$0.0087$ &$0.0132$\\
			\bottomrule%\hline
		\end{tabular}
		\caption{$u_{\text{err}}$, the relative error between the solution of the learned parameter and the noiseless solution.}
		\label{tab:unoiseless-noise-poisson-65}
	\end{subtable}
	\caption{Effects of noise on Tikhonov-based inverse problems for the Poisson problem and using datasets with a resolution of $65 \times 65$. The errors shown here are averaged over 100 test samples. }
	\label{tab:xxnoise-poisson-65}
\end{table}

\begin{table}[htp]\small %\footnotesize%
	\begin{subtable}[htp]{0.48\textwidth}
		\centering    
		\begin{tabular}{@{\hskip 0.0in} r @{\hskip 0.05in} | @{\hskip 0.05in} c @{\hskip 0.08in} c @{\hskip 0.08in} c @{\hskip 0.08in} c @{\hskip 0.08in} c @{\hskip 0.0in}}
			& $0 \%$ & $0.1 \%$ & $1 \%$ & $5 \%$& $10 \%$ \\
			\hlineB{3}\midrule%\hline
			% PINN &  &  &  & &\\
			% QRES &  &  &  & &\\ \midrule 
			PCANN & $77.74$ & $77.70$ & $77.83$ & $76.79$ & $74.19$ \\
			PCALin & $90.58$ & $90.58$ & $90.57$ & $90.55$ & $90.42$\\
			FNO & $96.92$ & $96.92$ & $96.71$ & $95.45$ & $94.30$ \\
			U-FNO & $94.23$ & $93.83$ & $95.20$ & $94.14$ & $92.48$ \\
			MWT & \cellcolor{gray!20}$98.26$ & \cellcolor{gray!20}$98.33$ & \cellcolor{gray!20}$97.84$ & \cellcolor{gray!20}$96.42$ & $95.04$ \\
			DeepONet &$93.88$ & $93.88$& $93.95$& $93.49$& $92.70$ \\\midrule%\hline
			PINO & $97.56$ & $97.59$ & $97.45$ & $96.34$ & \cellcolor{gray!20}$95.07$ \\
			PI-DeepONet &$92.03$ &$91.93$ &$91.84$ &$91.80$ &$91.60$\\
			%PI-DeepONet-Pi &$89.48$ &$89.68$ &$89.69$ &$89.34$ &$88.84$\\
			\bottomrule%\hline
		\end{tabular}
		\caption{$\lambda_{\text{acc}} (\%)$, the accuracy of the learned parameter}
		\label{tab:lambdaAcc-noise-darcypwc-65}
	\end{subtable} \hfill
	\begin{subtable}[htp]{0.48\textwidth}
		\centering    
		\begin{tabular}{@{\hskip 0.0in} r @{\hskip 0.05in} | @{\hskip 0.05in} c @{\hskip 0.08in} c @{\hskip 0.08in} c @{\hskip 0.08in} c @{\hskip 0.08in} c @{\hskip 0.0in}}
			& $0 \%$ & $0.1 \%$ & $1 \%$ & $5 \%$& $10 \%$ \\
			\hlineB{3}\midrule%\hline
			% PINN &  &  &  & &\\
			% QRES &  &  &  & &\\ \midrule 
			PCANN & $0.4684$ & $0.4692$ & $0.4680$ & $0.4828$ & $0.5148$ \\
			PCALin & $0.3157$ & $0.3157$ & $0.3159$ & $0.3163$ & $0.3183$ \\
			FNO & $0.1764$ & $0.1764$ & $0.1823$ & $0.2143$ & $0.2402$ \\
			U-FNO & $0.2140$ & $0.2205$ & $0.2017$ & $0.2371$ & $0.2734$ \\
			MWT & \cellcolor{gray!20}$0.1287$ & \cellcolor{gray!20}$0.1250$ & \cellcolor{gray!20}$0.1435$ & \cellcolor{gray!20}$0.1880$ & \cellcolor{gray!20}$0.2209$ \\
			DeepONet & $0.2484$ & $0.2483$ & $0.2469$ & $0.2547$ & $0.2691$ \\\midrule%\hline
			PINO & $0.1559$ & $0.1549$ & $0.1607$ & $0.1932$ & $0.2244$ \\
			PI-DeepONet &$0.2792$ &$0.2801$ &$0.2820$ &$0.2822$&$0.2861$\\
			%PI-DeepONet-Pi &$0.3259$ &$0.3228$ &$0.3230$ &$0.3288$&$0.3373$\\
			\bottomrule%\hline
		\end{tabular}
		\caption{$\lambda_{\text{err}}$, the relative error of the learned parameter}
		\label{tab:lambdaErr-noise-darcypwc-65}
	\end{subtable}
	\newline
	\vspace*{0.5cm}
	\newline
	\begin{subtable}[htp]{0.48\textwidth}
		\centering    
		\begin{tabular}{@{\hskip 0.0in} r @{\hskip 0.05in} | @{\hskip 0.05in} c @{\hskip 0.08in} c @{\hskip 0.08in} c @{\hskip 0.08in} c @{\hskip 0.08in} c @{\hskip 0.0in}}
			& $0 \%$ & $0.1 \%$ & $1 \%$ & $5 \%$& $10 \%$ \\
			\hlineB{3}\midrule%\hline
			% PINN &  &  &  & &\\
			% QRES &  &  &  & &\\ \midrule 
			PCANN & $0.1151$ & $0.1152$ & $0.1144$ & $0.1211$ & $0.1373$ \\
			PCALin & $0.0929$ & $0.0929$ & $0.0929$ & $0.0931$ & \cellcolor{gray!20}$0.0934$ \\
			FNO & \cellcolor{gray!20}$0.0104$ & \cellcolor{gray!20}$0.0105$ & \cellcolor{gray!20}$0.0174$ & $0.0711$ & $0.1362$ \\
			U-FNO & $0.0477$ & $0.0517$ & $0.0419$ & $0.0698$ & $0.1145$ \\
			MWT & $0.0129$ & $0.0125$ & $0.0177$ & \cellcolor{gray!20}$0.0544$ & $0.1032$ \\
			DeepONet &$0.0480$ &$0.0484$ &$0.0495$ &$0.0773$&$0.1190$  \\\midrule%\hline
			PINO & $0.0167$ & $0.0164$ & $0.0211$ & $0.0567$ & $0.1065$ \\
			PI-DeepONet &$0.0586$ &$0.0593$ &$0.0604$ &$0.0816$&$0.1244$\\ 
			%PI-DeepONet-Pi &$0.0707$ &$0.0697$ &$0.0719$ &$0.0894$&$0.1272$\\
			\bottomrule%\hline
		\end{tabular}
		\caption{$\tilde u_{\text{err}}$, the relative error between the solution of the learned parameter and the noisy solution.}
		\label{tab:unoisy-noise-darcypwc-65}
	\end{subtable}
	\hfill
	\begin{subtable}[htp]{0.48\textwidth}
		\centering    
		\begin{tabular}{@{\hskip 0.0in} r @{\hskip 0.05in} | @{\hskip 0.05in} c @{\hskip 0.08in} c @{\hskip 0.08in} c @{\hskip 0.08in} c @{\hskip 0.08in} c @{\hskip 0.0in}}
			& $0 \%$ & $0.1 \%$ & $1 \%$ & $5 \%$& $10 \%$ \\
			\hlineB{3}\midrule%\hline
			% PINN &  &  &  & &\\
			% QRES &  &  &  & &\\ \midrule 
			PCANN & $0.1151$ & $0.1152$ & $0.1144$ & $0.1207$ & $0.1363$ \\
			PCALin & $0.0929$ & $0.0929$ & $0.0929$ & $0.0928$ & $0.0923$ \\
			FNO & $0.0210$ & $0.0212$ & $0.0223$ & $0.0329$ & $0.0421$ \\
			U-FNO & $0.0477$ & $0.0517$ & $0.0395$ & $0.0437$ & $0.0496$ \\
			MWT & \cellcolor{gray!20}$0.0129$ & \cellcolor{gray!20}$0.0125$ & \cellcolor{gray!20}$0.0142$ & \cellcolor{gray!20}$0.0209$ & \cellcolor{gray!20}$0.0276$ \\
			DeepONet & $0.0480$&$0.0484$ &$0.0483$ &$0.0561$&$0.0597$ \\\midrule%\hline
			PINO & $0.0167$ & $0.0164$ & $0.0181$ & $0.0250$ & $0.0349$ \\
			PI-DeepONet &$0.0586$ &$0.0593$ &$0.0593$ &$0.0614$&$0.0677$\\
			%PI-DeepONet-Pi &$0.0707$ &$0.0697$ &$0.0711$ &$0.0719$&$0.0748$\\
			\bottomrule%\hline
		\end{tabular}
		\caption{$u_{\text{err}}$, the relative error between the solution of the learned parameter and the noiseless solution.}
		\label{tab:unoiseless-noise-darcypwc-65}
	\end{subtable}
	\caption{Effects of noise on Tikhonov-based Inverse Problems on the Darcy Flow problem with PWC coefficients, using a dataset of resolution of $65 \times 65$. The errors shown here are averaged over 100 test samples.}
	\label{tab:xxnoise-darcypwc-65}
\end{table}

\begin{table}[htp]\small
	\begin{center}
		\begin{tblr}
			{colspec = {X[c]X[c,h]X[c,h]X[c,h]X[c,h]X[c,h]},
				stretch = 0,
				rowsep = 0pt,}
			& & & Truth & & \\
			& & & \includegraphics[width=0.15\textwidth]{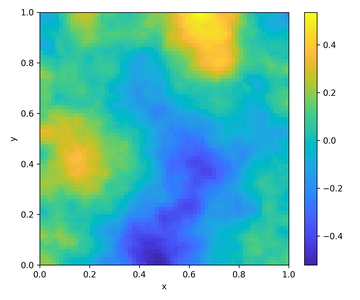} & & \\%\hline  
			& $0 \%$ & $0.1 \%$ & $1 \%$ & $5 \%$ & $10 \%$ \\
			PCANN & \includegraphics[width=0.15\textwidth]{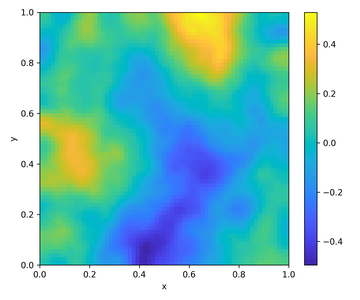} & \includegraphics[width=0.15\textwidth]{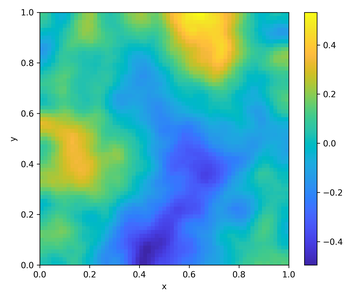} & \includegraphics[width=0.15\textwidth]{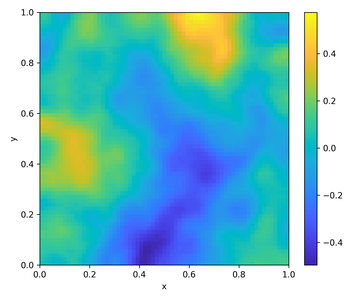} & \includegraphics[width=0.15\textwidth]{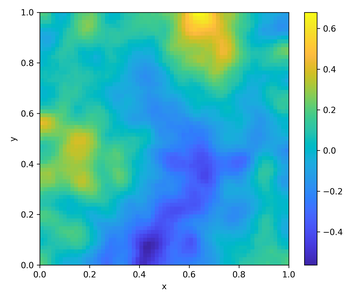} & \includegraphics[width=0.15\textwidth]{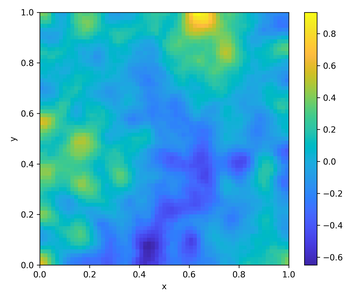}  \\
			PCALin & \includegraphics[width=0.15\textwidth]{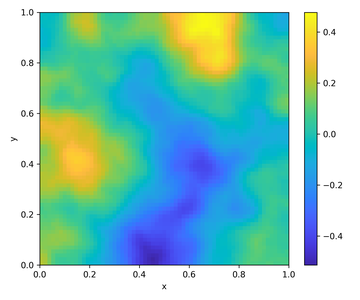} & \includegraphics[width=0.15\textwidth]{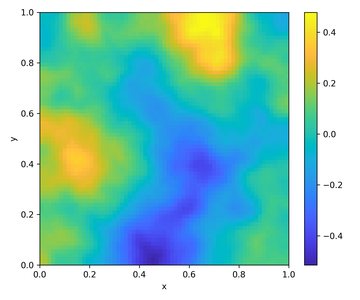} & \includegraphics[width=0.15\textwidth]{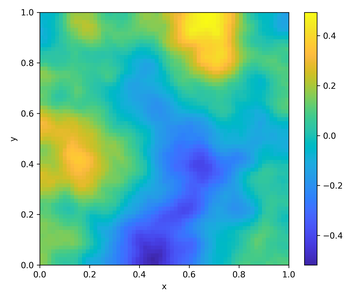} & \includegraphics[width=0.15\textwidth]{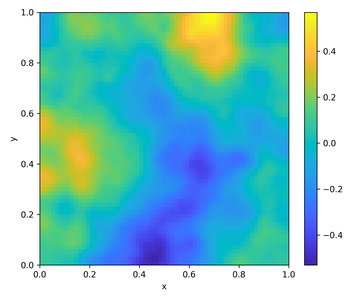} & \includegraphics[width=0.15\textwidth]{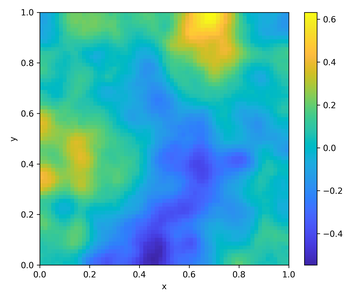}  \\
			FNO &  \includegraphics[width=0.15\textwidth]{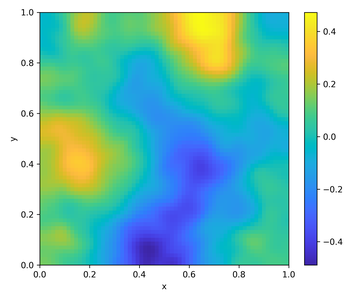} & \includegraphics[width=0.15\textwidth]{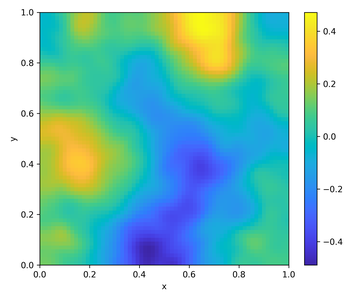} & \includegraphics[width=0.15\textwidth]{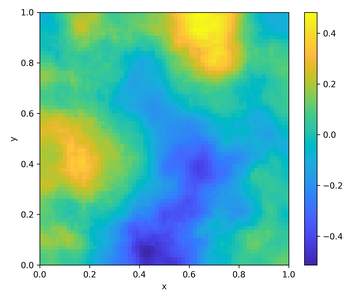} & \includegraphics[width=0.15\textwidth]{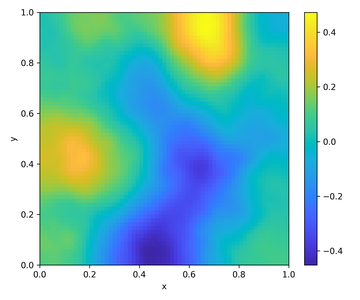} & \includegraphics[width=0.15\textwidth]{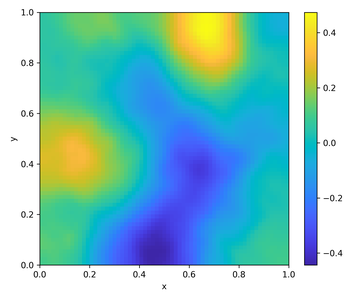}  \\
			U-FNO  & \includegraphics[width=0.15\textwidth]{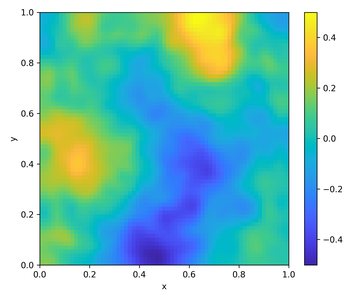} & \includegraphics[width=0.15\textwidth]{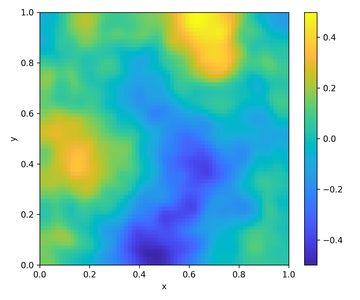} & \includegraphics[width=0.15\textwidth]{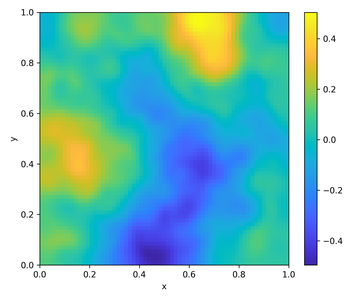} & \includegraphics[width=0.15\textwidth]{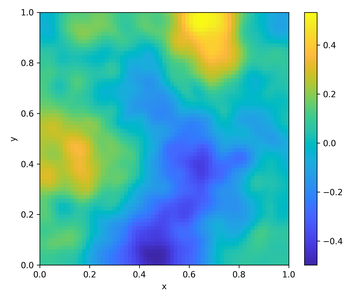} & \includegraphics[width=0.15\textwidth]{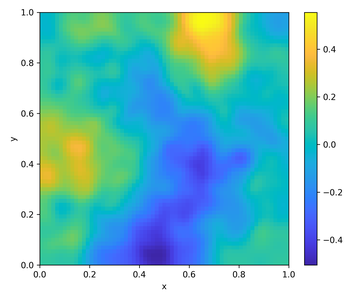}  \\
			MWT  & \includegraphics[width=0.15\textwidth]{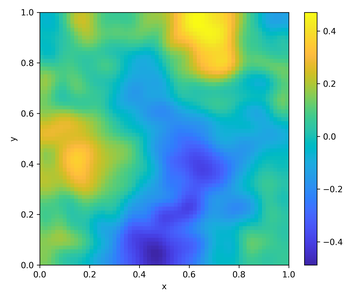} & \includegraphics[width=0.15\textwidth]{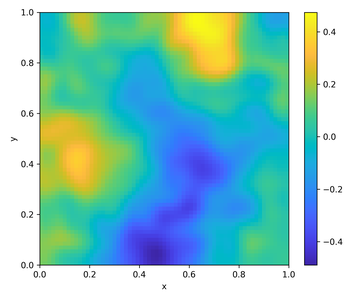} & \includegraphics[width=0.15\textwidth]{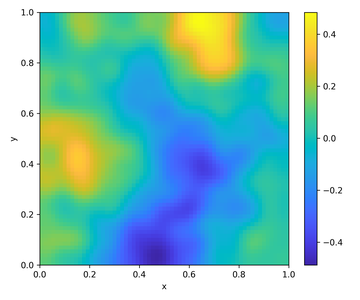} & \includegraphics[width=0.15\textwidth]{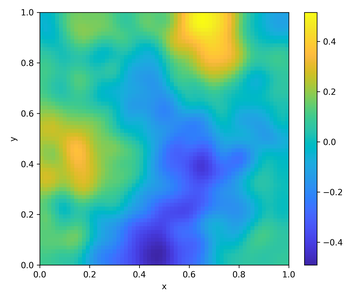} & \includegraphics[width=0.15\textwidth]{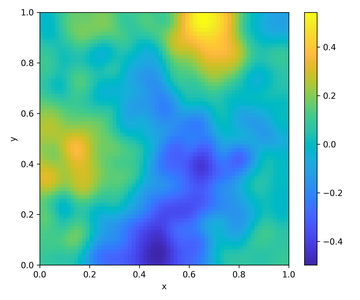}  \\
			DeepONet  & \includegraphics[width=0.15\textwidth]{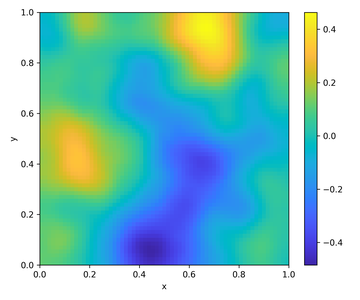} & \includegraphics[width=0.15\textwidth]{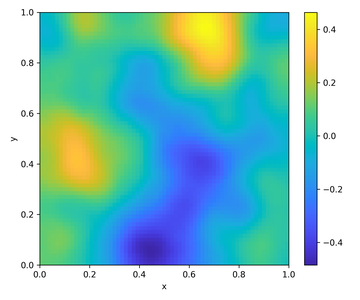} & \includegraphics[width=0.15\textwidth]{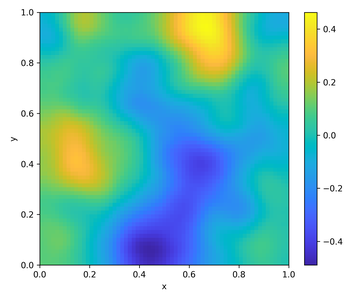} & \includegraphics[width=0.15\textwidth]{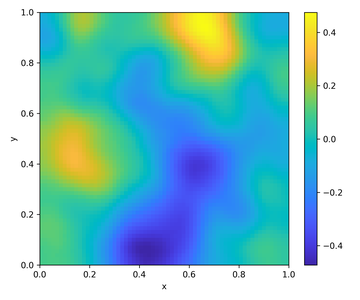} & \includegraphics[width=0.15\textwidth]{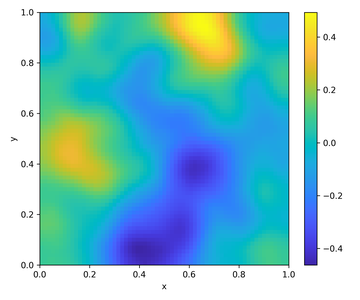}  \\
			PINO  & \includegraphics[width=0.15\textwidth]{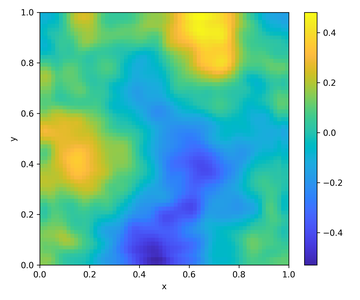} & \includegraphics[width=0.15\textwidth]{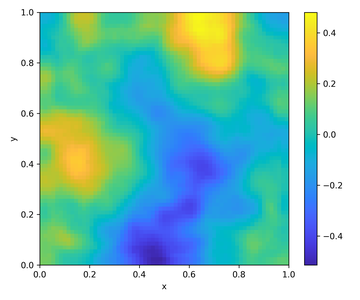} & \includegraphics[width=0.15\textwidth]{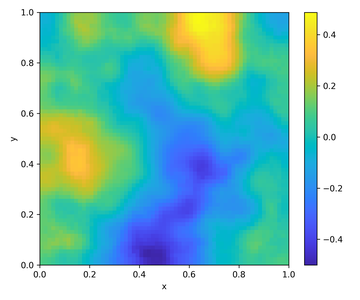} & \includegraphics[width=0.15\textwidth]{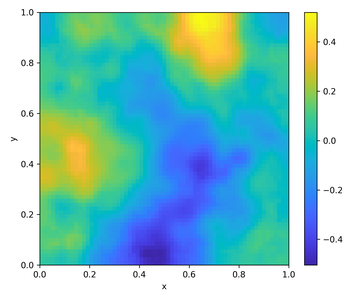} & \includegraphics[width=0.15\textwidth]{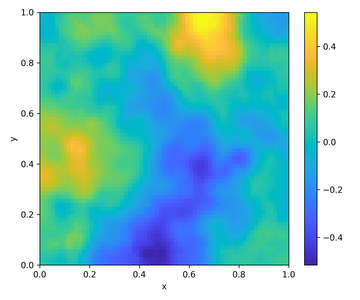}  \\
			PI-DeepONet  & \includegraphics[width=0.15\textwidth]{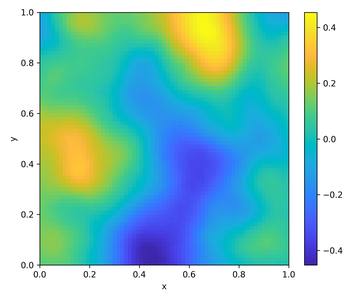} & \includegraphics[width=0.15\textwidth]{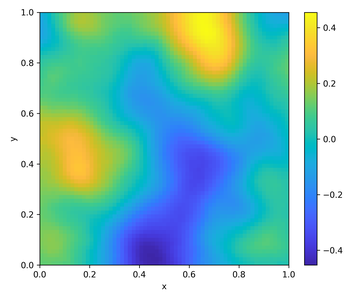} & \includegraphics[width=0.15\textwidth]{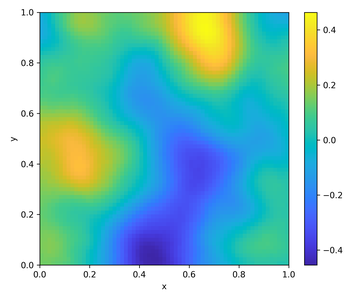} & \includegraphics[width=0.15\textwidth]{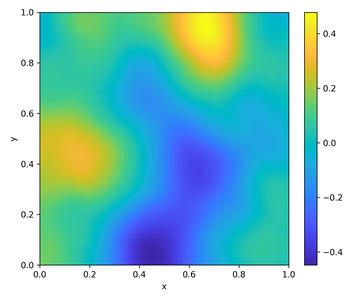} & \includegraphics[width=0.15\textwidth]{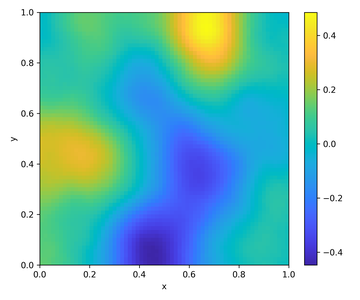}
		\end{tblr}
		\caption{Effects of noise on Tikhonov-based Inverse Problems on the Poisson problem, using a dataset of, resolution of $65 \times 65$.}
		\label{tab:fig-xxnoise-poisson-65}
	\end{center}
\end{table}

\begin{table}[htp]\small
	\begin{center}
		\begin{tblr}
			{colspec = {X[c]X[c,h]X[c,h]X[c,h]X[c,h]X[c,h]},
				stretch = 0,
				rowsep = 0pt,
				colsep = 0pt,}
			& & & Truth & & \\
			& & & \includegraphics[width=0.15\textwidth]{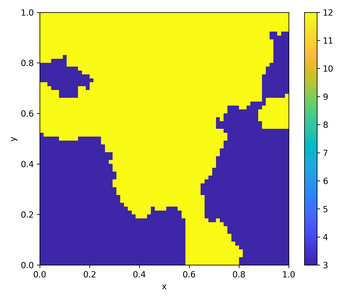} & & \\%\hline  
			& $0 \%$ & $0.1 \%$ & $1 \%$ & $5 \%$ & $10 \%$ \\
			PCANN & \includegraphics[width=0.15\textwidth]{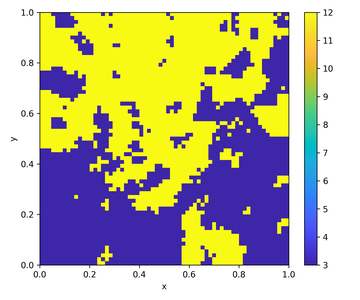} & \includegraphics[width=0.15\textwidth]{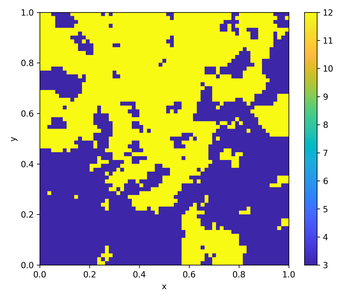} & \includegraphics[width=0.15\textwidth]{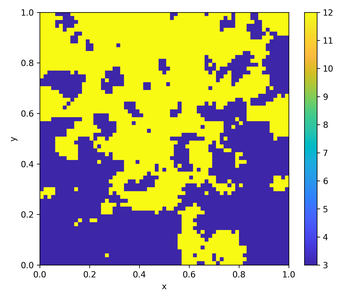} & \includegraphics[width=0.15\textwidth]{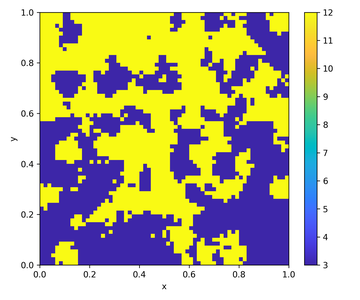} & \includegraphics[width=0.15\textwidth]{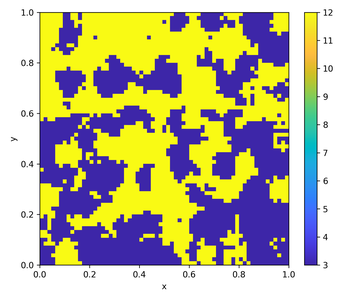}  \\
			PCALin & \includegraphics[width=0.15\textwidth]{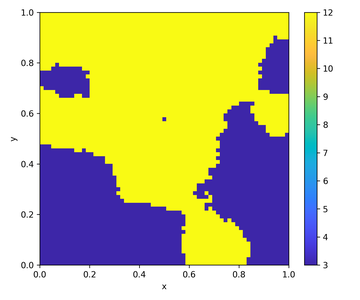} & \includegraphics[width=0.15\textwidth]{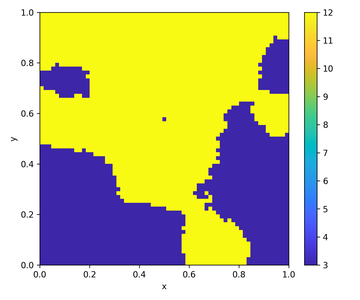} & \includegraphics[width=0.15\textwidth]{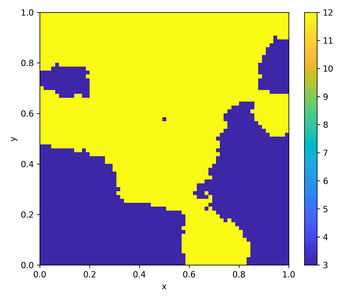} & \includegraphics[width=0.15\textwidth]{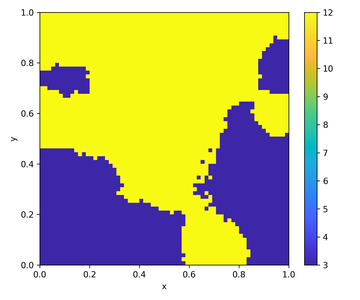} & \includegraphics[width=0.15\textwidth]{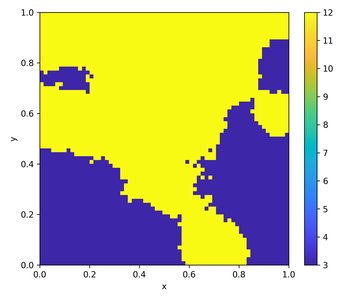}  \\
			FNO & \includegraphics[width=0.15\textwidth]{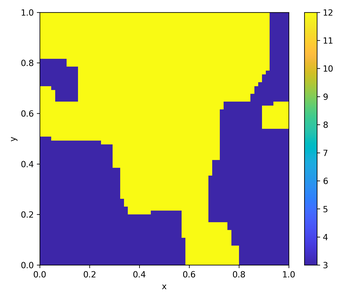} & \includegraphics[width=0.15\textwidth]{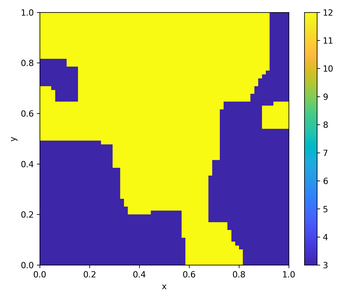} & \includegraphics[width=0.15\textwidth]{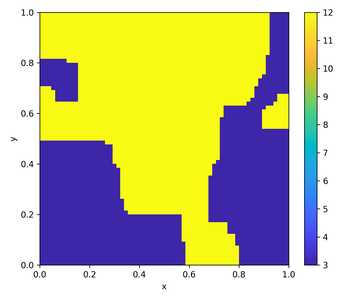} & \includegraphics[width=0.15\textwidth]{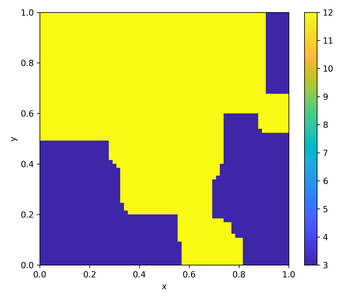} & \includegraphics[width=0.15\textwidth]{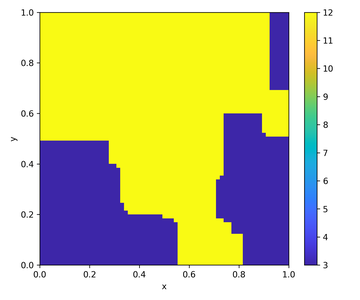}  \\
			U-FNO & \includegraphics[width=0.15\textwidth]{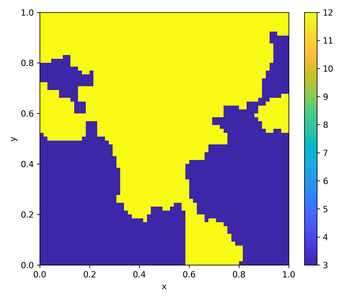} & \includegraphics[width=0.15\textwidth]{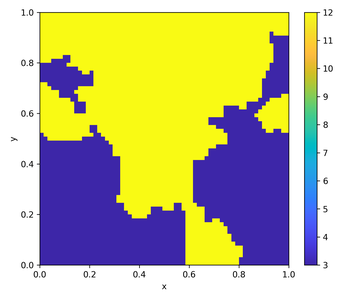} & \includegraphics[width=0.15\textwidth]{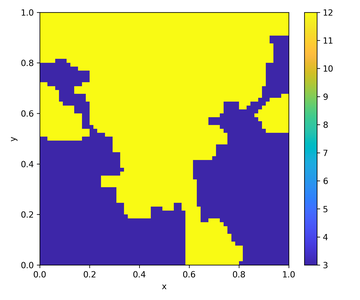} & \includegraphics[width=0.15\textwidth]{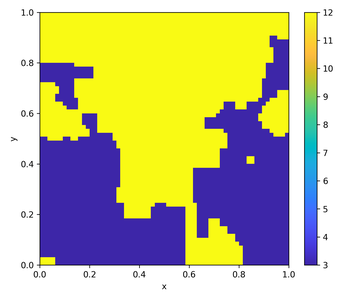} & \includegraphics[width=0.15\textwidth]{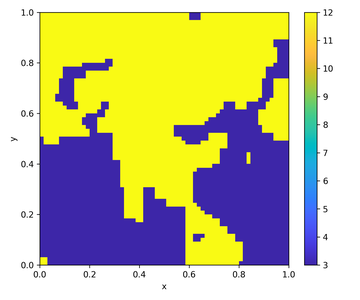}  \\
			MWT & \includegraphics[width=0.15\textwidth]{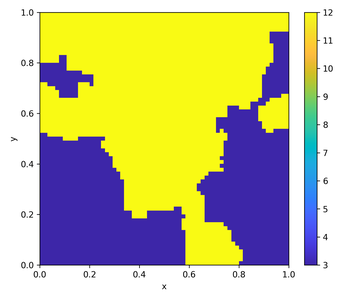} & \includegraphics[width=0.15\textwidth]{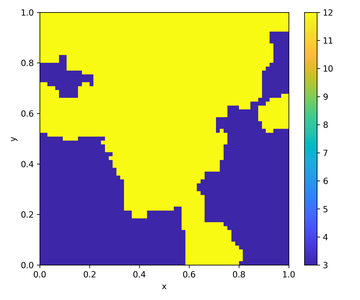} & \includegraphics[width=0.15\textwidth]{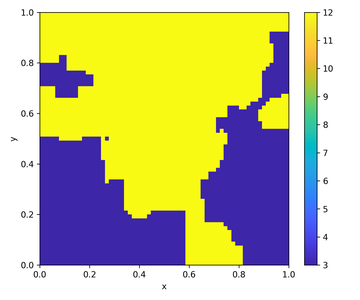} & \includegraphics[width=0.15\textwidth]{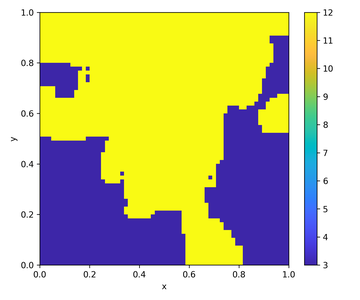} & \includegraphics[width=0.15\textwidth]{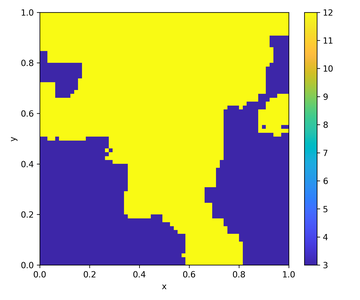}  \\
			DeepONet & \includegraphics[width=0.15\textwidth]{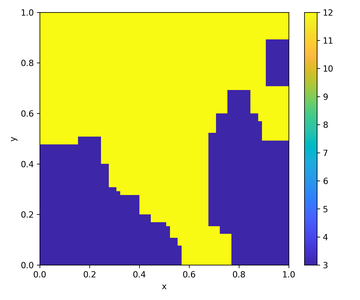} & \includegraphics[width=0.15\textwidth]{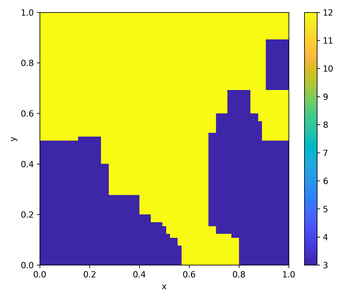} & \includegraphics[width=0.15\textwidth]{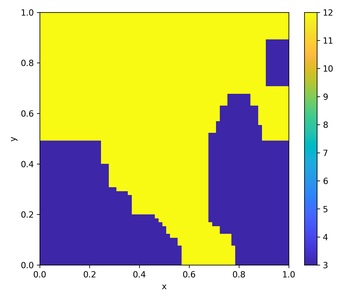} & \includegraphics[width=0.15\textwidth]{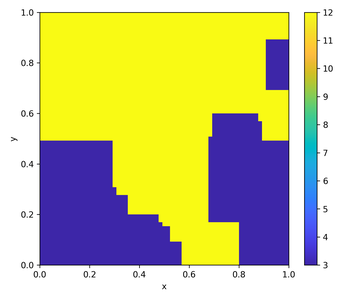} & \includegraphics[width=0.15\textwidth]{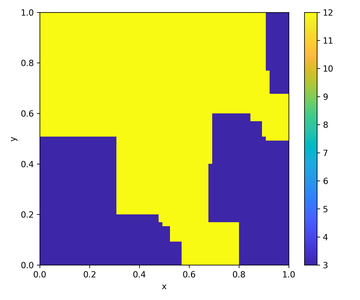}  \\
			PINO & \includegraphics[width=0.15\textwidth]{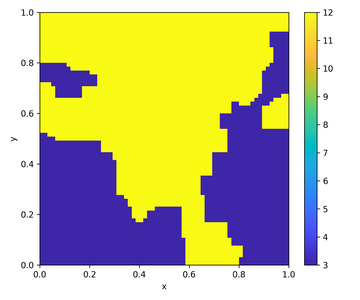} & \includegraphics[width=0.15\textwidth]{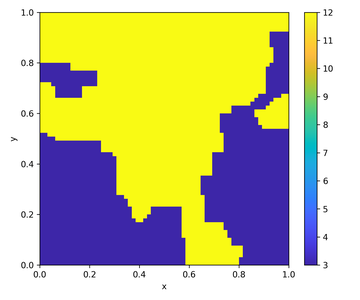} & \includegraphics[width=0.15\textwidth]{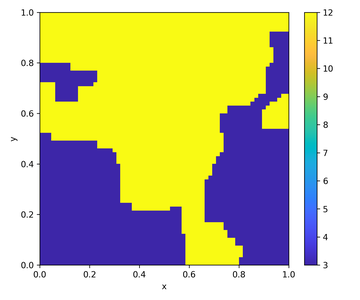} & \includegraphics[width=0.15\textwidth]{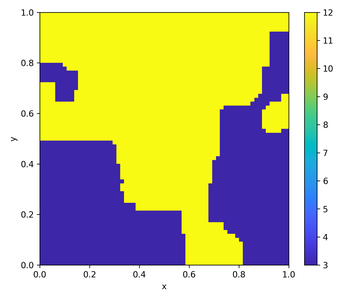} & \includegraphics[width=0.15\textwidth]{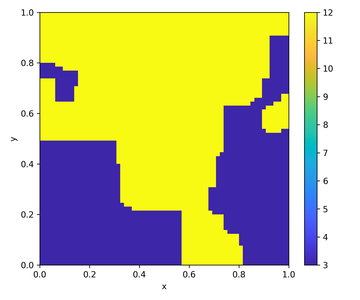}  \\
			PI-DeepONet & \includegraphics[width=0.15\textwidth]{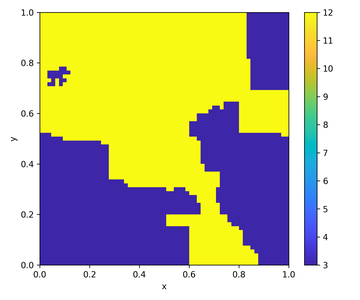} & \includegraphics[width=0.15\textwidth]{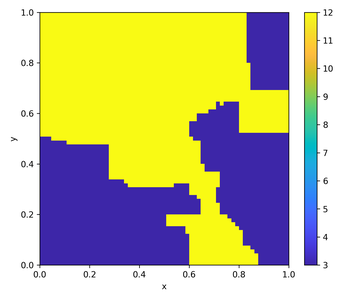} & \includegraphics[width=0.15\textwidth]{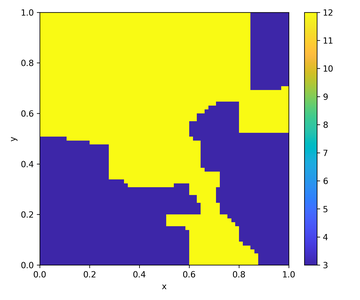} & \includegraphics[width=0.15\textwidth]{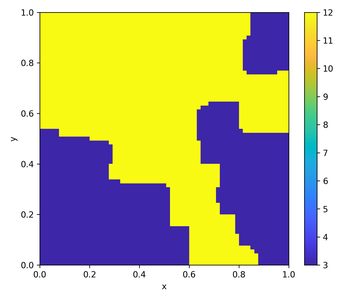} & \includegraphics[width=0.15\textwidth]{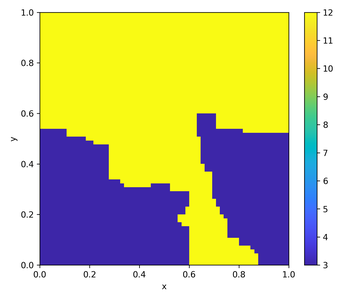}
		\end{tblr}
		\caption{Effects of noise on Tikhonov-based Inverse Problems on the Darcy Flow pwc problem, using dataset of, resolution of $65 \times 65$.}
		\label{tab:fig-xxnoise-darcypwc-65}
	\end{center}
\end{table}

% \begin{table}[htp]\small
	% 	\centering
	% 	%\resizebox{\textwidth}{!}{
		%     \begin{tabular}{r|ccccc|ccccc}
			%           \multicolumn{1}{c}{} & \multicolumn{5}{c|}{\textit{Poisson}}& \multicolumn{5}{c}{\textit{Darcy Flow PWC}}\\ \cline{2-11}
			%           Noise level & $0 \%$ & $0.1 \%$ & $1 \%$ & $5 \%$ & $10 \%$ & $0 \%$ & $0.1 \%$ & $1 \%$ & $5 \%$  & $10 \%$\\
			%         \hlineB{3}\midrule%\hline
			%         PCANN & & & & & & & & & \\
			%         PCALin & & & & & & & & & \\
			%         FNO & & & & & & & & & \\
			%         U-FNO & & & & & & & & & \\
			%         MWT &  & & & & & & & & \\
			%         DeepONet & & & & & & & & & \\\midrule%\hline
			%         PINO & & & & & & & & & \\
			%         PI-DeepONet & & & & & & & & & \\
			%         \bottomrule%\hline
			%     \end{tabular}
		%     	%}
	% 	\caption{Effects of noise on Tikhonov-based Inverse Problems, using a resolution of  $513 \times 513$ }
	%     \label{tab:noise-poisson-513}
	% \end{table}

\subsection{Industrial Application Problems}
\subsubsection{A moving domain problem}
Digital twins are ranked among the most influential technological
developments of our decade, e.g. the Gartner report ranks digital twins or AI engineering amongst the  Top 10   strategic development trends in all reports since 2018, \cite{gartner2018, gartner2019}. These digital twins combine data-driven concepts with domain-specific expert knowledge, most often in the form of DL solvers for PDE models.

One particular characteristic, which sets industrial applications apart from typical academic examples, is the domain of definition.
In an industrial setting, these geometries tend to be complex, e.g. given by the shape of a production device or restricted measurement geometries.

Particularly challenging are applications, where differential equations have to be solved on moving domains, e.g. the heat development inside a rotating electrical engine. The implementation of classical mesh-based methods is rather involved, e.g. computing the system matrix typically requires non-standard concepts,   and they are expensive to solve. To tackle these problems, and other differential equations, with deep learning approaches, we developed, in collaboration with Bosch, the software package \textsc{TorchPhysics}, cf.~\cite{TorchPhysics}. 
One characteristic feature of TorchPhysics is an easy-to-use generator for complex domains and geometries: Based on subtraction, addition and cross-product operations more and more complex geometries can be created from basic constituents. Another characteristic is the unbiased choice amongst several DL coders for solving forward or inverse problems.

In this Section, we give a small insight into a problem we are currently investigating in collaboration with Bosch. 

\begin{figure}[htp]
	\centering
	\includegraphics[width=0.8\linewidth]{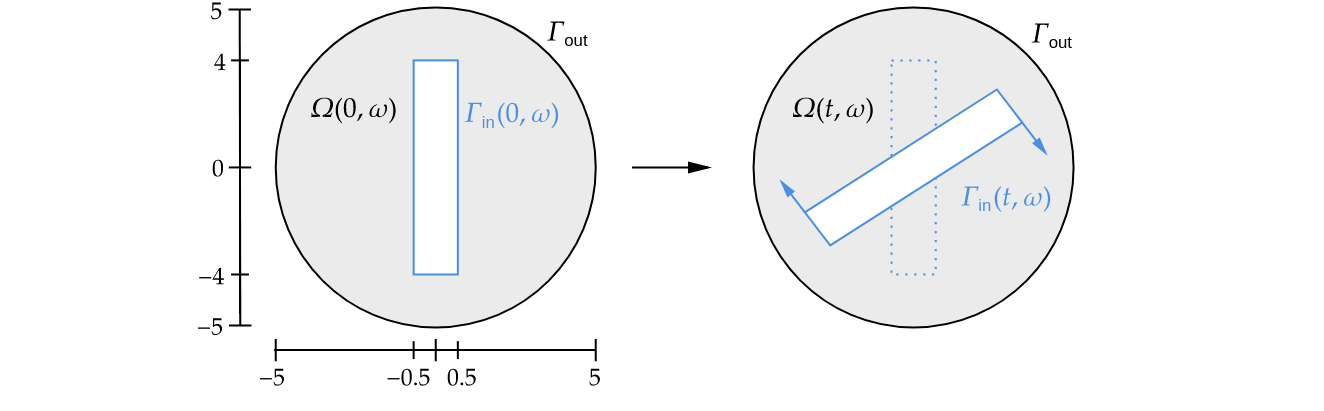}
	\caption{Time dependent rotating geometry in a circular housing}    \label{fig:bosch_domain}
\end{figure}

As an example of a moving domain, we consider the two-dimensional geometry is shown in Fig.~\ref{fig:bosch_domain}, which can be seen as a simplified sub-problem inside an electrical engine. The inside of the domain $\Omega(t, \omega) \subset \R^2$ is filled with fluid. We assume that the density of the incompressible fluid is $\rho = 1$ for simplicity. In the following, all physical values are in SI dimensions and they are not used. The blue outlined bar is rotating in time, inducing a circular motion onto the fluid. At the same time, the bar will heat up, and we also have to compute the temperature profile. Additionally, we are not only interested in one solution for a fixed angular velocity $\omega$. Instead, we aim to learn the solution operator, to approximate the system for different angular velocities, with one single neural network. 

We consider the time interval $I_t \coloneqq [0.0, 1.0]$ and interval for angular velocities $I_\omega \coloneqq [\pi, 2\pi]$. To model the fluid inside the domain we use the incompressible Navier–Stokes equations in two dimensions, cf.~\cite{LK16}:
\begin{equation}\label{eq:Bosch_ns}
	\begin{split}
		\partial_t u + (u \cdot \nabla) u &= \nu \Delta u - \nabla p \\ 
		\nabla \cdot u &= 0\,.
	\end{split}
\end{equation}
Here, $u = u(x,y,t;\omega) \in \R^2$ is the fluid velocity, $p = p(x,y,t;\omega) \in \R$ the pressure and $\nu \in \R^+$ the kinematic viscosity. 
For the modelling of the temperature $\theta = \theta(x,y,t;\omega) \in \R$ we consider a heat equation with forced convection due to rotating geometry, cf.~\cite{El88}:
\begin{align}\label{eq:Bosch_heat}
	\partial_t \theta + u \cdot \nabla \theta = a \Delta \theta
\end{align}
with thermal diffusion coefficient $a \in \R^+$.  Convection due to a temperature depending density is not considered due the assumption of incompressibility of the fluid. Initially, at $t=0$, we set $u=(0, 0)$, $p=0$ and assume a uniform temperature distribution $\theta = 270$ in the domain $\Omega(0, \omega)$. At the outer boundary $\Gamma_\text{out}$ we set a no-slip condition for the fluid flow and a Dirichlet condition for the temperature $\theta$:
\begin{alignat*}{2}
	u(x,y,t; \omega) &= (0, 0), \quad &&\text{for } (x,y,t,\omega) \in \Gamma_\text{out} \times I_t\times I_\omega \\
	\theta(x,y,t; \omega) &= 270\,, \quad &&\text{for } (x,y,t,\omega) \in \Gamma_\text{out} \times I_t \times I_\omega
\end{alignat*}
The rotation of the rod is given by the matrix
\begin{equation*}
	A(t; \omega) \coloneqq
	\left( 
	\begin{array}{cc}
		\cos(-\omega t (1 - e^{-\omega t})) & -\sin(-\omega t (1 - e^{-\omega t}))\\ \sin(-\omega t (1 - e^{-\omega t})) & \cos(-\omega t (1 - e^{-\omega t}))
	\end{array}
	\right),
\end{equation*} 
the factor $(1 - e^{-\omega t})$ is included to get a smooth start, which we scale with the rotation speed. Additionally  $-\omega$ is used to get a clockwise rotation. As a heat source, a Dirichlet condition for the temperature on $\Gamma_\text{in}$ is considered:
\begin{alignat}{2}\label{eq:Bosch_temp_bc}
	\theta(x,y,t; \omega) &= 270\ + 60 (1 - e^{-\omega t}), &&\text{ for } (x,y,t,\omega) \in \Gamma_\text{in} \times I_t \times I_\omega
\end{alignat}
For the fluid on $\Gamma_\text{in}$ we again apply a no-slip condition. Because of the moving boundary, this translates to
\begin{equation}\label{eq:Bosch_velocity_bc}
	u(x, y, t; \omega) = \omega \, d(x, y, t; \omega) \left( \begin{array}{cc}
		\cos(-\omega t (1 - e^{-\omega t}))\\ \sin(-\omega t (1 - e^{-\omega t})) 
	\end{array} \right) (1 - e^{-\omega t} + \omega t e^{-\omega t}), 
\end{equation}
for all $(x,y) \in \Gamma_\text{in}(t, \omega) \text{ and } (t,\omega) \in I_t \times I_\omega$. Here $d(x, y, t; \omega)$ corresponds to the signed distance of the point $(x, y)$ to the line perpendicular to the bar, which is running through the origin. This distance is given by:
\begin{equation*}
	d(x, y, t; \omega) \coloneqq  \left( \begin{array}{cc}
		x\\y
	\end{array} \right) \cdot \left(
	\begin{array}{cc}
		- \sin(-\omega t (1 - e^{-\omega t}))\\
		\cos(-\omega t (1 - e^{-\omega t}))
	\end{array} \right)
\end{equation*}
The distance has to be included, since points further away from the center have a higher radial velocity. 

This concludes the general description of the application problem. If classical PDE solvers such as FEM or FDM solvers are used, we have to compute a new solution for each velocity value $\omega$, that leads to costly simulations for each new parameter value. Furthermore, the most challenging part would be the implementation of discretisation meshes for the moving domain. Especially, to correctly transfer the discrete solution data between different time steps to guarantee conservation of mass, momentum and energy.  

By using a deep learning approach, we can try to learn the solutions for different values of $\omega$ in one single training process. Thus eliminating the need for new simulations for each $\omega$. 

This leads us to the crucial step of choosing a suitable DL method. 
In our case only the underlying differential equation is known, no data such as a solution for a particular parameter setting is available.  
Moreover, only scalar parameters appear, hence we can use the classical  PINN to learn the solution operator. Additionally, since PINNs are mesh-free, the implementation of the domain is easier, compared to the classical methods. One only has to correctly sample points inside the domain $\Omega(t, \omega)$ and on the boundary $\Gamma_\text{in}(t, \omega)$, for each combination of $(t,\omega) \in I_t \times I_\omega$, to train the needed conditions.

\begin{figure}[htp]
	\centering
	\includegraphics[width=0.7\linewidth]{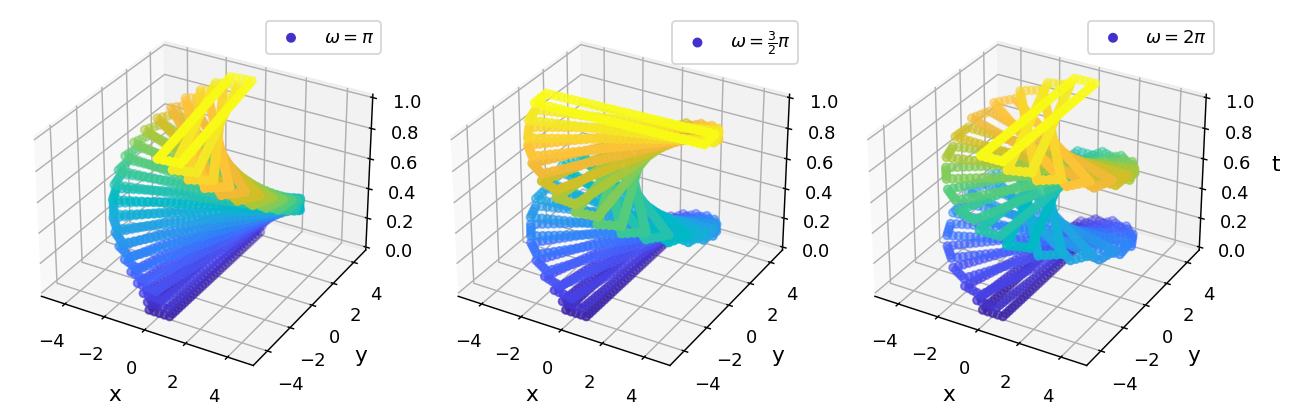}
	\caption{Example of sampled inner boundary points for the application problem. The points are only coloured for visual representation.}
	\label{fig:points_torchphysics}
\end{figure}

The library \textsc{TorchPhysics} was developed with this time (and parameter) dependent domains in mind. Therefore, the previously explained domain can be easily defined in  \textsc{TorchPhysics} and the needed points sampled. An example for sampled points on $\Gamma_\text{in}(t, \omega)$, for different $\omega$, is shown in Fig.~\ref{fig:points_torchphysics}. These points can then be created and used in each training step. The same is possible inside $\Omega(t, \omega)$. However, the visualisation of inner points is  more elaborate, so we do not show them here and refer any reader interested in more details to the TorchPhysics tutorial.
To learn the problem's solution operators, we define three different neural networks $u_\Theta, p_\Theta, \theta_\Theta$, for fluid velocity, pressure, and temperature respectively. Every network has space and time variables as input  $(x, y, t)$ and, since we want to train the solution for different $\omega$, also the angular velocity $\omega$. The network $u_\Theta$  has two output neurons, for the two velocity components.

Since the neural networks have to fulfil multiple conditions, this can lead to problems while training. I.e. different loss terms may have varying preferred minimisation directions. To somewhat simplify the optimisation process, we apply hard constraints to naturally fulfil some of the conditions and simplify the output range of our neural networks. These constrained networks are denoted by $\Tilde{u}_\Theta, \Tilde{p}_\Theta, \Tilde{\theta}_\Theta$. For the temperature we consider:
\begin{equation*}
	\Tilde{\theta}_\Theta(x, y, t; \omega) = 
	60 (1 - e^{-\omega t})\left(1- \tfrac{\sqrt{x^2+y^2}}{r}\right)\theta_\Theta(x, y, t; \omega) + 270\,.
\end{equation*}
In the following and the above equation, we omit the temperature unit. Our restricted network $\Tilde{\theta}_\Theta$ fulfils the initial and outer boundary conditions. Here $r = 5$, describing the radius of the whole circle. 
The only constrain for the pressure is the time scale: 
\begin{equation*}
	\Tilde{p}_\Theta(x, y, t; \omega) = (1 - e^{-\omega t})p_\Theta(x, y, t; \omega)\,.
\end{equation*}
To somewhat implement the movement of the rotating bar into our velocity network, we choose the constraints:
\begin{equation*}
	\Tilde{u}_\Theta(x, y, t; \omega) = r\omega (1 - e^{-\omega t} + \omega t e^{-\omega t}) \left(1- \tfrac{\sqrt{x^2+y^2}}{r}\right) \left( \begin{array}{cc}
		\cos(-\omega t (1 - e^{-\omega t})) u_{\Theta,1}(x, y, t; \omega)\\ \sin(-\omega t (1 - e^{-\omega t})) u_{\Theta,2}(x, y, t; \omega)
	\end{array} \right)  \,.
\end{equation*}
Again, $\Tilde{u}_\Theta$ naturally satisfies the initial and boundary conditions on $\Gamma_\text{out}$. All neural networks are FCNs, with hyperbolic tangent as activation. The networks $u_\Theta, \theta_\Theta$ consist of seven hidden layers with 100 neurons each and $p_\Theta$ use networks with 5 layers and 80 neurons.  

The only conditions that have to be learned are the Eqs.~\eqref{eq:Bosch_ns}--\eqref{eq:Bosch_velocity_bc} This is done via the classical PINN approach, as described in Sec.~\ref{ch:PINNs}, and applied to the constrained networks.
Since the temperature does not influence the fluid profile, we first train the solution of the Navier-Stokes equations. Then, the results are used in the training of the heat equation. For the optimisation process, Adam \cite{kingma2014adam} is used and randomly distributed training points are resampled in each iteration, for both the inner part and the moving boundary.

\begin{table}[htp]
	\centering
	\begin{tabular}{c|c|c}
		$\rho$  & $\nu$ & $a$ \\\hline
		1.0 &  2.0 & 0.5 \\ 
	\end{tabular}
	\caption{Material and operation parameter}
	\label{tab:Bosch}
\end{table}

\begin{figure}[!ht]%[htp]
	\centering
	\includegraphics[width=0.75\linewidth]{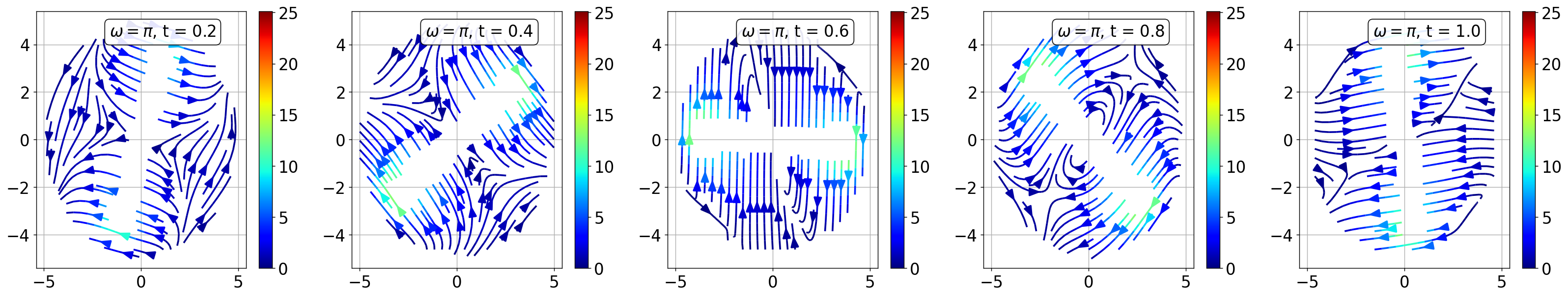}
	\\
	\includegraphics[width=0.75\linewidth]{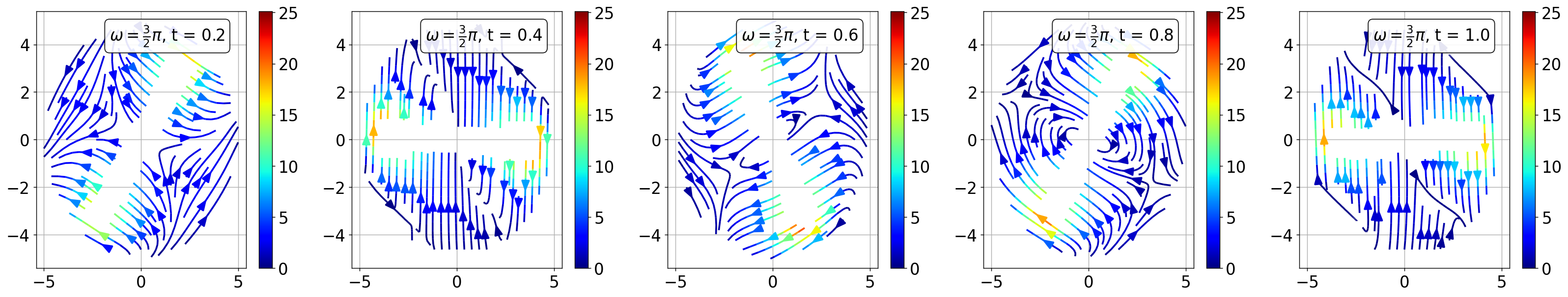}
	\\
	\includegraphics[width=0.75\linewidth]{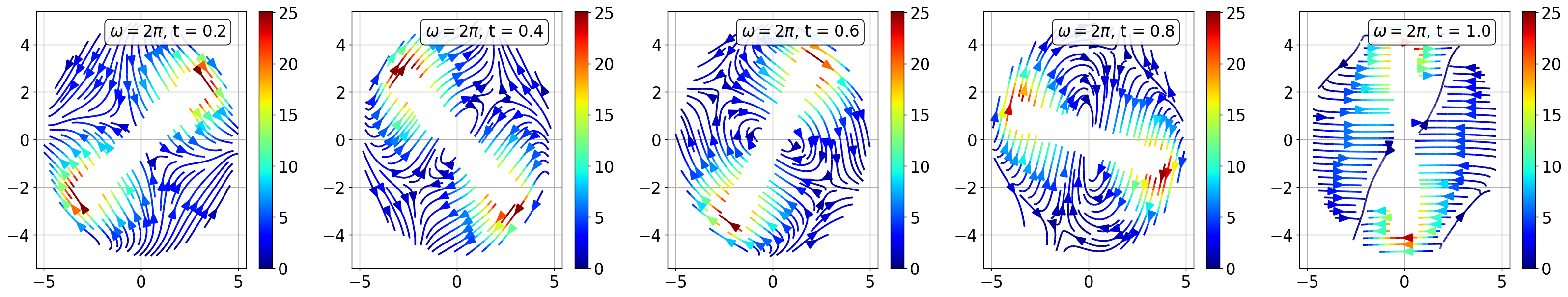}
	\caption{Learned velocity profile for different angular velocities $\omega$, at some discrete time steps.}
	\label{fig:bosch_u_solution}
\end{figure}

\begin{figure}[!ht]%[htp]
	\centering
	\includegraphics[width=0.9\linewidth]{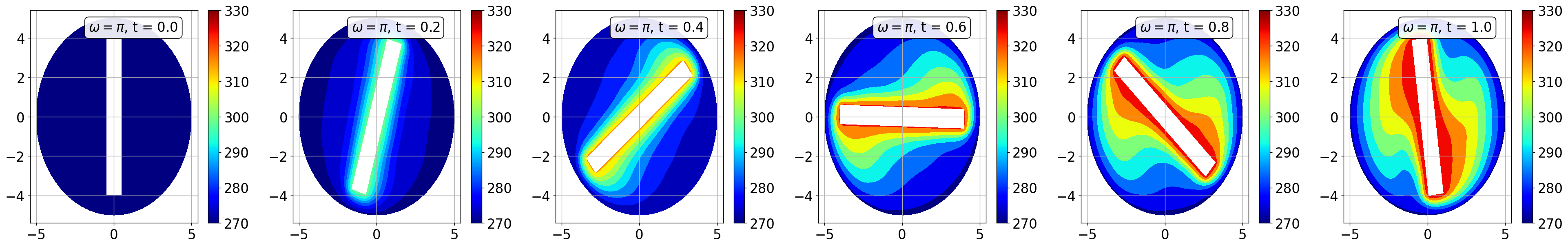}
	\\
	\includegraphics[width=0.9\linewidth]{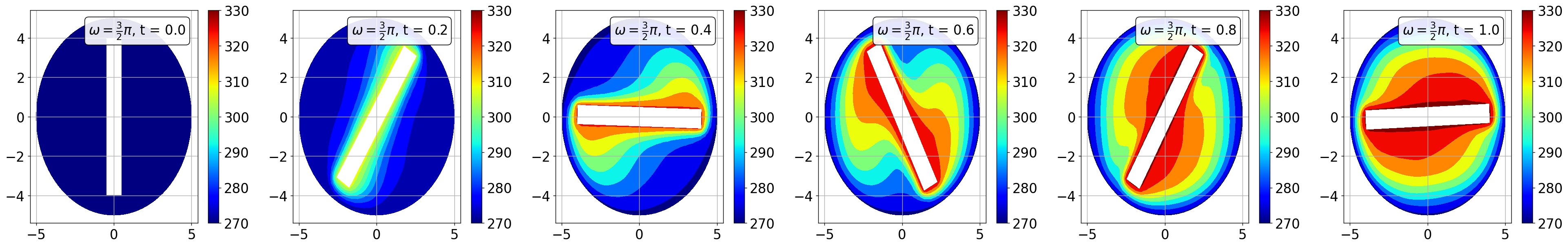}
	\\
	\includegraphics[width=0.9\linewidth]{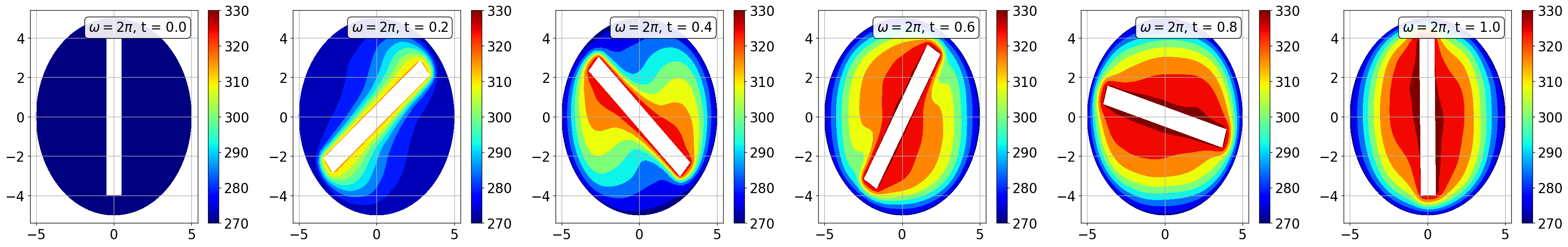}
	\caption{Learned temperature profile for different angular velocities $\omega$, at some discrete time steps.}
	\label{fig:bosch_temp_solution}
\end{figure}

The results are presented in Figs.~\ref{fig:bosch_u_solution} and \ref{fig:bosch_temp_solution}. Here two aspects have to be mentioned. Firstly, getting convergence of the training loss was rather challenging, and relatively large loss values for the divergence and heat term were obtained. Secondly, no reference solution is currently available. Therefore, we can not directly measure the accuracy of the learned approximation. Solely from a physical point of view, the results appear  reasonable and would allow for a first step in a rapid prototyping cycle. However, this aspect has to be further investigated, in general, the research for this problem is, from our side, in a rather early stage. This example is shown primarily to present a new application area (moving domains) for the deep learning approaches, where they may be advantageous compared to classical methods, and to demonstrate the potential of \textsc{TorchPhysics}.

\subsubsection{Identification of material model parameters from stress tensor (Volkswagen, Fraunhofer ITWM)}
To shorten the design cycle of vehicles and to reduce the cost of development, automotive industries employ numerical simulation tools in the vehicle development process for testing and analysis. In this application example, we are interested in the interaction of vehicles with various roadbeds such as sand, snow, mud, etc. Vehicle stability depends largely on this interaction, and the safety of the passengers is thus a concern. 
To approach this problem, a full-body model of the vehicle dynamics is needed as well as proper modelling of the roadbed. Of interest to us, is the modelling of the roadbed consisting of granular material. This is a continuum mechanics problem that involves not only the well-known conservation equations of mass, momentum, and energy but also, supplementary phenomenological material models. While the former specifies the process conditions and is generally well understood, the latter relates the applied strain to the resulting stress and comes with uncertainties as well as non-linearities. Obviously, the overall goal is for the simulations to match the real-life experiments, thus the selected material model is of great importance. 

Material models have parameters that are specific to the considered material as well as its reaction to external conditions, and these models range from simple to complex. By using single-parametric models for the granular material (roadbed), for example, one notices an increasing deviation between simulations and experiments as the simulation time progresses. Also, the simulations have long running times. As a result, complex material models with many more parameters are used. Such parameters are usually determined by a great wealth of expert knowledge, and costly experiments. The barodesy model \cite{kolymbas2012barodesy1, kolymbas2012barodesy2} is one of such complex material model which conforms to the basic mechanical properties of the material. It is formulated in tensorial form by Equations ~\eqref{eqn:barodesy_S}--\eqref{eqn:barodesy_e}.
\begin{eqnarray}
	\frac{d \mathbf{S}}{d t} &=& \mathbf{W S}-\mathbf{S} \mathbf{W}+\mathbf{H}(\mathbf{S}, \mathbf{D}, e) \label{eqn:barodesy_S} \\
	\frac{d e}{d t} &= &(1+e) \cdot \operatorname{tr}(\mathbf{D}) \label{eqn:barodesy_e} ,     
\end{eqnarray}
with %$\mathbf{W}=\frac{1}{2}(\nabla \mathbf{v}^{\mathrm{T}}-(\nabla \mathbf{v}^{\mathrm{T}})^{\mathrm{T}})$, the antisymmetric part of the velocity gradient and $\mathbf{D}=\frac{1}{2}(\nabla \mathbf{v}^{\mathrm{T}}+(\nabla \mathbf{v}^{\mathrm{T}})^{\mathrm{T}})$, the stretching tensor (the symmetric part of the velocity gradient)
%$$\mathbf{D}=\frac{1}{2}(\nabla \mathbf{v}^{\mathrm{T}}+(\nabla \mathbf{v}^{\mathrm{T}})^{\mathrm{T}}) \text{ and }  \mathbf{W}=\frac{1}{2}(\nabla \mathbf{v}^{\mathrm{T}}-(\nabla \mathbf{v}^{\mathrm{T}})^{\mathrm{T}}),$$ such that
% \begin{eqnarray*}
	% \mathbf{D} &=& \frac{1}{2}(\nabla \mathbf{v}^{\mathrm{T}}+(\nabla \mathbf{v}^{\mathrm{T}})^{\mathrm{T}})\\% \text{, the stretching tensor (the symmetric part of the velocity gradient)}\\
	% \mathbf{W} &=& \frac{1}{2}(\nabla \mathbf{v}^{\mathrm{T}}-(\nabla \mathbf{v}^{\mathrm{T}})^{\mathrm{T}})\\% \text{, the antisymmetric part of the velocity gradient}\\
	% \mathbf{H}(\mathbf{S}, \mathbf{D}, e) &=& h_b(\sigma) \cdot\left(f_b \mathbf{R}^0+g_b \mathbf{S}^0\right) \cdot|\mathbf{D}|\\
	% \sigma &=& |\mathbf{S}| = \sqrt{\operatorname{tr}\left(\mathbf{S}^2\right)} \\
	% \mathbf{R} &=& \operatorname{tr}\left(\mathbf{D}^0\right) \cdot \mathbf{I}+c_1 \cdot \exp \left(c_2 \cdot \mathbf{D}^0\right) \\
	% h_b &=& \sigma^{c_3} \\
	% f_b &=&  c_4 \cdot \operatorname{tr}\left(\mathbf{D}^0\right)+c_5 \cdot\left(e-e_c\right)+c_6 \\
	% g_b &=& -c_6 \\
	% e_c &=& \left(1+e_{c 0}\right) \cdot \exp \left(\frac{\sigma^{1-c_3}}{c_4 \cdot\left(1-c_3\right)}\right)-1, 
	% \end{eqnarray*}

$$\begin{aligned}
	\mathbf{D} &=\frac{1}{2}\left(\nabla \mathbf{v}^{\mathrm{T}}+\left(\nabla \mathbf{v}^{\mathrm{T}}\right)^{\mathrm{T}}\right) &  \mathbf{R} &=\operatorname{tr}\left(\mathbf{D}^0\right) \cdot \mathbf{I}+c_1 \cdot \exp \left(c_2 \cdot \mathbf{D}^0\right)\\
	\mathbf{W} &=\frac{1}{2}\left(\nabla \mathbf{v}^{\mathrm{T}}-\left(\nabla \mathbf{v}^{\mathrm{T}}\right)^{\mathrm{T}}\right) &  h_b &=\sigma^{c_3} \\
	\mathbf{H}(\mathbf{S}, \mathbf{D}, e) &=h_b(\sigma) \cdot\left(f_b \mathbf{R}^0+g_b \mathbf{S}^0\right) \cdot|\mathbf{D}| & f_b &=c_4 \cdot \operatorname{tr}\left(\mathbf{D}^0\right)+c_5 \cdot\left(e-e_c\right)+c_6 \\
	\sigma &=|\mathbf{S}|=\sqrt{\operatorname{tr}\left(\mathbf{S}^2\right)} & g_b &=-c_6\\ 
	\mathbf{S}^0 &=\mathbf{S}/|\mathbf{S}|, \mathbf{D}^0=\mathbf{D}/|\mathbf{D}|, \mathbf{R}^0=\mathbf{R}/|\mathbf{R}|  & e_c &=\left(1+e_{c 0}\right) \cdot \exp \left(\frac{\sigma^{1-c_3}}{c_4 \cdot\left(1-c_3\right)}\right)-1
\end{aligned}$$

where 
\begin{itemize}
	%\item[-] $\mathbf{S}^0=\mathbf{S}/|\mathbf{S}|$, $\mathbf{D}^0=\mathbf{D}/|\mathbf{D}|$, and $\mathbf{R}^0=\mathbf{R}/|\mathbf{R}|$
	\item[-] $\mathbf{S}$ is the Cauchy stress tensor (with principal stresses $\sigma_1, \sigma_2, \sigma_3$ in axial and lateral directions),
	\item[-] $\mathbf{W}$ is the anti-symmetric part of the velocity ($\mathbf{v}$) gradient, 
	\item[-] $\mathbf{D}$ is the stretching tensor (the symmetric part of the velocity gradient), 
	\item[-] $e= V_p/V_s$ is the void ratio, with a critical void ratio $e_c$, where $V_p$ and $V_s$ are the volumes of pores and solids (grains).
	\item[-] $t$ is the time.
\end{itemize}
The highly non-linear function $\mathbf{H}$ introduces the material parameters $c_1, c_2, c_3, c_4, c_5, c_6,$ and $ec_0$, (as well as other constants defined above) which we will identify via deep learning in a supervised task. 

%\paragraph*{Setting and Data Generation}
Using the MESHFREE software \cite{kuhnert2021meshfree} we generate parameters-stress pairs to train our neural network. MESHFREE employs a Generalised Finite Difference Method (GFDM): a meshless approach to numerically solve (coupled) PDEs based on their strong formulation on a sufficiently dense cloud of points carrying the physical information (such as velocity, pressure, etc.). A weighted moving least squares method is used to approximate the required spatial partial derivatives on a finite point set \cite{ostermann2013meshfree, michel2017meshfree}. MESHFREE has successfully been applied for the simulation of complex continuum mechanics problems in industry, like mining \cite{michel2021meshfree}, flow inside a turbine \cite{kuhnert2017fluid}, and vehicles travelling through water \cite{jefferies2015finite}, just to name a few.

\begin{figure}[!ht]
	\begin{subfigure}{0.5\textwidth}
		\centering
		\includegraphics[width=\textwidth]{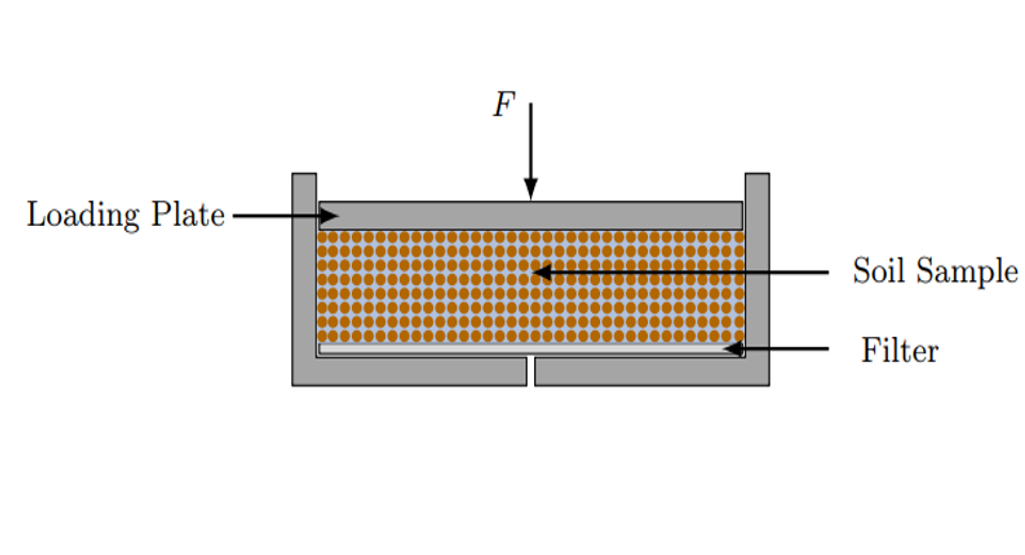}
		\caption{}
		\label{fig:oedometric_test}
	\end{subfigure}\hfill
	\begin{subfigure}{0.4\textwidth}
		\centering
		\includegraphics[width=\textwidth]{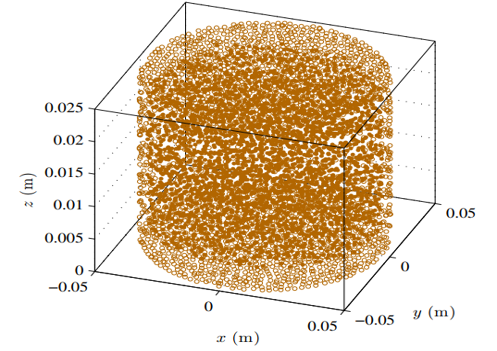}
		\caption{}
		\label{fig:meshfree_t0}
	\end{subfigure}
	\caption{(a) Schematic illustration of the oedometric test. (b) Initial FPM point cloud configuration for the oedometric test (filled circles: interior points, non-filled circles: boundary points) using MESHFREE (see \cite{ostermann2013meshfree})} 
\end{figure}

%\paragraph*{Problem/Simulation Setting}
In our case, the simulation setting is that of the oedometric test: one of two benchmark problems usually considered in soil mechanics to, e.g., check and calibrate material models. It does so by simulating a one-dimensional compression; the soil (material) sample is loaded in the axial direction and rigid side walls prevent any lateral expansion as demonstrated in Figure \ref{fig:oedometric_test}. Figure \ref{fig:meshfree_t0} shows the initial point cloud configuration for this test. 

% \begin{figure}[htp]
	%     \centering
	%     \includegraphics[width=0.60\textwidth]{figures/material_model/oedometric_test.png}
	%     \caption{Schematic illustration of the oedometric test.}
	%     \label{fig:oedometric_test}
	% \end{figure}

\begin{figure}[!ht]
	\centering
	\includegraphics[width=0.65\textwidth]{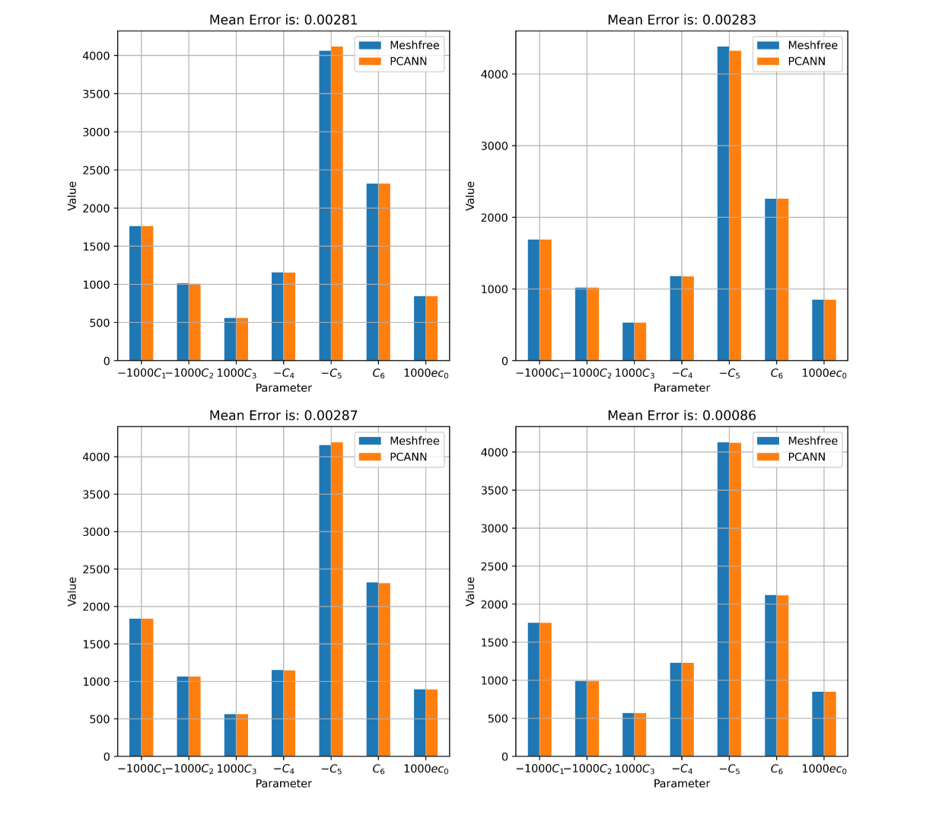}
	\caption{Comparison of ground truth(MESHFREE simulation in blue)  and learned parameters (PCANN in orange) for four randomly chosen examples of the oedometric test}
	\label{fig:meshfree_err}
\end{figure}

%\paragraph*{Procedure and Results}
Randomised PCA as in Section \ref{sec:pcann} is used to reduce the dimensions of the principal stress component from $675$ to $50$. This reduced dimension is then used as input to the FCN (fully connected network) similar to that in Section \ref{sec:pcann}. Since the parameter space is low enough, no PCA is used after the FCN. As a result, the output of the FCN yields the target parameters directly. Of the $6000$ data pairs generated, $75\%$ is used for training. During training, the relative L2 error of the individual parameters is calculated and their average is the loss function minimised for training the network. However, due to the nature of this loss function, the learning of the parameters of higher magnitude is favoured. As a remedy, the parameters of lower magnitude are scaled so that they are of the same order (of thousands) as the parameters of higher magnitude. In this way, learning of all the individual parameters is achieved. We obtained an average relative L2 test error of $2.63 \times 10^{-3}$ on the test data set. Figure \ref{fig:meshfree_err} shows the comparison of the ground truth (MESHFREE simulation in blue) and the learned parameters (PCANN in orange). A detailed discussion of this application is currently in preparation \cite{nganyutanyu2023pidl}.

% \begin{figure}[htp]
	%     \centering
	%     \includegraphics[width=\textwidth]{figures/material_model/meshfree_err1.png}
	%     \caption{Comparison of ground truth(MESHFREE simulation in blue)  and learned parameters (PCANN in orange) for four randomly chosen examples of the oedometric test}
	%     \label{fig:meshfree_err}
	% \end{figure}

% \begin{figure}[htp]
	% 	\begin{subfigure}{0.5\textwidth}
		% 		\centering
		% 		\includegraphics[width=\textwidth]{figures/material_model/meshfree_data.png}
		% 		\caption{Dataset examples}
		% 		\label{fig:meshfree_data}
		% 	\end{subfigure}\hfill
	% 	\begin{subfigure}{0.5\textwidth}
		% 		\centering
		% 		\includegraphics[width=\textwidth]{figures/material_model/meshfree_loss.png}
		% 		\caption{Evolution of loss during training}
		% 		\label{fig:meshfree_loss}
		% 	\end{subfigure}\\
	% 	\begin{subfigure}{\textwidth}
		% 		\centering
		% 		\includegraphics[width=0.5\textwidth]{figures/material_model/meshfree_err.png}
		% 		\caption{Comparism of ground truth and learned parameters for 4 randomly chosen examples}
		% 		\label{fig:meshfree_err}
		% 	\end{subfigure}
	% 	\caption{ }
	% \end{figure}
\section{Conclusions}
The presented survey aims at developing a guideline on whether and how to use deep learning concepts for PDE-based forward solvers and related inverse problems. To this end, we have implemented and tested the most widely used neural network concepts as well as several of their extensions for the two most common PDE problems (Poisson, Darcy flow). 
In addition, we have applied neural network concepts to two industrial projects. 

Naturally, the present text tries to give a complete-as-possible snapshot of the situation at the time we started writing up the results of our test.
We are well aware that several additional concepts have been proposed since and we apologise to those whose contributions are missing.

There is no unique criterion for determining the best method, this very much depends on the specific task at hand. Hence, we presented a survey of numerical values for each method, which should allow the user to evaluate the particular criterion relevant to its specific application and to determine a suitable method accordingly.

Nevertheless, we want to highlight some general findings.
Within the restricted scope of our test environment, we have found the following conclusions. DL concepts for solely solving forward problems lack precision as compared with classical FEM methods. This finding has been confirmed by other authors as well, see e.g. a numerical study for PINNs in  \cite{grossmann2023can}. However, in general, they achieve reasonable accuracy and offer great potential for efficient parameter studies, e.g. for applications in rapid prototyping.

Our findings differ depending on whether the underlying PDE is linear or non-linear and whether training data, i.e. sufficiently many parameter-state pairs are known, is available. Several methods such as Ritz-method or PINN only use the PDE and do not use training data. Hence the comparison has to distinguish these cases

For a linear PDE and 
if only the PDE is known but no data is available, we propose to use PINN for solving the forward problem. If data, i.e. sufficiently many parameter-state pairs are known, then PCA or even better PCALin are an excellent starting point. As mentioned, this is only the first indication, our tests were restricted to the Poisson problem.
Surprisingly, see Table
\ref{tab:noise-backward-operator-trained-noise-65},
there is not much difference between different concepts for solving the inverse Poisson problem by backward operator training.
All methods considered produce reliable results e.g. for $5 \% $
noise in the data. Hence, run times might be the criterion for the decision. Obviously, concepts for operator training are advantageous and  DeepONet and PI-DeepONet are the overall winners in terms of run-time for testing.

The Poisson forward problem is a rather simple linear elliptic  Hence, the numerical test should be seen as some preliminary test giving some limited insight into the behaviour of DL concepts for forward linear PDEs. As we will see from the more evolved examples, the findings in this section have only a limited meaning for more general cases.

Concerning solvers for the  Darcy problem,  MWT is the overall winner for solving the forward problem with noise-free data, at least in our experiments. For solving the inverse Darcy problem
we propose to use inverse training in combination with  
U-FNO and MWT  for small noise levels and PCANN for solving the non-linear inverse Darcy flow problem with higher noise levels.

As a general remark, we observe that concepts which combine data-driven learning with mathematical concepts such as reduced order models or incorporating the PDE directly, e.g. PCANN/PCALin, PINO, FNO or MWT, have an advantage in terms of stability and performance. Again, we want to emphasise that it is premature to make this a general rule, this is rather an observation restricted to our numerical setting.

Naturally, there are many meaningful extensions of the chosen test scenario. E.g. inserting a classical FDM/FEM forward solver into 
a Tikhonov regularisation scheme or including $H^1$ - error measures for comparison would be most interesting. However, we prefer to stay with the present scope of focusing on comparing DL methods for inverse problems and restrict ourselves to the tests provided.

Concerning industrial applications, additional features gain influence. E.g. the simulation of heat inside a rotating engine requires working with a time-dependent domain of definition. Most codes will require substantial additional work when dealing with such complex domains. TorchPhysics offers an accessible concept for working with such domains, which in connection with a standard PINN concept did yield good performance for a parametric study in this context.

The other industrial problem of identifying parameters in a problem of mechanical engineering reduces to a rather low-dimensional task and can be solved by PCANN. Hence, choosing a method suitable for a given industrial problem requires problem-specific considerations, which have to be evaluated task by task.

We should remark, that the size of the network differs considerably. We have performed rather extensive hyper-parameter searches for each method. The reported network sizes did yield the best results. E.g. the DRM does not benefit from extending the network to larger sizes for the mentioned application.

\begin{figure}[htp]
	\centering
	\includegraphics[width=0.68\linewidth]{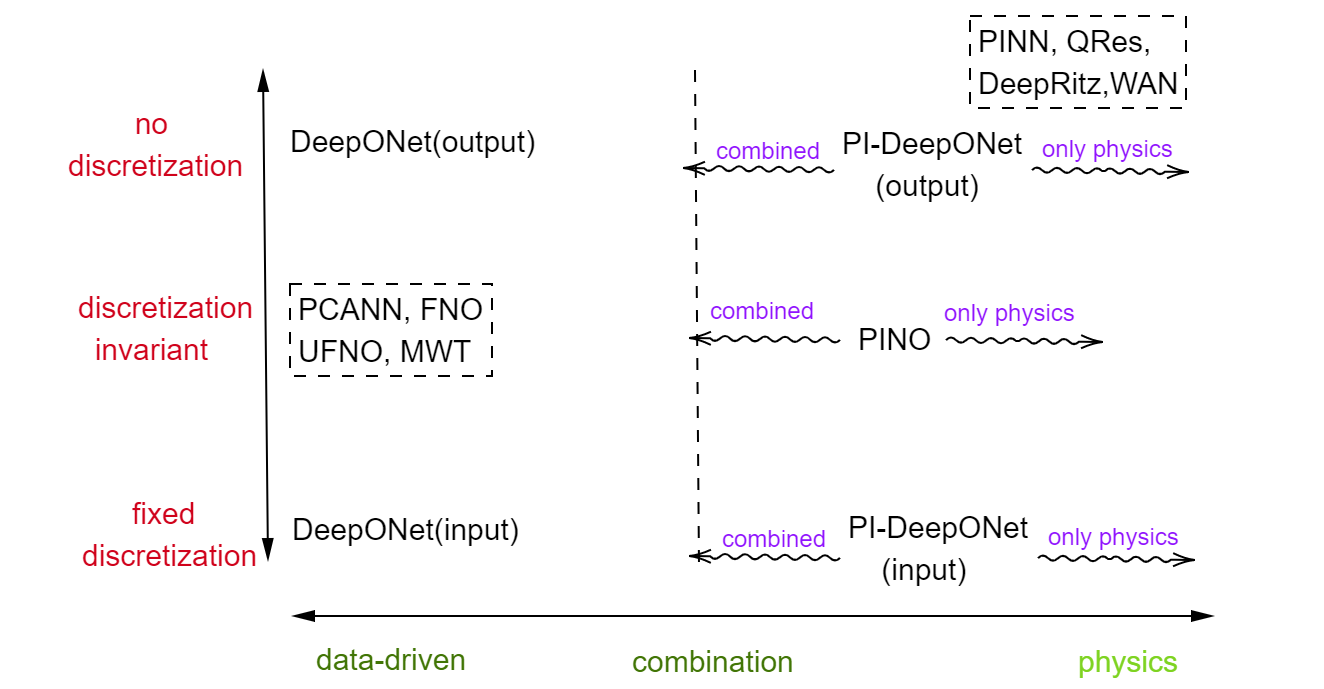} %[width=0.68\linewidth]
	\caption{Properties of deep learning methods for PDEs.}
	\label{fig:dlforpde}
\end{figure}

In Table \ref{tab:methods}, we list the properties of classical methods, Deep Ritz method in \cite{yu2017deep}, PINN in \cite{raissi2019physics}, model reduction and Neural Network in \cite{bhattacharya2020model}, DeepONets in \cite{lu2019deeponet} and Fourier Neural Operator \cite{li2020fourier}. We should mention that the properties only correspond to the methods in these mentioned papers, since by some modifications, they can have different properties. For example, the PINN may be used to model parameter-solution operators. In Figure\ref{fig:dlforpde}, we schematically illustrate the properties of different deep learning methods for PDEs.

The methods discussed above have shown their high potential for solving PDEs as well as their related parametric studies and inverse problems. Whenever efficiency is an issue, these methods perform most convincingly. However,
there are still many open questions and problems that are common in the deep learning community.  Hyperparameter search can be very time-consuming and would benefit from a sound theoretical foundation. Also, the choice of activation functions and loss functions can be more refined.

\bibliography{bibfile}

\appendix

\section{Appendix} \label{sec:appendix}

\subsection{Application Problems}

\subsubsection{Darcy Flow}
For this problem, we consider the steady state of the 2-d Darcy Flow equation on the unit box which is the second
order, linear, elliptic PDE

\begin{equation}
	\begin{aligned}
		-\nabla \cdot(\lambda(s) \nabla u(s)) &=f(s) & & s \in(0,1)^{2} \\
		u(s) &=0 & & s \in \partial(0,1)^{2}
	\end{aligned}
	%\label{eqn:darcy}
\end{equation}
with a Dirichlet boundary where $a \in L^{\infty}\left((0,1)^{2} ; \mathbb{R}_{+}\right)$ is the diffusion coefficient,  $f = 1 \in L^{2}\left((0,1)^{2} ; \mathbb{R}_{+}\right)$ is the forcing function and $u \in H_{0}^{1}((0,1)^{2} ; \mathbb{R})$ is the unique solution of the  Some few applications of this PDE are seen in the domains of modelling	the pressure of the subsurface flow, the deformation of linearly elastic materials, and the electric potential in conductive materials.

\subsubsection{Poisson Equation}
We are interested in the PDE
\begin{equation}
	\begin{aligned}
		-\Delta u(s) &=\lambda(s) & & s \in(0,1)^{2} \\
		u(s) &=0 & & s \in \partial(0,1)^{2}
	\end{aligned}
	\label{eqn:poisson-append}
\end{equation}
which is a special case of Equation \ref{eqn:darcy} where the diffusion coefficient is constant and equals to $1$ everywhere in the domain $\Omega = (0, 1)^2$.

\subsection{Setting of Numerical Implementations} \label{apdx:num_sett}
The details of the experiments are provided in Section \ref{sec:Numerical experiments}. The computing environment is described below: 
\begin{itemize}
	\item GPU: 8x nVidia GeForce GTX 2080 Ti
	\item CPU: Intel Xeon Gold 6234 (8 cores, 32 threads)
	\item RAM: 1500 GB of system RAM
	\item OS: Ubuntu 16.04.7 LTS
\end{itemize}

While the parameters for the Poisson problem are sampled from the distribution $\mu_G$, those for the Darcy Flow are sampled on $\mu_P$ (piece-wise constant). These distributions are described in Section \ref{sec:Numerical experiments}.

For the operator learning scenarios (PCANN, FNO, UFNO, DeepONet), for both forward and inverse problems of the Poisson and Darcy Flow problems, we train the neural operator on $1000$ training data-set pairs we evaluate on $5000$ test pairs. This also holds for the pre-training phase of the PINO.

The experiments for DRM, PINN, QRES, DeepONet and PI-DeepONet were all done in the deep learning library \textsc{TorchPhysics} \cite{TorchPhysics}

All codes will be available online at \href{https://github.com/dericknganyu/dl_for_pdes}{https://github.com/dericknganyu/dl\_for\_pdes}.

\subsubsection{DRM, PINNs and QRES}
\begin{table}[htp]\small
	\centering
	%\resizebox{\textwidth}{!}{%
		\begin{tabular}{c|c|cc|cc|cc}
			% & \multicolumn{6}{c|}{\textbf{Poisson}}\\ \cline{2-7} %\hline
			\multicolumn{2}{c}{}  & \multicolumn{4}{c|}{PINNs \& QRES} & \multicolumn{2}{c}{DRM}  \\  \cline{3-8} %\hline
			\multicolumn{2}{c|}{}  & \multicolumn{2}{c|}{\textit{Forward} } & \multicolumn{2}{c|}{\textit{Inverse}} & \multicolumn{2}{c}{\textit{Forward}} \\  \cline{3-8} % \hline
			\multicolumn{1}{c}{}  &     Hyperparameter & ADAM & \text{LBFGS} & \text{ADAM} & \text{LBFGS} & \text{ADAM} & \text{LBFGS}\\ \hline
			\multirow{4}{*}{\textbf{Poisson}} &    Learning rate & 0.001 & 0.2 & 0.001 & 0.5 & 0.001 & 0.2\\
			&    Iterations & 5,000 & 8,000 & 5,000 & 10,000 & 5,000 & 5,000 \\
			&   Scale factor $s$ & \multicolumn{2}{c|}{100} & \multicolumn{2}{c|}{100} & \multicolumn{2}{c}{100} \\
			&   $(\omega_\mathcal{L}, \omega_\mathcal{B}, \omega_\mathcal{D})$ & \multicolumn{2}{c|}{(1, 10,000, 0)} & \multicolumn{2}{c|}{(1, 0, 10,000)}  & \multicolumn{2}{c}{(1, 10,000, 0)}
			\\ \hline
			% & \multicolumn{6}{c|}{\textbf{Darcy Flow}}\\ \cline{2-7} %\hline
			% & \multicolumn{4}{c|}{PINNs \&
				% QRES} & \multicolumn{2}{c|}{DRM}  \\  \cline{2-7} %\hline
			% & \multicolumn{2}{c|}{\textit{Forward} } & \multicolumn{2}{c|}{\textit{Inverse}} & \multicolumn{2}{c|}{\textit{Forward}} \\  \cline{2-7} % \hline
			% Parameter & ADAM & \text{LBFGS} &
			% \text{ADAM} & \text{LBFGS} & \text{ADAM} &
			% \text{LBFGS}\\\hline
			\multirow{4}{*}{\textbf{Darcy Flow}} &  Learning rate & 0.001 & 0.25 & 0.005 & 0.5 & 0.002 & 0.25\\
			&  Iterations & 5,000 & 12,000 & 5,000 & 15,000 & 5,000 & 8,000 \\
			& Scale factor $s$ & \multicolumn{2}{c|}{0.01} & \multicolumn{2}{c|}{100} & \multicolumn{2}{c}{0.01} \\
			& $(\omega_\mathcal{L}, \omega_\mathcal{B}, \omega_\mathcal{D})$ & \multicolumn{2}{c|}{(1, 100, 0)} & \multicolumn{2}{c|}{(1, 0, 100)}  & \multicolumn{2}{c}{(1, 100, 0)}
		\end{tabular}%\label{comparion}
		%}
	\caption{Used parameters for the DRM, PINNs and QRES in the simulations.}
	\label{tab:pinn-hparams}
\end{table}

Since the general training procedures of DRM, PINNs and QRES are closely related, we explain the implementation of all 3 approaches at the same time.

The accuracy of the results mainly depends on the size of the network, the number of training points and the range of the solution. To get a fair comparison between PINN and QRES, we choose the networks in such a way that they have roughly the same number of trainable parameters. We considered both accuracy and training time while choosing the network architecture. All other Hyperparameters for PINNs and QRES are the same.
In the inverse problem, we use two networks $(u_\Theta, f_\Theta)$. Where $u_\Theta$ interpolates the given data and $f_\Theta$ approximates the searched data function.

In the Poisson problem for PINNs, we use a FCN with 3 hidden layers and 50 neurons. For QRES the architecture \ref{fig:network_overview} with 3 layers and 36 neurons is utilised. In the inverse problem, $(u_\Theta, f_\Theta)$ have both the same structure as in the forward problem.
The used neural network in the Deep Ritz method consists of 2 residual blocks and the initial and last FC layer. Where each layer has 40 neurons.   

For the Darcy equation, deeper neural networks were in general more advantageous. Therefore, the Deep Ritz network consisted of 3 residual blocks with 60 neurons. For the forward problems with PINNs, we applied an architecture with 4 layers, where the number of neurons of layer $k$ was the $k$-th entry in $(50, 50, 50, 20)$. Analogue, for QRES a network with the structure $(36, 36, 36, 15)$ was used. For the inverse problem, the PINN $u_\Theta$ was given by $(50, 50, 20)$ and for QRES by $(36, 36, 15)$. For the network of the data function $f_\Theta$, we choose for PINNs 
$(30, 40, 40, 20, 20)$ and for QRES $(25, 35, 25, 15, 15)$.

The strong formulation of the Darcy equation (\ref{eqn:darcy}) with a piece wise constant coefficient $\lambda$ is ill-posed at the points where the jump occurs. This leads to an additional challenge for the training of the forward problem with PINNs and QRES. When the derivatives of $\lambda$ become large, so does the loss function, and the optimisation is mainly focused on reducing the influence of the derivatives. Which leads overall to a worse approximation of the real solution. To avoid this problem, in \cite{darcy_weighted} a special weight function is introduced. The function is constructed in such a way, that the weight is zero at the interfaces where the discontinuities are and $\mathcal{O}(1)$ away from the interfaces. If we apply this idea to our problem, we have to remove all points where the discrete jumps occur and the equation reduces everywhere else to $\lambda \Delta u = f$, since our parameter is piece-wise constant. 
For the DRM we can use the original PDE (\ref{eqn:darcy}). This is also possible in the inverse case, since there the unknown parameter $\lambda$ is approximated by a smooth neural network.

Additionally, neural networks can best work with values in the range of $[-1, 1]$, because of this we scale our data by a factor $s$. The used values are listed in table \ref{tab:pinn-hparams}. For the forward problem the given data function is scaled, for the inverse problem, the known solution $u$ is multiplied by $s$.

Training is done with a combination of Adam \cite{kingma2014adam} and LBFGS \cite{liu1989limited}. We start with a fixed number of iterations of Adam and then switch over to LBFGS, like usually done in the literature. Since two different optimisation algorithms are used, the needed time per iteration (epoch) in table \ref{tab:errors-513-poisson} is an average over both algorithms. For LBFGS we use the PyTorch implementation and set the \textit{max\_iteration} variable to 2. In the inverse case, both data fitting and the minimisation of the PDE loss are trained at the same time. For LBFGS all available points have to be used in each training step, this leads to a rather slow training process for higher resolutions. Using lower resolutions leads to roughly the same accuracy, but speeds up the training considerably.

To control the influence of each condition we add additional weights $(\omega_\mathcal{L}, \omega_\mathcal{B}, \omega_\mathcal{D})$. The used values are presented in table \ref{tab:pinn-hparams}. Since the magnitude of the loss of the PDE condition is much larger than the loss of the other conditions, we use a large weight for the boundary and data conditions. If the weight is zero, in table \ref{tab:pinn-hparams}, this means the condition did not appear in the training. E.g. no additional boundary condition in the inverse case and data of the solution in the forward problem. Lastly, we want to mention, that if the boundary condition is built into the network architecture better results can be achieved.

\subsubsection{PCANN, PCALin}

The obtained error for the PCANN depends largely on the reduced/latent dimension used for the PCA. For each reduced dimension $d_\mathcal{X} = d_\mathcal{Y} \in [30, 50, 70, 100, 150, 200, 250]$, the hyperparameters were grid searched. The choice of this range was based on \cite{bhattacharya2020model}, which showed how larger dimensions do not necessarily guarantee better approximations for the PCANN. However, larger dimensions could lead to better results for the PCALin. The results shown in Tables \ref{tab:errors-513-poisson} and \ref{tab:errors-darcypwc} are for the best-performing cases. We present the hyperparameter setting for each separate case in Table \ref{tab:pcann-hparams}. 
ADAM was used in PCALin while SGD is was used in PCANN.

\begin{table}[htp]\small
	\centering
	%\resizebox{\textwidth}{!}{%
		\begin{tabular}{r|cc|cc|cc|cc}
			\multicolumn{1}{c}{} & \multicolumn{4}{c|}{\textbf{Poisson}} & \multicolumn{4}{c}{\textbf{Darcy Flow}}\\ \cline{2-9} %\hline
			& \multicolumn{2}{c}{PCANN} & \multicolumn{2}{c|}{PCALin} & \multicolumn{2}{c}{PCANN} & \multicolumn{2}{c}{PCALin}\\ \cline{2-9} %\hline
			Hyperparameter & \textit{Forward} & \textit{Inverse} & \textit{Forward} & \textit{Inverse} & \textit{Forward} & \textit{Inverse} & \textit{Forward} & \textit{Inverse}\\ \hline
			Latent dimension & 150 & 200 & 250 & 250 & 150  & 30 & 100 & 100 \\
			Weight decay & 0.1 & 0.1 & 0.015 & 0.1 & 0.5  & 0.1 & 0.1 & 0.15\\
			Batch size & 500 & 500 & 1000 & 1000 & 500  & 100 & 1000 & 200\\
			Learning rate & $5 \times 10^{-7}$ & $1 \times 10^{-5}$ & \multicolumn{2}{c|}{0.01, 0.001, 0.0025} & $1 \times 10^{-6}$ & $1 \times 10^{-4}$ & $1 \times 10^{-3}$ & $1 \times 10^{-3}$ \\
			Step size & 2000 & 2500 & \multicolumn{2}{c|}{2500, 5000} &  7500 & 2500 & - & 2500\\
			Gamma & 0.5 & 0.1 & \multicolumn{2}{c|}{0.01, 0.1} & 0.5 & 0.01 & - & 0.01\\
			Epochs & 20000 & 10000 & \multicolumn{2}{c|}{6000} & 20000  & 5000 &  2000 & 1000\\
			Dropout & 0 & 0.001 & \multicolumn{2}{c|}{0}  &  0 & 0.01 &  0 & 0  \\
			Weight Initialisation &  \multicolumn{2}{c|}{kaiming normal}  &  \multicolumn{2}{c|}{xavier}  &  \multicolumn{2}{c|}{kaiming normal}  &  \multicolumn{2}{c}{xavier} \\
		\end{tabular}%\label{comparion}
		%}
	\caption{Best performing hyperparameters for the PCANN method and its linear version-PCALin}
	\label{tab:pcann-hparams}
\end{table}

\subsubsection{FNO, UFNO, MWT, PINO}

We use the same hyperparameters for the 2D FNO/MWT problem as in the paper, and the code made available. Notably, training is done for $500$ \textit{epochs}, with a \textit{batch-size} of $10$. The ADAM algorithm is used for optimisation, with a \textit{learning rate} of $0.001$ which is reduced by a factor of $0.5$ (a.k.a. \textit{gamma}) after every $100$ (a.k.a. \textit{step size}) epochs. An L-2 regularisation is added to the loss function with a \textit{weight decay} of  $1 \times 10^{-4}$. Specific to the FNO, UFNO and PINO methods are the Fourier modes and width, we truncate $12$ \textit{Fourier modes} and use a \textit{width} size of $32$. 

On the other hand, in the MWT, Convolutional Neural Networks (CNNs) are used to parameterise the networks $A, B, C$ and $T$ is a single $k \times k$ linear layer with $k=4$. The Legendre orthogonal polynomial basis is used as it produces better results as reported in \cite{gupta2021multiwavelet}. 

For the PINO, we optimise the loss function $L$, given by:
$$L = L_{\text{data}} + 0.2 \cdot L_{\text{pde}}$$

\subsubsection{DeepONet}
For the possion problem, we use a linear model for the branch net, while the trunk net is a neural network with a width of 128 and 6 hidden layers and Tanh as the activation function. For Darcy-flow with piece-wise constant coefficients, we use the convolutional neural network for the branch net, while the trunk net is a neural network with a width of 128 and 4 hidden layers, and ReLU is the activation function. However, for the inverse training of the Darcy-flow problem, we use $\sin$ as the activation for the trunk net, which can achieve better results than using general activation functions.  We use the fast implementation to train the neural network as described in the algorithm. The training is done for 50000 epochs, with a batch size of $125$. The Adam algorithm is used for optimisation with a learning rate of $0.002$ which is reduced by a factor of 0.9 after every $1250$ epoch. 

We need to note that the resolution size will affect the size of the branch net of DeepONet. For example, for $513\times 513$ resolution case, the number of parameters in the branch net would be $513\times  513\times 128\approx 3.0\times 10^{7}$ and the server we use will face the memory limitation problem. Thus, for high-resolution cases, an adaptive average pooling is first used in the branch net to reduce the size of the data, which can reduce the number of parameters and training time.

\subsubsection{PI-DeepONet}
In our experiments on PI-DeepONet, we add an additional physics loss to the optimisation of a data-driven DeepONet described above. Therefore, we use architectures that are similar to the experiments done on DeepONet. In the case of inverse operator training for the Darcy-flow problem, we increase the size of the convolutional branch net and its adaptive average pooling.

In all cases, we re-scale the input and output data such that the values are roughly in the range of $[-1, 1]$ to improve the training performance. All terms in the physics loss are scaled accordingly.

The construction of a physics-based loss in the inverse operator training requires the derivatives of the discrete PDE solution. To obtain these, we come up with a concept that is inspired by the solution of inverse problems in PINN: We introduce a second trunk net that is optimised to interpolate the solution via mean squared error and shares the same branch network. The derivatives of the solution can then be computed via gradient descent and be included in the physics-based loss to reconstruct the parameter. As a result, in addition to the data- and physics-based loss, we have to minimise an interpolation loss $MSE_\mathcal{I}$. Both trunk nets are optimised simultaneously.

One has to choose the weights $\omega_\mathcal{L}, \omega_\mathcal{B}, \omega_\mathcal{D}$ (and $\omega_\mathcal{I}$) of the physics-informed, boundary condition, data-based (and interpolation) parts of the loss function. In most cases, we choose all weights to have the value $1$. Only in the Darcy-flow inverse operator learning, we scale the interpolation loss by $\omega_\mathcal{I}=5$ to improve the influence of the learned derivatives.

\begin{table}[!ht]\small
	\centering
	\begin{tabular}{c|ll|ll}
		\multicolumn{1}{c}{ } & \multicolumn{2}{c|}{ Poisson } & \multicolumn{2}{c}{ Darcy Flow } \\
		\cline { 2 - 5 } 
		Hyperparameter  & $\alpha_1$ & $\alpha_2$ & $\alpha_1$ & $\alpha_2$ \\
		\hline 
		PCANN & $0.0005$ & $0.05$ & $0.0001$ & $0.005$ \\
		PCALin & $0.00001$ & $0.025$ & $0.001$ & $0.001$ \\
		FNO & 0 & $0.25$ & 0 & $0.025$ \\
		UFNO & 0 & $0.05$ & 0 & $0.005$ \\
		MWT & 0 & $0.05$ & 0 & $0.005$ \\
		PINO & 0 & $0.05$ & 0 & $0.005$ \\
		DeepONet &0  &0.05  & 0 &0.025 \\
		PI-DeepONet &0  &0.05  & 0 &0.025  \\
	\end{tabular}
	%}
\caption{Hyperparameter configurations for Tikhnov-based methods for Inverse Problems.}
\label{tab:tikho-hparams}
\end{table}

In all settings, we train for $1, 650, 000$ iterations that consist of $128$ branch inputs and $100$ randomly sampled trunk input points. We reduce the learning rate in an iterative scheme for every $50,000$ steps. For the Poisson forward problem, we use an initial learning rate of $5e-4$ and reduce it iteratively by a factor of $0.8$. For the inverse operator training, we end up using an initial learning rate of $1.5e-4$ (Poisson) respectively $2.5e-4$ (Darcy) and a factor of $0.85$. %TODO: forward lr for Darcy?

During training, we use the full dataset, to ensure comparability to the other used methods. The advantages of a physics-informed approach might however come better into play in a setting in which only partial data is available, therefore, this might not be the optimal use case for PI-DeepONets.

\subsubsection{Tikhonov-based methods for Inverse Problems}

%\paragraph{PCANN, PCALin, FNO, U-FNO, MWT, PINO}

For learning the parameter, we used a learning rate of $0.1$ and trained for $10000$ epochs, decreasing the learning rate by a factor of $0.5$ every $2000$ epochs. The below loss functions are used.
\begin{eqnarray}
Loss(\lambda)_{\text{poisson}} &=& \Vert G_{\theta}\lambda - u_\delta\Vert_{2} + \alpha_1 \Vert \lambda\Vert_{2} + \alpha_2\Vert D (\lambda) \Vert_{2} \label{eqn:TikLoss1}\\
Loss(\lambda)_{\text{darcy~~}} &=& \Vert G_{\theta}\lambda - u_\delta\Vert_{2} + \alpha_1 \Vert \lambda\Vert_{2} + \alpha_2 \text{Var}(\lambda), \label{eqn:TikLoss2}
\end{eqnarray}
where $G_{\theta}$ is the forward operator, $D(\lambda)$ is the first difference quotient of $\lambda$, and Var is the Total Variation loss. 
Using same hyperparameter configuration, testing was done on 100 test examples and the results were reported in Tables \ref{tab:xxnoise-poisson-65} and \ref{tab:xxnoise-darcypwc-65}. For a visual appreciation of the results, we also show in Tables \ref{tab:fig-xxnoise-poisson-65} and \ref{tab:fig-xxnoise-darcypwc-65} the learned parameter for both Poisson and Darcy Flow with piece-wise constant parameter problems.

\end{document}